\documentclass[twocolumn,runningheads]{svjour3}

\usepackage{amssymb}
\usepackage{graphicx}

\usepackage{amsmath,amssymb,amscd} 
\def\bbbr{{\mathbb R}} 
\def\bbbz{{\mathbb Z}}

\newtheorem{mytheorem}[theorem]{Theorem}
\newtheorem{myproposition}[theorem]{Proposition}
\newtheorem{mydefinition}[theorem]{Definition}

\newenvironment{myproof}{\noindent{\bf Proof:} }{\hfill $\square$ \medskip}

\newcommand{\argmax}{\operatorname{argmax}}
\newcommand{\relu}{\operatorname{ReLU}}

\usepackage{url}

\usepackage{xcolor}
\usepackage{natbib}
\usepackage{graphicx}
\usepackage{subcaption} 
\usepackage{float}
\usepackage[switch]{lineno}
\usepackage{comment} 

\def\centre{\scriptsize\mbox{centre}}
\def\maximum{\scriptsize\mbox{max}}
\def\avg{\scriptsize\mbox{average}}
\def\logsumexp{\scriptsize\mbox{logsumexp}}

\journalname{arXiv preprint manuscript No.}

\begin{document}


\title{Scale-invariant Gaussian derivative residual networks%
\thanks{The support from the Swedish Research Council (contract 2022-02969) is gratefully acknowledged.}}

\titlerunning{Scale-invariant Gaussian derivative residual networks}

\author{Andrzej Perzanowski and Tony Lindeberg}

\institute{Andrzej Perzanowski (corresponding author)\\ amper@kth.se \medskip \\       
                Tony Lindeberg\\ tony@kth.se \medskip \\
                Computational Brain Science Lab,
                Department of Computational Science and Technology,
                KTH Royal Institute of Technology,
                SE-100 44 Stockholm, Sweden.}
              
\date{}          
\maketitle

\begin{abstract}
Generalisation across image scales remains a fundamental challenge 
for deep networks, which often fail to handle images at scales not seen 
during training (the out-of-distribution problem). In this paper, we present 
provably scale-invariant Gaussian derivative residual networks 
(GaussDerResNets), constructed out of scale-covariant Gaussian derivative 
residual blocks coupled in cascade, aimed at addressing this problem.

By adding residual skip connections to the previous notion of Gaussian 
derivative layers, deeper networks with substantially increased accuracy 
can be constructed, while preserving very good scale generalisation 
properties at the higher level of accuracy. Explicit proofs are provided 
regarding the underlying scale-covar\-iant and scale-invariant properties 
in arbitrary dimensions. We also conceptually
relate the functionality of the Gaussian derivative 
residual blocks to semi-discretis\-ations of the velocity-adapted affine 
diffusion equation.

To analyse the ability of GaussDerResNets to generalise to new scales, 
we apply them on a new rescaled version of the STL-10 dataset, 
where training is done at a single fixed scale and evaluation is performed 
on multiple copies of the test set, each rescaled to a single distinct spatial scale, 
with the scale factors extending over a range of 4. We also conduct similar 
systematic experiments on the rescaled versions of the Fashion-MNIST and 
CIFAR-10 datasets introduced in our previous work, and the previously 
existing STIR datasets.

Experimentally, we demonstrate that the GaussDerResNets have 
strong scale generalisation and scale selection properties, while also 
achieving good test accuracy, on all the four considered datasets with scaling variations. 
In our ablation studies, we investigate different architectural variants of 
GaussDerResNets, demonstrating that basing the architecture on 
depthwise-separable convolutions allows for decreasing both the 
number of parameters and the amount of computations, with reasonably 
maintained accuracy and scale generalisation. We also 
find that including a zero-order Gaussian term in the layer definition 
can sometimes be beneficial, as demonstrated for our
spatial-max-pooling-based networks trained on the rescaled STL-10 dataset.

In these ways, we demonstrate how deep networks can in a theoretically 
well-founded way handle variations in scale in the testing data that are 
not spanned by the training data.

We conclude by outlining how the proposed GaussDerResNets
can be extended to joint local spatial and scale selection, to
address the topic of multi-object detection in a provably
scale-invariant manner.

\end{abstract}

\section{Introduction}
\label{sec-introduction}

To effectively process real-world images acquired under general
viewing conditions, deep networks in computer vision must be
capable of handling scaling transformations over the spatial domain,
since the size of an object in the image domain may differ, because of
the different physical sizes of the 3-D objects in the world and
due to variations in the distance to the camera.

A theoretically well-founded way of incorporating priors about
spatial scaling transformations into convolutional neural networks
(CNNs) is by requiring the networks to be {\em scale covariant\/},
in the literature also referred to as scale equivariant.%
\footnote{According to its literal meaning, the use of the prefix
  ``equi'' in ``equivariant'' means that the transformation $T$ of the
  output of an operator $\Phi$ should be equal to the transformation applied
  to the input in a relationship of the form $\Phi(T(f)) = T(\Phi(f))$.
  In this paper, we instead use the terminology ``covariant'' in the
  meaning that the output of the operator should rather only be
  required to ``co-vary'' with the transformation $T$
  according to $\Phi(T(f)) = T'(\Phi(f))$ for some possibly other
  transformation $T'$ that still should be sufficiently similar to $T$.
  For the purpose of modelling (not necessarily scale-normalized)
  Gaussian derivative responses under
  spatial scaling transformations, this extension is necessary, since
  $T'$ will then be related to $T$ up to a scaling factor, which in
  turn directly depends upon the amount of scaling in the spatial scaling
  transformations. This terminology is specifically consistent with
  the terminology used in physics and classical scale-space theory
  (Lindeberg \citeyear{Lin13-ImPhys}).}
For defining a scale-covariant network, there are two main
types of approaches, either by building the network architecture on
a set of successively rescaled filters, or by applying the same
fixed network to a set of rescaled copies of the input image,
see Worrall and Welling (\citeyear{WorWel19-NeuroIPS}),
Bekkers (\citeyear{Bek20-ICLR}),
Sosnovik et al.\ (\citeyear{SosSzmSme20-ICLR}, \citeyear{SosMosSme21-BMVC}),
Zhu et al.\ (\citeyear{ZhuQiuCalSapChe22-JMLR}),
Jansson and Lindeberg (\citeyear{JanLin22-JMIV}),
Lindeberg (\citeyear{Lin22-JMIV}),
Zhan et al.\ (\citeyear{ZhaSunLi22-ICCRE}),
Wimmer et al.\ (\citeyear{WimGolDa23-arXiv}) and
Perzanowski and Lindeberg (\citeyear{PerLin25-JMIV-ScSelGaussDerNets})
for introductory overviews to this topic based on the two respective
dual types of approaches.

Such scale covariance extends the built-in inductive
bias of regular CNNs regarding translational covariance,
which enforces the application of the same filter at every
spatial location in an image, to also comprise structurally
similar processing operations at different scales,
by expanding the sizes of the receptive fields in the
networks over substantial scaling variations.
This guarantees that (i)~a spatially shifted input produces a
correspondingly shifted activation map in each layer and
that (ii)~a spatially rescaled image pattern
invokes a similar response in a scale-shifted layer in the
network.

A particular desirable property of a scale-covariant network is
the ability to perform {\em scale generalisation\/}, to perform testing
at scales that are not spanned by the training data.
A regular, not scale-covariant, network will, in general, perform
very poorly if exposed to such scale generalisation tasks.
A theoretical explanation for this is that regular deep networks
without complementary priors
merely perform interpolation between the training data,
with very poor abilities to perform extrapolation
(Domingos \citeyear{Dom20-arXiv},
Courtois et al.\ \citeyear{CurMorAri23-arXiv}).

To some extent, one could conceive the ability to process image 
data at different scales to be improved by performing data augmentation 
(Zhong et al.\ \citeyear{ZhoZhoWang25-NPL}), by substantially 
increasing the amount of training data with artificially rescaled 
images. It is, however, not clear how effective such an approach 
would be, when applied to testing data over wide scale ranges. For 
evaluating the scale-covariant properties of a deep network architecture, 
it is therefore essential to explore its scale generalisation abilities, when 
applied to testing data at scales that are not spanned by the training data.

Additionally, from the viewpoint of a vision system that learns to
recognise objects from visual data, one may question whether it should
be necessary to see a new object from all the distances from which it
is later going to be recognised.
Instead, the philosophy that we follow here is that it should be sufficient
to see a new object from a single distance, by providing priors about
scaling transformations to the deep network architecture, that enable
generalisation to new scales that are not spanned by the training data.

In previous work, we have developed a particular architecture for
scale-covariant deep networks with the layers defined from
linear combinations of Gaussian derivative operators over
multiple scales
(Lindeberg \citeyear{Lin22-JMIV},
Perza\-nowski and Lindeberg
\citeyear{PerLin25-JMIV-ScSelGaussDerNets}).
This approach extends ideas from classical scale-space theory,
where Gaussian smoothing and Gaussian derivatives at multiple
scales have been axiomatically shown to constitute a canonical
class of filters for processing image data
(Iijima \citeyear{Iij62},
Koenderink \citeyear{Koe84-BC},
Koenderink and van Doorn \citeyear{koenderink:92},
Lindeberg \citeyear{Lin93-Dis}, \citeyear{Lin10-JMIV},
Weickert et al.\ \citeyear{WeiIshImi99-JMIV}).
This approach is also closely related to other work on using
Gaussian derivative operators as the computational
primitives in deep networks, see
Jacobsen et al.\ (\citeyear{JacGemLouSme16-CVPR}),
Pintea et al.\ (\citeyear{PinTomGoeLooGem21-IP}),
Penaud-Polge et al.\ (\citeyear{PenVelAng22-ICIP}),
Gavilima-Pilataxi and Ibarra-Fiallo (\citeyear{GavIva23-ICPRS}) and
Basting et al.\ (\citeyear{Basting-2024-IJCCV}).

Concerning the more general use of Gaussian derivative operators as
computational primitives in deep networks, it is also
notable that by applying a greedy clustering technique to
the receptive fields learned from depthwise-separable deep networks,
Babaiee {\em et al.\/} (\citeyear{BabKiaRusGro24-ICLR},
\citeyear{BabKiaRusGro24-NeurIPSWS-fewer},
\citeyear{BabKiaRusGro25-AAAI-master}),
have extracted a set of ``8 master key filters''
from the ConvNeXt V2 Tiny model developed by Woo {\em et al.\/}
(\citeyear{WooDebHuCheLiuKweXi23-CVPR}) and Liu {\em et al.\/}
(\citeyear{liu2022-convnext-CVPR}), and which have been shown to
be well modelled in terms of discrete scale-space filters, 
see Lindeberg et al.\ (\citeyear{LinBabKia26-JMIV}).

By the provable scale-covariant and scale-invariant properties
of the Gaussian derivative networks (GaussDerNets) in
Lindeberg (\citeyear{Lin22-JMIV})
and Perzanowski and Lindeberg
(\citeyear{PerLin25-JMIV-ScSelGaussDerNets}),
these networks were demonstrated to have very good scale
generalisation properties, with structurally very close
similarities to classical methods for automatic scale selection
(Lindeberg \citeyear{Lin98-IJCV}, \citeyear{Lin97-IJCV},
\citeyear{Lin12-JMIV}, \citeyear{Lin15-JMIV}, \citeyear{Lin21-EncCompVis},
Bretzner and Lindeberg \citeyear{BL97-CVIU},
Lowe \citeyear{Low04-IJCV},
Bay et al.\ \citeyear{BayEssTuyGoo08-CVIU}).

The subject of this article is to extend this class of deep networks
to incorporate residual skip connections
(He et al.\ \citeyear{HeZhaRenSun16-CVPR}, \citeyear{HeZhaRenSun16-ECCV}),
which will make it possible
to construct substantially deeper GaussDerNets, without
encountering training problems due to vanishing gradients
when increasing the number of layers.

Experimentally, we will demonstrate that this approach leads
to a significant increase in the accuracy of the resulting
Gaussian derivative residual ResNets (GaussDerResNets),
to a level more comparable to modern ResNets.
We will also demonstrate that the presented approach
can be combined with the notion of depthwise-separable networks
(Chollet \citeyear{Cho17-CVPR}, Howard {\em et al.\/} 
\citeyear{HowZhuCheKalWanWeyAndAda17-arXiv}) to 
formulate depthwise-separable Gaussian derivative residual networks
(DSGaussDerResNets), that substantially reduce the amount 
of computations, by decoupling the convolution operations 
across the layers in the networks from the spatial convolutions.

Specifically, we will show that this extension from
GaussDerNets to GaussDerResNets, as well as further to DSGaussDerResNets
architectures, can be performed with maintained and even improved
scale covariance, scale invariance and scale generalisation
properties, for our resulting discrete implementations of the continuously
formulated network architecture.

Additionally, we will outline how the GaussDerResNets can be extended 
to joint local spatial and scale selection, to be able to detect and 
recognise multiple objects in the same scene.

\subsection{Contributions and novelty}
\label{sec-contributions-and-novelty}

To summarise, the main contributions of this paper comprise:
\begin{itemize}
\item
  Extension of the regular GaussDerNet architecture to
  GaussDerResNets, in which the residual skip connections enable
  much deeper networks with significantly better accuracy,
  as well as the ability to handle substantially more complex datasets.
\item
  A general proof in arbitrary dimensions that the resulting
  GaussDerResNet architecture is scale covariant
  for arbitrary orders of spatial differentiation, including showing
  that the scale covariance properties are
  preserved when residual skip connections are introduced into the
  networks.
\item
  Qualitative relationships between the proposed Gaussian derivative
  residual blocks and semi-discretisations of the velocity-adapted
  affine diffusion equation.
\item
  The development of an in-depth mathematical formalism for the
  entire end-to-end chain of image operations, from the input image to the final
  classification of the proposed GaussDerResNets, when applied to image
  classification/single object detection tasks, including formal
  proofs of their covariance and invariance properties, both
  without and with spatial pooling and scale pooling. In contrast to
  other approaches for covariant/equivariant deep networks, based on
  a formalisms based on group theory, we do not here make any specific
  use of group theory formalism. Instead, our formalism is based on
  the explicit transformation properties of the image representations
  under geometric image transformation, in our case spatial
  scaling transformations.%
\footnote{Implicitly, it is throughout understood that the spatial
  scaling transformations form a group structure and that the transformation
  properties of the Gaussian derivative kernels constituting the
  primitives in the filter weight kernels underlying this work are
  also closely related to the semi-group property of the Gaussian
  kernel. In these respects, there is an underlying group structure to
  the work, that we, however, do not explicitly reflect in the
  formalism. For our purposes, the transformation properties of the
  image representations under scaling transformations constitute the
  sufficient information that is necessary for expressing and proving the
  covariance and invariance properties in this treatment.}
\item
  Experimental evaluation of single-scale GaussDerResNets
  on the regular STL-10 dataset.
\item
  Experimental evaluation of the scale generalisation properties
  of multi-scale GaussDerResNets, when applied to the previously
  existing Rescaled Fashion-MNIST and CIFAR-10 datasets, 
  as well as the STIR datasets with traffic signs and aerial images.
\item
  Experimental evaluation of the scale generalisation properties
  of the GaussDerResNets when applied to new Rescaled STL-10 dataset
  proposed here, which constitutes a significantly harder dataset
  compared to those for which our previous Gaussian derivative
  networks have been experimentally evaluated.
\item
  Evaluation of the scale generalisation properties for three
  different scale pooling methods, in terms of max pooling,
  logsumexp pooling and average pooling over scales, finding that the
  logsumexp pooling method appears to constitute the most
  reliable scale pooling method for scale generalisation out
  of these three methods.
\item
  Ablation studies regarding:
  \begin{itemize}
  \item
    the extension of GaussDerResNets to depthwise-sep\-arable
    GaussDerResNets, which reduce both the number of parameters
    and the amount of computations
    with reasonably maintained accuracy and scale generalisation properties,
  \item
    including a zero-order term in the Gaussian derivative residual
    layers in the higher layers of the GaussDerResNets, which improves
    the performance on the rescaled STL-10 dataset,
  \item
    effects of pre-training with a single-scale-channel network to
    improve the efficiency and convergence of the training process, 
    while significantly improving the scale generalisation at finer scales,
  \item 
    showing that weight transfer to a network with a denser scale-channel 
    count can provide minor improvements to scale generalisation, without 
    any additional training cost, and
  \item
    using label smoothing to improve the convergence properties of the
    training process.
  \end{itemize}
\end{itemize}
Finally, we 
\begin{itemize}
\item
  inspect the activation maps and the learned filters in the 
  network, to better understand how the GaussDerResNets operate, 
  as the architecture makes these visualisations very interpretable, and
\item
  outline how the proposed GaussDerResNets can be extended to
  local-extended GaussDerResNets for generating multiple object
  hypotheses at different local image positions and at different local
  scales, when applied to image data in which there are multiple objects
  within the same scene.
\end{itemize}
In these ways, we substantially extend the framework for GaussDerNets
in Lindeberg (\citeyear{Lin22-JMIV}) as well as
the extended GaussDerNets in Perzanowski and Lindeberg
(\citeyear{PerLin25-JMIV-ScSelGaussDerNets}) to wider applicability
in deeper GaussDerResNets, which have significantly
better performance in terms of accuracy and scale generalisation, when
applied to datasets with more complex image structures. Specifically,
with the deeper GaussDerResNets, we are able to handle more complex
datasets compared to our previously studied GaussDerNets.

\section{Related work}

Initial efforts to develop scale-invariant deep networks for vision
were undertaken by Xu et al.\ (\citeyear{XuXiaZhaYanZha14-arXiv}),
Kanazawa et al.\ (\citeyear{KanShaJac14-arXiv}),
Marcos et al.\ (\citeyear{MarKelLobTui18-arXiv}) and
Ghosh and Gupta (\citeyear{GhoGup19-arXiv}).
Hierarchical networks that handle spatial scaling variations have
also been developed by Bruna and Mallat (\citeyear{BruMal13-PAMI})
and Mallat (\citeyear{Mal16-RoySoc}) based on wavelet
operations in terms of scattering transforms,
and by Lindeberg (\citeyear{Lin20-JMIV})
based on structurally related idealised models of complex cells, in
terms of quasi
quadrature combinations of first- and second-order Gaussian
derivatives.

Other approaches to involving multiple spatial scales
in deep networks have been developed by
Chen et al.\ (\citeyear{CheFanXuYanKalRohYanFen19-ICCV})
in terms of octave convolution,
by Xu et al.\ (\citeyear{XuVenSun20-CVPR}) in terms of
blur integrated gradients,
by Wang et al.\ (\citeyear{WanZhaYuFenZha20-CVPR})
in terms of scale-equalising pyramid convolution,
by Singh et al.\ (\citeyear{SinNajShaDav22-PAMI})
regarding scale-normalised image pyramids,
by Fu et al.\ (\citeyear{FuZhaLiuRonWu22-CircSystVidTech})
to reduce scale differences for large scale image matching,
by Wolski et al.\ (\citeyear{WolDjeJavSeiTheCorMysDerPanLei25-ACMGraph})
using learning across scales with adversarial training, and
by Zhang et al.\ (\citeyear{ZhaLiSuCaoTiaXu25-NeurNetw})
using a scale- and rotation-handling retina-like image transformation.
Rahman et al.\ (\citeyear{RahYanCheHaoJiaLimYeh25-ICCV})
proposed to include canonicalisers, to make pretrained
networks scale equivariant. Xu et al.\ (\citeyear{XuWanLuoYooQia25-CVPR}) 
proposed a method for object counting to handle size variations in
the image data. Cho et al.\ (\citeyear{ChNaCH25-IEEE}) proposed
a scale-equivariant network for object perception in autonomous driving.

Beyond the scale-covariant deep network architectures by
Worrall and Welling (\citeyear{WorWel19-NeuroIPS}),
Bekkers (\citeyear{Bek20-ICLR}),
Sosnovik et al.\ (\citeyear{SosSzmSme20-ICLR}, \citeyear{SosMosSme21-BMVC}),
Zhu et al.\ (\citeyear{ZhuQiuCalSapChe22-JMLR}),
Jansson and Lindeberg (\citeyear{JanLin22-JMIV}),
Lindeberg (\citeyear{Lin22-JMIV}),
Zhan et al.\ (\citeyear{ZhaSunLi22-ICCRE}),
Wimmer et al.\ (\citeyear{WimGolDa23-arXiv}) and
Perzanowski and Lindeberg (\citeyear{PerLin25-JMIV-ScSelGaussDerNets})
mentioned in the introduction,
scale-covariant U-Nets have been developed by
Sangalli et al.\ (\citeyear{SanBluVelAng22-BMVC}) and
Yang et al.\ (\citeyear{YanDasMah23-PMLRes}),
scale-invariant Riesz networks by
Barisin et al.\ (\citeyear{BarAngSchRed24-SIIMS, BarSchRed24-JMIV}),
and scale-invariant HMAX networks based on trainable
filters by Pant et al.\ (\citeyear{pant2024-hmax-CCN}). 

There have also been approaches for equivariant or invariant learning developed by
Bruintjes et al.\ (\citeyear{BruMotGem23-CVPR}),
Mondal et al.\ (\citeyear{MonPanKabMudRav23-NeurIPS}),
van~der~Ouderaa et al.\ (\citeyear{OudImmWil23-NeurIPS}),
Moskalev et al.\ (\citeyear{MosSepBekMe23-PMLR}),
Shewmake et al.\ (\citeyear{SheBurLilShiBekMioOls23-NeurIPSWS}),
Pertigkiozoglou et al.\ (\citeyear{PerChaTriDan24-NeurIPS}),
Nordenfors and Flinth (\citeyear{NorFli24-arXiv}),
Perin and Deny (\citeyear{PerDen24-JMLRes}) and
Manolache et al.\ (\citeyear{ManChaNie25-NeurIPS}).
It is, however, not clear to what extent such learning-based
approaches to covariance or invariance properties
would allow for scale generalisation to scale levels that are not
spanned in the training stage of the network.

The Gaussian smoothing and Gaussian derivative filters in the
deep networks developed by
Jacobsen et al.\ (\citeyear{JacGemLouSme16-CVPR}),
Pintea et al.\ (\citeyear{PinTomGoeLooGem21-IP}),
Lindeberg (\citeyear{Lin22-JMIV}),
Sangalli et al.\
(\citeyear{SanBluVelAng22-BMVC}, \citeyear{SangaliBluVel22--ICIP}),
Penaud-Polge et al.\ (\citeyear{PenVelAng22-ICIP}),
Gavilima-Pilataxi and Ibarra-Fiallo (\citeyear{GavIva23-ICPRS}),
Basting et al.\ (\citeyear{Basting-2024-IJCCV})
and Perza\-nowski and Lindeberg
(\citeyear{PerLin25-JMIV-ScSelGaussDerNets}) satisfy the diffusion
equation.
In this respect, these methods have close structural similarities
to other deep networks modelled by partial differential equations.

Indeed, deep learning has seen a shift in perspective, with architectures,
such as ResNets
(He et al.\ \citeyear{HeZhaRenSun16-CVPR}, \citeyear{HeZhaRenSun16-ECCV}),
having an alternative interpretation of
being numerical discretisations of partial differential equations,
see Ruthotto and Haber (\citeyear{RutHab20-JMIV})
and Lu et al.\ (\citeyear{lu2018-PMLR}).
This connection has been further explored by
Alt et al.\ (\citeyear{alt2023-JMIV}),
who show that there are close structural
parallels between numerical algorithms for partial differential
equations (PDEs) and
certain neural network architectures, such as ResNets.
Furthermore, PDE-based networks that generate scale spaces have been
been proposed by Smets et al.\ (\citeyear{SmePorBekDui23-JMIV}) and
Bellaard et al.\ (\citeyear{BellSak-2025-JMIV}).

Recently, ResNets have been developed to mimic the properties of
visual transformers, see Dai et al.\ (\citeyear{dai2021-coatnet-NIPS}),
Liu et al.\ (\citeyear{liu2022-convnext-CVPR}) and
Rao et al.\ (\citeyear{dai2021-coatnet-NIPS}).

More generally, the development of scale-covariant architectures
for deep learning is part of a wider effort of geometric deep learning
(Bronstein et al.\ \citeyear{BroBruCohVel21-arXiv},
Gerken et al.\ \citeyear{GerAroCarLinOhlPetPer23-AIRev}),
by incorporating explicit modelling of the influence of
geometric image transformations on the receptive field responses
(Lindeberg \citeyear{Lin21-Heliyon},
\citeyear{Lin23-FrontCompNeuroSci},
\citeyear{Lin25-JMIV}).

\section{Gaussian derivative residual networks}
\label{sec-GaussDerResNets-theory}

In this section, we will first define the computational primitives in
the scale-covariant 
and scale-invariant GaussDerResNets, as combinations of notions in scale-covariant 
and scale-invariant GaussDerNets (Lindeberg \citeyear{Lin22-JMIV},
Per\-zanowski and Lindeberg \citeyear{PerLin25-JMIV-ScSelGaussDerNets})
and ResNets (He et al.\ \citeyear{HeZhaRenSun16-CVPR}, \citeyear{HeZhaRenSun16-ECCV}).
Then, we will show that the resulting networks,
obtained by coupling these computational primitives in cascade,
will be provably scale covariant. Finally, we will combine these
network primitives with joint selection mechanisms over
space and scale, to express scale-invariant deep network
architectures for image classification tasks similar to single-object
detection and recognition.

\subsection{Scale covariance property for Gaussian derivative
  layers with residual skip connections}
\label{sec-scale-covariance-properties}

In Lindeberg (\citeyear{Lin22-JMIV}), it was shown that regular 
GaussDerNets, obtained by coupling layers based on linear 
combinations of Gaussian derivative operators in cascade, are 
provably scale covariant. In this section, we show that these
scale covariance properties do also hold under the introduction
of residual skip connections. We additionally generalize
from the previous study of GaussDerNets of order 2 for 2-D image data to
a more general proof that applies to GaussDerResNets for
any maximum order $N$ of differentiation and for any dimensionality $D$
of the input data.

\begin{figure}[hbt]
  \begin{center}
    \begin{tabular}{c}
       \includegraphics[width=0.48\textwidth]{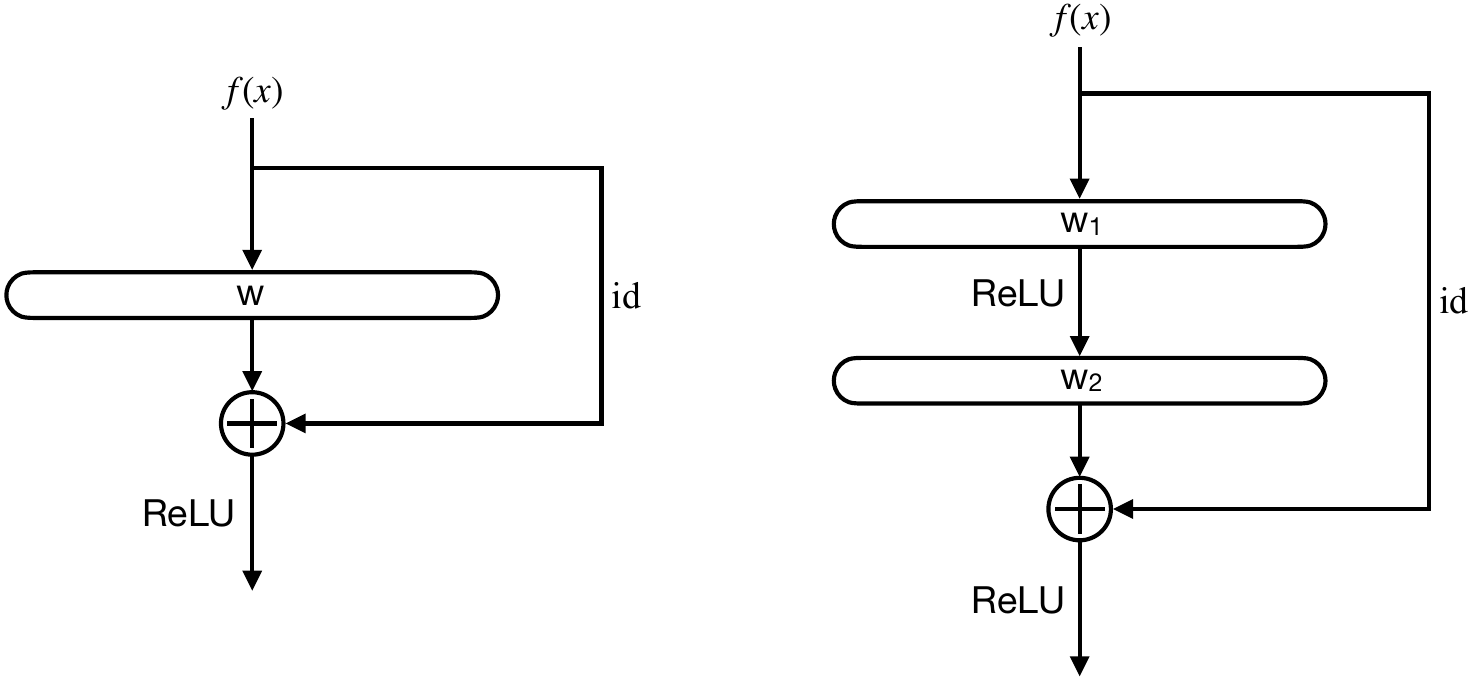}
    \end{tabular}
  \end{center}
  \caption{Schematic illustrations of simplified residual blocks:
   {\bf  (left)} the simplest possible residual block of the form
    (\ref{eq-simplest-resnet-module}), based on a single convolution
    kernel $w(x)$,
   {\bf (right)} the basic residual block of the form
    (\ref{eq-basic-resnet-module}),
    based on two convolution kernels $w_1(x)$ and $w_2(x)$. 
    In both the diagrams, the skip connections perform identity
    mappings on the input $f(x)$, denoted as $id$.}
  \label{fig-residual-blocks}
\end{figure}

\subsubsection{Prerequisites}

The fundamental building block in a GaussDerResNet is the
combination of an input function $f \colon \bbbr^D \rightarrow \bbbr$
denoted $f(x)$ for $x = (x_1, x_2, \dots, x_D)^T \in \bbbr^D$ with a
convolution kernel $w \colon \bbbr^D \rightarrow \bbbr$,
a pointwise ReLU function
$\operatorname{ReLU} \colon \bbbr \rightarrow \bbbr$
and a skip connection of the form illustrated in
Figure~\ref{fig-residual-blocks}~(left)
\begin{equation}
  \label{eq-simplest-resnet-module}
  M_w(x) = \relu(f(x) + (w * f)(x)).
\end{equation}
In practice, one often combines two convolution kernels,
say $w_1(x)$ and $w_2(x)$, of the form illustrated in
Figure~\ref{fig-residual-blocks}~(right)
\begin{equation}
  \label{eq-basic-resnet-module}
  M_{w_1,w_2}(x) = \relu(f(x) + w_2 * \relu((w_1 * f))(x)).
\end{equation}
Since the logical structure is essentially the same, we will in the logical steps of the proof to be given below consider residual skip connections of the
form (\ref{eq-simplest-resnet-module}).

Let us next, inspired by the structure of regular GaussDerNets,
assume that the convolution kernel $w(x)$ in (\ref{eq-simplest-resnet-module})
is a linear combination of Gaussian derivative operators of the
following form:

\begin{mydefinition}
  (Gaussian derivative weight function)
  A Gaussian derivative weight function
  $w \colon \bbbr^N \times \bbbr_+ \rightarrow \bbbr$
  at scale $\sigma \in \bbbr_+$ depends on the filter weights
  $C_{\alpha}$ according to%
\footnote{Regarding the first layer, we will in this summation exclude
  the zero-order term, in order to make the response invariant under
  intensity transformations of the form $f(x) \mapsto f(x) + a$,
  which model the effect of multiplicative illumination variations on a logarithmic
  brightness scale. In the higher layers in the deep hierarchy, we
  will, however, consider also including the zeroth-order term, as
  will be detailed further in Sections~\ref{sec-zero-order-based-net-theory}
  and~\ref{sec-zero-order-term-study}.}
  \begin{equation}
    \label{eq-gauss-der-layer}
    w(x;\; \sigma)
    = \sum_{\alpha : |\alpha| \leq N}
            m(\alpha) \, C_{\alpha }
             \, \sigma^{|\alpha|} \, g_{x^{\alpha}}(x;\; \sigma),
  \end{equation}
  where
  \begin{itemize}
  \item
    $\alpha = (\alpha_1, \alpha_2, \dots, \alpha_D) \in \bbbz^D$ is a multi-index
    in $D$ dimensions,
  \item
    $|\alpha| = \alpha_1 + \alpha_2 + \dots + \alpha_D \in \bbbz$
    represents the total order of differentiation as the sum of the indices,
  \item
    $m(\alpha)
    = {|\alpha| \choose {\alpha_1 \,  \alpha_2 \, \dots \, \alpha_D}}
    \in \bbbr$
    is a multinomial normalisation factor between different orders $\alpha$
    of spatial differentiation, similar to the coefficients in a Taylor expansion
    of the same order $|\alpha|$,
  \item
    $g_{x^{\alpha}}(x;\; \sigma)$ is a $D$-dimensional Gaussian derivative
    kernel of order $\alpha$ at the scale $\sigma \in \bbbr_+$ according to
    \begin{equation}
      g_{x^{\alpha}}(x;\; \sigma) = \partial_{x^{\alpha}} g(x;\; \sigma)
    \end{equation}
    for
    \begin{equation}
      g(x;\; \sigma) = \frac{1}{\sqrt{2 \pi}^D} \, e^{-x^Tx/2\sigma^2}
    \end{equation}
    and
    $\partial_{x^\alpha} = \partial_{x_1^{\alpha_1} x_2^{\alpha_2} \dots x_D^{\alpha_D}}$,
  \item
    $N \in \bbbz_+$ denotes the maximum order of spatial differentiation for the Gaussian derivative kernels, and
  \item
    the coefficients $C_{\alpha} \in \bbbr$ denote the filter weights%
    \footnote{Note that, in order to simplify the notation, we here normalise
      the filter weights in a different way than in the original work on scale-covariant
      and scale-invariant Gaussian derivative networks (Lindeberg \citeyear{Lin22-JMIV}).
      Since the filter weights are anyway to be learned, the choice of the normalisation
      method for the filter weights does, however, not affect the functional properties
      of the network.}
  \end{itemize}
\end{mydefinition}
In general, the weights $C_{\alpha}$ will be different for
different layers in the network. The weights will, however,
be shared between different scale channels, corresponding
to using different values of the initial scale level $\sigma_0 \in \bbbr_+$
in the first layer of the network.

\begin{figure*}[hbt]
 \begin{subfigure}{0.47\textwidth}
  \begin{center}
    \begin{tabular}{c}
      \includegraphics[width=0.20\textwidth]{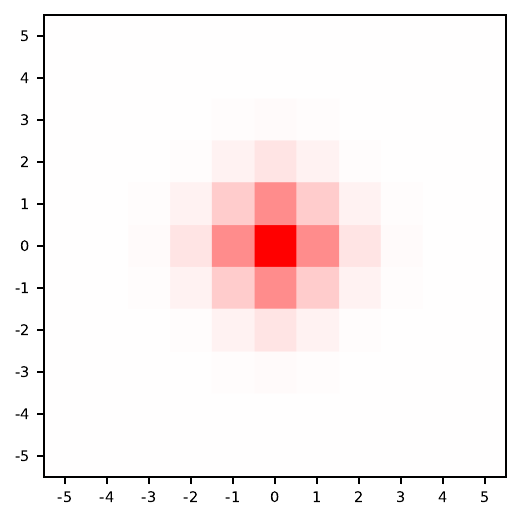} \\
    \end{tabular} 
  \end{center}

  \vspace{-5mm}
  
  \begin{center}
    \begin{tabular}{cc}
      \includegraphics[width=0.20\textwidth]{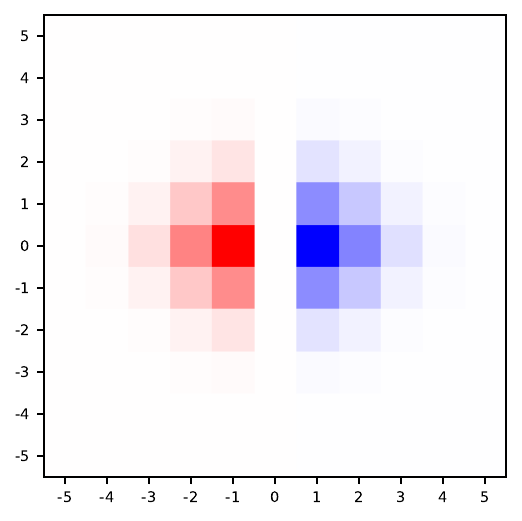} &
      \hspace{-3mm}
      \includegraphics[width=0.20\textwidth]{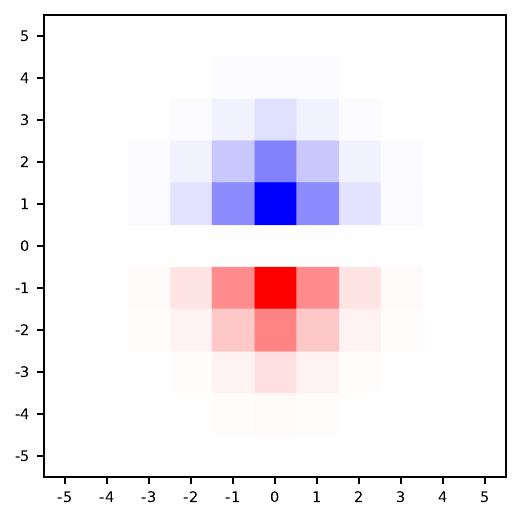} \\
    \end{tabular} 
  \end{center}

  \vspace{-5mm}

  \begin{center}
    \begin{tabular}{ccc}
      \includegraphics[width=0.20\textwidth]{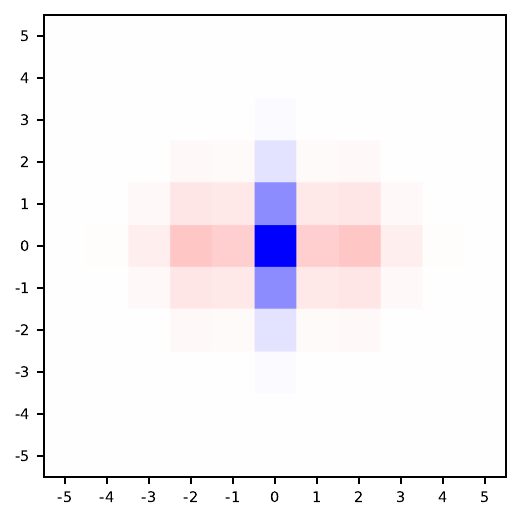} &
      \hspace{-3mm}
      \includegraphics[width=0.20\textwidth]{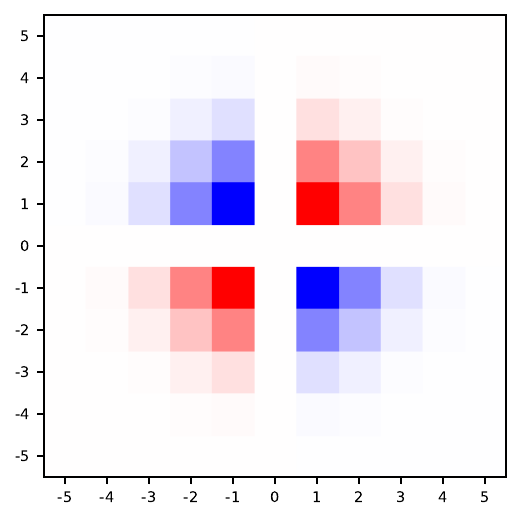} &
      \hspace{-3mm}
      \includegraphics[width=0.20\textwidth]{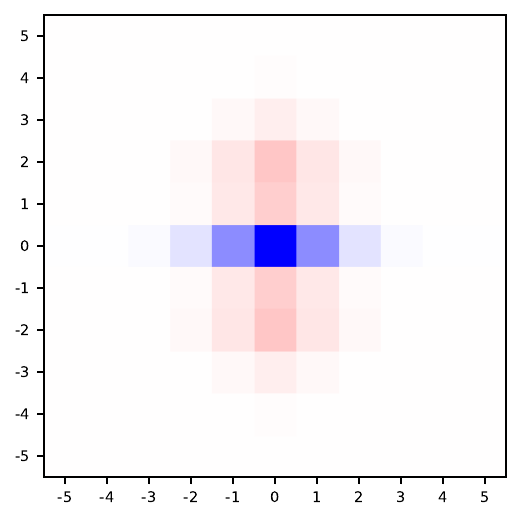} \\
    \end{tabular} 
  \end{center}

  \vspace{-5mm}

  \begin{center}
    \begin{tabular}{cccc}
      \includegraphics[width=0.20\textwidth]{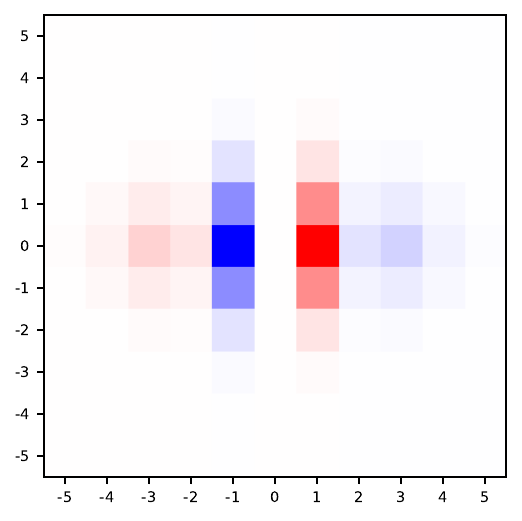} &
      \hspace{-3mm}
      \includegraphics[width=0.20\textwidth]{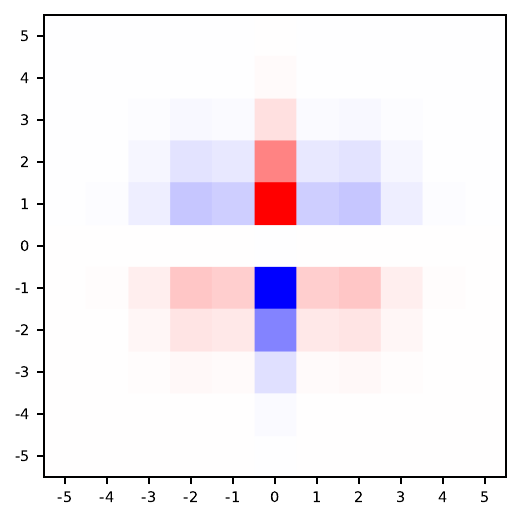} &
      \hspace{-3mm}
      \includegraphics[width=0.20\textwidth]{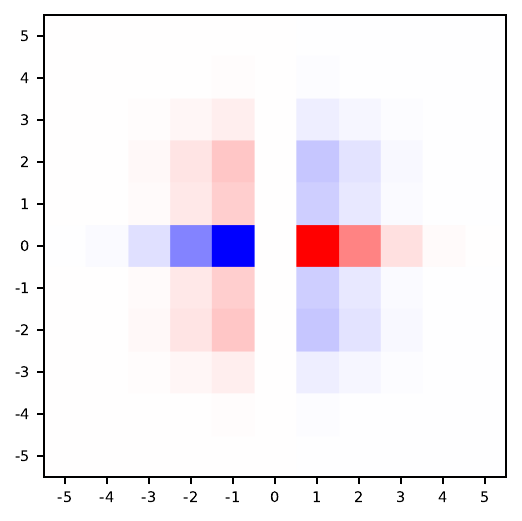} &
      \hspace{-3mm}
      \includegraphics[width=0.20\textwidth]{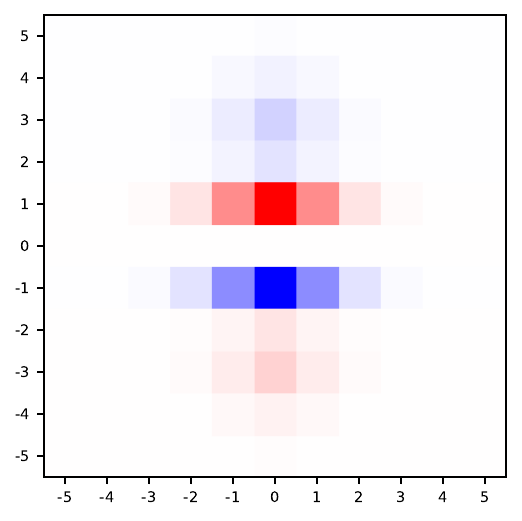} \\
    \end{tabular} 
   \end{center}
   \caption{Discrete Gaussian derivative kernels for $\sigma = 1$.}
 \end{subfigure}
 \hfill
 \begin{subfigure}{0.47\textwidth}
  \begin{center}
    \begin{tabular}{c}
      \includegraphics[width=0.20\textwidth]{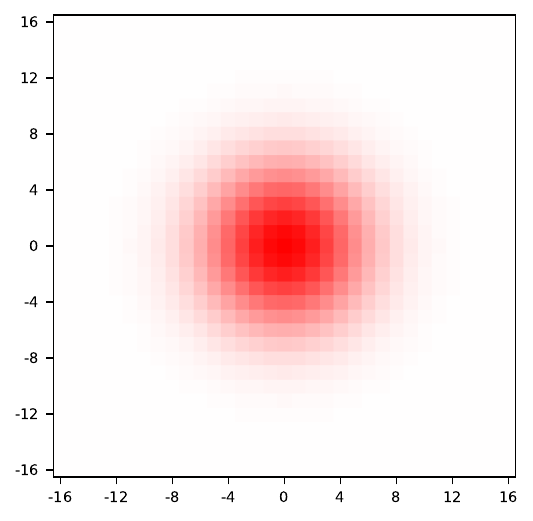} \\
    \end{tabular} 
  \end{center}

 \vspace{-5mm}
   
  \begin{center}
    \begin{tabular}{cc}
      \includegraphics[width=0.20\textwidth]{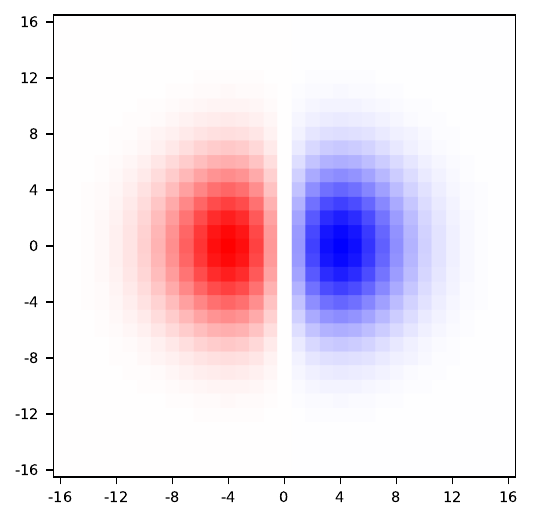} &
      \hspace{-3mm}
      \includegraphics[width=0.20\textwidth]{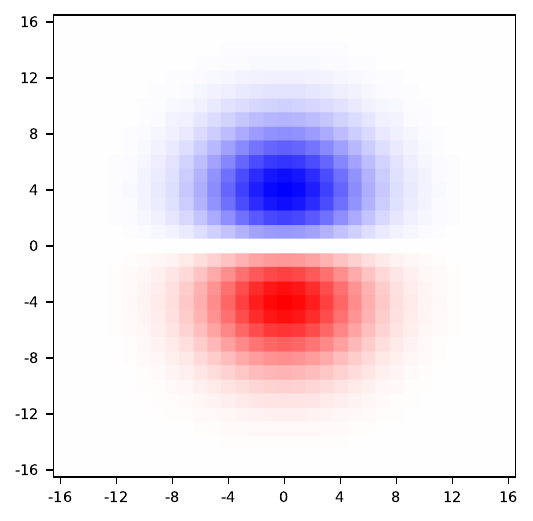} \\
    \end{tabular} 
  \end{center}

  \vspace{-5mm}

  \begin{center}
    \begin{tabular}{ccc}
      \includegraphics[width=0.20\textwidth]{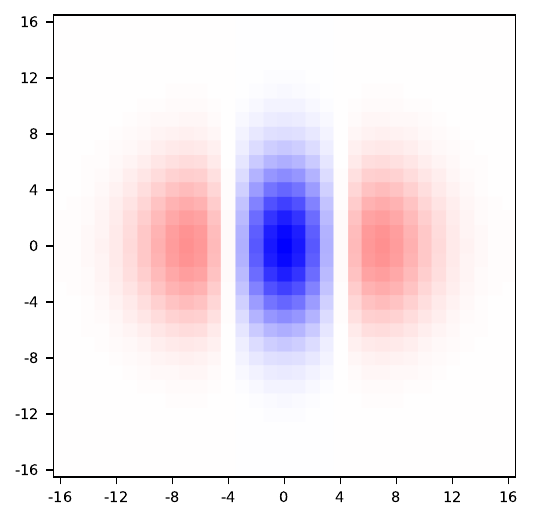} &
      \hspace{-3mm}
      \includegraphics[width=0.20\textwidth]{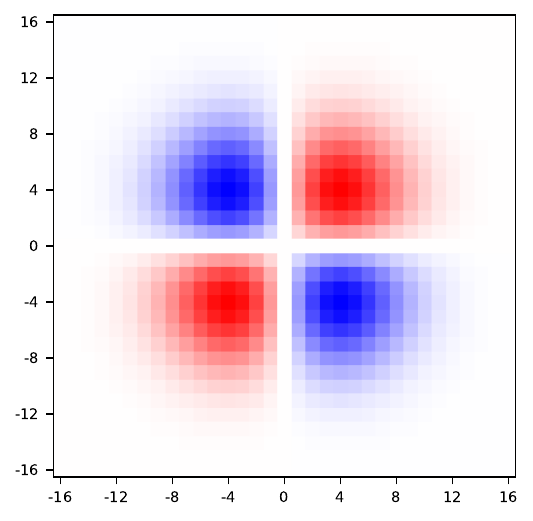} &
      \hspace{-3mm}
      \includegraphics[width=0.20\textwidth]{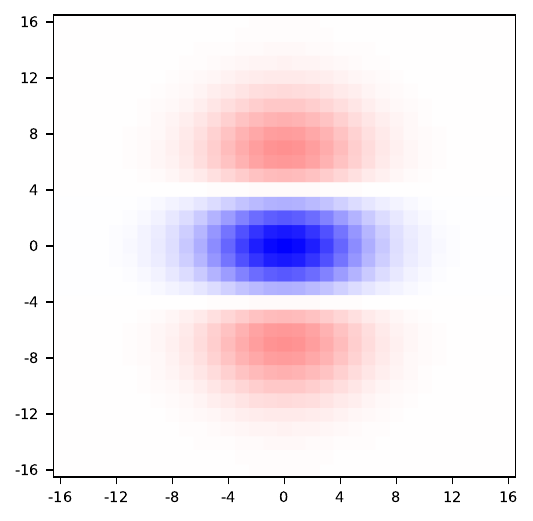} \\
    \end{tabular} 
  \end{center}

  \vspace{-5mm}

  \begin{center}
    \begin{tabular}{cccc}
      \includegraphics[width=0.20\textwidth]{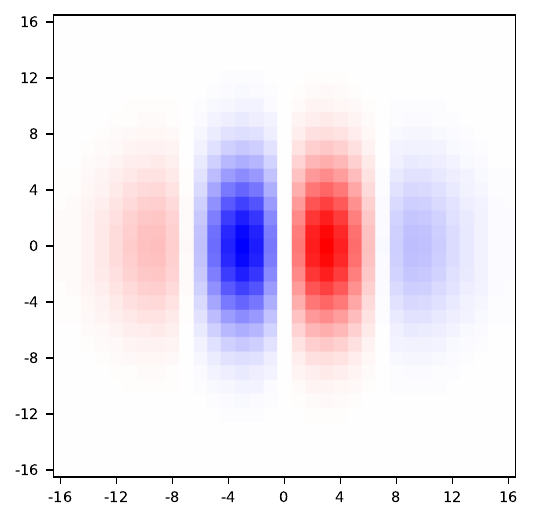} &
      \hspace{-3mm}
      \includegraphics[width=0.20\textwidth]{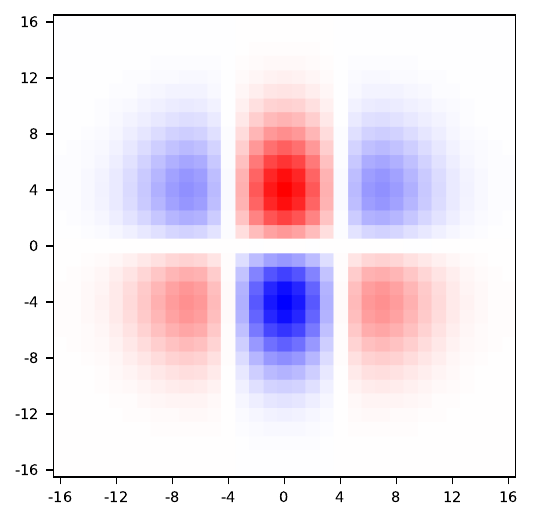} &
      \hspace{-3mm}
      \includegraphics[width=0.20\textwidth]{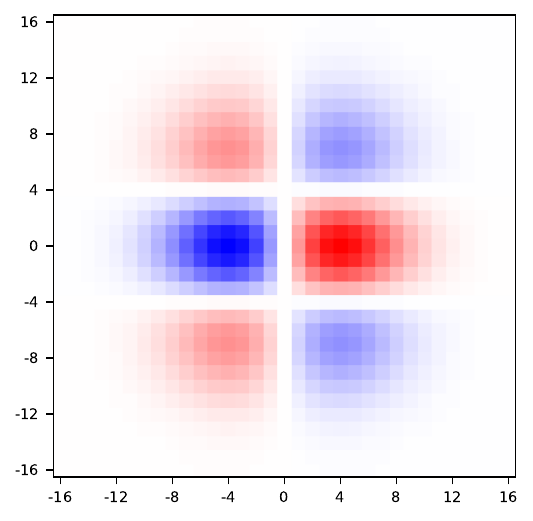} &
      \hspace{-3mm}
      \includegraphics[width=0.20\textwidth]{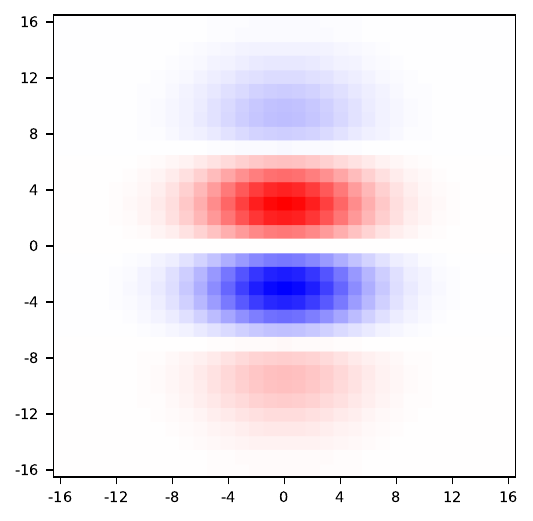} \\
    \end{tabular} 
   \end{center}
   \caption{Discrete Gaussian derivative kernels for $\sigma = 4$.}
 \end{subfigure}
 \caption{Discretised Gaussian derivative kernels up to order 3, 
   for (left) the scale parameter $\sigma = 1$ and for (right) the scale 
   parameter $\sigma = 4$. These receptive fields are obtained by 
   applying central difference operators (corresponding to the given 
   derivative order) to the discrete analogue of the Gaussian kernel, 
   as defined in Equation~(\ref{eq:disc-gauss+central-diff}). 
   The ranges of both the axes of the filters are for (the left figure) in
   the interval $[-5, 5]$, and for (the right figure) in the interval $[-16, 16]$. 
   {\bf (Top row)} the zero-order Gaussian kernel $g(x_1, x_2;\; \sigma)$,
   {\bf (second row)} first-order Gaussian derivatives
   $g_{x_1}(x_1, x_2;\; \sigma)$ and $g_{x_2}(x_1, x_2;\; \sigma)$,
   {\bf (third row)} second-order Gaussian derivatives
   $g_{x_1 x_1}(x_1, x_2;\; \sigma)$, $g_{x_1 x_2}(x_1, x_2;\; \sigma)$,
   $g_{x_2 x_2}(x_1, x_2;\; \sigma)$,
   {\bf (bottom row)} third-order Gaussian derivatives
   $g_{x_1 x_1 x_1}(x_1, x_2;\; \sigma)$, $g_{x_1 x_1 x_2}(x_1, x_2;\; \sigma)$,
   $g_{x_1 x_2 x_2}(x_1, x_2;\; \sigma)$ and $g_{x_2 x_2 x_2}(x_1, x_2;\; \sigma)$.}
 \label{fig-gauss-der-kernels}
\end{figure*}

For example, for a Gaussian derivative layer of the form (\ref{eq-gauss-der-layer})
of order 2, we should for 2-D input data let the multi-index $\alpha$ loop over
the set
\begin{equation}
  \label{eq-alpha-vals-2nd-order-jet}
  \alpha \in \{ (1, 0), (0, 1), (2, 0), (1, 1), (0, 2) \}.
\end{equation}
Similarly, for a Gaussian derivative layer 
of order 3, we should for 2-D input data let the multi-index $\alpha$ loop over
the set
\begin{align}
  \begin{split}
    \alpha \in
        & \left\{ (1, 0), (0, 1), (2, 0), (1, 1), (0, 2), \dots \right.
  \end{split}\nonumber\\
  \begin{split}
    \label{eq-alpha-vals-3rd-order-jet}
      & \hphantom{\left\{\right.} \left. (3, 0), (2, 1), (1, 2), (0, 3) \right\}.
  \end{split}
\end{align} 
Figure~\ref{fig-gauss-der-kernels} shows examples of such
primitive Gaussian derivative
kernels up to order 3 in the 2-D case.

The functionality of the scale normalisation factor $\sigma^{|\alpha|}$
in (\ref{eq-gauss-der-layer}) is to ensure that the Gaussian derivative 
operators are properly normalised over scale according to 
Lindeberg (\citeyear{Lin97-IJCV}) for the most scale-invariant choice of the 
scale normalisation power $\gamma = 1$, as the scale parameter $\sigma$ 
may be varied between different scale channels in the network for different 
values of $\sigma$.

\subsubsection{Transformation property under spatial scaling transformations}
\label{sec-trans-under-scaling}

Let us next consider the effect of scaling transformations applied to 
a minimal Gaussian derivative residual block of the following form:
\begin{mydefinition}
  (Gaussian derivative residual block based on a single weight function)
  Given a Gaussian derivative weight function $w(\cdot;\; \sigma)$
  at scale $\sigma \in \bbbr_+$ of the form
  (\ref{eq-gauss-der-layer}), its corresponding Gaussian derivative
  residual block $M_w(x;\; \sigma)$ is given by
  \begin{multline}
    \label{eq-simplest-resnet-module-sigma}
    M_w(x;\; \sigma)
    = ({\cal M}_{w,\sigma} f)(x;\; \sigma) = \\
    = \relu( f(x) + (w(\cdot;\; \sigma) * f(\cdot))(x)),
  \end{multline}
  where
  \begin{itemize}
  \item
    $M_w(\cdot;\; \sigma)$ denotes the layer at scale $\sigma$,
  \item
    ${\cal M}_{w,\sigma}$ denotes the operator that computes the layer from the
    input $f$.
  \end{itemize}
\end{mydefinition}

\noindent
A main basis for our further treatment is that such Gaussian
derivative residual blocks are provably scale covariant:

\begin{myproposition}
  \label{prop-sc-cov-single-resid-block}
  (Covariance property of Gaussian derivative residual block based on
  a single weight function)
  Gaussian derivative residual blocks of the form
  (\ref{eq-simplest-resnet-module-sigma}) with the
  scale-normalized Gaussian derivative weight functions
  $w(\cdot;\; \sigma)$ of the form
  (\ref{eq-gauss-der-layer}) are scale covariant
  and transform according to
  \begin{equation}
    M'_w(x';\; \sigma') = M_w(x;\; \sigma)
  \end{equation}
  under spatial scaling transformations of the input image of the form
  $f'(x') = f(x)$ with $x' = S_x \, x$,
  where the spatial scaling factor $S_x \in \bbbr_+$, with the
  associated scale parameter transforming according to
  $\sigma' = S_x \, \sigma$.
\end{myproposition}

\begin{myproof}
  Given rescaled input data $f'(x')$ defined according to
  \begin{equation}
    f'(x') = f(x) \quad\quad\mbox{for}\quad\quad x' = S_x \, x,
  \end{equation}
  where $S_x \in \bbbr_+$ is a spatial scaling factor,
  we define the output from a corresponding transformed
  Gaussian derivative residual layer as
  \begin{multline}
    \label{eq-simplest-resnet-module-sigma-prim}
    M'_w(x';\; \sigma') = ({\cal M}_{w,\sigma'} f')(x';\; \sigma') = \\
    = \relu( f'(x') + (w'(\cdot;\; \sigma') * f'(\cdot))(x')),
  \end{multline}
  where the transformed convolution kernel is of the form
  \begin{equation}
    \label{eq-gauss-der-layer-prim}
    w'(x';\; \sigma')
    = \sum_{\alpha : |\alpha| \leq N}
          m(\alpha) \, C_{\alpha } \, {\sigma'}^{|\alpha|} \,   g_{{x'}^{\alpha}}(x';\; \sigma'),
  \end{equation}
  specifically using the same filter weights $C_{\alpha }$ as in the definition of
  $w(x;\; \sigma)$ in (\ref{eq-gauss-der-layer}), to enable spatial scale covariance.
  
  Then, from a previously established property
  (Lindeberg \citeyear{Lin97-IJCV}), further detailed in
  Equation~(\ref{eq-equal-sc-norm-ders-sc-transf-mod}) below,
  that scale-normalised Gaussian derivative responses are equal
  at corresponding image points
  \begin{multline}
    \label{eq-equal-sc-norm-ders-sc-transf}
    \left(\left( {\sigma'}^{|\alpha|} \, g_{{x'}^{\alpha}}(\cdot;\; \sigma') \right) * f'(\cdot) \right)(x';\; \sigma') = \\
    = \left( \left( \sigma^{|\alpha|} \, g_{x^{\alpha}}(\cdot;\; \sigma) \right) * f(\cdot) \right)(x;\; \sigma),
  \end{multline}
  provided that the scale values $\sigma$ and $\sigma'$ are properly matched according to
  \begin{equation}
    \label{eq-match-rel-sigma-sc-transf}
    \sigma' = S_x \, \sigma,
  \end{equation}
  it follows that the output from the Gaussian derivative residual layers will be equal
  \begin{equation}
    \label{eq-match-rel-resnet-layers-sc-transf}
    M'_w(x';\; \sigma') = M_w(x;\; \sigma),
  \end{equation}
  provided that the scale parameters $\sigma$ and $\sigma'$ are properly 
  matched according to (\ref{eq-match-rel-sigma-sc-transf}).
  
  The relationship (\ref{eq-equal-sc-norm-ders-sc-transf}) can also with 
  more standard notation for Gaussian derivative responses be written as
  \begin{equation}
    \label{eq-equal-sc-norm-ders-sc-transf-mod}
    L'_{{\xi'}^{\alpha}}(x';\; \sigma') = L_{\xi^{\alpha}}(x;\; \sigma)
  \end{equation}
  with
  \begin{align}
    \begin{split}
      L_{\xi^{\alpha}}(x;\; \sigma)
      = \left( \left( \sigma^{|\alpha|} \, g_{x^{\alpha}}(\cdot;\; \sigma) \right) * f(\cdot) \right)(x;\; \sigma),
    \end{split}\\
    \begin{split}
      L'_{{\xi'}^{\alpha}}(x';\; \sigma')
      = \left( \left( {\sigma'}^{|\alpha|} \, g_{{x'}^{\alpha}}(\cdot;\; \sigma') \right) * f'(\cdot) \right)(x';\; \sigma'),
    \end{split}
  \end{align}
  where $L_{\xi^{\alpha}}(x;\; \sigma)$ and $L'_{\xi^{\alpha}}(x';\; \sigma')$
  denote the regular scale-normalised Gaussian derivative responses 
  computed for the functions $f(x)$ and $f'(x')$, respectively.
  With this notation, the equality in Equation~(\ref{eq-equal-sc-norm-ders-sc-transf-mod})
  follows from Equation~(25) in Lindeberg (\citeyear{Lin97-IJCV})
  for the choice of the scale normalisation power $\gamma = 1$.
\end{myproof}

\noindent
For later use in composed GaussDerResNets, we also state the notion of
regular Gaussian derivative layer (Lindeberg \citeyear{Lin22-JMIV}):

\begin{mydefinition}
  (Regular Gaussian derivative layer)
  Given a Gaussian derivative weight function $w(\cdot;\; \sigma)$
  at scale $\sigma \in \bbbr_+$ of the form
  (\ref{eq-gauss-der-layer}), its corresponding regular Gaussian derivative
  layer $G_w(x;\; \sigma)$ is given by
  \begin{multline}
    \label{eq-gauss-der-module-sigma}
    G_w(x;\; \sigma)
    = ({\cal G}_{w,\sigma} f)(x;\; \sigma) = \\
    = \relu(w(\cdot;\; \sigma) * f(\cdot))(x)),
  \end{multline}
  where
  \begin{itemize}
  \item
    $G_w(\cdot;\; \sigma)$ denotes the layer at scale $\sigma$,
  \item
    ${\cal G}_{w,\sigma}$ denotes the operator that computes the layer from the
    input $f$.
  \end{itemize}
\end{mydefinition}

\noindent
Such Gaussian derivative layers are also scale covariant
(Lindeberg \citeyear{Lin22-JMIV}):

\begin{myproposition}
  \label{prop-sc-cov-single-gauss-der-block}
  (Covariance property of regular Gaussian derivative layer)
  Gaussian layers of the form
  (\ref{eq-simplest-resnet-module-sigma}), with the
  scale-normalized Gaussian derivative weight functions
  $w(\cdot;\; \sigma)$ of the form
  (\ref{eq-gauss-der-layer}), are scale covariant
  and transform according to
  \begin{equation}
    G'_w(x';\; \sigma') = G_w(x;\; \sigma)
  \end{equation}
  under spatial scaling transformations of the input image of the form
  $f'(x') = f(x)$ with $x' = S_x \, x$,
  where the spatial scaling factor $S_x \in \bbbr_+$, with the
  associated scale parameter transforming according to
  $\sigma' = S_x \, \sigma$.
\end{myproposition}

\begin{myproof}
  Follows from structurally similar derivations as
  the proof of Proposition~\ref{prop-sc-cov-single-resid-block}.
\end{myproof}

\subsubsection{Generalisation to more general forms of skip connections in composed Gaussian derivative residual networks}

In Proposition~\ref{prop-sc-cov-single-resid-block}, we showed
that a maximally simplified Gaussian 
derivative residual block of the form (\ref{eq-simplest-resnet-module-sigma}) is scale covariant.
Since the pointwise $\relu$ function does not affect the scale
covariance properties, and additionally all the primitive components
in the residual block (\ref{eq-simplest-resnet-module-sigma}) are
provably scale covariant, it follows that also a more composed
residual block of the form illustrated in Figure~\ref{fig-residual-blocks}~(right)
\begin{multline}
  \label{eq-resnet-module-2-stages}
  M_{w_1,w_2}(x;\; \sigma) = ({\cal M}_{w_1,w_2,\sigma} f)(x;\; \sigma)  = \\
  = \relu( f(x) + (w_2(\cdot;\; \sigma) * \relu((w_1(\cdot;\; \sigma) * f(\cdot))))(x))
\end{multline}
with the two convolution kernels $w_1(x;\; \sigma)$ and $w_2(x;\; \sigma)$
of the form (\ref{eq-gauss-der-layer}) will be scale covariant:

\begin{mydefinition}
  (Standard Gaussian derivative residual block based on two weight
  functions)
  Given two Gaussian derivative weight functions $w_1(\cdot;\; \sigma)$
  and $w_2(\cdot;\; \sigma)$
  at scale $\sigma \in \bbbr_+$ of the form
  (\ref{eq-gauss-der-layer}), their corresponding standard Gaussian derivative
  block $M_{w_1,w_2}(x;\; \sigma)$ is given by
  \begin{multline}
    \label{eq-resnet-module-2-stages}
    M_{w_1,w_2}(x;\; \sigma) = ({\cal M}_{w_1,w_2,\sigma} f)(x;\; \sigma)  = \\
    = \relu( f(x) + (w_2(\cdot;\; \sigma) * \relu((w_1(\cdot;\; \sigma) * f(\cdot))))(x))
  \end{multline}
  where
  \begin{itemize}
  \item
    $M_{w_1,w_2}(\cdot;\; \sigma)$ denotes the layer at scale $\sigma$,
  \item
    ${\cal M}_{w_1,w_2,\sigma}$ denotes the operator that computes the layer from the
    input $f$.
  \end{itemize}
\end{mydefinition}

\begin{myproposition}
  \label{prop-sc-cov-resid-block-2-weight-functions} 
  (Covariance property of the standard form of Gaussian derivative
  residual block with two weight functions)
  The standard form of a Gaussian derivative residual block
  (\ref{eq-resnet-module-2-stages}), with the
  scale-normalized Gaussian derivative weight functions
  $w_1(\cdot;\; \sigma)$ and $w_2(\cdot;\; \sigma)$
  of the form (\ref{eq-gauss-der-layer}), is scale covariant
  and transforms according to
  \begin{equation}
    M'_{w_1,w_2}(x';\; \sigma') = M_{w_1,w_2}(x;\; \sigma)
  \end{equation}
  under spatial scaling transformations of the input image of the form
  $f'(x') = f(x)$ with $x' = S_x \, x$,
  where the spatial scaling factor $S_x \in \bbbr_+$, with the
  associated scale parameter transforming according to
  $\sigma' = S_x \, \sigma$.
\end{myproposition}

\begin{myproof}
  Follows from structurally similar derivations as
  the proof of Proposition~\ref{prop-sc-cov-single-resid-block}.
\end{myproof}

\noindent
Thereby, we can freely make use of residual skip connections in the 
proposed class of GaussDerResNets, with maintained scale covariance
properties.

Additionally, because of the scale covariance property of a single
composed Gaussian derivative residual block, it holds that the scale
covariance property also extends to the coupling of multiple Gaussian
derivative residual blocks in cascade, provided that the values of the
scale parameters are appropriately matched between the layers in the
corresponding scale channels.

Finally, concerning the generality of this scale covariance property,
it should be remarked that this result is not specific to
pointwise transformations of the $\relu$ form, but applies to more
general classes of pointwise transformations of the output from each
convolution stage. Furthermore, since a batch normalisation stage does
neither affect the scale covariance properties, the scale covariance
property does also apply with batch normalisation stages between the layers.

\subsection{Connections between Gaussian derivative residual blocks 
and semi-discretisations of the diffusion equation}
\label{sec-diffusion-eq-and-pdes}

Smoothing of any image $f$ with the Gaussian kernel,
as used for computing the Gaussian
derivative responses in the Gaussian derivative layers and the
Gaussian derivative residual layers, according to
\begin{equation}
  L(\cdot;\; s) = g(\cdot;\; s) * f(\cdot),
\end{equation}
for the scale parameter in units of the variance $s = \sigma^2$
of the Gaussian kernel,
corresponds to the solution of the diffusion equation
\begin{equation}
  \partial_s L
  = \frac{1}{2} \, \nabla^2 L
  = \frac{1}{2} \,
  \left( \partial_{x_1 x_1} + \partial_{x_2 x_2} + \dots + \partial_{x_D x_D} \right)
\end{equation}
with initial condition $L(\cdot;\; 0) = f(\cdot)$.

If we discretise this diffusion equation along the scale direction $s$
only, according to Euler's forward method
\begin{equation}
  (\partial_s L)(\cdot;\, s)
  \approx \frac{L(\cdot;\; s + \Delta s) - L(\cdot;\; s)}
                {\Delta s},
\end{equation}
then this corresponds to a forward iteration of the form
\begin{equation}
  L(\cdot;\; s + \Delta s)
  = \left( 1 + \frac{\Delta s}{2} \,  \nabla^2 \right)
   L(\cdot;\, s).
\end{equation}
As earlier pointed out in more general terms by Alt et al.\ (\citeyear{alt2023-JMIV}),
there is a close structural similarity between this form of computational operation
and the skip connections in a ResNet.

In our case with Gaussian derivative residual layers,
this computational operation is, however, different, in the sense that
the representations that we compute are, however, not mere Gaussian
smoothing operations or computations of the form ``1 + a Laplacian increment''.
Instead, we use linear combinations of Gaussian derivative responses of different
orders, which then constitute a way of parametrising a set of degrees
of freedom in a computational operation of the form
``1 + correction'',
with these degrees of freedom specifically corresponding
to a Taylor expansion of the local image structure at a certain scale
for a non-linear operation of the simplest possible form of a Gaussian
derivative residual block according to
\begin{multline}
  \label{eq-expl-1stage-gaussderresnet-block}
  M_w(x;\; \sigma) = \\
  = \relu
        \left(
          \left(
            \left( 1 + \sum_{\alpha : |\alpha| \leq N} m(\alpha) \, C_{\alpha }
                   \, \sigma^{|\alpha|} \, \partial_{x^{\alpha}} g(\cdot;\; \sigma)
           \right)
         \right.
       \right. \\
       \left.
         \left. \vphantom{\sum_{\alpha : |\alpha| \leq N}}
           * M_w(\cdot;\, 0)
         \right)(x;\; \sigma)
     \right).
\end{multline}
For the special case when the maximum differentiation order $N = 2$,
and if we disregard the effect of the non-linear $\relu$ operation
and the Gaussian smoothing operation $g(\cdot;\, \sigma)$, it holds that
this evolution operation corresponds to the first step of evolution
in a semi-discretisation
in the scale direction of the $D$-dimensional velocity-adapted
affine diffusion equation
(Lindeberg \citeyear{Lin10-JMIV} Equation~(27))
\begin{equation}
  \label{eq-gen-diff-eq}
  \partial_s L
  = \frac{1}{2} \, \nabla^T ( \Sigma \, \nabla L) - v^T \nabla L
\end{equation}
with the spatial covariance matrix $\Sigma$ of size $D \times D$,
the $D$-dimensional velocity vector $v$, and
the step size $\Delta s$,
leading to the an iteration in the forward scale direction of the form
\begin{multline}
  \label{eq-euler-forward}
  L(\cdot;\; s + \Delta s)
  = \left( 1 + \Delta s \left( \frac{1}{2} \, \nabla^T \Sigma \, \nabla
    - v^T \nabla \right) \right) L(\cdot;\, s)
\end{multline}
for the specific choice of the elements of the spatial covariance matrix
$\Sigma = \{ \Sigma_{i,j} \}_{i=1..D, j= 1..D}$ and the velocity
vector $v = \{ v_i \}_{i=1..D}$ obeying
\begin{align}
  \begin{split}
    \label{eq-rel-aff-vel-pars-to-2nd-order-weights}
    \Delta s \, \Sigma_{i,j} = C_{i,j} \, \sigma^2,
  \end{split}\\
  \begin{split}
    \label{eq-rel-aff-vel-pars-to-1s-order-weights}
    \Delta s \, v_i = C_i \, \sigma.
  \end{split}
\end{align}
Here, we have implicitly assumed natural rules for converting the multi-index notion for
the first-order Taylor coefficients with multi-index notation
$C_{\alpha}$ for $|\alpha| = 1$
to explicitly indexed values $C_i$ and for converting
the second-order Taylor coefficients with multi-index notation
$C_{\alpha}$ for $|\alpha| = 2$
to explicitly indexed values $C_{i,j}$.

In this respect, there is a direct connection between the differential
structure in the maximally simplified
primitive Gaussian derivative residual block
of the general form (\ref{eq-simplest-resnet-module-sigma}),
specialised to the form (\ref{eq-expl-1stage-gaussderresnet-block}),
and generalised models of
spatial receptive fields according to the theory in Lindeberg
(\citeyear{Lin10-JMIV}), which can be precisely shown to correspond to
constitute solutions of generalised diffusion equations of the
form (\ref{eq-gen-diff-eq}):%
\footnote{Additionally, the representations in the implementation are further
  regularised by a complementary spatial discretisation, as well as
  complementary discrete spatial smoothing steps with isotropic discrete
  Gaussian kernels, when computing the spatial derivatives underlying the
  discrete iteration over Gaussian derivative residual layers.
  By the use of such significantly non-infinitesimal amounts of
  spatial smoothing in the computation of the spatial derivative responses,
  the influence of fine-scale and high-frequency image structures on the
  scale steps will be fundamentally different from the bounds on the
  scale step $\Delta s$ obtained when not using such complementary
  spatial smoothing in the computation of spatial derivatives in more
  classical explicit forward schemes for the diffusion equation.
  Due to the use of isotropic Gaussian smoothing as the spatial
  smoothing operation in these iterations, the resulting
  GaussDerResNet architecture implied
  by the Gaussian derivative residual blocks of the form
  (\ref{eq-expl-1stage-gaussderresnet-block}) will, however,
  not obey similar affine covariant properties as the velocity-adapted
  affine diffusion equation (\ref{eq-gen-diff-eq}).}

\begin{myproposition}
  (Structural relationship between Gaussian derivative residual blocks
  up to order 2 and the velocity-adapted affine diffusion equation)
  In the special case, when the maximum order of differentiation
  is $N = 2$, the spatial differentiation operations in a single Gaussian
  derivative residual block of the form
  (\ref{eq-expl-1stage-gaussderresnet-block}), excluding both the
  non-linear ReLU operation and the effect of the Gaussian smoothing
  operation, correspond to the first step of an Euler forward
  discretisation (\ref{eq-euler-forward}) of the velocity-adapted affine diffusion equation
  (\ref{eq-gen-diff-eq}) with the affine and velocity parameters
  $\Sigma_{i,j}$ and $v_i$ related to the second- and first-order
  filter weights $C_{i,j}$ and $C_i$, respectively, according to
  (\ref{eq-rel-aff-vel-pars-to-2nd-order-weights}) and
  (\ref{eq-rel-aff-vel-pars-to-1s-order-weights}) in the associated Gaussian
  derivative weight function of the form (\ref{eq-gauss-der-layer}).
\end{myproposition}

\begin{myproof}
  Follows from direct identification of the parameters
  $\Sigma_{i,j}$ and $v_i$ in the Euler forward
  discretisation (\ref{eq-euler-forward})
  with the second- and first-order filter weights
  $C_{i,j}$ and $C_i$, respectively, in the Gaussian derivative
  weight function $w$ according to (\ref{eq-gauss-der-layer}),
  when the scale parameter $\sigma \downarrow 0^+$,
  leading to the following asymptotic form, when applied to any
  sufficiently regular test function $f$:
  \begin{multline}
    (w(\cdot;\; \sigma) * f(\cdot))(x) \approx \\
    \left( \frac{1}{2} \sum_{(i,j) \in [1, D]^2}C_{i,j} \, \sigma^2 \, \partial_{x_i x_j}
    - \sum_{i \in [1, D]}  C_i \, \sigma \, \partial_{x_i} \right) f(x).
  \end{multline}
  Note that due
  to the spatial mirroring operation in the associated convolution operation, we
  have flipped the signs of the underlying odd-shaped
  first-order Gaussian derivative terms, which
  thereby leads to the relationships in
  Equations~(\ref{eq-rel-aff-vel-pars-to-2nd-order-weights}) and
  (\ref{eq-rel-aff-vel-pars-to-1s-order-weights}),
  by means of comparison to the notion of a Gaussian residual block
  based on a single weight function of the form
  (\ref{eq-simplest-resnet-module}).
\end{myproof}

\noindent
With regard to the interpretation of this result, it should be noted
that the velocity-adapted diffusion equation
(\ref{eq-gen-diff-eq}) describes the evolution
properties over scale of the most general form of continuous
scale-space kernels, that obey the scale-space axioms in terms of
non-enhancement of local extrema
with a semi-group property over scale
(Lindeberg \citeyear{Lin10-JMIV}, Theorems~5 and~6).
In this respect, we therefore have a very close match between the
variabilities of the shapes of the receptive fields
in Gaussian derivative blocks up to order 2
and the variabilities in the receptive
fields, that are spanned by
axiomatically determined scale-space theory.

\subsection{Resulting Gaussian derivative ResNet architecture}
\label{sec-gaussderresnets}

Based on the above theory, we propose to define GaussDerResNets by
coupling sets of Gaussian derivative residual layers in cascade, with
non-linearities in between.

\subsubsection{Definition of single-scale-channel networks}
\label{def-single-sc-chan-network}

For a fixed value of the initial scale level
$\sigma_0$, with its associated coarser scale levels
$\sigma_k$ at any level $k$ for $k \in [1, K]$
in the hierarchy,%
\footnote{For the GaussDerResNet architecture with $K$ levels:
  (i)~the first level with $k = 1$ corresponds to the first layer~1,
  (ii)~the intermediate levels with $k \in [2, K-1]$ correspond to the
  layers~$2k-2$ and $2k-1$, and
  (iii)~the last level with $k = K$ corresponds to the last layer
  $2K-2$.}
we construct a single-scale-channel network
by combining the following types of operations in cascade:
\begin{itemize}
\item
  {\bf First layer:} for $k = 1$ a Gaussian derivative layer of the form
  (\ref{eq-gauss-der-module-sigma}) complemented with
  a batch normalization stage:
  \begin{equation}
    \label{eq-first-layer-single-scale-network}
    F_{\sigma_0,1}(\cdot) = {\cal G}_{w_1,\sigma_1}(f(\cdot)).
  \end{equation}
\item
  {\bf Intermediate layers:} for $k \in [2, K-1]$ a set of following standard Gaussian
  derivative blocks of the form (\ref{eq-resnet-module-2-stages}),
  complemented with batch normalization layers, and coupled in cascade:
  \begin{equation}
    \label{eq-intermed-layer-single-scale-network}    
    F_{\sigma_0,k}(\cdot) = 
      {\cal M}_{w_{2k-2},w_{2k-1},\sigma_k}(F_{\sigma_0,k-1}(\cdot)).
  \end{equation}
\item
  {\bf Final layer:} for $k = K$ a Gaussian derivative layer of the form
  (\ref{eq-gauss-der-module-sigma}) complemented with
  an optional complementary
  batch normalization stage:
  \begin{equation}
    \label{eq-final-layer-single-scale-network}    
    F_{\sigma_0,K}(\cdot) =
    {\cal G}_{w_{2K-2},\sigma_K}(F_{\sigma_0,K-1}(\cdot)).
  \end{equation}
\end{itemize}
\noindent
In this cascade of Gaussian derivative layers and residual
blocks, the values of the scale parameter $\sigma_k$
across the levels $k$ increases%
\footnote{The motivation for increasing the size of the receptive
  fields with the layer depth in this way is to enable integration from local to
  regional to possibly global information from lower to higher layers
  in the deep hierarchy. Specifically, the use of a geometric progression
  corresponds to a self-similar distribution of the scale values
  across the scales. If using the same size of the receptive
  fields in all the layers in the network, then either the lowest
  layers would be operating at too coarse scale levels or
  the receptive fields in the highest layers would be too small.
  Note that this is in contrast to the more common types of CNNs
  that may use the same fixed size of the filters in all the layers.
  In such more common deep networks,
  there are, however, usually also spatial subsampling
  operations between a significant number of the layers, which thereby
  implicitly increase the effective size of the receptive fields from
  lower to higher layers in the deep hierarchy. For such a more
  common type of CNN, the effective size of the receptive field is
  determined by the product of the size of the filter weight matrix multiplied
  by the product of the current and all spatial subsampling factors in the
  previous layers.}
geometrically for $k \geq 1$ according to
\begin{equation}
  \label{eq:sigma-geometric}
  \sigma_k = \sigma_0 \, r^{k-1}
\end{equation}
for some initial scale level $\sigma_0 \in \bbbr_+$ and some
suitably chosen scale ratio $r > 1$.

This choice is similar to the construction of regular Gaussian
derivative networks in Lindeberg (\citeyear{Lin22-JMIV}), and
is also motivated by the further findings in
Perzanowski and Lindeberg
(\citeyear{PerLin25-JMIV-ScSelGaussDerNets}),
where such a logarithmic distribution of the scale levels
was empirically demonstrated to perform very well compared to
a more general approach of learning the scale levels from training
data.%
\footnote{Since learning of the scale levels did not lead to any
  significant improvements in accuracy in our previous studies with
  regular GaussDerNets
  (Perzanowski and Lindeberg
  \citeyear{PerLin25-JMIV-ScSelGaussDerNets})
  and furthermore complementary learning of the scale levels only lead
  to a marginal improvement in accuracy for the discrete scale-space
  formulation of ConvNeXt networks in Lindeberg et al.\
  (\citeyear{LinBabKia26-JMIV}), we do not explicitly explore the option of
  complementing the GaussDerResNets with learning of the scale levels in
  this paper.}

\begin{mydefinition}
  (Pure single scale channel in a GaussDerResNet)
  Given an image $f \colon \bbbr^D \rightarrow \bbbr$
  and an initial scale level $\sigma_0 \in \bbbr_+$,
  the corresponding single scale channel in a GaussDerResNet
  is defined by the combined effect of the operations
  (\ref{eq-first-layer-single-scale-network}),
  (\ref{eq-intermed-layer-single-scale-network}) and
  (\ref{eq-final-layer-single-scale-network})
  expanded over a set of classes $c$ in the final
  layer according to
  \begin{equation}
    \label{pure-single-chan-gaussderresnet}
    F_{\sigma_{0},K}^c(\cdot) = \Lambda_{\sigma_{0}}^c(f(\cdot)),
  \end{equation}
  with $K$ denoting the total number of levels and
  $\Lambda^c_{\sigma_{0}}$ denoting the operator that computes the
  combined effect of these operations for the class $c$,
  when applied to any input image $f$.
\end{mydefinition}
A more detailed mathematical definition of the layer operations
in such a the pure single scale channel, as defined by the operator
$\Lambda^c_{\sigma_{0}}(f(\cdot))$ applied to the input image
$f$, and then also including the above suppressed batch normalization stages,
as well as the notation for combining multiple feature channels between
the different layers, is given in Appendix~A.

Then, by additionally combining these operations with
\begin{itemize}
\item
  {\bf Spatial selection}:%
\footnote{Regarding the rescaled datasets that we use for evaluation
    the scale generalisation propertied of GaussDerResNets experimentally,
    the underlying images from the Fashion-MNIST and CIFAR-10 datasets
    have a very minimal image size and are indeed very tightly
    cropped. For those datasets, we here throughout make use of central pixel
    extraction, to ensure that we may not risk possible artefacts
    caused by boundary effects.
    If the spatial max pooling operation would not assume its maximum
    response at the image centre, then the response of the activation
    map at such an image point could be strongly affected by the
    receptive fields picking up effects of the spatial mirroring
    operations at the image boundaries prior to the convolution operations.
    By choosing the central pixel for these datasets,
    we do thereby ensure to make maximum use of the
    available image information for these minimal image sizes,
    without risk for possible artefacts.
    For the rescaled dataset
    based on images from the STL-10 dataset, where the object are not
    centred, the spatial max pooling operation is on the other hand
    really essential. Let us, however, also complement by saying that
    we have in our previous work
    (Perzanowski and Lindeberg
    \citeyear{PerLin25-JMIV-ScSelGaussDerNets})
    successfully applied global spatial max pooling to the
    Rescaled Fashion-MNIST
    with translations dataset
     (Perzanowski and Lindeberg
     \citeyear{PerLin-2025-fashionmnist-transl-zenodo}), where the
     objects in the Fashion-MNIST dataset are, however, embedded in
     larger images with certain margins to the boundaries.}
   in terms of either
  {\em central pixel extraction\/}%
  \begin{equation}
    \label{eq:central-pixel-extraction}
    \operatorname{SpatSel}_{\centre}(f) = f(x_{\centre}),
  \end{equation}
  or {\em global spatial max pooling\/}
  \begin{equation}
    \label{eq:spat-max-pool}
    \operatorname{SpatSel}_{\max}(f) = \sup_{x \in \bbbr^2} f(x),
  \end{equation}
\end{itemize}
we can construct a {\em single-scale-channel network\/} with global
spatial max pooling of the following form:

\begin{mydefinition}
  (Single scale channel with global spatial pooling in a
  GaussDerResNet)
  Given an image $f \colon \bbbr^D \rightarrow \bbbr$
  and an initial scale level $\sigma_0 \in \bbbr_+$,
  the corresponding single scale channel with global spatial pooling
  for the class $c$ in a GaussDerResNet is defined by
  \begin{equation}
    \label{fig-single-chan-gaussderresnet-architecture}
    \Gamma_{\sigma_{0}}^c(f(\cdot))
    = \operatorname{SpatSel}(\Lambda_{\sigma_{0}}^c(f(\cdot))).
  \end{equation}
\end{mydefinition}

\begin{figure*}[hbtp]
  \begin{center}
    \begin{tabular}{c}
       \includegraphics[width=0.99\textwidth]{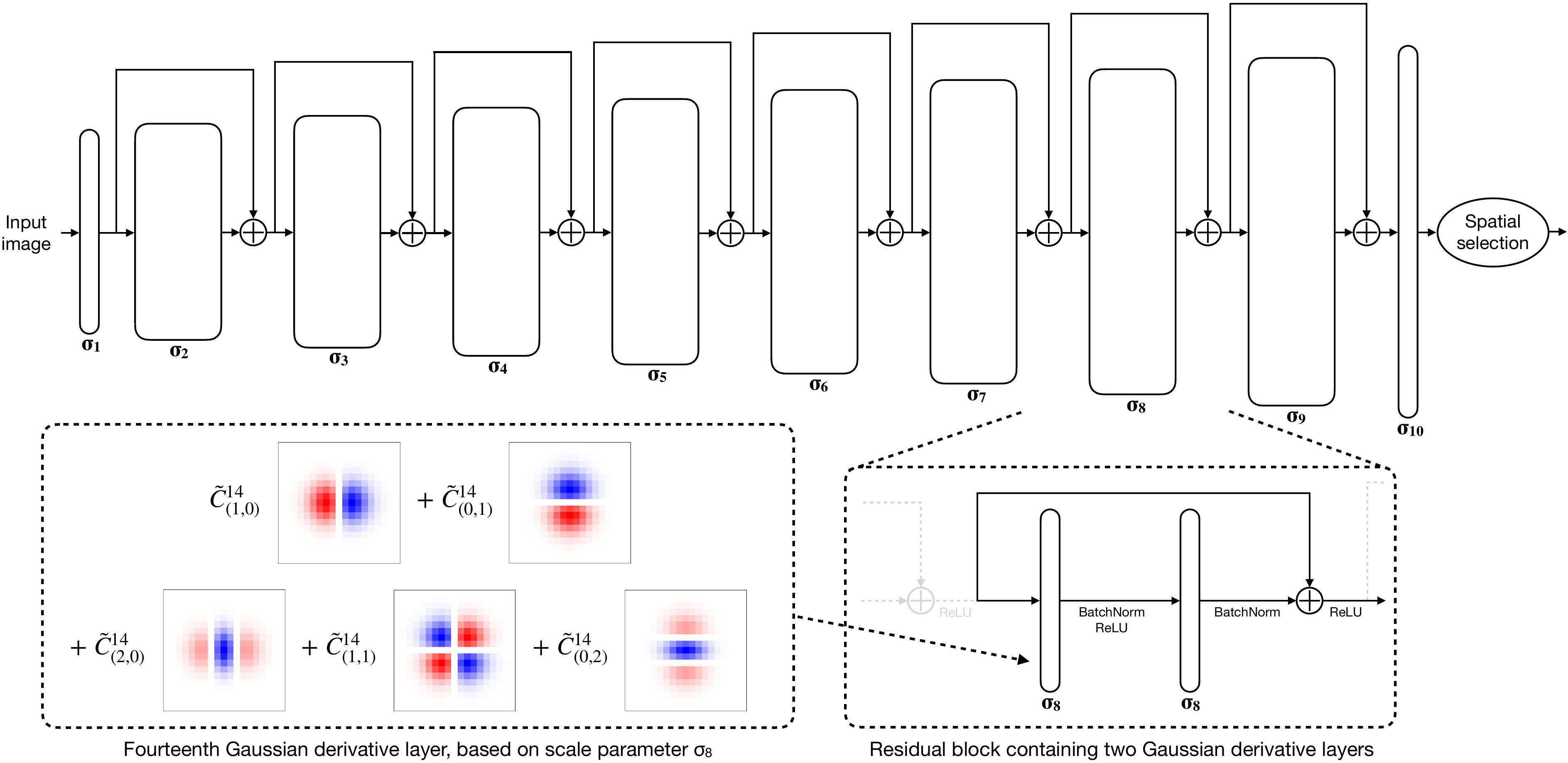}
    \end{tabular}
  \end{center}
\caption{Illustration of a single-scale-channel Gaussian derivative
  residual network consisting of 18 Gaussian derivative layers,
  together with the internal structure of a representative residual
  block, as well as an inside look into the composition of a Gaussian
  derivative layer within this residual block.
  The top part of the diagram illustrates the architecture of a
  single-scale-channel GaussDerResNet, with the layers
  $F_{\sigma_0,k}$ organised into
  residual blocks according to (\ref{eq-resnet-module-2-stages})
  and (\ref{eq-intermed-layer-single-scale-network},
  with the exception of the first and the last
  layers, which are regular Gaussian derivative layers
  according to (\ref{eq-gauss-der-module-sigma}),
  (\ref{eq-first-layer-single-scale-network}),
  and (\ref{eq-final-layer-single-scale-network}).
  Each residual block is constructed out of two Gaussian derivative
  layers together with a skip connection, and is shown in
  detail for the residual block ${\cal M}_{w_{14},w_{15},\sigma_{8}}$ in the bottom
  right of the diagram, with the light grey dotted lines not part of
  the residual block and shown only for context.
  The pure single-scale-channel GaussDerResNet
  $\Lambda_{\sigma_{i,0}}^c(f)$ applied the input image $f$
  according to (\ref{pure-single-chan-gaussderresnet})
  is parametrised by an initial scale value $\sigma_{i,0}$,
  with the scale parameter of each
  layer determined according to a geometric progression 
  according to (\ref{eq:sigma-geometric}), resulting in the
  sizes of the receptive fields increasing with the network depth, with each residual
  block using the same scale parameter value for both of its layers.
  Furthermore, each Gaussian derivative layer in the network consists
  of a convolution with a linear combination of discretised Gaussian
  derivative basis kernels $w(x;\; \sigma)$ according to
  (\ref{eq-gauss-der-layer}), as visualised in the
  bottom left of the diagram for layer $14$, which is the first layer within
  the Gaussian derivative residual block at the level $k = 8$.
  In the top row of that bottom-left figure,
  we see the contributions from the first order Gaussian
  derivatives, and in the bottom row the contributions from the second
  order Gaussian derivatives.
  Each basis kernel in the layer is computed by applying a
  corresponding central difference operator to a discrete analogue of
  the Gaussian kernel according to (\ref{eq:disc-gauss+central-diff}).
  The parameters $\tilde{C_{\alpha}} = m(\alpha) \, C_{\alpha } \, \sigma^{|\alpha|}$
  represent the learned weights of the
  layer, with the tilde symbol over the weight parameter $C_{\alpha}$
  including the effects of the corresponding scale
  normalisation factor $\sigma^{|\alpha|}$ and the
  multinomial normalisation factor $m(\alpha)$, both defined according
  (\ref{eq-gauss-der-layer})
  and~(\ref{eq-gauss-der-layer-in-depthsep}).
  Finally, the output of the single-scale-channel GaussDerResNet is
  obtained by applying a spatial selection stage $\Gamma_{\sigma_{i,0}}^c(f)
  = \operatorname{SpatSel}(\Lambda_{\sigma_{i,0}}^c(f))$
  according to (\ref{fig-single-chan-gaussderresnet-architecture})
  to the output of the final layer $F_{\sigma_0,K}$.}
\label{fig-single-gaussderresnet-architecture}
\end{figure*}

\noindent
Notably, when constructing the GaussDerResNets, the first and the last layers
are regular Gaussian derivative layers without skip
connections, while all the intermediate layers are Gaussian derivative
residual layers with skip connections,
as also illustrated in
Figure~\ref{fig-single-gaussderresnet-architecture}.%
\footnote{The reason for using a regular Gaussian derivative layer, instead of a
Gaussian derivative residual layer in the first layer, is that regular
Gaussian derivative kernels have by axiomatic derivations been shown
to constitute a canonical choice for defining the first layer of
visual processing, based on general symmetry arguments by
Iijima (\citeyear{Iij62}),
Koenderink (\citeyear{Koe84-BC}),
Koenderink and van Doorn (\citeyear{koenderink:92}),
Lindeberg (\citeyear{Lin93-Dis, Lin10-JMIV}) and
Weickert et al.\ (\citeyear{WeiIshImi99-JMIV}).
By this choice, we also prevent the pure input
image from being forwarded to the higher layers by skip connections.
The reason, for using a regular Gaussian derivative layer instead of a
Gaussian derivative residual layer in the last layer, is to provide
maximum flexibility with respect to the construction of the output
from each single scale channel.
These choices are also consistent with how regular ResNets, not based
on Gaussian derivatives, are constructed, by not using any skip
connections in the first or the last layers.}

\subsubsection{Definition of multi-scale-channel networks}
\label{def-multi-sc-chan-network}

By in addition, extending this definition over multiple values of the
initial scale parameter $\sigma_0$,
we can construct a {\em multi-scale-channel network\/}
for image classification/single object detection
over the class indices $c$ in the following way:

\begin{mydefinition}
  (Multi-scale-channel GaussDerResNet with global spatial and scale
  pooling)
  Given an image $f \colon \bbbr^D \rightarrow \bbbr$ and a set of
  geometrically distributed initial scale levels $\sigma_{i ,0}\in \bbbr_+$, its
  GaussDerResNet for global image classification/single object detection over
  the classes $c$ is defined by
  \begin{equation}
    \label{eq:max-pooling-over-scale-rot}
    \operatorname{logits}^{c}(f(\cdot)) =
    \operatorname{ScalePool}_{i\in I}(\Gamma_{\sigma_{i,0}}^{c}(f(\cdot))),
  \end{equation}
  where
  \begin{itemize}
  \item
    $\operatorname{ScalePool}_{i\in I}$ denotes the scale pooling%
\footnote{With regard to scale pooling, permutation invariance of the
  scale pooling operation, as max pooling, logsumexp pooling and
  average pooling obey, constitutes a {\em sufficient\/} condition for
  provable scale invariance. Permutation invariance is, however, not a
  necessary condition. More generally, one could conceive also
  explicitly measuring how stable the responses in the activation
  maps in the different scale channels are to variations in the
  initial scale parameter $\sigma_{i,0}$ of the different scale
  channels as indexed by $i$, by linking similar features across scales and
  then also including the life-length of these structures as one
  component that forms the output scale pooling mechanism,
  based on the heuristic principle that features that are stable across scales should be
  likely to correspond to significant image structures, see Lindeberg
  (\citeyear{Lin92-IJCV,Lin92-PAMI,Lin12-JMIV,Lin15-JMIV}) for
  previous applications of such ideas for detecting regions of
  interest for focus-of-attention and for performing interest point detection.}
  mechanism over the set of scale channels $i \in I$, with $|I|$ denoting
  the number of scale channels, in terms of either
    \begin{itemize}
    \item
      max pooling over scales
      \begin{multline}
        \label{eq:max-pooling-over-scales}
        \operatorname{ScalePool}_{\maximum}(\{ \Gamma_{\sigma_{i,0}}^{c}(f(\cdot)) \}_{i=1}^{|I|}) = \\
        = \sup_{i \in I} \operatorname{SpatSel}(\Lambda_{\sigma_{i,0}}^{c}(f(\cdot))),
      \end{multline}
    \item
      logsumexp pooling (a smooth approximation to the maximum)
      over scales
      \begin{multline}
        \label{eq:logsumexp-pooling-over-scales}
        \operatorname{ScalePool}_{\logsumexp}(\{ \Gamma_{\sigma_{i,0}}^{c}(f(\cdot)) \}_{i=1}^{|I|}) = \\
        = \log \left( {\sum_{i \in I}
            \exp\left(\operatorname{SpatSel}(\Lambda_{\sigma_{i,0}}^{c}(f(\cdot)))\right)}
          \right),
      \end{multline}
    \item
      or average pooling over scales
      \begin{multline}
        \label{eq:avg-pooling-over-scales}
        \operatorname{ScalePool}_{\avg}(\{ \Gamma_{\sigma_{i,0}}^{c}(f(\cdot)) \}_{i=1}^{|I|}) = \\
        = \frac{1}{|I|} \sum_{i \in I}\operatorname{SpatSel}(\Lambda_{\sigma_{i,0}}^{c}(f(\cdot))).
      \end{multline}
    \end{itemize}
  \end{itemize} 
\end{mydefinition}
In this setting, the different scale channels are copies of each
other, while at different scales, meaning that they
share the same weights and are parametrised by the same relative scale
ratio $r$ between the adjacent scale channels. This means that the 
receptive field sizes of the Gaussian derivative basis kernels in each layer 
of a scale channel steadily increase, proportional to the initial scale 
value of the scale channel.
Specifically, the initial scale parameters $\sigma_{i,0}$ in the different scale
channels are related according to
\begin{equation}
  \label{eq:sigma0-multichan}
  \sigma_{i,0} = \lambda^{i-1} \sigma_{1,0}
\end{equation}
where $\lambda \in \{ \sqrt{2}, \sqrt[4]{2} \}$ is the ratio between
the initial scale levels between adjacent scale channels.

Thus, when further combining the use of different initial scale values
for different scale channels with increasing scale values according to
(\ref{eq:sigma-geometric}) towards higher layers in the hierarchy for 
any scale channel, we obtain that the scale
value at level $k$ in the scale channel with index $i$ is
for $k \geq 1$ given by
\begin{equation}
  \sigma_{i,k} = \sigma_0 \, \lambda^{i-1}  \, r^{k-1}.
\end{equation}

\begin{figure}[hbt]
  \begin{center}
    \begin{tabular}{c}
       \includegraphics[width=0.48\textwidth]{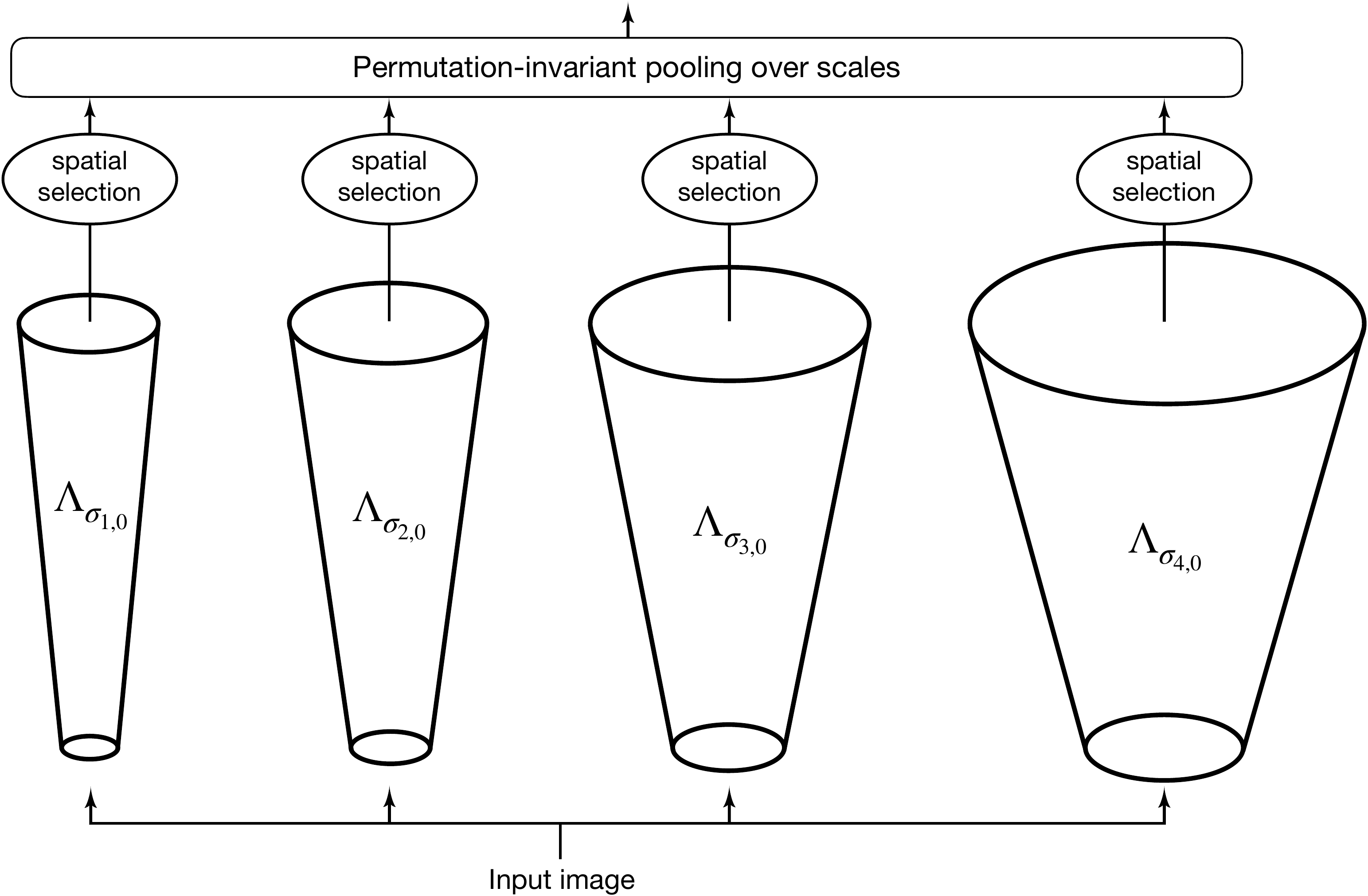}
    \end{tabular}
  \end{center}
\caption{Conceptual illustration of a multi-scale-channel Gaussian derivative
  residual network, composed of 4 parallel single-scale-channel
  GaussDerResNets, referred to as scale channels.
  Each scale channel is constructed as shown in the top part of
  Figure~\ref{fig-single-gaussderresnet-architecture}, consisting of the
  (i)~convolutional stage $\Lambda_{\sigma_{i,0}}^c$, composed of 18
  Gaussian derivative layers, represented by the conical frustums
  (meaning the tapered cylinders) in the diagram, indicating the
  increase of receptive field size with depth, and (ii)~the spatial
  selection stage.
  There is a geometric distribution of the spacing between the initial
  scale values $\sigma_{i,0}$, that the scale channels are based on,
  set by a fixed ratio, as expressed in
  Equation~(\ref{eq:sigma0-multichan}).
  Each scale channel processes a copy of the input image, and
  crucially, in order to achieve scale covariance, the scale channels
  share the same weights, with all the scale channels parametrised by the same
  relative scale ratio $r$. Finally, instead of a fully connected
  classification layer, permutation-invariant pooling over the scale
  channels is performed as a final processing stage in the network.}
\label{fig-multi-gaussderresnet-architecture}
\end{figure}

\subsubsection{Resulting network architecture}
\label{sec-gaussderresnet-arch}

The entire multi-scale-channel GaussDerResNet architecture is
illustrated in Figure~\ref{fig-multi-gaussderresnet-architecture}.
Due to the above described scale-covariant properties of the
underlying Gaussian derivative weight functions, with the
corresponding associated covariance properties of their associated
Gaussian residual blocks, according to
Propositions~\ref{prop-sc-cov-single-resid-block}
and~\ref{prop-sc-cov-resid-block-2-weight-functions}, it follows that
the resulting multi-scale-channel GaussDerNets will be provably scale
covariant:

\begin{mytheorem}
  \label{prop-sc-cov-gaussderresnets}
  (Scale covariance properties of GaussDerResNets without scale pooling)
  Consider two images $f \colon \bbbr^D \rightarrow \bbbr$
  and $f' \colon \bbbr^D \rightarrow \bbbr$, that are related
  according to a uniform scaling transformation such that
  $f'(x') = f(x)$ for $x' = S_x \, x$,
  where $S_x \in \bbbr_+$ is the spatial scaling factor.
  Define the pure scale channels $\Lambda_{\sigma_0}^{c}(f(\cdot))$ and
  $\Lambda_{\sigma_0'}^{c}(f'(\cdot))$ over a continuum of the initial
  scale parameter $\sigma_0 \in \bbbr$ without any spatial pooling or
  scale pooling of $f$ and $f'$, respectively, by coupling a first
  Gaussian derivative layer, a set of standard Gaussian derivative blocks and a
  final Gaussian derivative layer in cascade for each pure scale
  channel, as described further in Section~\ref{def-single-sc-chan-network}.
  Then, these pure scale channels obey the scale
  covariance property
  \begin{equation}
    \label{sec-sc-cov-prop-pure-sc-channels}
    \Lambda_{\sigma'_{0}}^{c}(f'(\cdot))(x') = \Lambda_{\sigma_{0}}^{c}(f(\cdot))(x),
  \end{equation}
  provided that the scale parameters $\sigma_{0}$ and $\sigma'_{0}$ in the
  respective first layers are related according to
  $\sigma'_{0} = S_x \, \sigma_{0}$.
  When additionally combined with spatial selection based on spatial
  max pooling according to
  (\ref{fig-single-chan-gaussderresnet-architecture})
  \begin{align}
    \begin{split}
      \Gamma_{\sigma_{0}}^c(f(\cdot))
      & = \operatorname{SpatSel}_{\max}(\Lambda_{\sigma_{0}}^c(f(\cdot))),
    \end{split}\\
    \begin{split}
      \Gamma_{\sigma_{0}'}^c(f'(\cdot))
      & = \operatorname{SpatSel}_{\max}(\Lambda_{\sigma'_{0}}^c(f'(\cdot))),
      \end{split}
  \end{align}
  the resulting scale channels
  $\Gamma_{\sigma_{0}}^c(f(\cdot))$
  and $\Gamma_{\sigma'_{0}}^c(f'(\cdot))$ with spatial max
  pooling are also scale covariant with
  \begin{equation}
    \label{sec-sc-cov-prop-max-pool-sc-channels}
    \Gamma_{\sigma'_{0}}^c(f'(\cdot)) = \Gamma_{\sigma_{0}}^c(f(\cdot)).
  \end{equation}
\end{mytheorem}

\begin{myproof}
  The result (\ref{sec-sc-cov-prop-pure-sc-channels})
  follows directly by combining (i)~the scale covariance
  property of the standard Gaussian derivative blocks in
  Proposition~\ref{prop-sc-cov-resid-block-2-weight-functions}
  with (ii)~the corresponding scale covariance property of the first
  and the last Gaussian derivative layers in
  Lindeberg (\citeyear{Lin22-JMIV}) Section~4.3
  extended to the arbitrary dimensions in
  Proposition~\ref{prop-sc-cov-single-gauss-der-block},
  and finally
  (iii)~applying the transformation properties of these image
  representations under scaling transformation in cascade across the
  layers.
  
  The extension (\ref{sec-sc-cov-prop-max-pool-sc-channels}) to
  spatial max pooling follows directly
  from the fact that the result of applying spatial max
  pooling to an image is not affected by the scaling transformations.
\end{myproof}

\noindent
Note, however, that if the initial scale levels $\sigma_0$
would on the other hand only be discrete
\begin{equation}
  \sigma_{i,0} = \lambda^{i-1} \, \sigma_{1,0}
\end{equation}
for some global initial scale level $\sigma_{1,0} \in \bbbr_+$, then the scale covariance
properties will only hold for scaling factors $S_x$ that are integer
power of the ratio $\lambda$ between adjacent scale levels
\begin{equation}
  S_x = \lambda^j \quad\mbox{for some} \quad j \in \bbbz.
\end{equation}
When combined with scale selection, the network architecture is,
as we will described further in
Section~\ref{sec-sc-inv-gaussderresnets},
also scale invariant. Experimentally, we will demonstrate that the network is
capable of scale generalisation, meaning that it can achieve high
performance even when tested at scales not encountered during
training.

The range of scales, that this model is able to handle, is
determined by the scale range spanned by the initial scale values
$\sigma_{i,0}$ of the scale channels. Due to the weight sharing, each
scale channel is responsible for handling a single scale level
corresponding to its initial scale level $\sigma_{i,0}$, meaning that if an image
featuring a pattern, that has been learned during training, is rescaled
and then provided as input, then the scale channel at a matching scale
level will be strongly activated in response to it, even if the given
scale level has not been previously seen during training.

Additionally, with a dense enough distribution of the initial scale
levels $\sigma_{i,0}$, as defined according to
Equation~(\ref{eq:sigma0-multichan}), the multi-scale-channel
GaussDerResNet can automatically compute good approximations
to smooth interpolation between the scales in
between the scale channels, resulting in a flat and smooth scale
generalisation curve across the entire operational scale
interval. This means that neighbouring scale channels can
approximately cover the entire in-between scale range, due to the
discretised nature of the filter in the Gaussian derivative layers and the
scale selection properties of the network, effectively allowing for
scale-invariant processing, despite using only a finite number of scale
channels when implemented in the real-world setting.

Finally, we also include \textit{boundary} scale channels at both 
the ends of the scale interval, to reduce the possible artefacts
that could otherwise occur due to image structures falling outside 
the operational scale range.

\begin{figure*}[ht]
\begin{center}
    \[
      \begin{CD} 
       f' @>{\mathcal{G}_{\sigma'_{1}}}>> F_{1}' @>{\mathcal{M}_{\sigma'_{2}}}>>  F_{2}' @>{\mathcal{M}_{\sigma'_{3}}}>> F_{3}' @>{\mathcal{M}_{\sigma'_{4}}}>> F_{4}' @>{\mathcal{M}_{\sigma'_{5}}}>> F_{5}' @>{\mathcal{M}_{\sigma'_{6}}}>> F_{6}' @>{\mathcal{M}_{\sigma'_{7}}}>>  F_{7}' @>{\mathcal{M}_{\sigma'_{8}}}>>  F_{8}'  @>{\mathcal{M}_{\sigma'_{9}}}>>  F_{9}'  @>{\mathcal{G}_{\sigma'_{10}}}>> F_{10}' \\     
       @AA{\mathcal{S}}A @AA{\mathcal{S}}A @AA{\mathcal{S}}A @AA{\mathcal{S}}A @AA{\mathcal{S}}A @AA{\mathcal{S}}A @AA{\mathcal{S}}A @AA{\mathcal{S}}A @AA{\mathcal{S}}A @AA{\mathcal{S}}A @AA{\mathcal{S}}A \\
       f @>{\mathcal{G}_{\sigma_{1}}}>> F_{1} @>{\mathcal{M}_{\sigma_{2}}}>> F_{2} @>{\mathcal{M}_{\sigma_{3}}}>> F_{3} @>{\mathcal{M}_{\sigma_{4}}}>> F_{4} @>{\mathcal{M}_{\sigma_{5}}}>> F_{5} @>{\mathcal{M}_{\sigma_{6}}}>> F_{6} @>{\mathcal{M}_{\sigma_{7}}}>>  F_{7} @>{\mathcal{M}_{\sigma_{8}}}>>  F_{8}  @>{\mathcal{M}_{\sigma_{9}}}>>  F_{9}  @>{\mathcal{G}_{\sigma_{10}}}>> F_{10}
    \end{CD}
    \]
  \end{center}
  \caption{Commutative diagram for the entire 18-layer Gaussian
    derivative residual network, illustrating the scale-covariant
    properties of the architecture. In the diagram,
    $\mathcal{G}_{\sigma_k}= \mathcal{G}_{w_k,\sigma_k}$
    represents Gaussian derivative layer at the scale $\sigma_k$
    according to (\ref{eq-gauss-der-module-sigma}),
    $\mathcal{M}_{\sigma_k} = \mathcal{M}_{w_{2k-2,2k-1,\sigma_k}}$
    represents a standard Gaussian derivative
    residual block at the scale $\sigma_k$
    according to (\ref{eq-resnet-module-2-stages})
    and (\ref{eq-intermed-layer-single-scale-network}),
    where both there operators are based on Gaussian derivative primitives up to
    a given order $N$ of spatial differentiation according to (\ref{eq-gauss-der-layer}).
    The scaling operator $\mathcal{S}$ represents a uniform
    scaling transformation with a scaling factor
    $S_x \in \mathbb{R}_+$ of the form $x' = S_x \, x$,
    with the associated transformation of the scale parameters
    according to $\sigma' = S_x \, \sigma$.
    This scaling operator acts on both the input image $f$ or any
    layer output $F_k = F_{\sigma_0,k}$ from the residual blocks,
    according to $f'(x') = f(x)$ and
    $F'_{\sigma'_0,k}(x') = F_{\sigma_0,k}(x)$
    for $\sigma'_0 = S_x \, \sigma_0$, leading to the 
    abbreviated notation $F'_{k}(x')$ = ${F}_{k}(x)$ above.
    The commutative diagram should be read from the bottom left to
    the top right, and shows that each level in the cascade of the
    network can be matched under arbitrary scaling transformations
    ${\cal S}$, provided that
    such matching has also been done in the same manner at every
    previous step in the hierarchy, including the input image. As
    described in Section~\ref{sec-scale-covariance-properties}, these
    scale covariance properties are not affected by the use of
    residual connections, or the presence non-linear $\relu$ or
    batch normalisation stages in the network.}
  \label{fig-comm-diag-gaussderresnet}
\end{figure*}

\subsubsection{Scale invariance and translation covariance properties
  of multi-scale-channel GaussDerResNets with combined spatial and scale selection}
\label{sec-sc-inv-gaussderresnets}

Due to the use of permutation-invariant pooling mechanisms over
scales, it follows that the results of scale selection using either
max pooling, average pooling or logsumexp pooling are
provably scale invariant:

\begin{mytheorem}
  (Scale invariant properties of GaussDerResNets with combined spatial
  pooling and scale pooling)
  Consider two images $f \colon \bbbr^D \rightarrow \bbbr$
  and $f' \colon \bbbr^D \rightarrow \bbbr$, that are related
  according to a uniform scaling transformation such that
  $f'(x') = f(x)$ for $x' = S_x \, x$,
  where $S_x \in \bbbr_+$ is the spatial scaling factor.
  Define the jointly spatially and scale pooled GaussDerResNets
  \begin{align}
    \begin{split}
      \operatorname{logits}^{c}(f(\cdot)) =
      & \operatorname{ScalePool}_{\sigma_0 \in \bbbr_+}(\Gamma_{\sigma_0}^{c}(f(\cdot))),
    \end{split}\\
    \begin{split}
      \operatorname{logits}^{c}(f'(\cdot)) =
      & \operatorname{ScalePool}_{\sigma_0 \in \bbbr_+}(\Gamma_{\sigma_0}^{c}(f'(\cdot))),
    \end{split}
  \end{align}
  of $f$ and $f'$, respectively, with the scale pooling methods extended
  to over a continuum of the initial scale parameter
  $\sigma_0 \in \bbbr_+$: (i)~for max pooling over scale according to
  \begin{multline}
    \label{eq:max-pooling-over-scales-cont}
    \operatorname{ScalePool}_{\text{max},\sigma_0 \in \bbbr_+}
    (\Gamma_{\sigma_{0}}^{c}(f(\cdot))= \\
    = \sup_{\sigma_0 \in \bbbr_+} \operatorname{SpatSel}(\Lambda_{\sigma_0}^{c}(f(\cdot))),
  \end{multline}
  (ii)~for logsumexp pooling over scale according to
  \begin{multline}
    \label{eq:logsumexp-pooling-over-scales-cont}
    \operatorname{ScalePool}_{\text{logsumexp}, \sigma_0 \in \bbbr_+}
    ( \Gamma_{\sigma_{0}}^{c}(f(\cdot)) = \\
    =  \log{\int_{\sigma_0 = 0+}^{\infty}
        \exp\left(\operatorname{SpatSel}(\Lambda_{\sigma_{0}}^{c}(f(\cdot)))\right)}
       \, \frac{d\sigma_0}{\sigma_0},
  \end{multline}
  and (iii)~for average pooling over scale according to
  \begin{multline}
    \label{eq:avg-pooling-over-scales-cont}
    \operatorname{ScalePool}_{\text{average},\sigma_0 \in \bbbr_+}
    ( \Gamma_{\sigma_{0}}^{c}(f(\cdot)) = \\
    = \int_{\sigma_0 = 0^+}^{\infty}
        \operatorname{SpatSel}(\Lambda_{\sigma_{0}}^{c}(f(\cdot)))
      \, \frac{d\sigma_0}{\sigma_0}.
  \end{multline}
  Then, the resulting scale-pooled multi-scale-channel networks are scale invariant
  in the sense that
  \begin{equation}
    \operatorname{logits}^{c}(f'(\cdot)) = \operatorname{logits}^{c}(f(\cdot)),
  \end{equation}
  implying that the results of image classification/single object
  detection
  \begin{align}
    \begin{split}
      \hat{c}(f(\cdot)) = \argmax_c \operatorname{logits}^{c}(f(\cdot)),
    \end{split}\\
    \begin{split}
      \hat{c}(f'(\cdot)) = \argmax_c \operatorname{logits}^{c}(f'(\cdot)),
    \end{split}
  \end{align}
  will be the same
  \begin{equation}
    \hat{c}(f'(\cdot)) = \hat{c}(f(\cdot))
  \end{equation}
  under spatial scaling transformations.
\end{mytheorem}

\begin{myproof}
  In Theorem~\ref{prop-sc-cov-gaussderresnets}, we have already
  proved that the underlying GaussDerResNets without scale pooling are
  scale covariant.
  The result of (i)~max pooling over scales according to
  (\ref{eq:max-pooling-over-scales-cont}) will then be scale invariant
  by construction.
  Regarding (ii)~logsumexp pooling according to
  (\ref{eq:logsumexp-pooling-over-scales-cont}) or
  (iii)~average pooling over scales according to
  (\ref{eq:avg-pooling-over-scales-cont}),
  the change of variables $x' = S \, x$ for $S_x \in \bbbr_+$ with the
  associated change of the scale parameter
  $\sigma_0' = S_x \, \sigma_0$ and $d\sigma_0' = S_x \, d\sigma_0$,
  leaves the form of the integral intact, under reasonable regularity
  on the input image $f$ such that the integrals converge.
  Thereby, it follows that the output from
  either of these scale pooling stages will be scale invariant.
\end{myproof}

\noindent
Again, if the initial scale levels $\sigma_0$
would on the other hand only be discrete 
$\sigma_{i,0} = \lambda^{i-1} \, \sigma_{1,0}$,
for some global intial scale level $\sigma_{1,0} \in \bbbr_+$, then the scale invariance
properties will only hold for scaling factors $S_x$ that are integer
power of the ratio $\lambda$ between adjacent scale levels
$S_x = \lambda^j$ for some $j \in \bbbz$.

A key role of using the spatial max pooling operation in
GaussDerResNets is that it ensures the translation covariance
properties within these networks, as it facilitates the feature 
maps to be shifted proportionally to the shift in the input.

In the ideal continuous case with an infinite image domain, it holds
that the spatial maxima translate with the translation of the input, given an 
image transformation consisting of a scaling followed by a
translation. When applied to discrete image data of finite size,
interference with boundary effects can, however, occur for objects that
are too close to the image boundary.

\subsubsection{Relationships to classical methods for interest point
  detection}
\label{sec-rel-int-pt-det}

Notably, the application of scale pooling mechanisms used here
bears close similarities to how scale
selection mechanisms are constructed in classical computer vision,
see Lindeberg (\citeyear{Lin97-IJCV}, \citeyear{Lin98-IJCV},
\citeyear{Lin12-JMIV}, \citeyear{Lin15-JMIV},
\citeyear{Lin21-EncCompVis}).
Specifically, the combination of spatial max pooling with max pooling
over scales can be seen as a generalisation of the notion of
scale-space extrema in Lindeberg (\citeyear{Lin97-IJCV}, \citeyear{Lin12-JMIV}, 
\citeyear{Lin15-JMIV}) and Lowe (\citeyear{Low04-IJCV}).
Average pooling over scale can also be seen as
a generalisation of the notion of scale selection by weighted
averaging over scales in Lindeberg (\citeyear{Lin12-JMIV},
\citeyear{Lin15-JMIV}).

\subsection{Additional extensions regarding design options}

In addition to the above general architecture for GaussDerNets,
we will also consider a few modifications:

\subsubsection{The use of a zero-order term in the Gaussian derivative
  residual layers}
\label{sec-zero-order-based-net-theory}

In the original formulation of GaussDerNets in Lindeberg
(\citeyear{Lin22-JMIV}), the linear combinations of Gaussian derivative
responses were performed for strictly positive orders of
differentiation, motivated by the theoretically desirable property of
having the responses in the deep network being unaffected
by intensity transformations of the form $f(x) \mapsto f(x) + a$
for any constant $a \in \bbbr$, which model local multiplicative illumination
transformations on a logarithmic intensity scale,
see Lindeberg (\citeyear{Lin13-BICY}) Section~2.3
for a more extensive treatment.
Therefore, the use of a term in the Gaussian derivative layer that
depends on the zero-order Gaussian smoothed image information was
excluded, since such a term would depend on the absolute value
of the image intensity and not just on relative differences in intensity.

Given that the first layer in the GaussDerResNet has been computed 
using only strictly positive orders of spatial differentiation, one could, 
however, conceive including the zero-order terms in the higher layers.%
\footnote{In the deep learning literature, most other networks, that parametrise their 
receptive fields using an N-jet Gaussian derivative basis, typically 
include a zero-order term in all the layers, see, e.g., Jacobsen et 
al.\ (\citeyear{JacGemLouSme16-CVPR}) and Pintea et 
al.\ (\citeyear{PinTomGoeLooGem21-IP}).}
This corresponds to extending the set of Gaussian derivative 
responses for a second-order Gaussian derivative residual layer in 
the higher layers (\ref{eq-gauss-der-layer}) according to 
(\ref{eq-alpha-vals-2nd-order-jet}) to
\begin{equation}
  \label{eq-alpha-vals-2nd-order-jet+zero-order}
  \alpha \in \{ (0, 0), (1, 0), (0, 1), (2, 0), (1, 1), (0, 2)\}.
\end{equation}
and extending the set of Gaussian derivative
responses for a third-order Gaussian derivative residual layer
in the higher layers
according to (\ref{eq-alpha-vals-3rd-order-jet}) to
\begin{align}
  \begin{split}
    \alpha \in
        & \left\{ (0, 0), (1, 0), (0, 1), (2, 0), (1, 1), (0, 2), \dots \right.
  \end{split}\nonumber\\
  \begin{split}
    \label{eq-alpha-vals-3rd-order-jet+zero-order}
      & \hphantom{\left\{\right.} \left. (3, 0), (2, 1), (1, 2), (0, 3) \right\}.
  \end{split}
\end{align}
In Section~\ref{sec-zero-order-term-study}, we will compare these two approaches for
defining the higher layers in the GaussDerResNets.

\subsubsection{Depthwise-separable Gaussian derivative residual networks}
\label{sec-dwsc-based-net-theory}

Work by Chollet (\citeyear{Cho17-CVPR}) and Howard {\em et al.\/}
(\citeyear{HowZhuCheKalWanWeyAndAda17-arXiv}),
with continued work on ConvNeXt by
Liu et al.\ (\citeyear{liu2022-convnext-CVPR}) and Woo {\em et al.\/}
(\citeyear{WooDebHuCheLiuKweXi23-CVPR}),
has demonstrated that the amount of computational work
in deep networks can be substantially decreased, by decoupling the
convolution operations across the layers from the spatial
convolutions.

For the GaussDerResNet architecture described in
Section~\ref{sec-gaussderresnets},
this corresponds to replacing the operation for computing the
next layer from the previous layer for the
simplest form of a residual block of the form
(\ref{eq-simplest-resnet-module-sigma})
with the channels $c_{\text{in}}$ in the input layer
and the channels $c_{\text{out}}$ in the output layer
\begin{multline}
  \label{def-gaussderresnet-layer-k+1-simpler}
  F_{\sigma_0,k+1}^{c_{\text{out}}} (\cdot) = 
  \sum_{c_{\text{in}} \in [1, N_k]}
  {\cal M}_{w_{k+1},\sigma_{k+1}}^{c_{\text{out}},c_{\text{in}}}
  (F^{c_{\text{in}}}_{\sigma_0,k}(\cdot)) = \\
  = \relu\left(  F^{c_{\text{in}}}_{\sigma_0,k}(\cdot) +
    \vphantom{\sum_{c_{\text{in}} \in [1, N_k]}} \right. \\
  \hphantom{=}
  + \left. \sum_{c_{\text{in}} \in [1, N_k]}
    w_{k+1}^{c_{\text{out}},c_{\text{in}}}(\cdot;\; \sigma_{k+1}) * F_{\sigma_0,k}^{c_{\text{in}}}(\cdot) \right)
\end{multline}
to instead applying a two-stage computation of the form
\begin{multline}
  \label{def-gaussderresnet-layer-k+1-simpler-depthsep}
  F_{\sigma_0,k+1}^{c_{\text{out}}} (\cdot) = 
  \sum_{c_{\text{in}} \in [1, N_k]}
  {\cal N}_{\beta_{k+1},w_{k+1},\sigma_{k+1}}^{c_{\text{out}},c_{\text{in}}}
  (F^{c_{\text{in}}}_{\sigma_0,k}(\cdot)) = \\
  = \relu\left( F^{c_{\text{in}}}_{\sigma_0,k}(\cdot) +
    \vphantom{\sum_{c_{\text{in}} \in [1, N_k]}} 
      \sum_{c_{\text{in}} \in [1, N_k]} \beta_{k+1}^{c_{\text{out}},c_{\text{in}}} \,
    h_{\sigma_0,k+1}^{c_{\text{in}}}(\cdot)\right)
\end{multline}
for some set of scalar weights $\beta_{k+1}^{c_{\text{out}},c_{\text{in}}} \in \bbbr$
that are to be learned,
where the intermediate functions
$h_{\sigma_0,k}^{c_{\text{in}}} \colon \bbbr^D \rightarrow \bbbr$
are defined according to
\begin{equation}
  \label{eq-def-intermed-stage-depthwise-sep-conv}
  h_{\sigma_0,k+1}^{c_{\text{in}}}(\cdot) =
 w_{k+1}^{c_{\text{in}}}(\cdot;\; \sigma_{k+1}) * F_{\sigma_0,k}^{c_{\text{in}}}(\cdot)
\end{equation}
for some set of filter weight functions
$w_{k+1}^{_{\text{in}}} \colon \bbbr^D \times \bbbr_+ \rightarrow \bbbr$,
that may be either specified {\em a priori\/} or be learned.%
\footnote{The difference between the weight functions
  $w_{k+1}^{c_{\text{out}},c_{\text{in}}}(\cdot;\; \sigma_{k+1})$
  in (\ref{def-gaussderresnet-layer-k+1-simpler}) and the
  weight functions $w_{k+1}^{c_{\text{in}}}(\cdot;\; \sigma_{k+1})$
  in (\ref{eq-def-intermed-stage-depthwise-sep-conv}) is that the
  latter set of weight functions is not expanded over the indices
  $c_{\text{out}}$ of the output channels.
  Thereby, the effect of the depthwise-separable convolutions,
  according to the combined effect of 
  Equations~(\ref{def-gaussderresnet-layer-k+1-simpler-depthsep}) 
  and~(\ref{eq-gauss-der-layer-in-depthsep}),
  requires a significantly lower amount of computational work than
  the convolutions in the regular convolution layer
  (\ref{def-gaussderresnet-layer-k+1-simpler}) in a combined software
  and hardware environment that is able to take advantage of
  depthwise-separable convolutions in an efficient manner.}
  
In this way, the computational work needed to perform the spatial
convolutions can be reused when computing the different output in
the different channels of the next layer,
thereby substantially decreasing the amount of computations.
In this work, we will specifically consider
linear combinations of Gaussian derivative operators of the form
  (\ref{eq-gauss-der-layer})
  \begin{multline}
    \label{eq-gauss-der-layer-in-depthsep}
    w_{k+1}^{c_{\text{in}}} (\cdot;\; \sigma) 
    = \sum_{\alpha : |\alpha| \leq N}
    m(\alpha) \, C_{\alpha ,k+1}^{c_{\text{in}}} \, \sigma^{|\alpha|} \,   g_{x^{\alpha}}(\cdot;\; \sigma),
  \end{multline}
  where the coefficients $C_{\alpha,k+1}^{c_{\text{in}}}$ are to be learned.

\subsection{Discrete implementation of Gaussian derivative residual networks}
\label{sec-discrete-gauss-implementation}
As the default method for discrete implementation of the Gaussian
derivative operators, we first use separable convolution%
\footnote{For smaller filter sizes, it is generally more
  efficient to perform explicit convolutions, while for larger filters
  a Fourier-based implementation can be computationally more
  efficient. The break point, where a Fourier-based implementation
  becomes more efficient than explicit spatial convolutions, depends on
  the combined efficiency of the software and hardware components that
  implement either the explicit convolutions or the Fourier-based
  computations, typically based on an FFT. In this work, we throughout
  make use of spatial convolutions, since the main focus of this work
  is not primarily computational efficiency but
  to provide a feasibility proof regarding the use of
  scale-invariant Gaussian derivative residual networks for image classification
  and object recognition tasks in settings where there may be
  substantial scaling transformations in the image data, and thereby
  warrant the ability for scale generalisation.
  With regard to Fourier-based implementations, it should, however, be
  noted that the discrete analogue of the Gaussian kernel
  (\ref{eq:discrete-gaussian-kernel}), that we throughout use for the
  spatial smoothing operations, has the following simple explicit expression
  for its Fourier transform
  $\psi_T(\theta) = \sum_{n = -\infty}^{\infty} T(n;\; \sigma) \,
  e^{-in\theta} = e^{\sigma^2(\cos \theta -1)}$
  (Lindeberg \citeyear{Lin93-Dis} Equation~(3.52)),
  which means that it is not necessary to algorithmically compute the
  forward Fourier transform of the smoothing kernel in a Fourier-based
  implementation. A Fourier-based implementation does, however, also
  carry practical issues, such as matching the image size to an
  embedding within an image size that can be efficiently handled by an
  FFT, typically a power of 2. For these reasons,
  we do not explore that direction further here and leave it to future work.}
  along each
dimension with the discrete analogue of the Gaussian kernel
(Lindeberg \citeyear{Lin90-PAMI})
\begin{equation}
  \label{eq:discrete-gaussian-kernel}
  T(n;\; \sigma) = e^{-\sigma^2} I_n(\sigma^2),
\end{equation}
where $I_n$ denotes the modified Bessel functions
of integer order $n$ (Abramowitz and Stegun, \citeyear{AS64}).
Then, we apply central difference
operators of the forms
\begin{align}
  \begin{split}
  \label{eq:central-difference-operators-1}
    \delta_x & = (-1/2, 0, 1/2),
  \end{split} \\
  \begin{split}
  \label{eq:central-difference-operators-2}
    \delta_{xx} & = (1, -2, 1),
  \end{split}
\end{align}
to the spatially smoothed image data, implying that any Gaussian
derivative operator $g_{x^{\alpha}}$ of order
$\alpha = (\alpha_1, \alpha_2, \dots, \alpha_D)$
will be approximated by
\begin{multline}
  g_{x^{\alpha}}(x;\; \sigma)
  = g_{x_1^{\alpha_1} x_2^{\alpha_2} \dots x_D^{\alpha_D}}
    (x;\; \sigma) \approx \\
  \approx \delta_{x_1^{\alpha_1}} \delta_{x_2^{\alpha_2}} \dots \delta_{x_D^{\alpha_D}}
    T_1(x_1;\; \sigma) \, T_2(x_2;\; \sigma) \dots T_D(x_D;\; \sigma),
    \label{eq:disc-gauss+central-diff}
\end{multline}
where $T_i(x_i;\; \sigma)$ denotes the 1-D discrete analogue
of the Gaussian kernel according to
(\ref{eq:discrete-gaussian-kernel})
along the $i$th coordinate direction, and
$\delta_{x_i^{\alpha_i}}$ denotes the composed difference
operator of order $\alpha_i$ along the $i$th coordinate direction
defined by
\begin{equation}
  \delta_{x_i^{\alpha_i}}
  = \left\{
      \begin{array}{ll}
        \delta_{x_i x_i}^{j} & \mbox{if $\alpha_i = 2 j$,} \\
        \delta_{x_i} \delta_{x_i x_i}^{j} & \mbox{if $\alpha_i = 2 j + 1$,} \\        
      \end{array}
    \right.
\end{equation}
for integer $j$, where $\delta_{x_i}$ denotes the
first-order central difference operator according to
(\ref{eq:central-difference-operators-1}) along the
$i$th coordinate direction and $\delta_{x_i x_i}$ denotes the
second-order central difference operator according to
(\ref{eq:central-difference-operators-2}) along the same direction.
        
The motivation for this choice is that this method has in
Lindeberg (\citeyear{Lin24-JMIV-discgaussder,Lin25-FrontSignProc})
and Perzanowski and Lindeberg (\citeyear{PerLin25-JMIV-ScSelGaussDerNets})
been demonstrated to constitute the best choice among a larger
set of discrete approximations in terms of (i)~sampled Gaussian
derivative kernels, (ii)~integrated Gaussian derivative kernels or
(iii)~convolution with the normalised sampled Gaussian kernel complemented
with central differences or (iv)~convolution with the integrated
Gaussian kernel complemented with central differences.
Specifically, by the use of multiple central difference
operators of small support for each order of spatial differentiation, the spatial
smoothing operation can be {\em shared\/} between the receptive fields
for the different orders of spatial differentiation in the Gaussian derivative
residual layers, thereby increasing the computational efficiency.

\section{Datasets} 
\label{sec-rescaled-datasets}

For evaluating the proposed GaussDerResNets experimentally,
we will consider two main types of
use cases:
\begin{itemize}
\item
  computing the accuracy of single-scale-channel
  GaussDerResNets on the regular STL-10 dataset, and
\item
  characterising the scale generalisation performance
  of multi-scale-channel GaussDerResNets on the
  previously existing Rescaled Fashion-MNIST and
  Rescaled CIFAR-10 datasets
  (Perzanowski and Lindeberg
  \citeyear{PerLin25-JMIV-ScSelGaussDerNets}),
  the STIR datasets with traffic signs and aerial images
  (Altstidl et al.\ \citeyear{AltNguSchKoeMutEskZan23-IJCNN})
  as well as on the here newly created Rescaled STL-10 dataset.
\end{itemize}
  
\subsection{Rescaled image datasets}
\label{sec-rescaled-image-datasets}

To assess the ability of the GaussDerResNets to generalise to scales 
not present in the training data, we evaluate them 
on three rescaled image datasets, which contain spatial scaling
variations up to a factor of 4 in the corresponding test datasets. 
In this study, we first investigate the scale generalisation properties 
of networks on the rescaled Fashion-MNIST and the rescaled 
CIFAR-10 datasets, with a complementary evaluations
on the structurally somewhat different STIR datasets with traffic signs
and aerial images. To evaluate the scale generalisation properties 
of networks on relatively higher resolution natural image data, 
we also introduce and investigate the rescaled STL-10 dataset.

The rescaled Fashion-MNIST, CIFAR-10 and STL-10 datasets
have been created using the same
design philosophy, where the images in the training datasets remain 
at the original size, referred to as the size factor 1, while multiple
copies of the test set are created, with each copy rescaled by a single 
size factor. The goal is to train the networks on images of the original size, 
and then evaluate them on identical sets of test datasets, that each 
have the images rescaled by a certain single size factor. 
For each rescaled dataset, the possible test dataset image scaling
factors $S$ that we consider are
\begin{equation}
  \label{eq:dataset-size-factors}
  S \in \{1/2, 2^{-\frac{3}{4}}, 2^{-\frac{2}{4}}, 2^{-\frac{1}{4}},1, 2^{\frac{1}{4}}, 2^{\frac{2}{4}}, 2^{\frac{3}{4}}, 2\},
\end{equation}
which are within the range of interest $[1/2, 2]$, each separated by a
relative scaling factor of $\sqrt[\leftroot{0}\uproot{1}4]{2}$.

The STIR datasets, described in Section~\ref{sec-stir-datasets}, 
which we also use in our experiments, follow a related
but also somewhat different design 
philosophy, where instead of using a single size factor during 
training they use a certain subset of the entire scale range. 

\subsubsection{The rescaled Fashion-MNIST dataset}
\label{sec-rescaled-fashion-dataset}

The rescaled Fashion-MNIST dataset%
\footnote{The rescaled Fashion-MNIST dataset is available for download at Zenodo, 
see Perzanowski and Lindeberg (\citeyear{PerLin-2025-fashionmnist-zenodo}).} 
(Perzanowski and Lindeberg \citeyear{PerLin25-JMIV-ScSelGaussDerNets}), 
was created by extending the Fashion-MNIST dataset (Xiao et al.\ \citeyear{ZiaRasVol-arXiv}) 
with spatial scaling variations in the test dataset, for size factors
according to Equation~(\ref{eq:dataset-size-factors}). The dataset
contains grayscale images of clothing items of size
72$\times$72 pixels and belonging to 10 different classes. 
Each sample depicts a clothing item centred in the frame, with a completely black background. 
The dataset comprises a total of 70~000 images, partitioned into 50~000 training samples, 
10~000 validation samples, and 10~000 test samples.

\subsubsection{The rescaled CIFAR-10 dataset}
\label{sec-rescaled-cifar-dataset}

The rescaled CIFAR-10 dataset%
\footnote{The rescaled CIFAR-10 dataset is available for download at Zenodo, 
  see Perzanowski and Lindeberg (\citeyear{PerLin-2025-cifar10-zenodo}).} 
(Perzanowski and Lindeberg \citeyear{PerLin25-JMIV-ScSelGaussDerNets})
was created by extending the CIFAR-10 dataset (Krizhevsky and Hinton \citeyear{KriHin09-CIFAR}) with 
spatial scaling variations in the test dataset for size factors given by Equation~(\ref{eq:dataset-size-factors}). 
The dataset contains RGB images of real-world photographs of animals
or vehicles of size 64$\times$64 pixels and belonging 
to 10 different classes, with the object in each sample being centred
in the frame. In order to have all the 
training and the test images of the same size,
mirror extension was used to extend the images to size 
64$\times$64 for scaling factors less than 2. The dataset comprises a total of 60~000 images, partitioned 
into 40~000 training samples, 10~000 validation samples, and 10~000 test samples.

\begin{figure*}[htbp]
   \centering 
    \begin{tabular}{ccccccccccc}
      \includegraphics[width=0.083\textwidth]{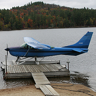} &
      \hspace{-2.7mm}
      \includegraphics[width=0.083\textwidth]{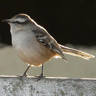} &
      \hspace{-2.7mm}
      \includegraphics[width=0.083\textwidth]{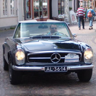} &
      \hspace{-2.7mm}
      \includegraphics[width=0.083\textwidth]{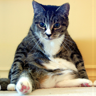} &
      \hspace{-2.7mm}
      \includegraphics[width=0.083\textwidth]{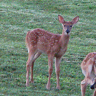} &
      \hspace{-2.7mm}
      \includegraphics[width=0.083\textwidth]{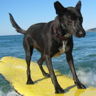} &
      \hspace{-2.7mm}
      \includegraphics[width=0.083\textwidth]{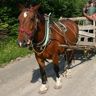} &
      \hspace{-2.7mm}
      \includegraphics[width=0.083\textwidth]{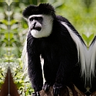} &
      \hspace{-2.7mm}
      \includegraphics[width=0.083\textwidth]{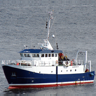} &
      \hspace{-2.7mm}
      \includegraphics[width=0.083\textwidth]{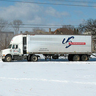} &
      \\
    \end{tabular} 
    \smallskip
    Samples from each class, from the training set (size factor 1)

  \vspace{1mm}
  
    \begin{tabular}{ccccccccccc}
      \includegraphics[width=0.083\textwidth]{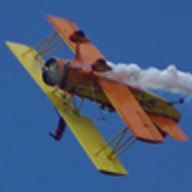} &
      \hspace{-2.7mm}
      \includegraphics[width=0.083\textwidth]{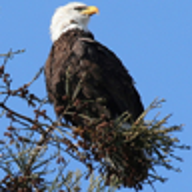} &
      \hspace{-2.7mm}
      \includegraphics[width=0.083\textwidth]{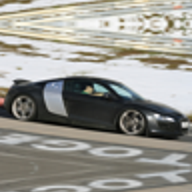} &
      \hspace{-2.7mm}
      \includegraphics[width=0.083\textwidth]{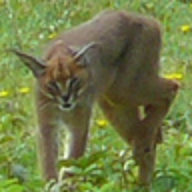} &
      \hspace{-2.7mm}
      \includegraphics[width=0.083\textwidth]{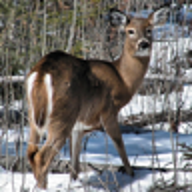} &
      \hspace{-2.7mm}
      \includegraphics[width=0.083\textwidth]{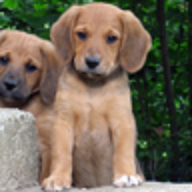} &
      \hspace{-2.7mm}
      \includegraphics[width=0.083\textwidth]{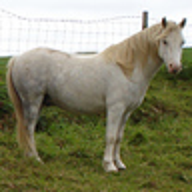} &
      \hspace{-2.7mm}
      \includegraphics[width=0.083\textwidth]{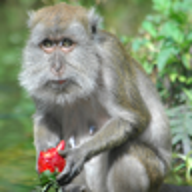} &
      \hspace{-2.7mm}
      \includegraphics[width=0.083\textwidth]{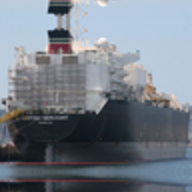} &
      \hspace{-2.7mm}
      \includegraphics[width=0.083\textwidth]{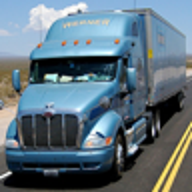} &
      \\
    \end{tabular} 
    \smallskip
    Samples from each class, from the test set at size factor 2
    
  \vspace{1mm}
  
    \begin{tabular}{cccccccccc}
      \includegraphics[width=0.083\textwidth]{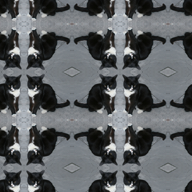} &
      \hspace{-2.7mm}
      \includegraphics[width=0.083\textwidth]{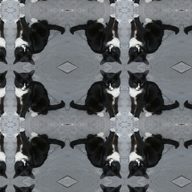} &
      \hspace{-2.7mm}
      \includegraphics[width=0.083\textwidth]{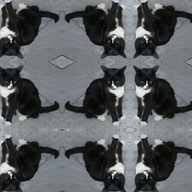} &
      \hspace{-2.7mm}
      \includegraphics[width=0.083\textwidth]{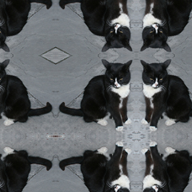} &
      \hspace{-2.7mm}
      \includegraphics[width=0.083\textwidth]{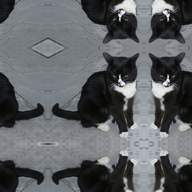} &
      \hspace{-2.7mm}
      \includegraphics[width=0.083\textwidth]{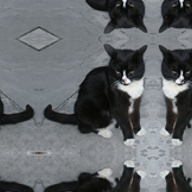} &
      \hspace{-2.7mm}
      \includegraphics[width=0.083\textwidth]{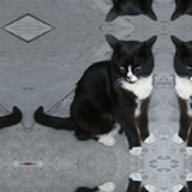} &
      \hspace{-2.7mm}
      \includegraphics[width=0.083\textwidth]{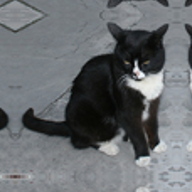} &
      \hspace{-2.7mm}
      \includegraphics[width=0.083\textwidth]{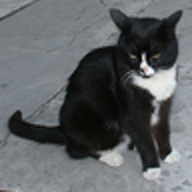} &
      \\
    \end{tabular}
    
    A single test sample for all scaling factors between 1/2 and 2
     
  \caption{A collage made up of a representative subset of images from the 
  rescaled STL-10 dataset. The {\bf (top row)} contains samples from the training
  set (size factor 1), each with $96 \times 96$ pixels (with the pixel size adapted to 
  fit the collage), with one example per class, where the 10 classes are: ``airplane'', 
  ``bird'', ``car'', ``cat'', ``deer'', ``dog'', ``horse'', ``monkey'', ``ship'' and ``truck''. 
  In a similar manner, the {\bf (middle row)} depicts samples from the size 
  factor 2 test set, each with $192 \times 192$ pixels. The {\bf (bottom row)} depicts 
  a single sample from the ``cat'' class in the test set, for each possible scaling 
  factor between 1/2 and 2, with the scaling factor increasing from left to right, 
  and the images at all the size factors extended (if needed) using mirroring to the 
  final image size of $192 \times 192$ pixels. Note that this means
  that the size factor 1 
  test set has mirror extensions and a larger image size compared to the training set.}
  \label{fig-stl10_collage}
\end{figure*}

\subsubsection{The STIR datasets with traffic signs and aerial images}
\label{sec-stir-datasets}
We also evaluate the scale-invariance properties of GaussDerResNets 
on the traffic sign and aerial image subsets of the existing Scaled and 
Translated Image Recognition (STIR) dataset, which was proposed in 
Altstidl et al. (\citeyear{AltNguSchKoeMutEskZan23-IJCNN}) for the 
task of evaluating scale generalisation properties of networks.

These datasets are constructed by extracting objects of given type 
within bounding boxes, from RGB images in existing datasets, and 
downscaling copies of them to one per each of 48 scales in the 
range $l \in [17,64]$. For each such scale, the objects are randomly 
positioned within 64$\times$64 images, together with their corresponding 
background. The traffic sign subset is derived from the Mapillary Traffic 
Sign Dataset by Ertler et al. (\citeyear{ertler2020-ECCV}), while the aerial 
image subset is derived from the Dataset for Object Detection in Aerial 
Images by Xia et al. (\citeyear{xia2018-CVPR}).

The traffic sign subdivision of the STIR dataset consist of 16 classes, 
while the arial subdivision contains 9 classes. Both datasets are small, 
using 25 instances per class for each of the training, validation and test 
sets, with the aforementioned 48 scale-specific copies existing, each 
containing objects at a single scale.

\subsubsection{The rescaled STL-10 dataset}
\label{sec-rescaled-stl-dataset}

We introduce a modified version of the labelled subset of the 
STL-10 dataset (Coates et al.\ \citeyear{CoatNgLee11-PMLR}), 
which is a subset of the ImageNet dataset, extended with scaling 
variations, to allow for evaluating the scale generalisation 
properties of the model in more challenging natural settings, and
at a much higher pixel resolution than in the CIFAR-10 dataset.

The original labelled subset of the STL-10 dataset, with 5~000 
training samples and 8~000 test samples, consists of real-world 
RGB images of vehicles and animals of size 96$\times$96 pixels from the 
10 distinct classes: ``airplane'', ``bird'', ``car'', ``cat'', ``deer'', 
``dog'', ``horse'', ``monkey'', ``ship'' and ``truck'', all of which, 
with the exception of the class ``monkey'', are also present 
in the CIFAR-10 dataset. A semi-automated cleaning stage was 
performed prior to the rescaling of this dataset, to remove images 
for which mirroring at the image boundaries would have substantial 
negative effects. This cleaning stage mainly involved removing 
the artificial constant padding in some of the images. In addition, 
most blurry, poorly lit, severely off-centre or substantially obscured 
images were manually removed, to improve the quality of the dataset, 
resulting in about 10\% of the original dataset being removed.

Just as for the rescaled Fashion-MNIST and CIFAR-10 datasets, 
bicubic interpolation with default anti-aliasing in Matlab was used 
for generating the scaled images, with the considered image size 
factors given by Equation~(\ref{eq:dataset-size-factors}). The 
training dataset still uses the original size factor 1, as well as the original 
image size of 96$\times$96 pixels to keep the training-time memory 
usage low, with mirror extension being used, to achieve the desired 
image size if de-padding was applied during the cleaning stage. 
Meanwhile, copies of the test dataset were created, each dedicated 
to one of the scaling factors $S_x$ in Equation~(\ref{eq:dataset-size-factors}). 
The images in each test set were extended with mirroring at the 
boundaries to an image size of 192$\times$192 pixels, meaning that 
the test sets use a larger image size compared to the training set, 
even at the size factor 1. See Figure~\ref{fig-stl10_collage} 
for an illustrative overview of example images from this dataset. 

The dataset comprises a total of 11~600 images, partitioned 
into 7~000 training samples, 1~000 validation samples, and 
3~600 test samples. The small size of this dataset also enables 
it to be used to evaluate the ability of networks to generalise 
from limited training data.

Note that the input images used for generating the rescaled 
STL-10 dataset also comprise a certain variability in the 
sizes of the objects in the image domain, due to factors such 
as the viewing distance or the direction of the camera. Furthermore, in 
contrast to the rescaled CIFAR-10 dataset, the objects in the images 
are not always centred, with a greater amount of background 
typically included within the frame, as the images are less cropped. 
Thus, the scale selection properties may be less clear-cut for 
this dataset than for the rescaled Fashion-MNIST or CIFAR-10 datasets.

\begin{table*}[hbtp]
  \centering
  \begin{tabular}{lcccc}
       \hline
       \textbf{Model Parameters} & \textbf{Symbol} & \textbf{rescaled Fashion-MNIST} & \textbf{rescaled CIFAR-10} & \textbf{rescaled STL-10}\\ 
       \hline
       Number of scale channels & $N$ & 6 & 6  & 6\\ 
       Initial scale value range & [$\sigma_{i,0}$] & [$\sqrt{2}^{-3},2$] & [$\sqrt{2}^{-3},2$] & [$\sqrt{2}^{-3},2$] \\ 
       Relative scale ratio & $r$ & 1.13 & 1.2 & 1.32 \\ 
       Relative truncation error bound & $\epsilon$ & 0.01 & 0.005 & 0.005 \\ 
       Number of layers & $K$ & 18 & 18 & 18 \\ 
       Intermediate feature channels & $c$ & 1-48-24-$\overline{32}$-$\overline{48}$-$\overline{64}$-128-10 & 3-$\overline{48}$-$\overline{\overline{64}}$-$\overline{\overline{192}}$-256-10 &  3-$\overline{48}$-$\overline{64}$-$\overline{128}$-$\overline{192}$-256-10 \\ 
       Spatial selection method & - & \text{central pixel} & \text{central pixel} & \text{spatial max pooling}\\ 
       Zeroth-order term & - & \text{without} & \text{without} & \text{with}\\ 
       Number of network parameters & - & 295k & 1.78M & 2.07M \\ 
       Input image size & - & 1$\times$72$\times$72 & 3$\times$64$\times$64 & 3$\times$96$\times$96\\ \hline
  \end{tabular}
  \caption{Key parameter values and mechanisms employed in our
    experiments with the multi-scale-channel GaussDerResNets,
    for the three rescaled datasets.
    In the row specifying the number of intermediate feature channels $c$, 
    a line over a number means that two blocks in a row use the same value, 
    while two lines means that three blocks in a row use the same value.}
  \label{tab:model-parameters}
\end{table*}

\begin{table*}[hbtp]
  \centering
  \begin{tabular}{lccc}
       \hline
       \textbf{Training Parameters} & \textbf{rescaled Fashion-MNIST} & \textbf{rescaled CIFAR-10} & \textbf{rescaled STL-10}\\ 
       \hline
       Number of training iterations (epochs) & 30000 (32) & 100000 (90) & 56000 (119) \\
       Batch size & 64 & 45 & 17 \\ 
       Initial learning rate & 0.01 & 0.01 & 0.01\\ 
       Number of warm-up epochs & 0 & 8 & 8 \\ 
       Weight decay & 0.025 & 0.025 & 0.025 \\ 
       Scale channel dropout factor & 0.2 & 0.3 & 0.3 \\ 
       Random cropping & \text{none} & \text{none} & \text{yes (padding=24)} \\ \hline
  \end{tabular}
  \caption{Training configurations for multi-scale-channel GaussDerResNet 
  employed in our experiments, for the three rescaled datasets.}
  \label{tab:training-parameters}
\end{table*}

\section{Network architectures and training configurations}
\label{sec-training-procedure}

This section describes the network architecture and the training
parameters used to evaluate the GaussDerResNets and their variants
on the rescaled datasets introduced in Section~\ref{sec-rescaled-image-datasets}. 
For each dataset, we used the predefined data splits, employing 
the validation sets to select the hyperparameter settings and the training schedule.

\subsection{Network architecture design}
\label{sec-network-arch-design}

The architectural specifications for the multi-scale-channel 
Gaussian derivative residual networks (GaussDerResNets) 
were chosen through model selection, to achieve strong 
scale generalisation properties and performance on the 
considered rescaled datasets, and are summarised in 
Table~\ref{tab:model-parameters}.%
\footnote{For training on the traffic sign and aerial image subsets of
  the STIR dataset, we 
    used similar network settings as for the rescaled CIFAR-10 dataset, but 
    with the number of intermediate feature channels set to 
    3-48-24-$\overline{32}$-$\overline{48}$-$\overline{64}$-128-numclasses, 
    with 16~classes for the traffic sign subset and 9~classes for the aerial image subset, 
    and using spatial max pooling for spatial selection. The training settings were 
    also in part similar, but with the batch size set to 20 and the
    number of training epochs 
    set to 32 with 5 warm-up epochs.
    Label smoothing was used by default during training.
    In the row specifying the number of intermediate feature channels $c$, 
    a line over a number means that two blocks in a row use the same value, 
    while two lines means that three blocks in a row use the same value.}
  \label{tab:model-parameters.}
The design of the GaussDerResNets
is formally defined in
Sections~\ref{def-single-sc-chan-network}--\ref{sec-gaussderresnet-arch}.

The initial scale values $\sigma_{i,0}$ are the main network 
parameters that determine the range of size factors, not seen 
during training, that the network can generalise to. For the 
rescaled datasets used in this work, the possible scales are 
given by Equation~(\ref{eq:dataset-size-factors}), which is 
what motivates the choice of $\sigma_{i,0} \in \{1/(2\sqrt{2}), 1/2, 1/\sqrt{2}, 1$, $\sqrt{2}, 2\}$, 
to cover that range with an adequate spacing by a factor of $\sqrt{2}$, 
while also including an additional boundary scale channel for 
the smaller size factors, based on $\sigma_{i,0} = 1/(2\sqrt{2})$. 
The value of the relative scale factor $r$ is chosen based on 
the original image size, with higher values picked for
larger image sizes.

The number of feature channels $c$ used by each network, 
provided in Table~\ref{tab:model-parameters}, specifies the 
values for each effective layer.
The first and the last values represent the input colour channels and the
number of output classes, respectively, while the intermediate 
values represent the input feature channel number in the 
intermediate residual blocks, with one value per block. In 
Table~\ref{tab:model-parameters}, when two residual blocks 
in a row use the same value, we denote that with a line over 
the number, and three residual blocks in a row with two lines 
over the number.

Before the permutation-invariant scale pooling stage, the 
GaussDerResNets and their variants explored in our experiments 
are configured to use a spatial selection mechanism that is based 
on either (i)~central pixel extraction, or (ii)~spatial max pooling. The
chosen design depends on whether the objects within the frame of the
images for a given dataset are centered or not. The possible approaches 
to permutation-invariant pooling over scales that we consider are 
either max pooling, logsumexp pooling, or average pooling.

\subsection{Training protocol}
\label{sec-training-protocol}

To train the GaussDerResNets and their variants on the rescaled datasets, 
we used the AdamW optimiser with the cross-entropy loss, with all the training 
settings summarised in Table~\ref{tab:training-parameters}. For all the datasets, 
the network weights were initialised using a uniform He
initialisation. The batch size 
was selected per dataset based on the network size, the input
resolution, and the single-GPU
memory limits, to maximise the model capacity. We trained the largest feasible 
networks, which in the case of the rescaled STL-10 dataset required a small 
batch size due to the memory constraints.

The number of epochs was also dataset 
dependent, because the more complex datasets require longer training. 
Regarding the learning rate, a cosine decay schedule was used, with the 
learning rate decreasing from the initial value to $10^{-5}$, over the course 
of the training epochs. Additionally, for the rescaled CIFAR-10 and STL-10 
datasets, the learning rate was warmed up from 10\% of its initial value to 
the full initial value over the first 8 epochs.

Since the selected training configurations have been found during the
hyperparameter search, the training and the validation datasets 
were combined to maximize the available training data in a final
training stage.

To enhance the generalisation ability of the networks and 
to prevent overfitting, several data augmentation and 
regularisation techniques were employed during training. 
Random horizontal flipping with a probability of 0.5 was 
used to enhance the sample diversity in the training set, 
while for the RGB datasets random colour jitter configured 
with the parameters (brightness=0.2, contrast=0.2, 
saturation=0.2, hue=0.05) was also used. Additionally, 
random cropping to the original size after padding was used 
when training on the rescaled STL-10 dataset.

We also used scale-channel dropout, see Perzanowski and Lindeberg (\citeyear{PerLin25-JMIV-ScSelGaussDerNets}), 
which involves applying dropout across the outputs from the scale channels before the permutation-invariant scale pooling stage, 
which enables the network to make better use of the information across
the different scales during training.

Additionally, we used label smoothing, unless specified otherwise,
which is a regularisation technique that prevents overconfidence during 
classification, by replacing hard one-hot target labels with softened (smoothed) probability distributions during training. 
This improves the generalisation and can help to reduce overfitting,
see M\"uller et. al (\citeyear{MuKorHin19-NIPS}), which is 
important for achieving good scale generalisation. A smoothing parameter of $0.1$ was used in all our experiments.

\section{Experiments}
\label{sec-experiments}

In this section, we will evaluate the properties of the
GaussDerResNet networks:
\begin{itemize}
\item
  regarding the accuracy of {\em single-scale-channel GaussDerResNets\/} on the
  regular STL-10 dataset in Section~\ref{sec-single-scale-exp}, and
\item
  regarding the scale generalisation and the scale selection
  properties for {\em multi-scale-channel GaussDerResNets\/} applied on
  (i)~the Rescaled Fashion-MNIST dataset,
  (ii)~the Rescaled CIFAR-10 dataset,
  (iii)~the STIR datasets with traffic signs and aerial images, and
  (iv)~the Rescaled STL-10 dataset,
  in Section~\ref{sec-scale-gen-experiments}.
\end{itemize}

\subsection{Experiments with single-scale-channel networks on the
  regular STL-10 dataset}
\label{sec-single-scale-exp}

The newly proposed Gaussian derivative residual network (GaussDerResNet) 
architecture is provably scale covariant and scale invariant, but also has 
greatly improved generalisation capabilities compared to the previously studied 
Gaussian derivative networks. To demonstrate that the single-scale-channel 
GaussDerResNet can achieve competitive performance on image 
classification tasks, we trained and evaluated the network on the labelled subset 
of the regular STL-10 dataset, and compared the results to several other well 
performing networks on this dataset. 

The training was for the most part performed according to the procedure 
outlined in Section~\ref{sec-training-procedure} for multi-scale-channel 
networks applied on the rescaled STL-10 dataset, with the main difference 
being that a single-scale-channel of the network based on $\sigma_{0}=0.5$ 
was used, and the relative scale ratio was set to $r=1.5$. With the exception 
of the first layer, the network layers included a zero-order term, and spatial 
max pooling was used after the final layer as the spatial selection 
mechanism. Additional regularisation was performed during training 
by the use of random cropping and cutout. 

In this comparison study, we considered two different variants of the
GaussDerResNets, one trained without label smoothing and the other 
trained with label smoothing. The test accuracy of these 
GaussDerResNets on the regular STL-10 dataset was compared to 
three strong baseline models, the Wide Residual Network (WideResNet) 
by Zagoruyko and Komodakis (\citeyear{ZagKom16-BMVC}), the 
Harmonic Wide Residual Network (Harm-WideResNet) by Ulicny et 
al.\ (\citeyear{UliKryDah19-EUSIPCO}), and the SESN-B network 
by Sosnovik et al.\ (\citeyear{SosSzmSme20-ICLR}).

\begin{table}[htbp]
  \centering
  \begin{tabular}{lcc}
    \hline
    \textbf{Network} & \textbf{\#Params} & \textbf{Test accuracy} \\ \hline
    GaussDerResNet (no LS) & 2.1M & $88.05\%$ \\
    WideResNet & 11M & $88.52\%$ \\
    GaussDerResNet (with LS) & 2.1M & $89.36\%$ \\
    Harm-WideResNet & 11M & $90.45\%$ \\
    SESN-B & 11M & $91.49\%$ \\ \hline
  \end{tabular}
  \caption{Comparison of the test accuracies and the numbers of parameters of networks trained on the 
    original STL-10 dataset, showing the performance obtained by (i)~the single-scale-channel 
    GaussDerResNet with zero-order terms trained without label
    smoothing (abbreviated as ``no LS''), (ii)~the single-scale-channel 
    GaussDerResNet with zero-order terms
    trained with label smoothing (abbreviated ``with LS''),
    (iii)~the Wide Residual (WideResNet) network 
    (Zagoruyko and Komodakis \citeyear{ZagKom16-BMVC}), (iv)~the Harmonic Wide Residual 
    (Harm WideResNet) network (Ulicny et al.\ \citeyear{UliKryDah19-EUSIPCO}), and (v)~the 
    scale-equivariant steerable (SESN-B) network (Sosnovik et al.\ \citeyear{SosSzmSme20-ICLR}).
    We can see that GaussDerResNets can achieve accuracies comparable to 
    the other well performing networks, though marginally lower to the best performing one, while 
    making use of far fewer network parameters.}
  \label{tab:original-stl10-performance}
\end{table}

The results from this comparative study are presented in 
Table~\ref{tab:original-stl10-performance}.
As can be seen from the table, the GaussDerResNets
using label smoothing during the training process achieve very good 
performance on the original STL-10 dataset, demonstrating near-parity
with several competing methods, and showing only a 2\% difference in 
accuracy relative to SESN-B, which is the best performing model. 
Notably, however, the GaussDerResNets achieve this level of classification 
accuracy while using five times less network parameters.

This demonstrates that the single-scale-channel GaussDerResNets are
capable of handling classification on complex image datasets at a level
comparable to other well performing networks, while additionally 
obeying scale covariance properties and a network design that 
makes it possible to extend the architecture to become scale invariant, 
by adding additional scale channels at different scale levels, with shared weights.

\subsection{Scale generalisation experiments with multi-scale-channel
  GaussDerResNets on rescaled datasets}
\label{sec-scale-gen-experiments}

In this section, we will compare the scale generalisation abilities of
the multi-scale-channel Gaussian derivative residual networks (GaussDerResNets) 
to the Gaussian derivative networks (GaussDerNets), where the latter 
were explored in our previous work in Perzanowski and 
Lindeberg (\citeyear{PerLin25-JMIV-ScSelGaussDerNets}) on the
rescaled Fashion-MNIST and CIFAR-10 datasets,
as well as on the STIR datasets with traffic signs and aerial images
(Altstidl et al.\ \citeyear{AltNguSchKoeMutEskZan23-IJCNN}).
We will also demonstrate the scale generalisation properties
of the GaussDerResNets on the new rescaled version of the
STL-10 dataset.

Using multi-scale-channel networks is key to 
achieving good scale generalisation, as regular not scale-covariant 
deep networks typically fail on this task, see Appendix~B.

Additionally, we will compare the influence of basing the networks on 
different permutation-invariant scale pooling methods on the scale 
generalisation performance. All these experiments are characterised 
by the fact that the networks are trained at only a single scale (object size), 
while the evaluation is performed on a set of copies of the test dataset, 
with each copy rescaled by a single size factor.

\subsubsection{Scale generalisation experiments with multi-scale-channel
  networks on the rescaled Fashion-MNIST dataset}
\label{sec-fashion-experiments}

For the rescaled Fashion-MNIST dataset, the multi-scale-channel 
GaussDerResNets were trained following the training configuration 
outlined in Section~\ref{sec-training-procedure}, using scale channels 
with $\sigma_{i,0} \in \{1/(2\sqrt{2}), 1/2, 1/\sqrt{2}, 1, \sqrt{2}, 2\}$. 
As the objects are centred within the image frame for this
dataset, we based all the networks on central pixel extraction as the 
spatial selection method.

\begin{figure}[ht]
  \begin{center}
    \textit{\quad\quad Scale generalisation on rescaled Fashion-MNIST}
    \includegraphics[width=0.48\textwidth]{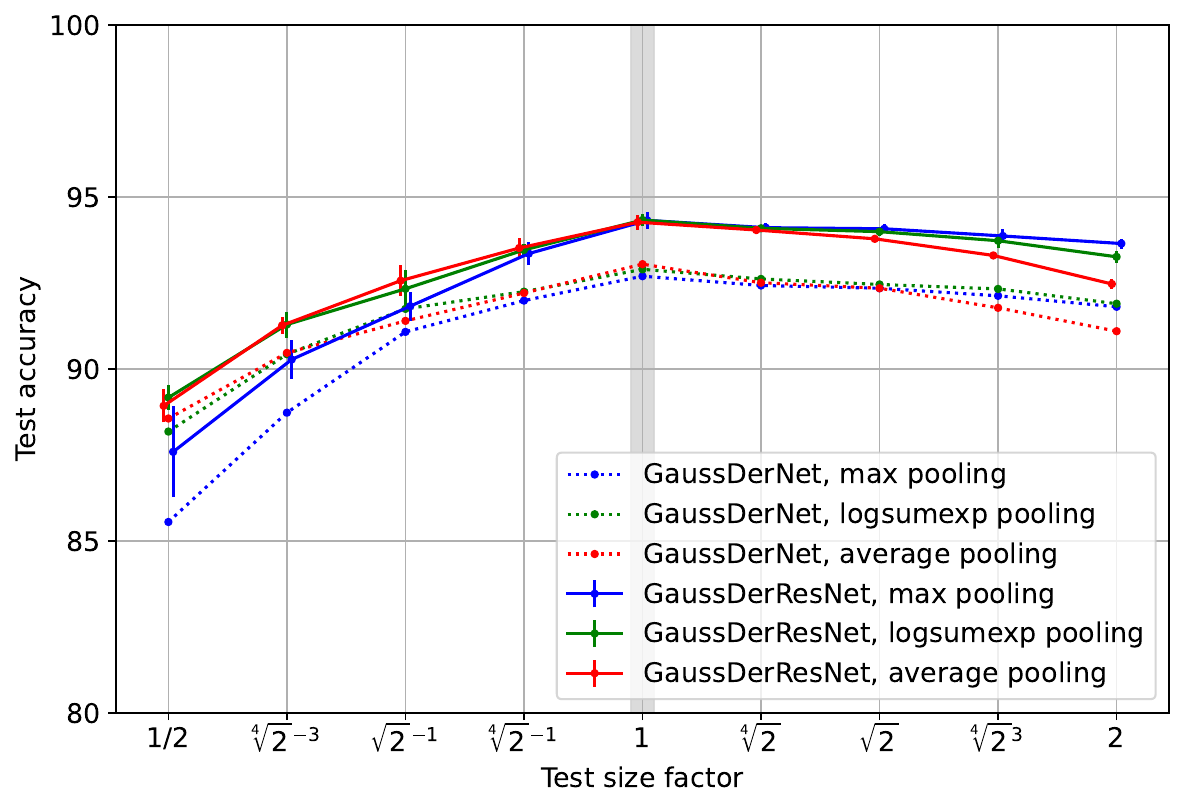}
  \end{center}
  \caption{Comparison between the scale generalisation curves
    for the rescaled Fashion-MNIST dataset, showing
    both the mean and the error bars for the standard deviation over 5~runs
    for multi-scale-channel GaussDerResNets based on either max, logsumexp or 
    average pooling over scale, also with comparisons to single run
    of multi-scale-channel GaussDerNets based on max, logsumexp or
    average pooling over scale (reproduced from 
    Perzanowski and Lindeberg (\citeyear{PerLin25-JMIV-ScSelGaussDerNets})).
    For the test data with the same size factor 1 as the training
    data, represented by the shaded grey region, the GaussDerResNets
    achieved a mean accuracy of 94.3\% (with standard deviation $\pm$ 0.2\%), 
    which is $\sim$1.5~ppt higher compared to the GaussDerNets. 
    From the graphs, we can see that the GaussDerResNets have overall 
    better performance compared to the GaussDerNets, while retaining 
    very good scale generalisation, in agreement with the theoretical analysis.}
  \label{fig-fashion-gaussresnet-scale-gen}
\end{figure}

The results from this experiment averaged across 5 different runs are presented 
in Figure~\ref{fig-fashion-gaussresnet-scale-gen}, with the networks
trained on the training image data for the original size factor 1 and 
evaluated on the rescaled copies of the test dataset for all the 
possible size factors ranging from 1/2 to 2, and using multi-scale-channel 
networks based on either max, logsumexp or average pooling over scales. 
The results in the graph show that the performance of the GaussDerResNets 
is mostly flat over scales, with almost no variability between the runs, demonstrating 
that the scale covariance properties derived in Section~\ref{sec-trans-under-scaling} 
consistently hold in practice.

We can also see that the GaussDerResNets achieve 
slightly flatter scale generalisation curves compared to the GaussDerNets 
for the size factors $> 1$, and becoming almost completely flat for the 
max pooling based network, and also outperforming the GaussDerNets 
by $\sim$1.5~ppt. The minor drop in the performance and slightly larger error 
bars for the GaussDerResNets present at the smaller size factors, resulting 
in a similar level of performance as for the GaussDerNets at the finest 
scales, is likely due to discretisation effects and loss of information, 
caused by the fact that the original dataset consists of small objects, 
which are challenging to distinguish at the resulting lower effective
resolution.%
\footnote{For the rescaled Fashion-MNIST dataset, we leave the analysis of the 
scale selection properties of the GaussDerResNets to Appendix~D, 
where the effect of the use of label smoothing on the scale selection 
properties is also investigated.}

\begin{figure}[ht]
  \begin{center}
    \textit{\quad\quad Scale generalisation on rescaled CIFAR-10}
    \includegraphics[width=0.48\textwidth]{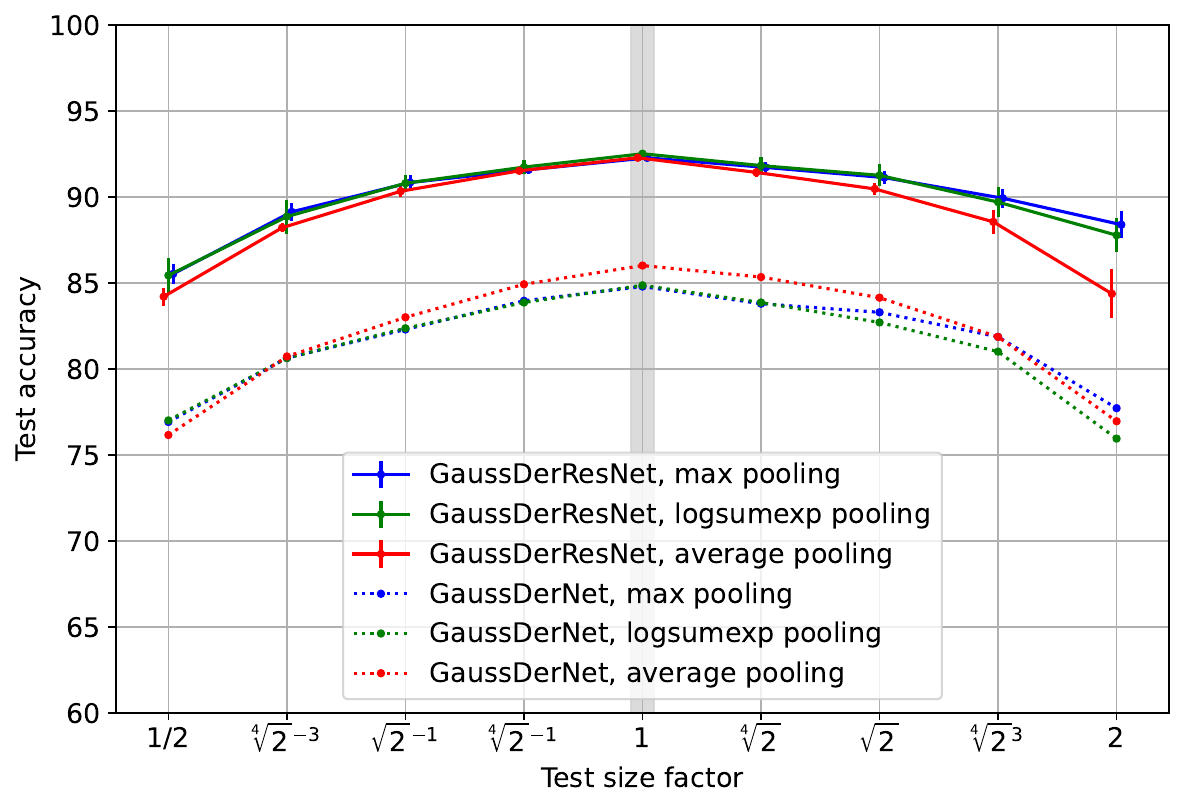}
  \end{center}
  \caption{Comparison between the scale generalisation curves
    for the rescaled CIFAR-10 dataset, showing
    both the mean accuracy and the error bars for the standard deviation over 5~runs
    for multi-scale-channel GaussDerResNets based on either max, logsumexp or 
    average pooling over scale, also with comparisons to single run
    of multi-scale-channel GaussDerNets based on max, logsumexp or
    average pooling over scale (reproduced from 
    Perzanowski and Lindeberg (\citeyear{PerLin25-JMIV-ScSelGaussDerNets})).
    Each network was trained on training data for the size factor 1,
    and then evaluated on test data covering all the size factors
    between 1/2 and 2, with each copy of the test set only 
    containing images at a single given scale. For the test data with 
    the same size factor 1 as the training data, represented by the 
    shaded grey region, the GaussDerResNets achieved a mean accuracy 
    of 92.2-92.5\% (with standard deviation $\pm$ 0.15-0.21\%), 
    which is $\sim$7~ppt higher compared to the GaussDerNets. 
    From the graphs, we can see that the scale generalisation properties 
    of the GaussDerResNets is significantly better compared to the GaussDerNets.}
  \label{fig-cifar10-gaussresnet-scale-gen}
\end{figure}

\subsubsection{Scale generalisation and scale selection
  experiments with multi-scale-channel
  GaussDerResNets on the rescaled CIFAR-10 dataset}
\label{sec-cifar10-experiments}

Figure~\ref{fig-cifar10-gaussresnet-scale-gen} shows 
corresponding scale generalisation experiments for the rescaled
CIFAR-10 dataset. As can be seen from the results, for the 
GaussDerResNets the test accuracy obtained on the training scale, 
meaning the size factor 1, shows substantial relative improvement 
in performance of about 7~ppt compared to the GaussDerNets. 
The scale generalisation curves obtained for the GaussDerResNets 
are significantly flatter compared to the GaussDerNets, 
especially for the max pooling and logsumexp pooling based methods,
showing up to $\sim$12~ppt relative improvement in accuracy for the 
larger size factors, and up to $\sim$8.5~ppt relative improvement in 
accuracy on the smaller size factors, with little variability between
the runs.

From the scale-covariant property of the GaussDerResNets,
it would be expected that the scale values associated with
the scale channels selected by the scale
selection mechanism would increase proportionally to scaling
transformations of the input data. This property can be experimentally 
verified by inspecting so-called scale selection histograms, constructed 
by accumulating information about the relative contribution of each 
scale channel to the classification of test samples, for each spatial 
scaling factor. 

Figure~\ref{fig-cifar10-gaussresnet-scale-selection-main} depicts 
the scale selection histograms averaged over 5 runs for the networks 
based on max, logsumexp and average pooling over scales for the 
experiments on the rescaled CIFAR-10 dataset, trained without 
label smoothing. As can be seen from the results, there is a clear and distinct linear 
trend for the networks based on max pooling and logsumexp pooling, 
a property similar to classical methods for automatic scale selection. 
For the network based on average pooling over scales, the trend is 
also linear, although less localised and more spread out. This 
behaviour is expected, as average pooling combines information 
across all the scale channels, thus enabling more scale channels to 
contribute to the classification.

Finally, we can note that the GaussDerResNets show roughly similar 
performance and parameter efficiency on the size factor 1 of the 
rescaled CIFAR-10 dataset%
\footnote{Training a single-scale-channel 
GaussDerResNet with 10M network parameters, using appropriate
settings, leads to about 94.5~\% accuracy on the regular CIFAR-10 dataset.}, 
compared to other networks based on Gaussian derivative operators designed 
for handling image data at different scales, see for example Pintea et 
al.\ (\citeyear{PinTomGoeLooGem21-IP}) for corresponding results on the 
regular CIFAR-10 dataset, while having the additional property 
of being scale invariant, and achieving very flat scale generalisation curves. 

\begin{figure*}[hbt]
  \begin{center}
    \begin{subfigure}{0.99\textwidth}      
      \begin{tabular}{ccc}
	\hspace{3mm} \textit{Training with small object sizes}
          & \hspace{3mm} \textit{Training with medium object sizes}
          & \hspace{3mm} \textit{Training with large object sizes} \\
          \includegraphics[width=0.32\textwidth]{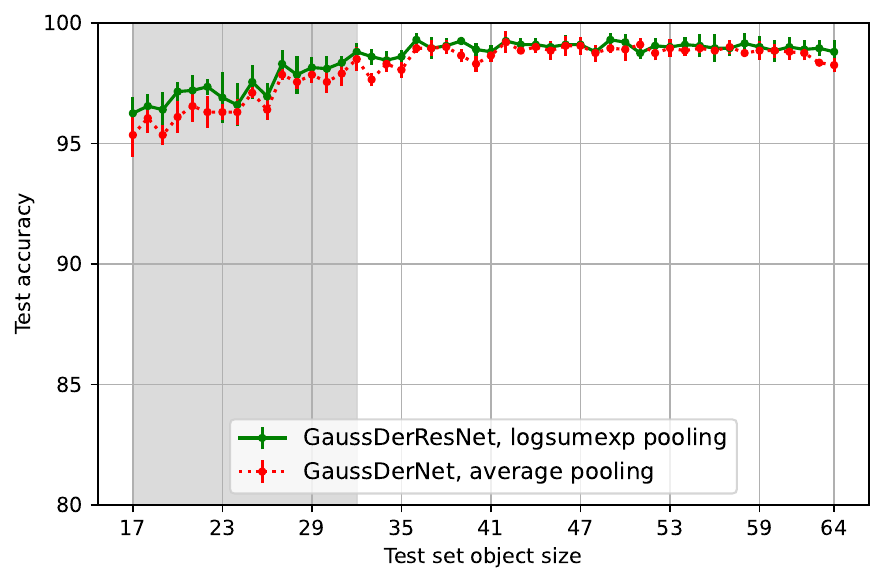}
          & \includegraphics[width=0.32\textwidth]{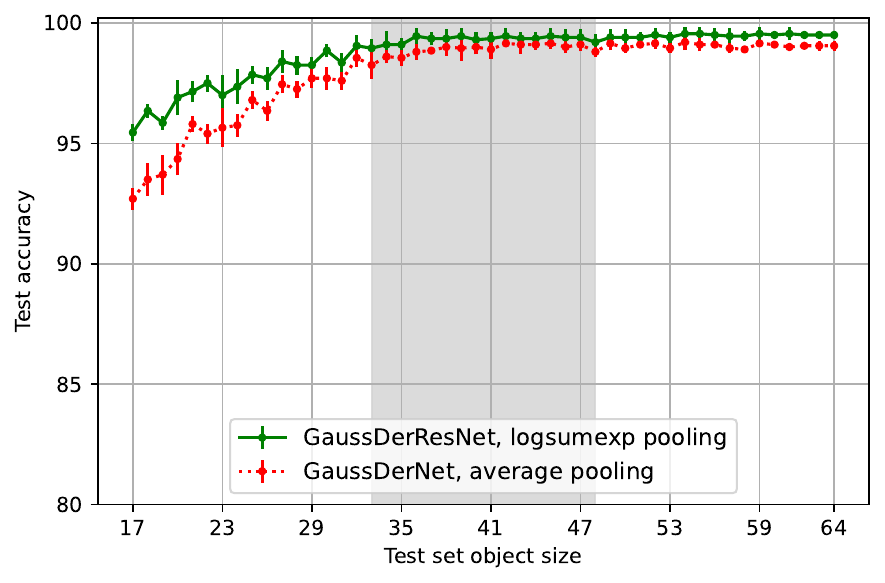}
          &  \includegraphics[width=0.32\textwidth]{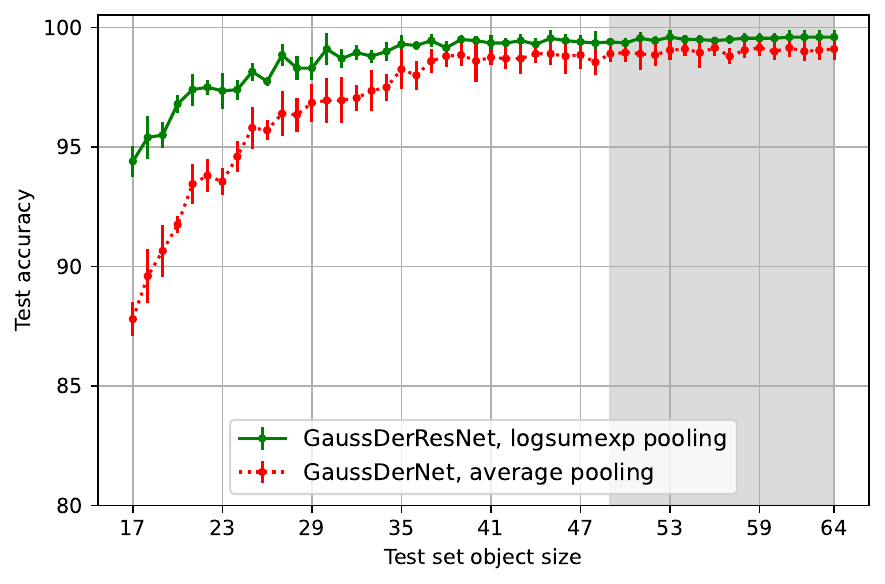}
      \end{tabular}
      \caption{Comparison of scale generalisation curves for the
        traffic sign subset of the STIR dataset}
      \label{fig:subfig-STIR-trafficsign}
    \end{subfigure}

    \bigskip
    
    \begin{subfigure}{0.99\textwidth}      
      \begin{tabular}{ccc}
          \hspace{3mm} \textit{Training with small object sizes}
          & \hspace{3mm} \textit{Training with medium object sizes}
          & \hspace{3mm} \textit{Training with large object sizes} \\
          \includegraphics[width=0.32\textwidth]{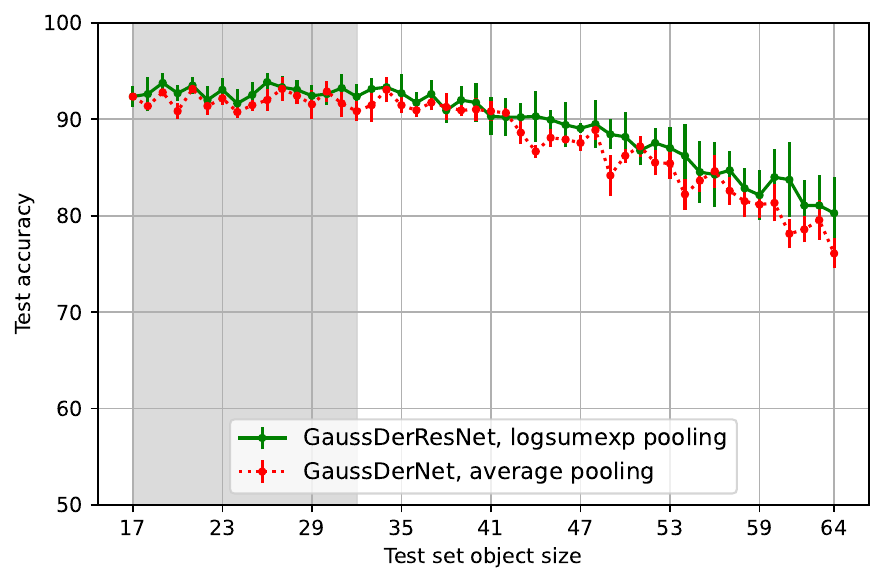}
          & \includegraphics[width=0.32\textwidth]{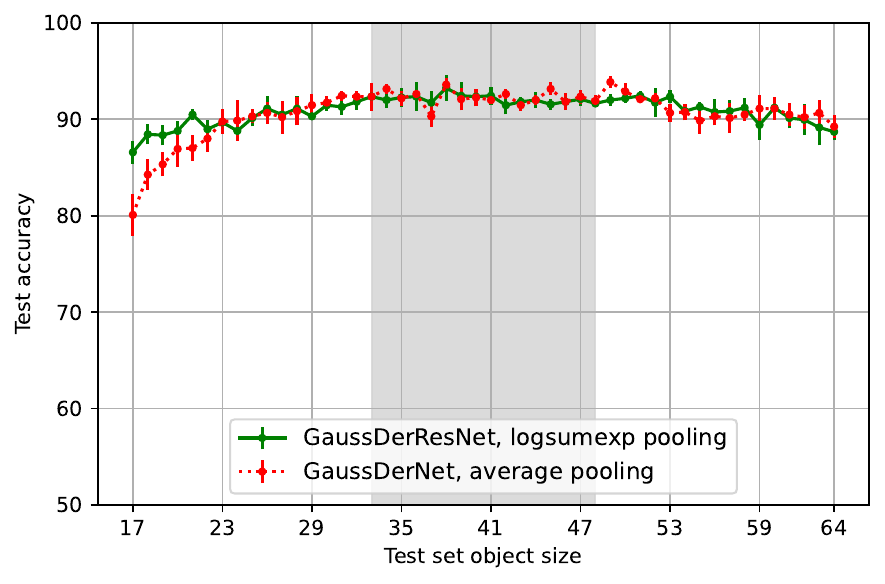}
          & \includegraphics[width=0.32\textwidth]{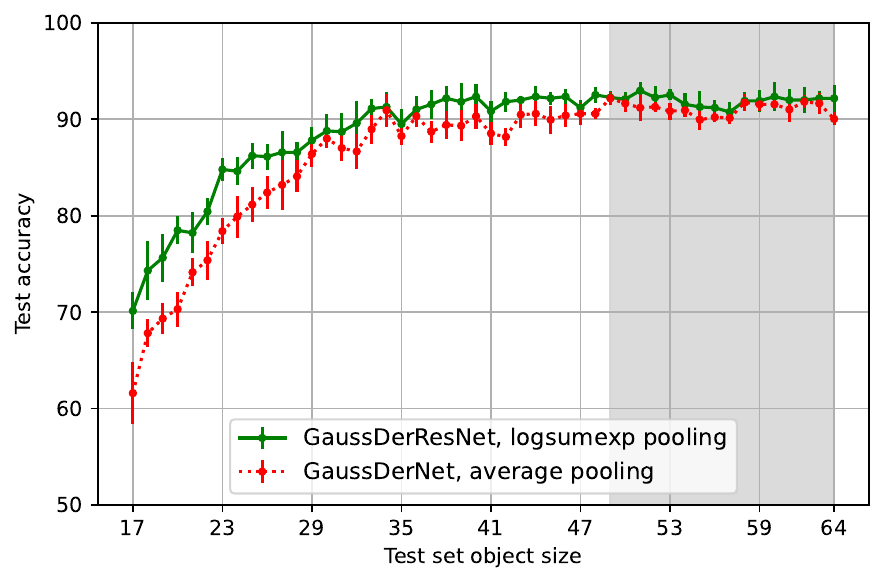}
      \end{tabular}
      \caption{Comparison of scale generalisation curves on the aerial
        image subset of the STIR dataset}
      \label{fig:subfig-STIR-aerial}
    \end{subfigure} 
  \end{center}
  \caption{Scale generalisation curves for the (a)~traffic sign and 
  (b)~aerial image subsets of the STIR dataset, comparing both the mean accuracy 
  and the error bars for the standard deviation over 5 runs for the 
  multi-scale-channel GaussDerResNets based on logsumexp pooling over 
  scale to the multi-scale-channel GaussDerNets based on average
  pooling over scale
  (reproduced from Perzanowski and Lindeberg (\citeyear{PerLin25-JMIV-ScSelGaussDerNets})). 
  Three different training regimes are shown for each dataset, indicated 
  by shaded grey regions, with the networks trained on data containing 
  (left)~small size objects (size $\in [17,32]$), (middle)~medium size objects 
  (size $\in [33,48]$), and (right)~large size objects (size $\in [49,64]$). All
  the networks 
  were evaluated on the test set containing all the possible object sizes 
  $l \in [17,64]$. From the graphs, we see that the GaussDerResNets have 
  better scale generalisation properties than the GaussDerNets. (Note 
  that the vertical axes for the test accuracy in the top and bottom 
  subfigures use different lower bounds.)}
  \label{fig-STIR-scale-selection-main}
\end{figure*}

\subsubsection{Scale generalisation experiments with multi-scale-channel
  networks on the STIR datasets with traffic signs and aerial images}
\label{sec-stir-experiments}

Experiments of a similar type were also performed on the traffic 
sign and aerial subsets of the STIR dataset, 
with Figure~\ref{fig-STIR-scale-selection-main} showing the 
corresponding scale generalisation results averaged over 5 runs. 
In this case, to investigate the effect of object size during training 
on the ability of the GaussDerResNets to generalise to any previously 
unseen scale, three different training settings were considered;
each scenario restricting the training data to a distinct range of 
object sizes, corresponding to the object bounding box length $l$ 
that represents the object scale. Training on small size object
covers the subset $l \in [17,32]$, training on medium size
objects covers the subset $l \in [33,48]$, 
and training on large size objects covers the subset $l \in [49,64]$.
The resulting networks were 
each evaluated on test data covering the full range of object sizes, 
$l \in [17,64]$.

From the results, we can see that for both of the STIR datasets the test 
accuracy for the object sizes seen during training is only marginally 
better for the GaussDerResNets compared to the GaussDerNets.
The scale generalisation performance is, however, clearly better, as the scale 
generalisation curves are flatter for every training setting, since their 
greater depth enables them to learn representations better suited 
for class discrimination.

This good performance on the aerial image dataset is consistent with 
the ability of the GaussDerResNets to handle images containing multiple objects, 
which we will demonstrate in Section~\ref{sec-loc-extended-gaussderresnets}. 
The network is able to detect multiple maxima that may correspond to different 
objects and base the final prediction on the strongest responses, as multiple 
object instances of a given class are sometimes present in the aerial images.
Furthermore, while the traffic sign samples in the subset with small object sizes
occasionally include signs belonging to other classes 
in the background, either partially visible or as a whole, the good 
performance demonstrates that GaussDerResNets may typically select 
the most dominant sign within each image, resulting in only a minor drop 
in accuracy for the smaller object sizes.

\begin{figure}[ht]
  \begin{center}
    \textit{\quad\quad Scale generalisation on rescaled STL-10}
    \includegraphics[width=0.48\textwidth]{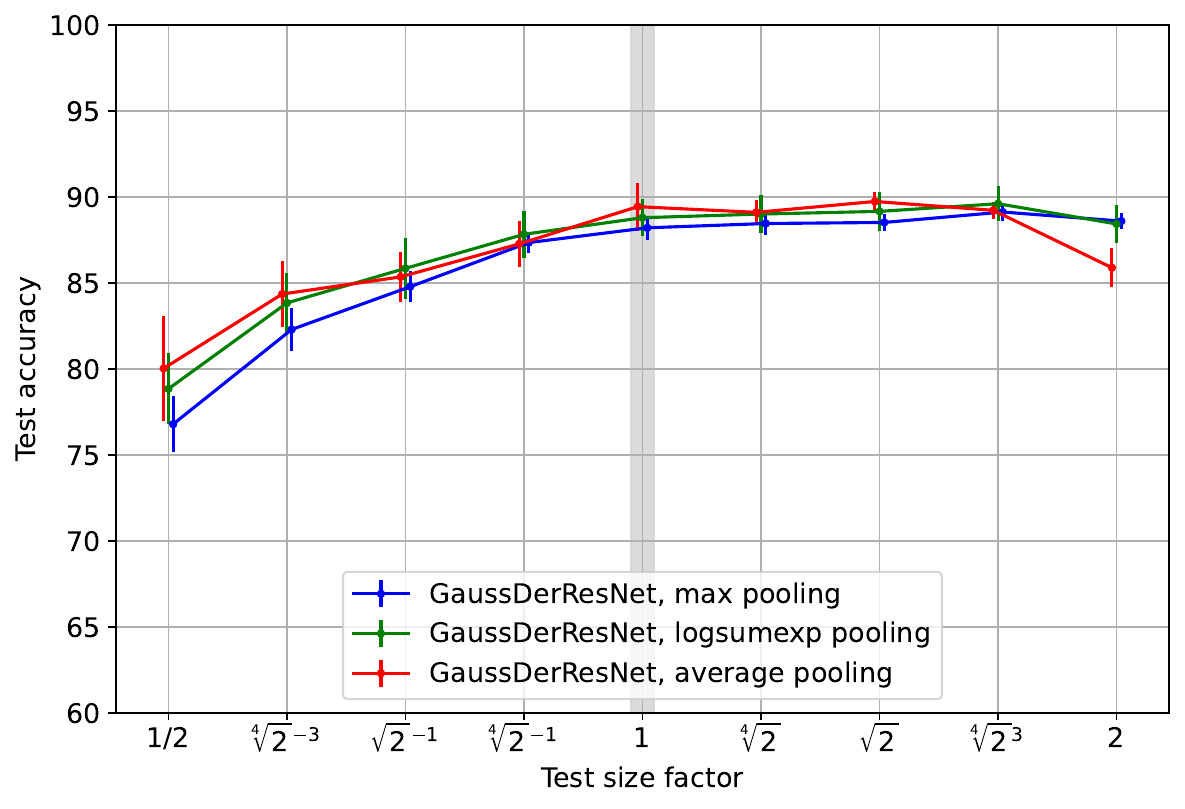}
  \end{center}
\caption{Comparison between the scale generalisation curves
    for the rescaled STL-10 dataset, showing both the mean accuracy and
    the error bars for the standard deviation over 5~runs for
    multi-scale-channel GaussDerResNets based on either
    max, logsumexp or average pooling over scale.
    In this experiment, the networks were trained on training data
    corresponding to the size factor 1, and were then evaluated on
    rescaled copies of the test set, covering all the size factors
    between 1/2 and 2, with each copy corresponding to a
    different single size factor.
    The multi-scale-channel networks achieved
    a mean accuracy of 88.2--89.5\% (with standard deviation $\pm$ 0.7-1.4\%) 
    on the test data of the same size (the size factor 1) as the training data, 
    represented by the shaded grey region in the graph.
    As can be seen in the figure, despite a portion of the
    STL-10 images containing non-centered objects,
    the multi-scale-channel GaussDerResNets based on the spatial max
    pooling mechanism lead to rather flat scale generalisation curves.}
  \label{fig-stl10-gaussresnet-scale-gen}
\end{figure}

\begin{figure*}[hbtp]
  \begin{center}
    \begin{subfigure}{0.32\textwidth}
        \centering
        \includegraphics[width=\linewidth]{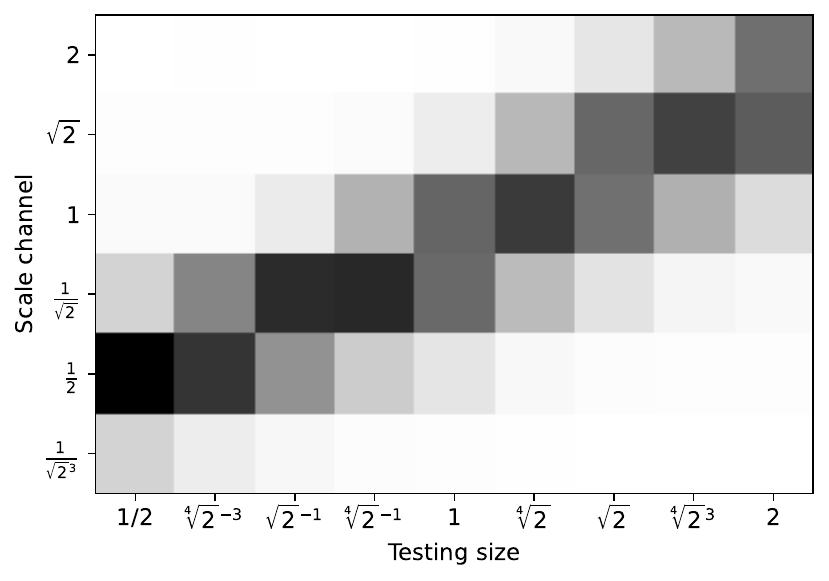}
        \caption{Max pooling}
        \label{fig:subfig-cifar-lbl-hist-1-main}
    \end{subfigure}
    \hfill
    \begin{subfigure}{0.32\textwidth}
        \centering
        \includegraphics[width=\linewidth]{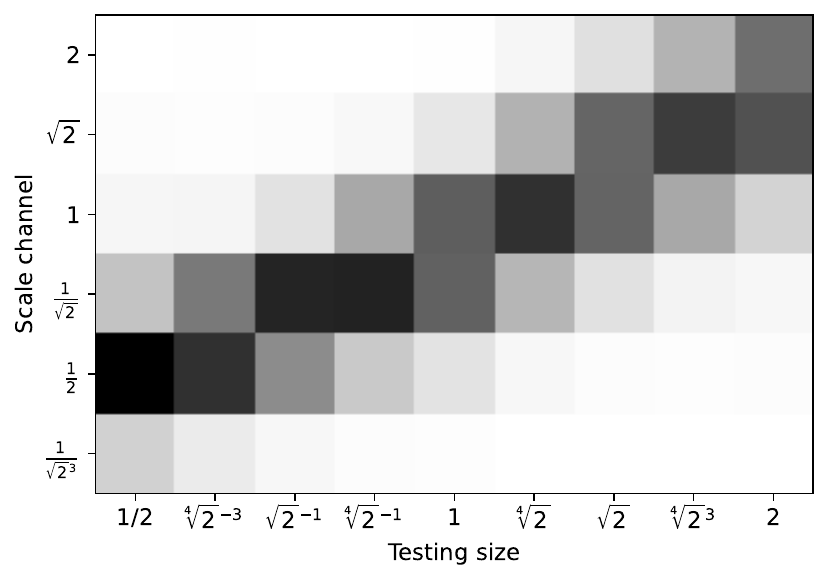}
        \caption{Logsumexp pooling}
        \label{fig:subfig-cifar-lbl-hist-2-main}
    \end{subfigure}
    \hfill
    \begin{subfigure}{0.32\textwidth}
        \centering
        \includegraphics[width=\linewidth]{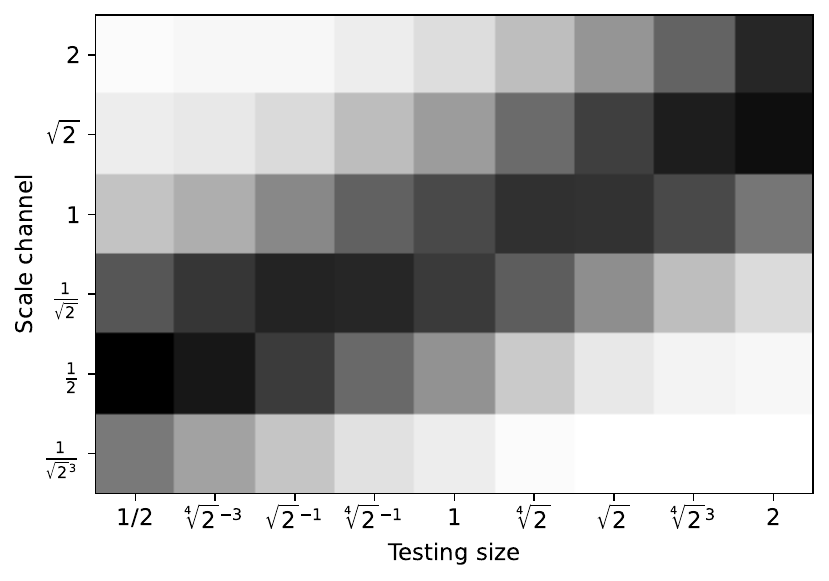}
        \caption{Average pooling}
        \label{fig:subfig-cifar-lbl-hist-3-main}
    \end{subfigure}
  \end{center}
  \caption{Scale selection histograms averaged over 5~runs for the 
    multi-scale-channel GaussDerResNets trained and evaluated 
    on the rescaled CIFAR-10 dataset, with no label smoothing 
    used during training, for the three different permutation invariant
    aggregation methods for pooling over scales, including
    (a)~max pooling, (b)~logsumexp pooling and (c)~average pooling.
    In these scale selection histograms, the vertical axis represents 
    the relative contribution of each scale channel to the final
    classification as function of the different size factors for the
    test dataset as mapped onto the horizontal axis. 
    A clear linear trend can be seen in all the three histograms,
    with the dominant selected scale levels, represented by the
    dark regions, found to be directly proportional to the 
    size factor of the test images.
    This behaviour is an explicit manifestation of the
    scale-covariant properties of the GaussDerResNet architecture.}
  \label{fig-cifar10-gaussresnet-scale-selection-main}

  \begin{center}
    \begin{subfigure}[t]{0.32\textwidth}
        \centering
        \includegraphics[width=\linewidth]{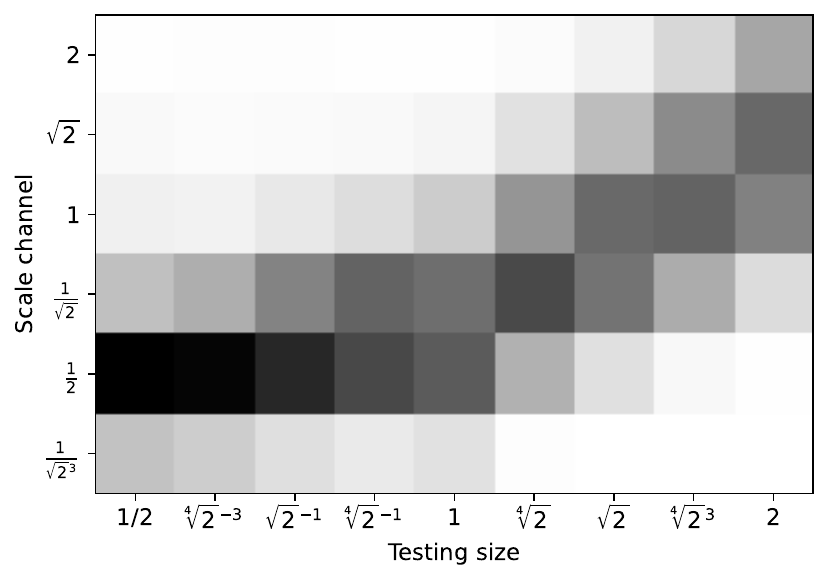}
        \caption{Max pooling}
        \label{fig:subfig-stl-lbl-hist-1-main}
    \end{subfigure}
    \hfill
    \begin{subfigure}[t]{0.32\textwidth}
        \centering
        \includegraphics[width=\linewidth]{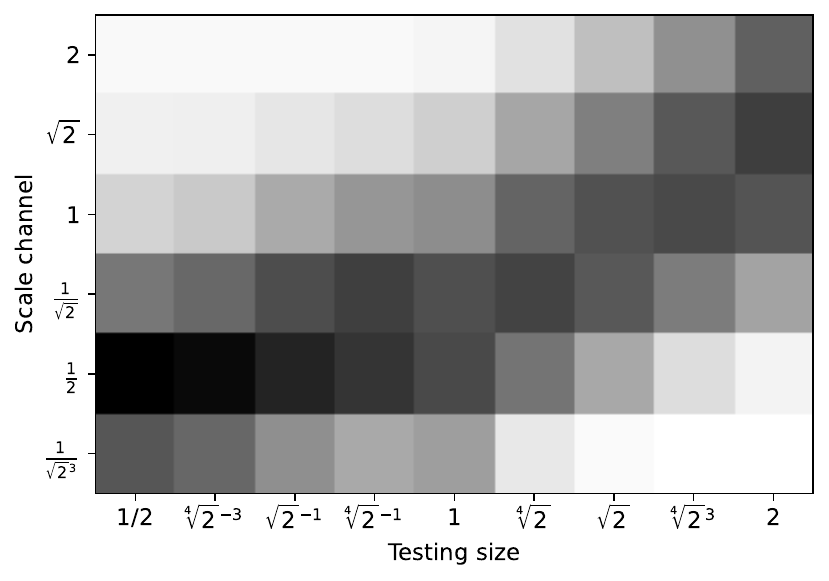}
        \caption{Logsumexp pooling}
        \label{fig:subfig-stl-lbl-hist-2-main}
    \end{subfigure}
    \hfill
    \begin{subfigure}[t]{0.32\textwidth}
        \centering
        \includegraphics[width=\linewidth]{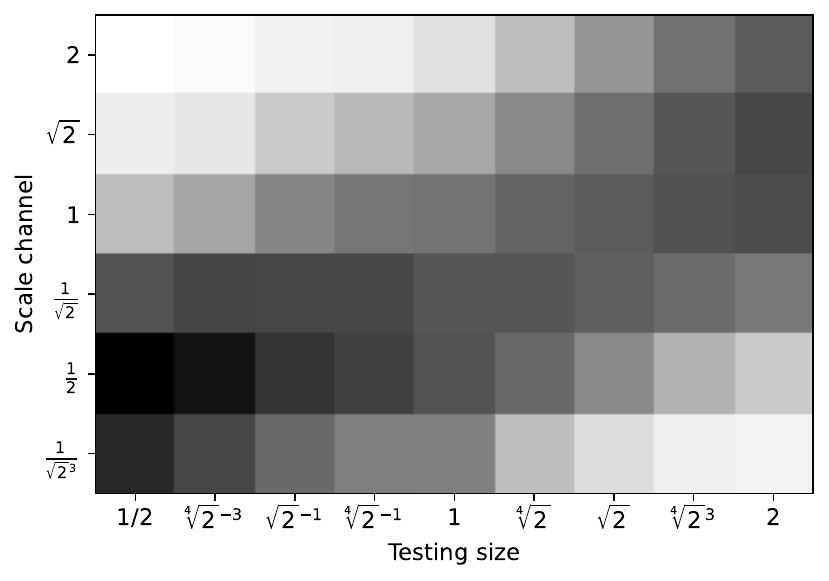}
        \caption{Average pooling}
        \label{fig:subfig-stl-lbl-hist-3-main}
    \end{subfigure}
  \end{center}
  \caption{Scale selection histograms averaged over 5~runs for the 
    multi-scale-channel GaussDerResNets trained and evaluated on 
    the rescaled STL-10 dataset, for the three different permutation 
    invariant methods for pooling over scales, including (a)~max pooling,
    (b)~logsumexp pooling and (c)~average pooling.
    In these scale selection histograms, the vertical axis represents
    the relative contribution of each scale channel to the final
    classification as function of the different size factors for
    the test dataset as mapped onto the horizontal axis.
    Similar to the results obtained for the rescaled CIFAR-10 
    dataset, all the three scale pooling methods show a linear trend,
    where a larger size in the test images leads to the selection
    of proportionally coarser scale levels.
    This confirms that incorporating a spatial max pooling mechanism into
    the network architecture preserves the scale-covariant properties
    of the GaussDerResNets.}
  \label{fig-stl10-gaussresnet-scale-selection-main}
\end{figure*}

\subsubsection{Scale generalisation and scale selection
  experiments with multi-scale-channel
  networks on the rescaled STL-10 dataset}
\label{sec-stl10-experiments}

Figure~\ref{fig-stl10-gaussresnet-scale-gen} shows the result of a
corresponding scale generalisation experiment on the new rescaled 
STL-10 dataset. Unlike the GaussDerResNet architectures 
used in the previous experiments, these experiments use 
networks with layers that include a zero-order term (except in the first layer), 
according to Equation~(\ref{eq-alpha-vals-2nd-order-jet+zero-order}),  
since that choice has been found to perform significantly better
on this dataset, as detailed further in Section~\ref{sec-zero-order-term-study}. 
Additionally, unlike the previously considered datasets, the rescaled STL-10 
test set contains images extended using mirror padding to twice the size of 
the training images, as described in Section~\ref{sec-rescaled-stl-dataset}.

Since the objects in the STL-10 dataset are not centered in the image
domain, we throughout made use of spatial max pooling as the
spatial selection mechanism.

As can be seen from the results, for the GaussDerResNets,
we obtain rather flat scale generalisation curves, with no 
drop in performance between the size factors 1 and 2 for 
the networks based on max or logsumexp pooling over scales, and only a 
slight drop in performance for the size factor 2 for the network%
\footnote{Using an additional large boundary scale channel 
during evaluation can improve performance for the size factor 2 
for average pooling, however, ideally it should be included also
during the training process to balance its influence.} 
based on average pooling over scales,
thereby demonstrating very good scale generalisation capability. 
For the smaller factors, 
we find a clear and gradual drop in performance in the range of 
up to 9--11.5~ppt towards the scale factor 1/2,
likely due to discretisation effects and possibly
also caused by the mirror extensions in the test images 
at the smaller size factors. The latter effect may, however,
largely be an artefact of how the image dataset has been
constructed and not because of the deep learning architecture.
 
Figure~\ref{fig-stl10-gaussresnet-scale-selection-main} shows the scale 
selection histograms for the corresponding networks. On this dataset, the 
networks show a somewhat less localised linear trend, with a shift to relatively 
finer scale channels being used at corresponding size factors, compared 
to the results for the networks trained on the rescaled CIFAR-10
dataset, as shown in 
Figure~\ref{fig-cifar10-gaussresnet-scale-selection-main}. This is likely due to 
the use of spatial max pooling as the spatial selection mechanism, 
which benefits from allowing different scales to be considered at different 
spatial locations, resulting in more neighbouring scale channels being
involved 
in the final prediction, and fine scale image structures being more impactful for 
computing the prediction.

In summary, these results demonstrate that incorporating a spatial 
max pooling mechanism into the GaussDerResNet architecture does 
preserve the desired scale covariance and scale selection properties 
of the network, while enabling the handling of non-centred 
image structures, by introducing the ability to focus on any spatial 
location within the image, and at different scales.

\begin{figure*}[ht]
  \begin{center}
    \begin{subfigure}{0.32\textwidth}
      \centering
      \textit{\quad\quad Scale generalisation on rescaled Fashion-MNIST}
      \includegraphics[width=\linewidth]{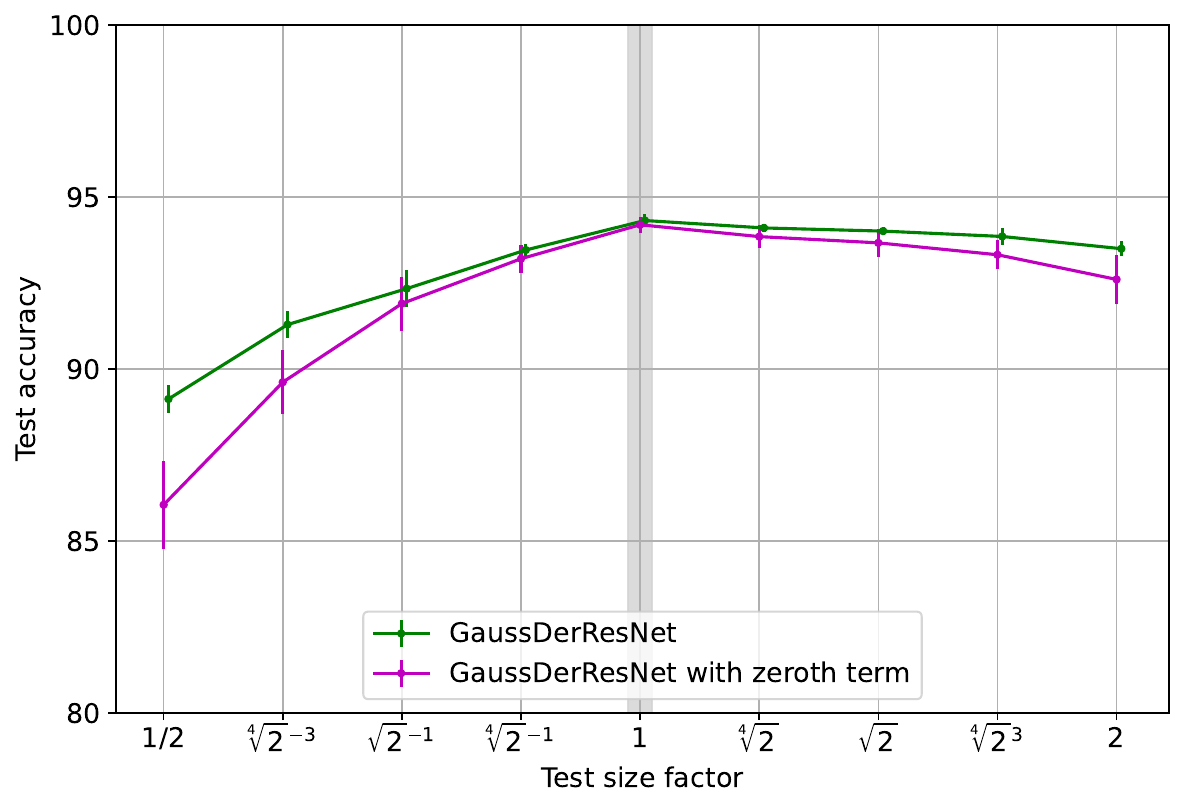}
      \caption{}
      \label{fig:subfig-zeroth-1}
    \end{subfigure}
    \hfill
    \begin{subfigure}{0.32\textwidth}
      \centering
      \textit{\quad \quad Scale generalisation on rescaled CIFAR-10}
      \includegraphics[width=\linewidth]{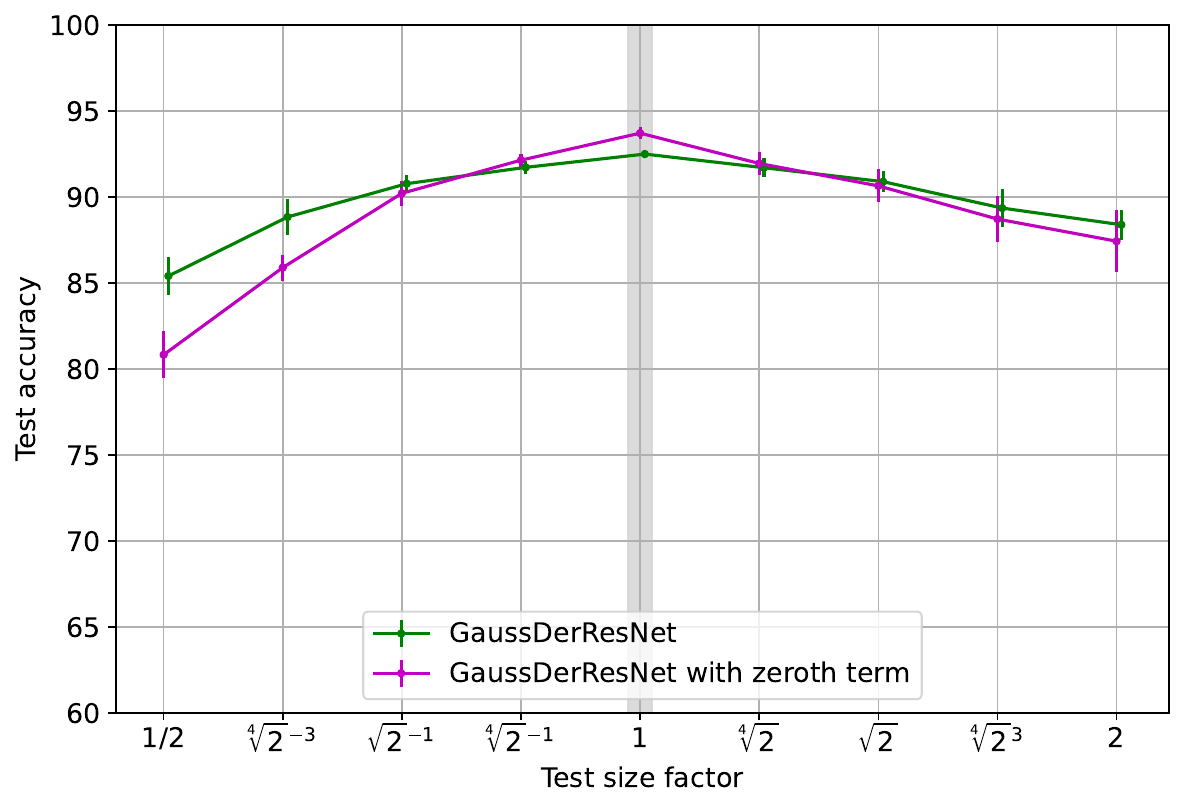}
      \caption{}
      \label{fig:subfig-zeroth-2}
    \end{subfigure}
    \hfill
    \begin{subfigure}{0.32\textwidth}
      \centering
      \textit{\quad \quad Scale generalisation on rescaled STL-10}
      \includegraphics[width=\linewidth]{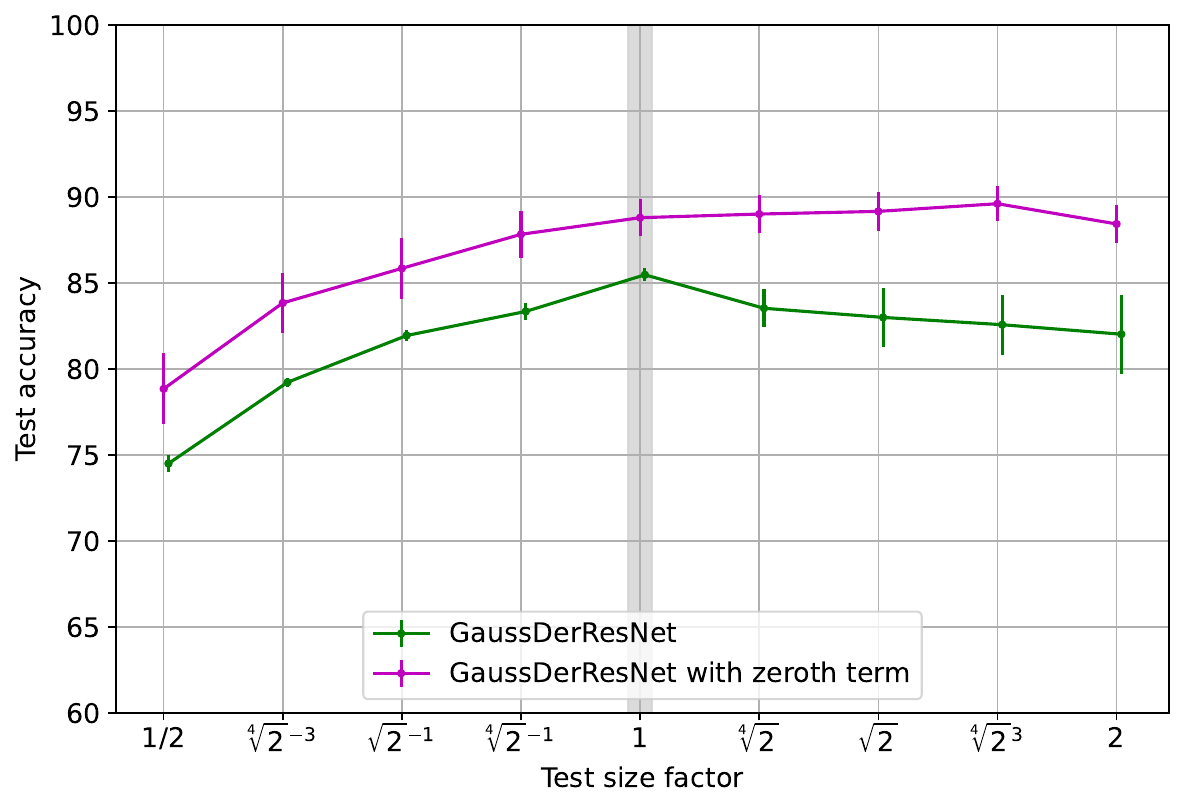}
      \caption{}
      \label{fig:subfig-zeroth-3}
    \end{subfigure}
    
    \vspace{0.4cm} 
    
    \begin{subfigure}{0.32\textwidth}
      \centering
      \textit{\quad \quad Scale selection histogram on rescaled Fashion-MNIST}
      \includegraphics[width=\linewidth]{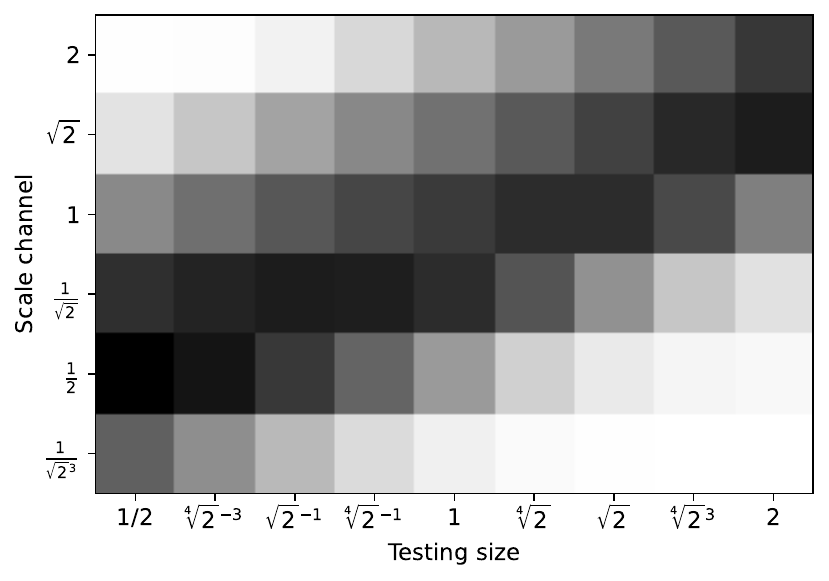}
      \caption{}
      \label{fig:subfig-zeroth-4}
    \end{subfigure}
    \hfill
    \begin{subfigure}{0.32\textwidth}
      \centering
      \textit{\quad \quad Scale selection histogram on rescaled CIFAR-10}
      \includegraphics[width=\linewidth]{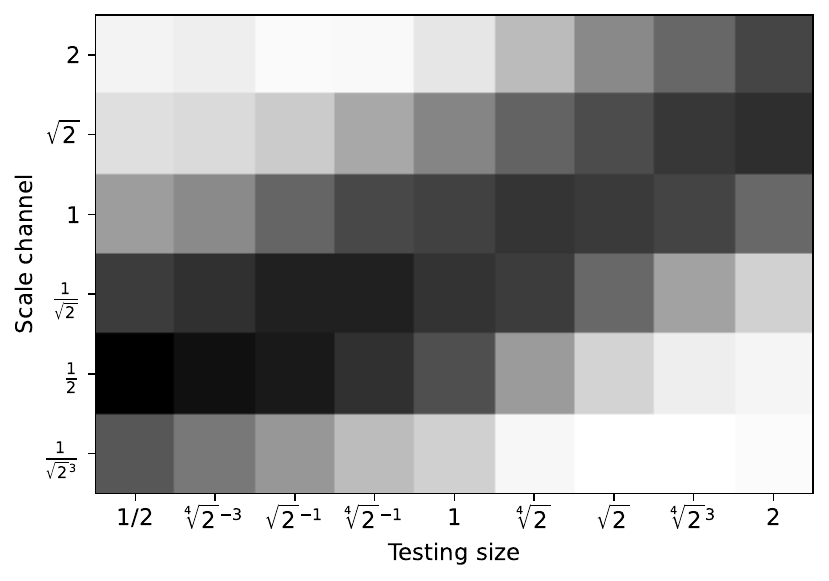}
      \caption{}
      \label{fig:subfig-zeroth-5}
    \end{subfigure}
    \hfill
    \begin{subfigure}{0.32\textwidth}
      \centering
      \textit{\quad \quad Scale selection histogram on rescaled STL-10}
      \includegraphics[width=\linewidth]{scale_hist_gaussderresnet_stl10_avgpool_multiruns.pdf}
      \caption{}
      \label{fig:subfig-zeroth-6}
    \end{subfigure}
  \end{center}
  \caption{Comparison between the scale generalisation curves 
    and the corresponding scale selection histograms for the 
    multi-scale-channel GaussDerResNets, averaged across 5~runs, based on 
    either standard 2-jet layers or 2-jet layers complemented with 
    a zero-order term, on the rescaled (left)~Fashion-MNIST, 
    (middle)~CIFAR-10 and (right)~STL-10 datasets. The subfigures use 
    networks based on logsumexp pooling over scales. 
    After being trained on the training set for the size factor 1, 
    the networks were tested on all the size factors between 1/2 and 2. 
    The scale selection histograms visualise the relative contribution 
    of each scale channel to the final classification. For the
    rescaled Fashion-MNIST and CIFAR-10 datasets,
    an extra boundary scale channel $\sigma_0 = 2^{5/4}$ 
    was added during inference, although we do not show its contribution 
    in the histograms for the sake of simplicity. As can be 
    seen from the graphs, including a zeroth-order term clearly improves 
    both the accuracy and the scale generalisation for the rescaled STL-10 
    dataset. For the rescaled Fashion-MNIST and CIFAR-10 datasets, 
    using a zeroth-order term does not lead to any improvement, with the performance at 
    the smaller size factors noticeably reduced when including 
    this term. Additionally, a clear linear trend can be seen in the 
    scale selection histograms for all the three datasets, demonstrating 
    that including a zero-order term in the computational primitives 
    of the GaussDerResNets leads to the selected scale levels 
    still being proportional to the size in the test image data, as 
    expected from theory. (Note that the vertical axes for the test 
    accuracy in the top subfigures use different lower bounds for 
    the different datasets.)}
  \label{fig-gaussresnet-zeroth-order}
\end{figure*}

\subsection{Conclusions from the scale generalisation experiments}
\label{sec-concl-sc-gen-exps}

From a joint analysis of the scale generalisation experiments for the
rescaled Fashion-MNIST, CIFAR-10 and STL-10 datasets presented
in Figures~\ref{fig-fashion-gaussresnet-scale-gen},
\ref{fig-cifar10-gaussresnet-scale-gen}
and~\ref{fig-stl10-gaussresnet-scale-gen}, respectively, it appears as if the scale
pooling method based on logsumexp pooling constitutes the
most {\em reliable\/} scale pooling method out of the three studied scale
pooling methods, consistently converging well across all the datasets,
unlike max pooling or average pooling.

The max pooling method over scales appears to be more sensitive to the loss of
effective resolution in the image data for the smaller size factors,
while the average pooling method appears to be in a higher need for
additional boundary scale channels towards the upper bound of the
range of scale channels, probably due its property of integrating
information over multiple scale channels, which the logsumexp
and the max pooling methods are not as critically dependent on.

The max pooling method does, however, constitute a very
straightforward method to use when extending the GaussDerResNets
to joint local spatial and scale selection, as we will do in
Section~\ref{sec-loc-extended-gaussderresnets}.

As we demonstrate in comparisons to other networks in 
Appendices~G and~H, the GaussDerResNets have very good scale generalisation 
properties compared to those of the DISCO network on the rescaled 
Fashion-MNIST, CIFAR-10 and STL-10 datasets, and does also 
compare very well to the DISCO, SESN and PixelPool 
networks on the STIR datasets with traffic signs and aerial 
images.

\section{Ablation and comparative studies}
\label{sec-ablation-comparative-studies}

This section presents ablation and comparative studies of GaussDerResNets 
and their variants, performed using the rescaled Fashion-MNIST, CIFAR-10 
and STL-10 datasets. Here, we will examine the results of experiments dealing 
with different architectural variations:
\begin{itemize}
\item
  investigating the influence of extending GaussDerResNets 
to include a zero-order term in the 2-jet.
\end{itemize}
In appendix sections, we additionally:
\begin{itemize}
  \item
    demonstrate that comparable accuracy and scale generalisation
    properties can be obtained from depth\-wise-separable-convolution-based 
    DSGaussDerResNets, which substantially decrease both the number
    of parameters and the computational cost for inference
    (Appendix~C),
  \item
    demonstrate that the use of label smoothing can significantly
    improve the scale generalisation properties of the networks,
    however, with the side effect of leading to less sharp scale
    selection histograms (Appendix~D),
\item
  demonstrate that pre-training, using a first phase of single-scale-channel 
  mode followed by a second phase of genuine multi-scale-channel
  training, can substantially decrease the computational work for training,
  and often leading to better convergence properties, compared 
  to training a multi-scale-channel GaussDerResNet from scratch 
  (Appendix~E), and
\item
  demonstrate that weight transfer to a denser scale-channel 
  spacing can slightly improve the scale generalisation properties
  (Appendix~F).
\end{itemize}
Additionally, Appendix~G shows a comparison to the results of
applying the official code for the DISCO network 
(Sosnovik et al.\ \citeyear{SosMosSme21-BMVC})
to the scale-generalisation tasks for
the rescaled Fashion-MNIST, CIFAR-10 and STL-10 datasets,
while Appendix~H gives a comparison to the results of the
previously presented scale generalisation experiments by
Altstidl et al.\ (\citeyear{AltNguSchKoeMutEskZan23-IJCNN}) on the
STIR datasets with traffic signs and aerial images.

\subsection{Benefit of using a zero-order term in higher layers
  of the networks}
\label{sec-zero-order-term-study}

The receptive fields in CNNs can be related to the concept of 
scale-space, by approximating the filters using a Gaussian $N$-Jet, 
which is accomplished by using a basis of Gaussian derivative
operators, up to a certain order $N$, to form a truncated local 
Taylor expansion, that represents the local image structure. This 
kind of approximation involves a Gaussian operator of order zero, 
which differs from the standard layer design of the GaussDerResNet 
architecture, which is purposely based only on first- and second-order 
(scale-normalised) Gaussian derivative operators.

In this section, we will investigate how including a zero-order term in the computational 
primitives for the layers of GaussDerResNets (except for in the first layer), 
according to Equation~(\ref{eq-alpha-vals-2nd-order-jet+zero-order}), 
affects the scale generalisation and the scale selection properties of 
the network. For this purpose, for the rescaled Fashion-MNIST, CIFAR-10 
and STL-10 datasets, we will perform comparative studies of the effect of 
using a standard vs. a zero-order based GaussDerResNet design.
For  the rescaled Fashion-MNIST and CIFAR-10 datasets,
an additional boundary scale channel 
with $\sigma_0 = 2^{5/4}$ was used during the inference stage, as it leads 
to better scale generalisation for the larger size factors.

For each dataset, we trained two multi-scale-channel GaussDerResNets, 
one with the layers defined using the standard design, 
according to Equation~(\ref{eq-alpha-vals-2nd-order-jet}), 
and the other defined using the Gaussian $N$-jet with a zero-order term, 
according to Equation~(\ref{eq-alpha-vals-2nd-order-jet+zero-order}). 
The models used logsumexp pooling over scales, since this pooling
method was judged as the most reliable scale pooling method
in Section~\ref{sec-concl-sc-gen-exps}.
The training was done 
using the protocol and the settings described in 
Section~\ref{sec-training-procedure}, for each dataset. 

The upper row in Figure~\ref{fig-gaussresnet-zeroth-order}
shows the resulting means of the scale generalisation curves over 5~runs. 
For the rescaled Fashion-MNIST dataset, we can see that omitting 
the zero-order terms results in slightly better performance, as well 
as improved scale generalisation for the smaller test size factors, 
with minor improvement also for the large size factors. For the 
rescaled CIFAR-10 dataset, we can see that the performance at the 
size factor 1 is better when basing the network on layers with the 
zero-order terms included. The scale generalisation curve is, 
however, less flat compared to the standard GaussDerResNet. 
The omission of zeroth-order terms results in more stable 
convergence during training for both of these datasets, with less 
variability in the test accuracy between runs.

The results for the rescaled STL-10 dataset show a notably different 
behaviour, where both the test accuracy and the scale generalisation 
is significantly better when including zero-order terms, converging to 
a much better solution during training. When basing the layers on zero-order 
terms, the performance for the size factor 1 improves by $\sim$3~ppt, 
and the performance at the large size factors is also clearly improved, 
with the performance at the size factor 2 improving by $\sim$6~ppt.

The corresponding averages of the scale selection histograms over 5 runs
are shown for the networks involving zero-order terms in the bottom 
row in Figure~\ref{fig-gaussresnet-zeroth-order}. A clear linear trend 
can be observed between the selected scale levels and the size of the 
images for each rescaled dataset, indicating that 
including the zero-order terms in the layers of the GaussDerResNets 
results in scale covariance properties consistent with the theoretical
analysis in Section~\ref{sec-scale-covariance-properties}.

\section{Visualisations of Gaussian derivative ResNets}
\label{sec-network-function-visualisation}

To facilitate a deeper understanding of the multi-scale-channel
GaussDerResNets in operation, we will in this section analyse 
visualisations of the activation maps in the final 
convolution layer before the spatial selection stage, 
for the rescaled STL-10 dataset.

In Appendix~I, we also analyse 
the effective filters learned by the GaussDerResNets with the 
layers based on zero-order terms.

\begin{figure*}[htbp]
  \begin{center}
    \begin{subfigure}{0.9\textwidth}      
        \centering
        \raggedright
        \textit{ \hspace{0.4cm} Input image \hspace{1.3cm} Scale channel $\sigma_{0}=1/2$ \hspace{0.9cm} Scale channel $\sigma_{0}=1$ \hspace{1.1cm} Scale channel $\sigma_{0}=2$}
        \raisebox{1.7mm}{\includegraphics[width=0.178\textwidth]{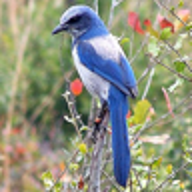}}
        \hspace{2.3mm}
        \includegraphics[width=0.245\textwidth]{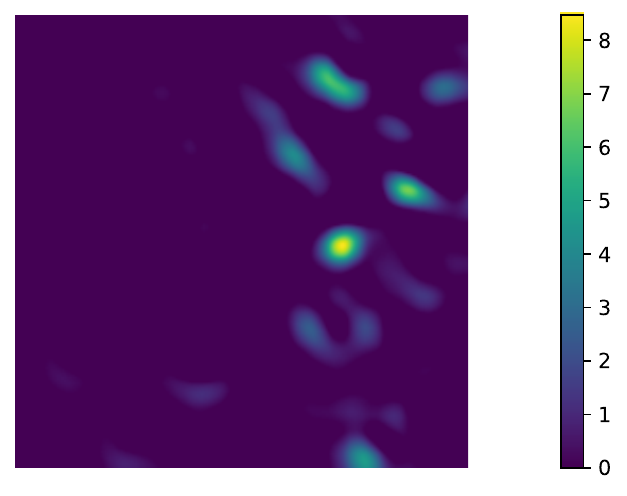}
        \hspace{1.1mm}
        \includegraphics[width=0.25\textwidth]{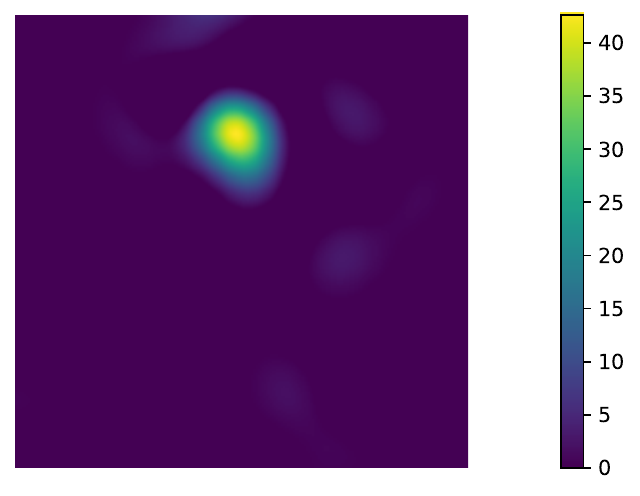}
        \hspace{2.2mm}
        \includegraphics[width=0.25\textwidth]{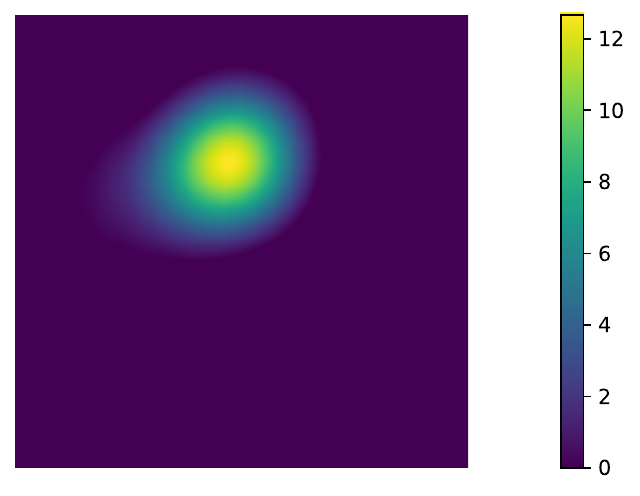}
        \caption{Activation maps from different scale channels for a sample image from the class ``bird''.}
        \label{fig:subfig-activation-stl-u}
    \end{subfigure} 
    
    \vspace{0.4cm} 
    
    \begin{subfigure}{0.9\textwidth}      
        \centering
        \raggedright
        \textit{ \hspace{0.4cm} Input image \hspace{1.3cm} Scale channel $\sigma_{0}=1/2$ \hspace{0.9cm} Scale channel $\sigma_{0}=1$ \hspace{1.1cm} Scale channel $\sigma_{0}=2$}
        \raisebox{1.7mm}{\includegraphics[width=0.178\textwidth]{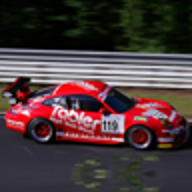}}
        \hspace{2.3mm}
        \includegraphics[width=0.245\textwidth]{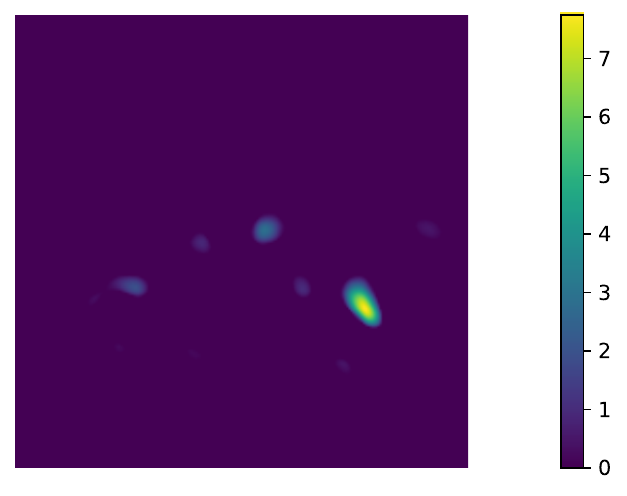}
        \hspace{1.1mm}
        \includegraphics[width=0.257\textwidth]{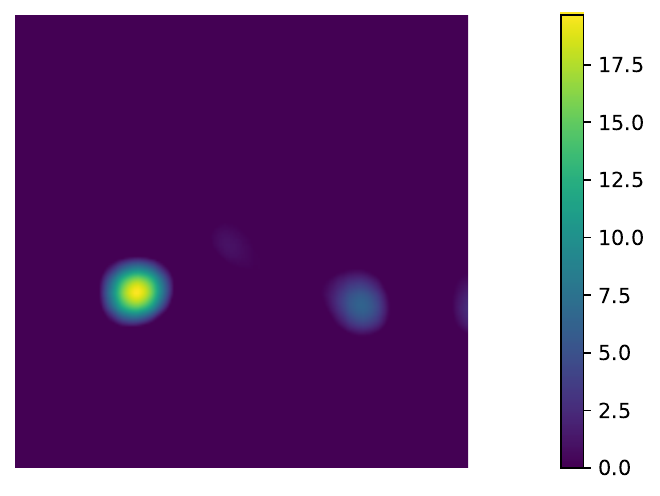}
        \hspace{1.2mm}
        \includegraphics[width=0.25\textwidth]{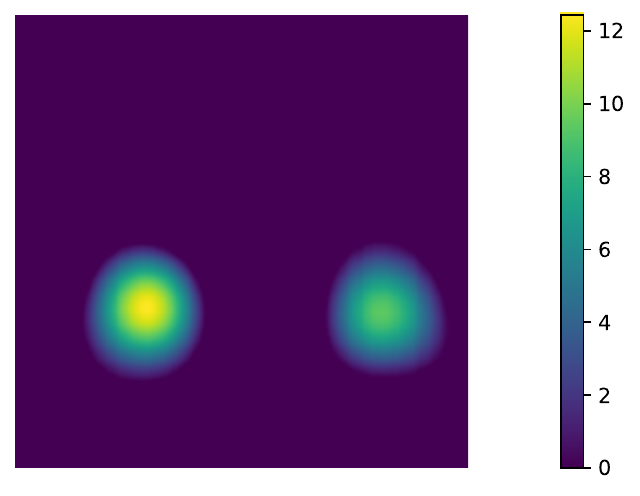}
        \caption{Activation maps from different scale channels for a sample image from the class ``car''.}
        \label{fig:subfig-activation-stl-v}
    \end{subfigure} 
    
    \vspace{0.4cm} 
    
    \begin{subfigure}{0.9\textwidth}      
        \centering
        \raggedright
        \textit{ \hspace{0.4cm} Input image \hspace{1.3cm} Scale channel $\sigma_{0}=1/2$ \hspace{0.9cm} Scale channel $\sigma_{0}=1$ \hspace{1.1cm} Scale channel $\sigma_{0}=2$}
        \raisebox{1.7mm}{\includegraphics[width=0.178\textwidth]{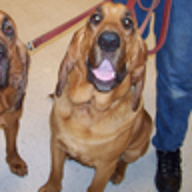}}
        \hspace{2.3mm}
        \includegraphics[width=0.257\textwidth]{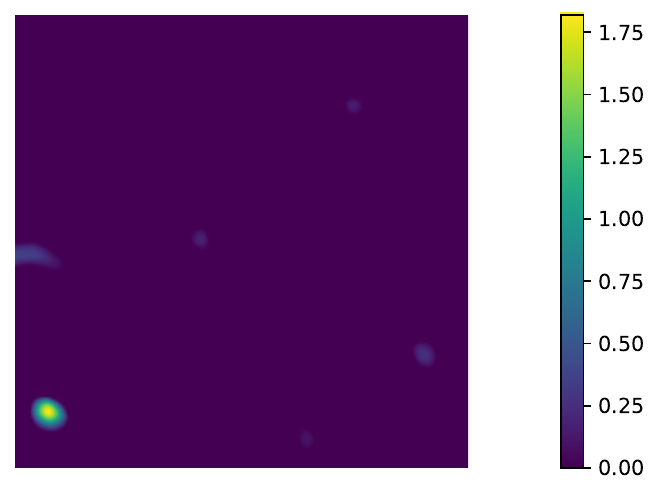}
        \hspace{-0.8mm}
        \includegraphics[width=0.257\textwidth]{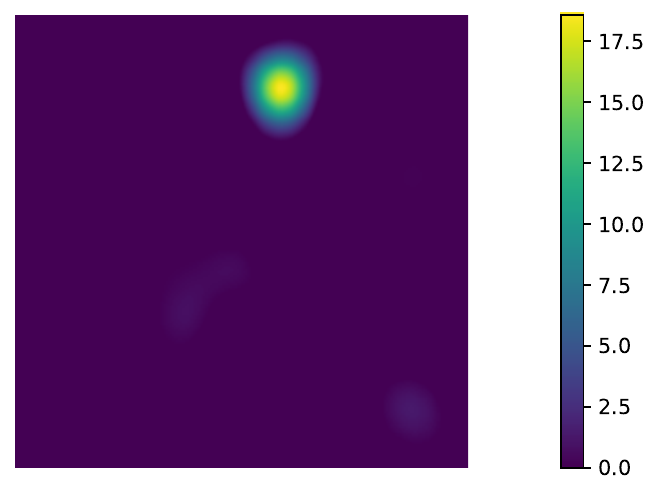}
        \hspace{1.1mm}
        \includegraphics[width=0.25\textwidth]{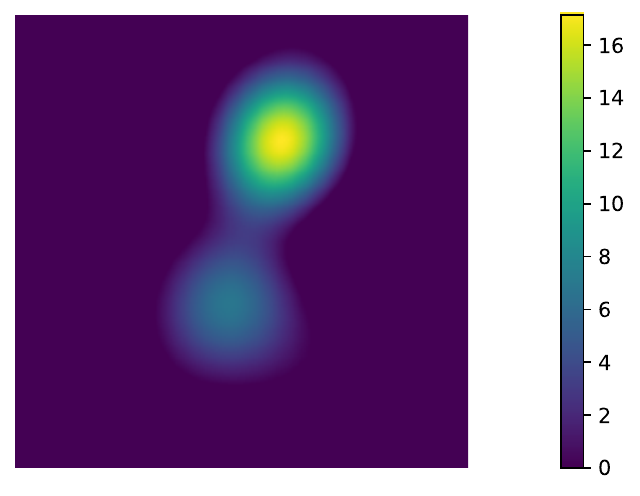}
        \caption{Activation maps from different scale channels for a sample image from the class ``dog''.}
        \label{fig:subfig-activation-stl-x}
    \end{subfigure} 
    
    \vspace{0.4cm} 
    
    \begin{subfigure}{0.9\textwidth}      
        \centering
        \raggedright
        \textit{ \hspace{0.4cm} Input image \hspace{1.3cm} Scale channel $\sigma_{0}=1/2$ \hspace{0.9cm} Scale channel $\sigma_{0}=1$ \hspace{1.1cm} Scale channel $\sigma_{0}=2$}
        \raisebox{1.7mm}{\includegraphics[width=0.178\textwidth]{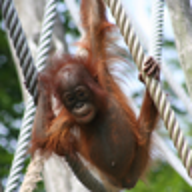}}
        \hspace{2.3mm}
        \includegraphics[width=0.253\textwidth]{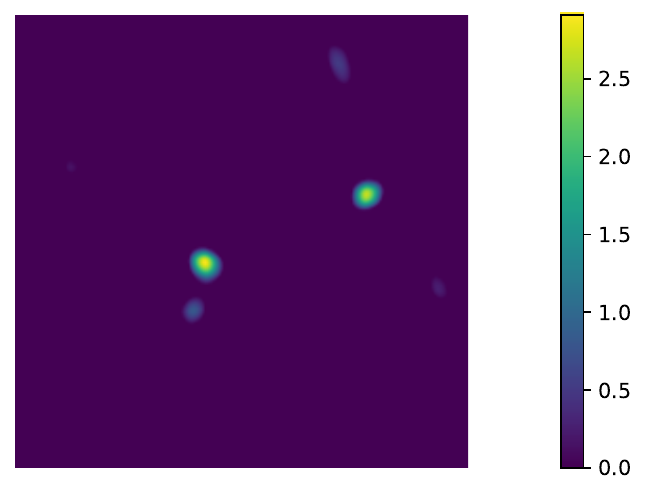}
        \hspace{-0.3mm}
        \includegraphics[width=0.251\textwidth]{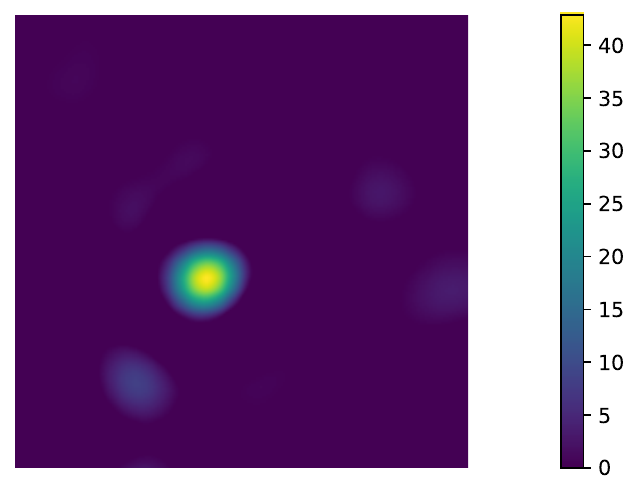}
        \hspace{2.2mm}
        \includegraphics[width=0.25\textwidth]{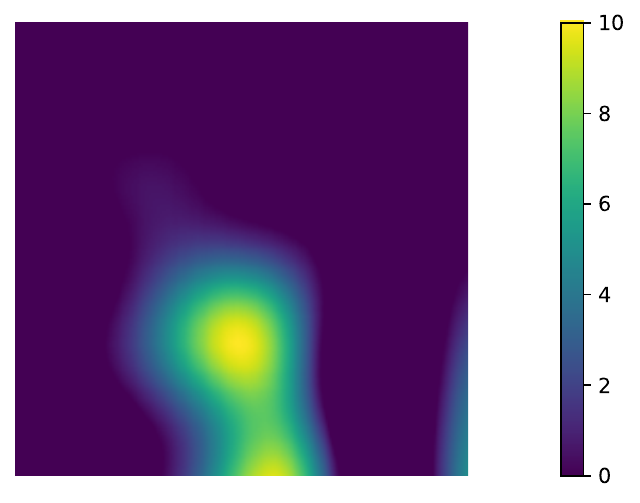}
        \caption{Activation maps from different scale channels for a sample image from the class ``monkey''.}
        \label{fig:subfig-activation-stl-y}
    \end{subfigure} 
    
    \vspace{0.4cm} 
    
    \begin{subfigure}{0.9\textwidth}      
        \centering
        \raggedright
        \textit{ \hspace{0.4cm} Input image \hspace{1.3cm} Scale channel $\sigma_{0}=1/2$ \hspace{0.9cm} Scale channel $\sigma_{0}=1$ \hspace{1.1cm} Scale channel $\sigma_{0}=2$}
        \raisebox{1.8mm}{\includegraphics[width=0.178\textwidth]{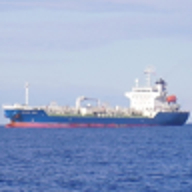}}
        \hspace{2.3mm}
        \includegraphics[width=0.25\textwidth]{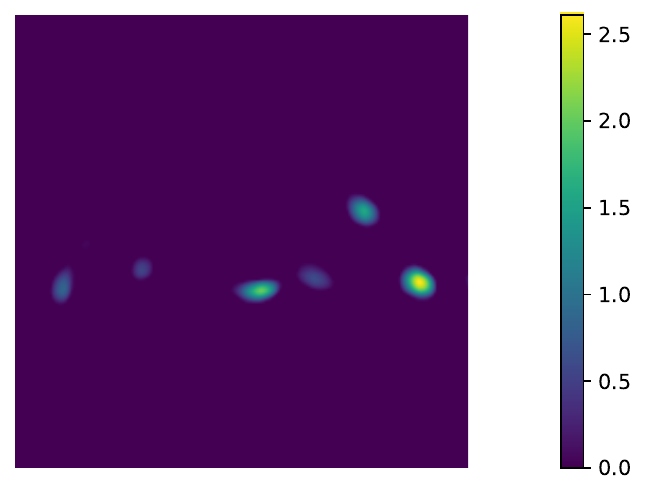}
        \hspace{0.5mm}
        \includegraphics[width=0.248\textwidth]{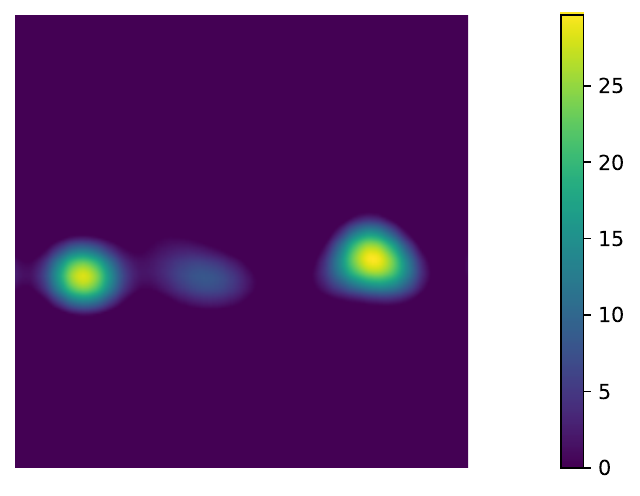}
        \hspace{2mm}
        \includegraphics[width=0.257\textwidth]{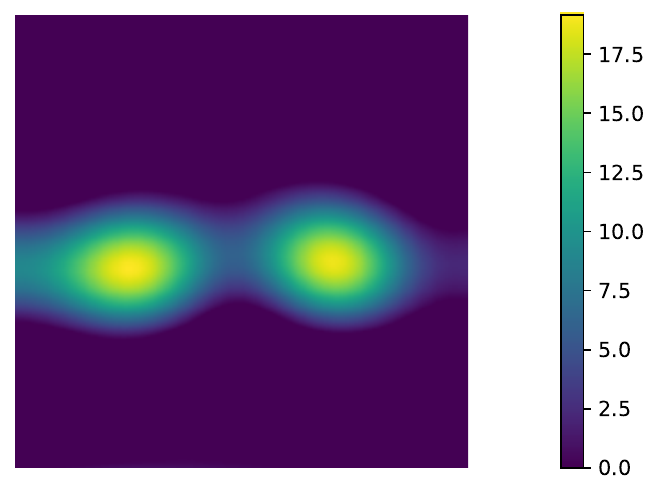}
        \caption{Activation maps from different scale channels for a sample image from the class ``ship''.}
        \label{fig:subfig-activation-stl-z}
    \end{subfigure} 
  \end{center}
\caption{Activation maps of correctly classified images from the rescaled STL-10 dataset,
extracted from the final layer of a multi-scale-channel GaussDerResNet with a spatial 
selection stage based on spatial max pooling and scale selection stage based on average
pooling over scales. The activation maps reveal which spatial areas within the image elicit 
strong responses at the feature channel corresponding to the correct class, for scale channels 
based on initial scale values $\sigma_{0} \in \{1/2, 1, 2\}$, given test samples chosen from 
size factor 2 test set. Each row depicts results from one of the five chosen classes: 
(a)~``bird'', (b)~``car'', (c)~``dog'', (d)~``monkey'' and (e)~``ship''. As can be seen from the 
subfigures, the network typically makes a prediction by identifying characteristic parts of objects, 
either at fine or coarse scales, which correspond to strong responses at finer scale channels 
or coarser scale channels, respectively.}
\label{fig-activation-maps-stl10}
\end{figure*}

\subsection{Activation maps of networks with zero-order terms}
\label{sec-act-map-visualisation}

In this section, we will inspect the feature maps in the final 
convolution layer, often referred to as activation maps, for the 
GaussDerResNets with the computational primitives of the 
layers extended to include zero-order terms and trained 
on the rescaled STL-10 dataset using spatial max pooling as 
the spatial selection mechanism. As previously mentioned, this
enables the network to focus on objects at any location in 
the image domain.

Notably the layers in the GaussDerResNets are directly interpretable,
due to the absence of spatial subsampling operations, such as strided convolutions, 
thus enabling a direct mapping between the image structures and 
their corresponding activations at different scales.

Five illustrative correctly classified images were selected from
the classes ``bird'', ``car'', ``dog'', ``monkey'' and ``ship'', 
and chosen from the test set for the size factor 2, where there is no
mirroring in the image data. The activations were 
computed at fine, medium or coarse scales, achieved by extracting the output 
of the final layers of the scale channels $\sigma_{0} = \{1/2, 1, 2\}$, respectively, 
for the feature channel corresponding to the ground-truth class of the input image.

Figure~\ref{fig-activation-maps-stl10} depicts the generated 
activation maps, along with colour bars for the strength of the 
response. We can see that, due to the use of spatial max pooling 
as the spatial selection method, the network is capable of 
localising characteristic image structures and features at any 
spatial position within the image, even close to the boundary. 
The finer scale channels typically focus on fine scale textures 
and body/machine parts, such as paws, mouths or car parts. 
The coarser scale channels focus on coarse scale image 
structures, such as heads, torsos or larger machine parts. 
In fact, we see that the network is capable of detecting multiple maxima 
corresponding to image structures key for making a prediction, 
despite the fact that only a single global maximum is selected 
during the classification stage. Additionally, the sizes of the activation 
regions are seen to be proportionally larger for the scale channels
with larger scale values, 
as expected from the scale-covariant properties of the network.

The depicted activation maps have quite intuitive explanations.
For the bird, the medium scale channel focuses on the head, while
the fine scale focuses on the background, which may help
classification. For the car, there is a main focus on the wheels.
For the dog, the main focus is on the head and the torso. However, the
finer scale channel focuses on the other paw of the nearby dog.
For the monkey, the main focus is on the mouth. The coarser scale
focuses on the torso, whereas the fine scales focus on smaller body
parts. For the boat, the medium and the coarser scale channels focus
on the bow and the stern of the boat. The finer scale channels focus
on smaller parts of the boat.

In this way, these results illustrate how the spatial selection scheme, 
in combination with the pooling method over scale, is able to 
automatically learn and classify characteristic structures for the 
objects in the respective classes in the dataset. This ability of the 
GaussDerResNets to respond to image structures at different 
local positions and scales will be further explored in the following section.

\section{Extension to local-extended Gaussian derivative ResNets with
  local spatial and scale selection}
\label{sec-loc-extended-gaussderresnets}

Concerning the conceptual framing of the type of recognition problems
that we considered, we have so far throughout been concerned with
image classification or single object recognition,
for which a single type of classification
is to be computed for each image. The motivation for this restriction
has been because of conceptual and algorithmic simplicity and due
to the fact that the Fashion-MNIST, CIFAR-10 and STL-10,
that we have built our rescaled datasets on,
each contain a single object that is to be
recognized for each image.

\subsection{Theory for joint local spatial and scale selection in
  GaussDerResNets}

From the visualisations of the underlying
activation maps in Figure~\ref{fig-activation-maps-stl10},
that are computed in the GaussDerNets, it is, however, clear that if we replace
the global spatial pooling and the global scale pooling stages in the
GaussDerNets by more general combined local scale selection and
spatial selection methods, then such local-extended GaussDerResNets can also
be used for generating multiple hypothesis of the learned objects at
different spatial positions and at different scales.

\subsubsection{Joint spatial and scale selection using deep
  scale-space maxima}

A straightforward approach for designing such a method is
by detecting the correspondence to scale-space maxima
(Lindeberg \citeyear{Lin97-IJCV}), in terms of points that assume values
in the 3-D space, spanned by the two spatial image dimensions and the
scale dimension, that assume values that are greater than their values
neighbour at adjacent points and scales, in the deep scale-space representation
of the input image, constructed from the feature maps in the
final layers of the different scale channels:

\begin{mydefinition}
  \label{def-deep-scsp-repr}
  (Deep scale-space representation based on a GaussDerResNet)
  Given a $D$-dimensional input image $f \colon \bbbr^D \rightarrow \bbbr$,
  its deep scale-space representation
  $\Pi^c \colon \bbbr^D \times \bbbr_+ \rightarrow \bbbr$
  for the class $c$
  is the $D+1$-dimensional volume, obtained by stacking the output from
  the pure scale channels $\Lambda^c_{\sigma_{0}}(f(\cdot))(x)$ for
  the class $c$ in their final layers
  without spatial pooling in (\ref{pure-single-chan-gaussderresnet})
  for all the possible values of the initial scale parameter
  $\sigma_0 \in \bbbr_+$. Thus,
  \begin{equation}
    \label{eq-deep-scsp-def}
    \Pi^c(x;\; \sigma) = \Lambda_{\sigma}^c(f(\cdot))(x)
  \end{equation}
  for $(x;\; \sigma) \in \bbbr^D \times \bbbr_+$.
\end{mydefinition}

\noindent
In this work, we will generalize the previous notion of scale-space
extrema detection presented in Lindeberg (\citeyear{Lin97-IJCV})
and taken further in Lindeberg (\citeyear{Lin12-JMIV}, \citeyear{Lin15-JMIV})
from the previous%
\footnote{For a condensed overview of the regular scale-space extrema
  detection algorithm applied to discrete approximations of
  scale-normalized differential invariants computed from discrete image data,
  see Appendix~A in Lindeberg (\citeyear{Lin18-JMIV}).}
application to a scale-normalized differential
invariant $({\cal D}_{\text{norm}} L)(\cdot;\; \sigma)$ computed at
different scales $\sigma$ in the
scale-space representation
$L(\cdot;\; \sigma) = g(\cdot;\; \sigma) * f(\cdot)$
of any image $f(\cdot)$ to, here, being applied to the deep scale-space
representation $\Pi^c(\cdot;\; \sigma)$ of the input image $f(\cdot)$, for discrete
practical implementation constructed as the union
of the feature maps in the final layers
$F_{\sigma_{i,0},K}^c(\cdot) = \Lambda_{\sigma_{i,0}}^c(f(\cdot))$
according to (\ref{pure-single-chan-gaussderresnet}) over the set of
initial scale levels $\sigma_{i,0}$ for $i \in [1, |I|]$
in the GaussDerResNet of the discrete input image $f(\cdot)$:

\begin{mydefinition}
  \label{def-deep-scsp-max}
  (Deep scale-space maxima based on a GaussDerResNet)
  Given the deep scale-space representation
  $\Pi^c \colon \bbbr^D \times \bbbr_+ \rightarrow \bbbr$
  for the class $c$ of a $D$-dimensional input image
  $f \colon \bbbr^D \rightarrow \bbbr$
  according to (\ref{eq-deep-scsp-def}), its deep scale-space maxima are
  defined as the set of points
  \begin{equation}
    \{ (x_{\nu};\; \sigma_{\nu}) \}^c
    = \operatorname{argmaxlocal}_{(x;\, \sigma)} (\Pi^c(x;\, \sigma))
  \end{equation}
  where the operator $\operatorname{argmaxlocal}()$ applied to a function
  $f$ returns the set of local maxima in $f$ over its domain.
\end{mydefinition}

\noindent
Due to the provably scale-covariant and
translation-covariant properties of the underlying image
representations, it follows that the deep scale-space maxima extracted in
such a way will be both scale covariant and translationally covariant:

\begin{myproposition}
  \label{prop-sc-cov-deep-scsp-extr}
  (Scale covariance of deep scale-space maxima)
  Consider two images $f \colon \bbbr^D \rightarrow \bbbr$
  and $f' \colon \bbbr^D \rightarrow \bbbr$, that are related
  according to a uniform scaling transformation, such that
  $f'(x') = f(x)$ for $x' = S_x \, x$,
  where $S_x \in \bbbr_+$ is the spatial scaling factor.
  Define the deep scale-space representations $\Pi^c$ and ${\Pi^c}'$ of $f$
  and $f'$, respectively, for the class $c$ according to (\ref{eq-deep-scsp-def})
  \begin{align}
    \begin{split}
      \Pi^c(x;\; \sigma) = \Lambda_{\sigma}^c(f(\cdot))(x),
    \end{split}\\
    \begin{split}
      {\Pi^c}'(x';\; \sigma') = \Lambda_{\sigma'}^c(f'(\cdot))(x').
    \end{split}
  \end{align}
  Then, the sets of deep scale-space maxima of $\Pi^c$ and ${\Pi^c}'$,
  respectively,
  \begin{align}
    \begin{split}
      \{ (x_{\nu};\; \sigma_{\nu}) \}^c
      & = \operatorname{argmaxlocal}_{(x;\, \sigma)} (\Pi^c(x;\, \sigma)),
    \end{split}\\
    \begin{split}
      \{ (x'_{\nu};\; \sigma'_{\nu}) \}^c
      & = \operatorname{argmaxlocal}_{(x';\, \sigma')} ({\Pi^c}'(x';\, \sigma')),
    \end{split}
  \end{align}
  are related according to
  \begin{equation}
    \label{eq-transf-prop-deep-scsp-max-sc-transf}
    \{ (x'_{\nu};\; \sigma'_{\nu}) \}^c
    = \{ (S_x \, x_{\nu};\; S_x \, \sigma_{\nu}) \}^c.
  \end{equation}
\end{myproposition}

\begin{myproof}
  By the scale-covariant property (\ref{sec-sc-cov-prop-pure-sc-channels})
  of the pure scale channels in
  Theorem~\ref{prop-sc-cov-gaussderresnets},
    \begin{equation}
    \Lambda_{\sigma'}^{c}(f'(\cdot))(x') = \Lambda_{\sigma}^{c}(f(\cdot))(x),
  \end{equation}
  we have that the deep scale-space representations $\Pi^c$ and ${\Pi^c}'$
  of $f$ and $f'$, respectively, are according to the definition of the
  notion of deep scale-space representation (\ref{eq-deep-scsp-def})
  related according to
  \begin{equation}
    \label{eq-transf-prop-deep-scsp-repr}
    {\Pi^c}'(x';\; \sigma') = \Pi^c(x;\; \sigma) 
  \end{equation}
  for $x' = S_x \, x$ and $\sigma' = S_x \, \sigma$.

  If a point $(x_{\nu};\; \sigma_{\nu})$ is a deep scale-space
  maximum of $\Pi^c$, it thereby follows that the point
  $(x'_{\nu};\; \sigma'_{\nu}) = (S_x \, x_{\nu};\; S_x \, \sigma_{\nu})$
  must be a scale-space maximum of ${\Pi^c}'$, which proves the result.
\end{myproof}

\subsubsection{Using post-smoothing to suppress spurious multiple
  responses to the same underlying image structures}

While a possible complication with these operations is that there
could be multiple local responses among the deep scale-space maxima, due to
noise or other local perturbations, there are also ways of handling
such problems using established mechanisms previously explored for interest point
detection by regular scale-space extrema of differential invariants
(Lindeberg \citeyear{Lin12-JMIV}):
\begin{itemize}
\item
  by performing post-smoothing of the local feature maps prior to
  detection of scale-space extrema, and
\item
  by thresholding the scale-space extrema
  on the magnitude of the response to suppress weak
  responses.
\end{itemize}

\noindent
Crucially, by for each scale layer performing smoothing with a
Gaussian kernel, with its scale parameter proportional to the current
level of scale, the resulting post-smoothed deep scale-space
representation will also support scale covariance.

\begin{mydefinition}
  (Post-smoothed deep scale-space representation)
  Given any post-smoothing factor $\gamma > 0$, the post-smoothed deep
  scale-space representation
  $\Pi^c_{\gamma} \colon \bbbr^D \times \bbbr_+ \rightarrow \bbbr$
  of the deep scale-space representation
  $\Pi^c \colon \bbbr^D \times \bbbr_+ \rightarrow \bbbr$
  of an image $f$ is defined by smoothing each scale layer
  $\Pi^c(\cdot;\; \sigma)$ with a Gaussian kernel
  $g(\cdot;\; \sigma_{\text{post-smooth}})$
  with its scale parameter%
\footnote{When expressing the post-smoothing operation, as determined
    \label{footnote-def-gamma-tot}
  by the post-smoothing factor $\gamma$, a more direct
  parameterisation consists of instead expressing the amount of
  post-smoothing as
    $\sigma_{\text{post-smooth}} = \gamma_{\text{tot}} \, \sigma_{\text{tot}}$
  instead of as
    $ \sigma_{\text{post-smooth}} = \gamma_{0} \, \sigma_{0}$,
  with $\sigma_{\text{tot}}$ denoting the total amount of Gaussian
  smoothing used for computing the final layer in the scale channel
  with the initial scale value $\sigma_0$, where
  $\sigma_{\text{tot}} =
      \sigma_{0} \sqrt{\sum_{k = 1}^K \left( r^{k-1} \right)^2} =
      \sigma_{0} \sqrt{(r^{2K} - 1)/(r^2 - 1)}$, as further explained
      in Footnote~\ref{footnote-def-sigma-tot}.}
  $ \sigma_{\text{post-smooth}} = \gamma \, \sigma$
  proportional to the current level of scale
  \begin{equation}
    \Pi^c_{\gamma}(\cdot;\; \sigma) = g(\cdot;\; \gamma \, \sigma) * \Pi^c(\cdot;\; \sigma).
  \end{equation}
\end{mydefinition}

\begin{myproposition}
  (Deep scale-space maxima extracted from a post-smoothed deep
  scale-space representation are scale covariant)
  Given any post-smoothing factor $\gamma > 0$,
  consider two images $f \colon \bbbr^D \rightarrow \bbbr$
  and $f' \colon \bbbr^D \rightarrow \bbbr$, that are related
  according to a uniform scaling transformation such that
  $f'(x') = f(x)$ for $x' = S_x \, x$,
  where $S_x \in \bbbr_+$ is the spatial scaling factor.
  Define the deep
  scale-space representations $\Pi^c$ and ${\Pi^c}'$ of $f$
  and $f'$, respectively, for the class $c$ according to (\ref{eq-deep-scsp-def})
  \begin{align}
    \begin{split}
      \Pi^c(x;\; \sigma) = \Lambda_{\sigma}^c(f(\cdot))(x),
    \end{split}\\
    \begin{split}
      {\Pi^c}'(x';\; \sigma') = \Lambda_{\sigma'}^c(f'(\cdot))(x'),
    \end{split}
  \end{align}
  and from these representations, then, in turn, the corresponding
  post-smoothed deep scale-space representations
  $\Pi^c_{\gamma}$ and ${\Pi^c}'_{\gamma}$, respectively, according to
  \begin{align}
    \begin{split}
      \label{eq-def-post-smooth-deep-scsp-repr-in-prop}
      \Pi^c_{\gamma}(\cdot;\; \sigma)
      = g(\cdot;\; \gamma \, \sigma) * \Pi^c(\cdot;\; \sigma),
    \end{split}\\
    \begin{split}
      \label{eq-def-post-smooth-deep-scsp-repr-in-prop-prim}
      {\Pi^c}'_{\gamma}(\cdot;\; \sigma')
      = g(\cdot;\; \gamma \, \sigma') * {\Pi^c}'(\cdot;\; \sigma').
    \end{split} 
  \end{align}
  Then, the sets of deep scale-space maxima of
  $\Pi^c_{\gamma}$ and ${\Pi^c}'_{\gamma}$,
  respectively,
  \begin{align}
    \begin{split}
      \{ (x_{\nu};\; \sigma_{\nu}) \}^c
      & = \operatorname{argmaxlocal}_{(x;\, \sigma)} (\Pi^c_{\gamma}(x;\, \sigma)),
    \end{split}\\
    \begin{split}
      \{ (x'_{\nu};\; \sigma'_{\nu}) \}^c
      & = \operatorname{argmaxlocal}_{(x';\, \sigma')} ({\Pi^c}'_{\gamma}(x';\, \sigma')),
    \end{split}
  \end{align}
  are related according to
  \begin{equation}
    \{ (x'_{\nu};\; \sigma'_{\nu}) \} = \{ (S_x \, x_{\nu};\; S_x \, \sigma_{\nu}) \}.
  \end{equation}
\end{myproposition}

\begin{myproof}
  As we already proved in
  Proposition~\ref{prop-sc-cov-deep-scsp-extr}, the underlying deep
  scale-space representations transform according
  to (\ref{eq-transf-prop-deep-scsp-repr})
  \begin{equation}
    {\Pi^c}'(x';\; \sigma') = \Pi^c(x;\; \sigma) 
  \end{equation}
  under the spatial scaling transformation.
  Additionally, due to the way the post-smoothed deep scale-space
  representations are defined from the unsmoothed deep scale-space
  representations according to
  (\ref{eq-def-post-smooth-deep-scsp-repr-in-prop}) and
  (\ref{eq-def-post-smooth-deep-scsp-repr-in-prop-prim}), it follows
  that also the post-smoothed deep scale-space representations of $f$
  and $f'$ are related according to
  \begin{equation}
    {\Pi^c}'_{\gamma}(x';\; \sigma') = \Pi^c_{\gamma}(x;\; \sigma) .
  \end{equation}
  If a point $(x_{\nu};\; \sigma_{\nu})$ is a deep scale-space
  maximum of $\Pi^c_{\gamma}$, it thereby follows that the point
  $(x'_{\nu};\; \sigma'_{\nu}) = (S_x \, x_{\nu};\; S_x \, \sigma_{\nu})$
  must be a deep scale-space maximum of ${\Pi^c}'_{\gamma}$, which proves the result.
\end{myproof}

\subsubsection{Thresholding on the scale-normalised magnitudes of the
  responses}

Notably, because of the use of scale-normalized derivatives in the
Gaussian derivative weight functions (\ref{eq-gauss-der-layer}),
which lead to the transformation properties over scales of the
pure scale channels of the form
(\ref{sec-sc-cov-prop-pure-sc-channels}), it follows that the
magnitude values will be directly comparable across scales, and we can therefore use
scale-independent thresholds values, when thresholding on the magnitude
of response $\Pi^c(x_{\nu};\; \sigma_{\nu})$ for the deep scale-space
maxima.

\begin{mydefinition}
  (Magnitude thresholding of deep scale-space maxima)
  Given a threshold $\Pi_{\text{thresh}}$, magnitude thresholding of a set of
  deep scale-space maxima $\{ (x_{\nu};\; \sigma_{\nu}) \}^c$ is
  performed according to
  \begin{multline}
    \operatorname{thresh}_{\Pi_{\text{thresh}}}(\{ (x_{\nu};\; \sigma_{\nu}) \}^c) = \\
    = \{ (x_{\nu};\; \sigma_{\nu}) \colon
            \Pi_{\gamma}^c(x_{\nu};\; \sigma_{\nu}) \geq \Pi_{\text{thresh}} \}.
  \end{multline}
\end{mydefinition}

\begin{myproposition}
  (Magnitude thresholding of deep scale-space maxima is scale
  invariant)
  Consider two images $f \colon \bbbr^D \rightarrow \bbbr$
  and $f' \colon \bbbr^D \rightarrow \bbbr$, that are related
  according to a uniform scaling transformation such that
  $f'(x') = f(x)$ for $x' = S_x \, x$,
  where $S_x \in \bbbr_+$ is the spatial scaling factor,
  and define the sets of deep scale-space maxima
  $\{ (x_{\nu};\; \sigma_{\nu}) \}^c$ and $\{ (x'_{\nu};\; \sigma'_{\nu}) \}^c$
  from the post-smoothed deep scale-space representations
  $\Pi_{\gamma}^c \colon \bbbr^D \times \bbbr_+ \rightarrow \bbbr$ and
  ${\Pi'}_{\gamma}^c \colon \bbbr^D \times \bbbr_+ \rightarrow \bbbr$ of
  $f$ and $f'$, respectively, for the class $c$.
  Then, the results of performing magnitude thresholding on these sets
  of deep scale-space maxima are in perfect correspondence,
  i.e.\ the respective sets of thresholded deep scale-space maxima
  \begin{align}
    \begin{split}
      \{ (x_{\mu};\; \sigma_{\mu}) \}^c =
      \operatorname{thresh}_{\Pi_{\text{thresh}}}(\{ (x_{\nu};\; \sigma_{\nu}) \}^c),
    \end{split}\\
    \begin{split}
     \{ (x'_{\mu};\; \sigma'_{\mu}) \}^c =
      \operatorname{thresh}_{\Pi_{\text{thresh}}}(\{ (x'_{\nu};\; \sigma'_{\nu}) \}^c),
    \end{split}
  \end{align}
  do perfectly match according to
  \begin{equation}
    (x'_{\mu};\; \sigma'_{\mu}) = (S_x \, x_{\mu};\; S_x \, \sigma_{\mu}).
  \end{equation}
\end{myproposition}

\begin{myproof}
  Follows directly from the scale-invariant property of the magnitude
  responses of the pure scale channels under scaling transformations
  according to (\ref{sec-sc-cov-prop-pure-sc-channels})
  \begin{equation}
    \Lambda_{\sigma'}^{c}(f'(\cdot))(x') = \Lambda_{\sigma}^{c}(f(\cdot))(x),
  \end{equation}
  which also carries over to the post-smoothed deep scale-space
  representations, because of their scale-covariant properties.
\end{myproof}

\noindent  
Indeed, a common approach for ranking regular interest points on
significance is by sorting them in reverse order relative to the
magnitude of the response. A corresponding ranking method can also be
applied to deep scale-space maxima $\{ (x_{\nu};\; \sigma_{\nu}) \}$,
by sorting them on the their magnitude responses.

With such extensions, it should therefore be feasible to express
algorithms for performing joint local spatial and scale selection
in such local-extended GaussDerResNets, by detecting deep scale-space
maxima in post-smoothed deep scale-space representations.

\subsubsection{Spatial circle/sphere representations of the deep
  scale-space maxima}

A possible issue regarding such extraction of object candidates, is
that there could still be multiple responses at different points
$(x_{\nu};\; \sigma_{\nu})^c$ corresponding to the same underlying
physical object in the deep scale-space maxima extracted from
the post-smoothed deep scale-space representation.
By adapting criteria used in the methodology for linking blob-like
image features across
scales described in Section~9.2.3 in Lindeberg (\citeyear{Lin93-Dis}),
it is, however, possible to formulate criteria for merging multiple
responses to the same underlying image structures, based on measurements
of their scale-space characteristics in relation to their relative
overlap, that operate on spatial surround regions defined in the following way:

\begin{mydefinition}
  \label{def-circle-rep-dscsp-max}
  (Circle/sphere representation of deep scale-space maxima)
  Given a scale-space maximum  $(x_{\nu};\; \sigma_{\nu})^c$
  extracted from the deep scale-space representation
  $\Pi^c$ for the class $c$ of any image $f$, its visualisation
  circle/sphere representation
  over the $D$-dimensional spatial image domain of $f$ is given
  by the circle/sphere
  \begin{equation}
    \label{eq-circ-repr-deep-scsp-max}
    \sum_{p = 1}^D (x_p - x_{\nu,p})^2 = R^2 \, \sigma_{\nu,\text{tot}}^2,
  \end{equation}
  where
  \begin{itemize}
  \item
    $p \in [1, D]$ denotes the indices for the spatial dimensions,
  \item
    $x = (x_1, x_2, ..., x_D)$, as indexed by $p$, represents the free
    spatial points over the spatial image domain,
  \item
    $\sigma_{\nu,\text{tot}}$ represents to total amount of Gaussian
    smoothing that has been used for computing the final layer for the
    scale channel with the initial scale level $\sigma_{\nu}$,
    which is defined from the additive property%
\footnote{Because of the semi-group property of the Gaussian kernel,
  \label{footnote-def-sigma-tot}
  the result of coupling all the Gaussian kernels
  $g(\cdot;\; \sigma_k)$ ($k = [1..K]$), used for reaching the
  representation in the final layer, in cascade corresponds to a
  Gaussian kernel $g(\cdot;\; \sigma_{\text{tot}})$
  with standard deviation
  $\sigma_{\text{tot}} = \sqrt{\sum_{k=1}^K \sigma_k^2}$.
  Note that for this specific reasoning, we ignore both the effects of the
  spatial differentiation operators, the activation functions and the
  batch normalisation stages in the layers of the network.}
    of the squares
    of the respective scale parameters in the layers
    according to (\ref{eq:sigma-geometric}) 
    \begin{equation}
      \label{eq-sigma-tot}
      \sigma_{\nu,\text{tot}} =
      \sigma_{\nu} \sqrt{\sum_{k = 1}^K \left( r^{k-1} \right)^2} =
      \sigma_{\nu} \sqrt{\frac{r^{2K} - 1}{r^2 - 1}},
    \end{equation}
  \item
    and the relative radius factor $R \in \bbbr_+$ is a parameter.
  \end{itemize}
\end{mydefinition}

\noindent
Due to the scale-covariant properties of the underlying deep
scale-space maxima, these visualisation circle/sphere
representations will follow any scaling transformations over the image
domain, and thereby appear as if they would be attached to the
objects in the image data $f$ under spatial scaling transformations,
analogous to the corresponding properties for regular scale-space as
extrema of differential invariants $({\cal D}_{\text{norm}} L) (\cdot;\; \sigma)$
previously demonstrated in Bretzner and Lindeberg (\citeyear{BL97-CVIU})
and Lindeberg (\citeyear{Lin97-IJCV}).
Thereby, matching relations defined based on such
circle/sphere representations of deep scale-space maxima
will specifically be scale invariant.

\subsubsection{Combined method for generating object hypotheses from deep
  scale-space maxima}

The overall approach for computing joint local deep interest points and
their scales could also be complemented with extending
such pointwise features in the resulting 3-D deep scale space
to regions in the joint 3-D deep scale space, as determined by competition
between neighbouring other features (Lindeberg \citeyear{Lin92-IJCV})
and stability over scales as registered by linking similar features
across scales
(Lindeberg \citeyear{Lin92-IJCV}, \citeyear{Lin12-JMIV}, \citeyear{Lin15-JMIV}).

In this way, and
considering the in Section~\ref{sec-rel-int-pt-det} described close
structural relationships to classical methods for interest point
detection, such a local-extended GaussDerResNet should be able
to serve as a substantially generalized
version of an interest point detector that, in contrast to previous
applications of interest point detectors for detecting more point-like
image features like corner and blobs at different locally determined
image positions and scales, instead detects object
candidates for the types of objects that the network has been trained on,
and specifically for each such object candidate outputs both a local
image position/region of interest and a local spatial scale for recognising
that specific object.

\subsection{Scale-invariant and order-invariant methods for suppressing
  multiple responses of deep scale-space maxima to the same object}
\label{sec-methods-merge-mult-responses}

Despite the use of post-smoothing to reduce the number of multiple
responses of deep scale-space maxima corresponding to the
same physical object in the image, there may, however, still be a
number of multiple responses to the same object.

To reduce such multiple responses further, we will in this section formulate a
set of criteria that can be used for this purpose:

\begin{mydefinition}
  \label{def-raw-match-cand}
  (Raw merging candidates for two different deep scale-space maxima of the
  same class)
  Given two deep scale-space maxima
  $P_{\nu_1}^c = (x_{\nu_1};\; \sigma_{\nu_1})^c$
  and $P_{\nu_2}^c = (x_{\nu_2};\; \sigma_{\nu_2})^c$
  extracted from the deep scale-space representation
  $\Pi^c$ of any image $f$ for the same%
\footnote{The reason for only attempting to merge multiple responses
  belonging to the {\em same\/} class, is that the merging algorithm should
  not in any way attempt to make decisions between whether an object
  candidate should be classified as belonging to either of two
  conflicting interpretations. Instead, those decisions should
  be passed on to later processing stages in the visual hierarchy,
  in such a way that the resulting module for
  generating object hypotheses should allow for multiple object
  hypotheses of different classes being present for the same
  underlying image structures in the image domain.}
  class $c$,
  these deep scale-space maxima are regarded as candidates to be
  merged
  \begin{itemize}
  \item
    if the spatial point $x_{\nu_1}$ is within the visualisation
    circle/sphere (\ref{eq-circ-repr-deep-scsp-max})
    of the deep scale-space maximum
    $(x_{\nu_2}; \sigma_{\nu_2})^c$, and 
  \item
    if the spatial point $x_{\nu_2}$ is within the visualisation
    circle/sphere  (\ref{eq-circ-repr-deep-scsp-max})
    of the deep scale-space maximum
    $(x_{\nu_1}; \sigma_{\nu_1})^c$,
  \end{itemize}
  thereby, if the merge is accepted given any set of complementary
  additional criteria for accepting the merging of merging candidates,
  keeping only the deep scale-space maximum with the
  strongest magnitude
  \begin{equation}
    \nu_{\text{keep}}
    = \argmax_{\nu \in \{ \nu_1, \nu_2 \}} \Pi^c(x_{\nu};\; \sigma_{\nu}).
  \end{equation}
\end{mydefinition}

\noindent
Concerning the suppression of the multiple responses to a single
object, in situations when a deep scale-space maximum has more than
one raw match candidate, the result of a merging process
based on a direct application of Definition~\ref{def-raw-match-cand} 
may, however, depend upon the order in which it is carried out,
unless special precaution is taken.
If that would the case for a given application scenario, then the
selection of the obtained features after suppression of multiple
responses may to a certain extent become arbitrary and less robust.

What one can do then, is to formulate additional complementary matching
criteria, that do not depend on the ordering of the features,
that is not depending upon the ordering by which successive merges
are carried out.

\subsubsection{Minimal size clusters of raw matching candidates}

What one could do first, is that given the pairwise raw matching 
candidates, obtained from the circle/sphere criterion in
Definition~\ref{def-raw-match-cand},
first cluster the detected deep scale-space maxima into groups of
deep scale-space maxima, in such a way that no matching candidate
has a raw matching candidate relation pointing to a
feature outside the group of features that point to each other.

Such clusters can be computed by picking one scale-space maximum from
each cluster, preferably the one with its strongest magnitude value,
and then in a recursive way follow all the successive raw merging 
candidates that can be reached from that seed. Thereby, we obtain the
first cluster of minimum size.

Within the set of remaining deep scale-space maxima, not part of the
first cluster, we then initiate a corresponding search using the
remaining deep scale-space maximum having the strongest magnitude,
and in a recursive way follow all the remaining raw merging candidates
that can then be reached from the seed. In that way, we obtain the
second cluster.

This procedure should then be repeated until there are no more
remaining deep scale-space maxima that have not yet been assigned to a
cluster:

\begin{mydefinition}
  (Minimum size clusters of spatially coinciding deep scale-space maxima from the
  same class)
  Given a set of deep scale-space maxima
  $\{ P_{\nu}^c \} = \{ (x_{\nu};\; \sigma_{\nu})^c \}$
  computed from the deep scale-space representation $\Pi^c$
  of an image $f$ for the same class $c$, form an
  equivalence relation $P_{\nu_1}^c \sim P_{\nu_2}^c$ between
  any pair $(P_{\nu_1}^c, P_{\nu_2}^c)$ of deep scale-space maxima
  that constitute raw matching candidates to each other according
  to Definition~\ref{def-raw-match-cand}.
  The set of disjunct equivalence classes formed by resolving these
  equivalence relations is referred to as the set of clusters of
  spatially coinciding deep scale-space maxima.
\end{mydefinition}
Such relationships can, in turn, be computed efficiently by algorithms
for resolving equivalence classes.

\subsubsection{Merging multiple responses within each cluster
  using the strongest magnitude response in each cluster}

Given such a cluster, computed according to the process described
above, a straightforward brute-force approach consists of picking the
deep scale-space maximum having the strongest response within
each cluster.

Clearly, such an approach leads to combined merging approach that does not depend 
on the ordering of the deep scale-space maxima, or the order 
by which the merging relations are performed.

\subsubsection{Merging multiple responses within each cluster
  using pairwise mutually best-attraction matching based on
  scale-normalized distances}

In case the above approach would in experiments turn out to lead
to a certain leakage between multiple objects in the image of
the same class, in such a way that the responses due to two physically
different objects in the world, while belonging to the same class, however,
would be accidentally merged by the above resulting merging algorithm,
then one can complement the above
approach with following additional criteria:
\begin{mydefinition}
  (Scale-normalized distance between deep scale-space maxima)
  Given two deep scale-space maxima
  $P_{\mu} = (x_{\mu};\; \sigma_{\mu})$
  and $P_{\nu} = (x_{\nu};\; \sigma_{\nu})$,
  their scale-normalised distance $d(P_{\mu}, P_{\nu})$ is defined as%
\footnote{Note that with this definition, the relative radius factor
  $R$ in the circle/sphere representation
  (\ref{eq-circ-repr-deep-scsp-max}) corresponds to a corresponding
  threshold on the scale-normalized distance $d(P_{\mu}, P_{\nu})$ in
  situations where the scale parameters of the two deep scale-space
  maxima are equal $\sigma_{\mu} = \sigma_{\nu}$.}
  \begin{equation}
    \label{eq-sc-norm-distance}
    d(P_{\mu}, P_{\nu}) =
    \sqrt{\frac{\sum_{p=1}^D (x_{\mu,p} - x_{\nu,p})^2}{\sigma_{\mu} * \sigma_{\nu}}},
  \end{equation}
  where the numerator denotes the Euclidean distance 
  between the spatial points $x_{\mu}$ and $x_{\nu}$, and which is
  then normalized with respect to the geometric mean of the respective
  scale parameters in the denominator.
\end{mydefinition}

\begin{myproposition}
  \label{prop-sc-inv-of-sc-norm-dist-deep-scsp-max}
  (Scale invariance of the scale-normalized distance measure)
  The scale-normalized distance (\ref{eq-sc-norm-distance})
  between any two deep scale-space maxima computed
  from the same image is scale-invariant.
\end{myproposition}

\begin{myproof}
  Follows directly from the transformation property
  (\ref{eq-transf-prop-deep-scsp-max-sc-transf}) of
  deep scale-space maxima, whereby the influence of the spatial scale
  factor $S_x$ disappears by the division operator.
\end{myproof}

\begin{mydefinition}
  (Mutual best-attraction matching of deep scale-space maxima)
  If two scale-space maxima $P_{\nu_1} = (x_{\nu_1};\; \sigma_{\nu_1})$
  and $P_{\nu_2} = (x_{\nu_2};\; \sigma_{\nu_2})$ in a set of deep
  scale-space maxima obey mutual best-attraction matching
  relative to the scale-normalised distances $d(P_{\nu_1}, P_q)$ and
  $d(P_{\nu_2}, P_q)$ in relation to all the other deep scale-space
  maxima $P_{q} = (x_{q};\; \sigma_{q})$ over the indices $q$ in the set
  in the following way:
  \begin{itemize}
  \item
    if $P_{\nu_1}$ has the nearest scale-normalised distance
    to $P_{\nu_2}$, compared to all the other deep scale-space maxima
    $P_{q}$  in the set, such that
    \begin{equation}
      d(P_{\nu_1}, P_{\nu_2}) \leq  d(P_{\nu_1}, P_q)
    \quad \forall q \neq \nu_1, \nu_2,
    \end{equation}
    and
  \item
    if $P_{\nu_2}$ has the nearest scale-normalised distance to $P_{\nu_1}$,
    compared to all the other deep scale-space maxima $P_{q}$ in the
    set, such that
       \begin{equation}
         d(P_{\nu_2}, P_{\nu_1}) \leq  d(P_{\nu_2}, P_q)
       \quad \forall q \neq \nu_1, \nu_2,
    \end{equation}
  \end{itemize}
  then we say that the deep scale-space maxima
  $P_{\nu_1}$ 
  and $P_{\nu_2}$ 
  obey a mutual best-attraction matching relation.%
\footnote{In this treatment, we do not explicitly handle the singular
  case that arises if the minimum distance would be obtained for multiple deep
  scale-space maxima, which is, however, unlikely to occur for real-world data,
  if the computations are performed with sufficient numerical
  precision. Such singular matching situations could, however, be
  handled by merging the set of multiple deep scale-space maxima
  that lead to the same minimum distance into mini-clusters,
  and then treating those mini-clusters as the primitives in
  the merging algorithm.}
\end{mydefinition}

\noindent
Thereby, it is also safe to merge these features, without the result
depending on the order of the merges.

By applying this methodology iteratively, we can then
perform a merging, that is independent of the numbering
of the features as well as by the ordering by which the merging
is performed:

\begin{myproposition}
  (Scale-invariant and order-invariant merging algorithm in any
  minimum-size cluster of spatially coinciding deep scale-space maxima,
  as clustered together based on raw merging candidates)
  A combined merging procedure applied to minimum-size
  clusters of spatially coinciding deep scale-space maxima
  and consisting of the following steps:
  \begin{itemize}
  \item
    compute all mutual best-attraction matches,
  \item
    perform those merges,
  \item
    compute a new set of best attraction matches,
    merge those, and
  \item
    iterate,
  \end{itemize}
  is both scale invariant and order invariant.
\end{myproposition}

\begin{myproof}
  Follows from the scale-invariant property of the scale-normalized
  distance between any pair of deep scale-space maxima under scaling
  transformations, according to
  Proposition~\ref{prop-sc-inv-of-sc-norm-dist-deep-scsp-max},
  in combination with the order-invariant property of any mutual
  best-attraction merging method that is based on an unambiguous
  formulation of the pairwise best-attraction matching candidates.
\end{myproof}

\subsubsection{Composed method for generating object hypotheses in
  multi-object detection scenarios}

In this way, we thereby obtain a both scale-invariant and
order-invariant method for suppressing multiple responses
to the same object from the set of deep scale-space maxima
computed from any image.

Notably, the resulting method for detecting object candidates in a
multi-object scenario only comprises three parameters to tune:
(i) the relative post-smoothing factor $\gamma$,
(ii)~the threshold $\Pi_{\text{thresh}}$ on the magnitude of the response,
and (iii)~the relative radius factor $R$ for determining the radii of the
circle/sphere representations (\ref{eq-circ-repr-deep-scsp-max}), given 
the total amount of smoothing $\sigma_{\nu,\text{tot}}$
according to (\ref{eq-sigma-tot}), in turn, given the scale level
$\sigma_{\nu}$ at which each deep scale-space maximum has been
detected.

\begin{figure*}[hbtp]
  \begin{center}
    \begin{tabular}{cccc}
      \textit{Input image}
        & \textit{10 strongest maxima}
        & \textit{Intra-class overlaps merged}
        & \textit{Cross-class overlaps absorbed} \\
      \includegraphics[width=0.22\textwidth]{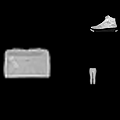}
        & \includegraphics[width=0.22\textwidth]{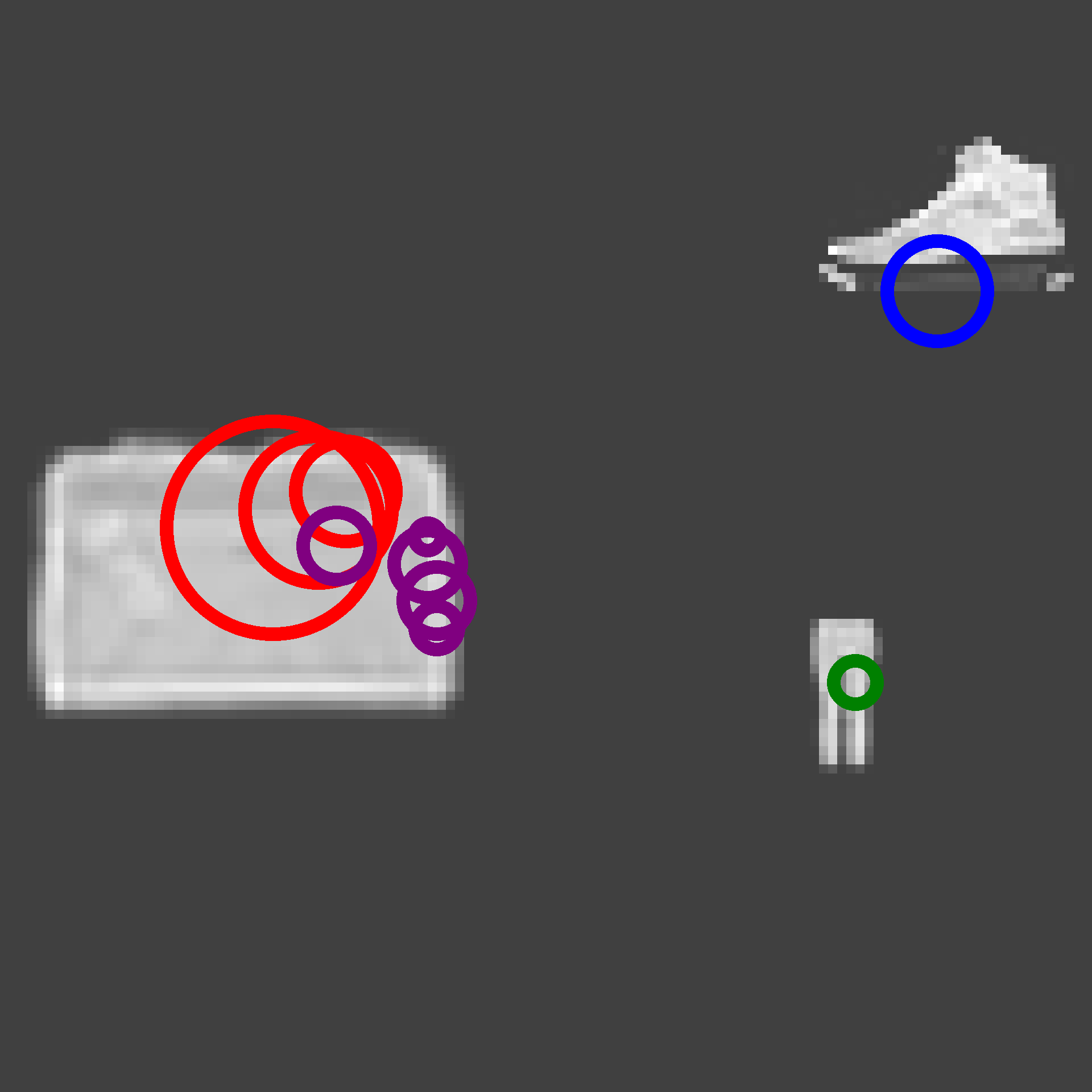}
        & \includegraphics[width=0.22\textwidth]{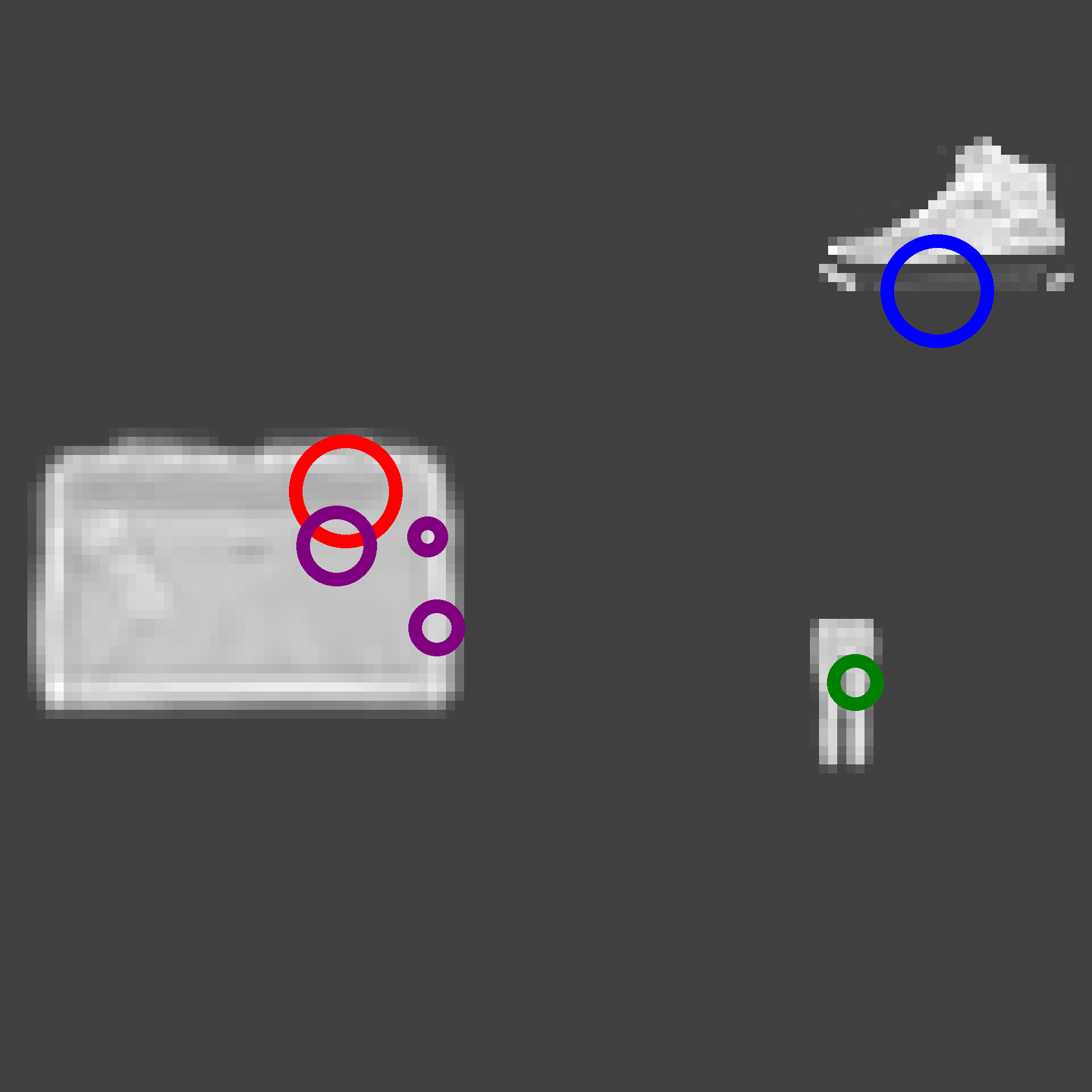}
        & \includegraphics[width=0.22\textwidth]{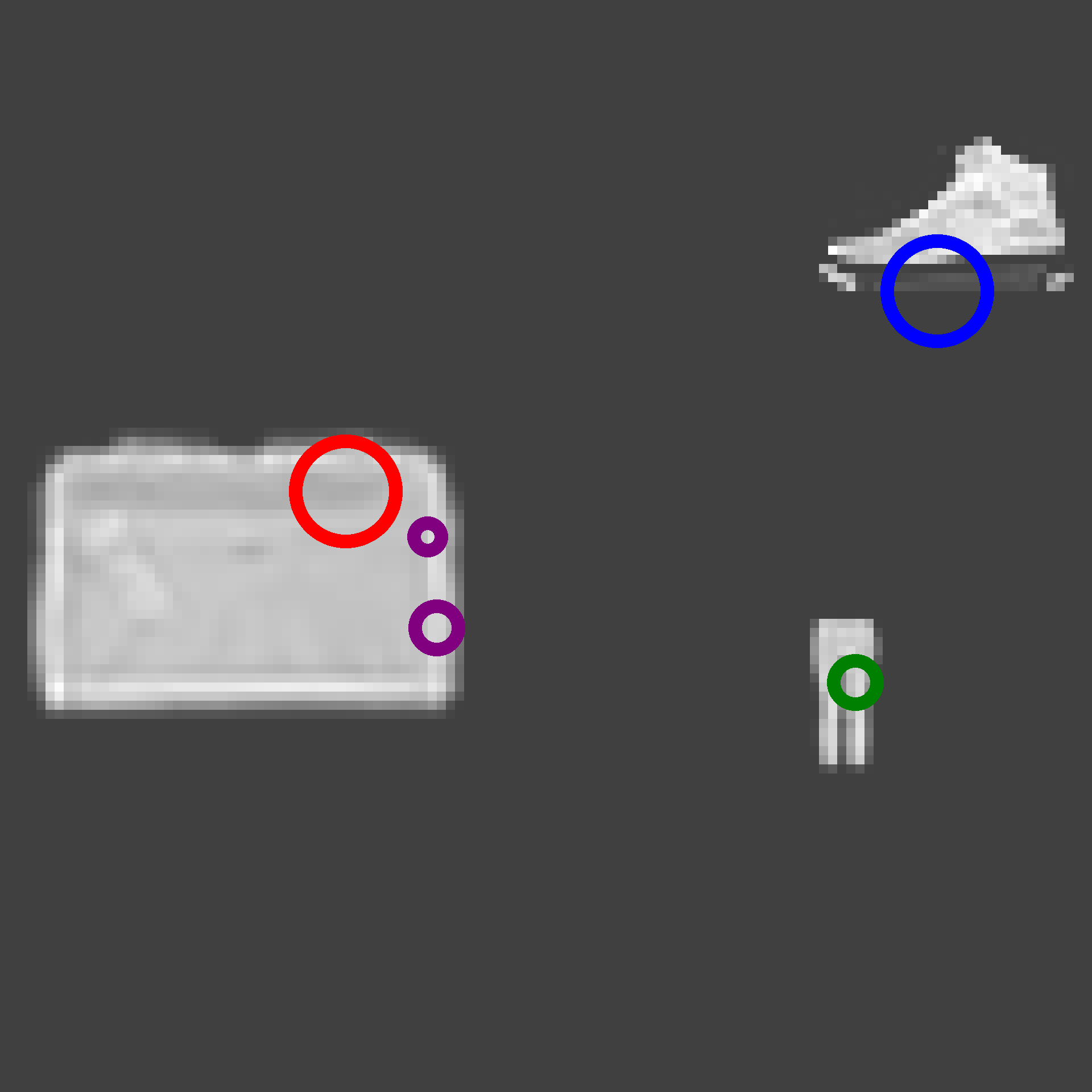}
    \end{tabular}
  \end{center}
  \caption{Illustration of the main steps in the generation of object
    hypotheses from deep scale-space maxima extracted from the deep
    scale-space representations for each class, for the given multi-object 
    test image generated out of Fashion-MNIST samples of different sizes.
    (a)~Original image.
    (b)~Circle representations of the 10 strongest deep scale-space
    maxima, with the radius proportional to the selected scale for
    each deep scale-space maximum, and using a relative post-smoothing 
    scale of $\gamma_{\text{tot}} = 0.2$ before the detection of
    deep scale-space maxima.
    (c)~The result of suppressing overlapping spatial
    representations within each class.
    (d)~The result of suppressing overlapping spatial
    representations across all the classes, preserving only the
    spatial representation having the strongest magnitude value, 
    purely for visualisation purposes.
    From the resulting circles we can see that the network produces 
    maxima over the objects, no matter the object scale.
    The colours of the circles indicate the classes of the object
    hypotheses, with \textcolor{red}{red = ``bag''}, 
    \textcolor{green!60!black}{green = ``trouser''},
    \textcolor{blue}{blue = ``sneaker''} and 
    \textcolor{violet}{purple for all the other classes}.
    The underlying image data in (b)--(d) have
    been brightened to improve clarity.}
  \label{fig-multi-obj-fashion-standard-main}
\end{figure*}

\begin{figure*}[hbt]
  \begin{center}
    \begin{subfigure}{0.99\textwidth}      
      \begin{tabular}{cccc}
        \textit{Input image}
          & \textit{10 strongest maxima}
          & \textit{Intra-class overlaps merged}
          & \textit{Cross-class overlaps absorbed} \\
        \includegraphics[width=0.22\textwidth]{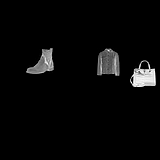}
          & \includegraphics[width=0.22\textwidth]{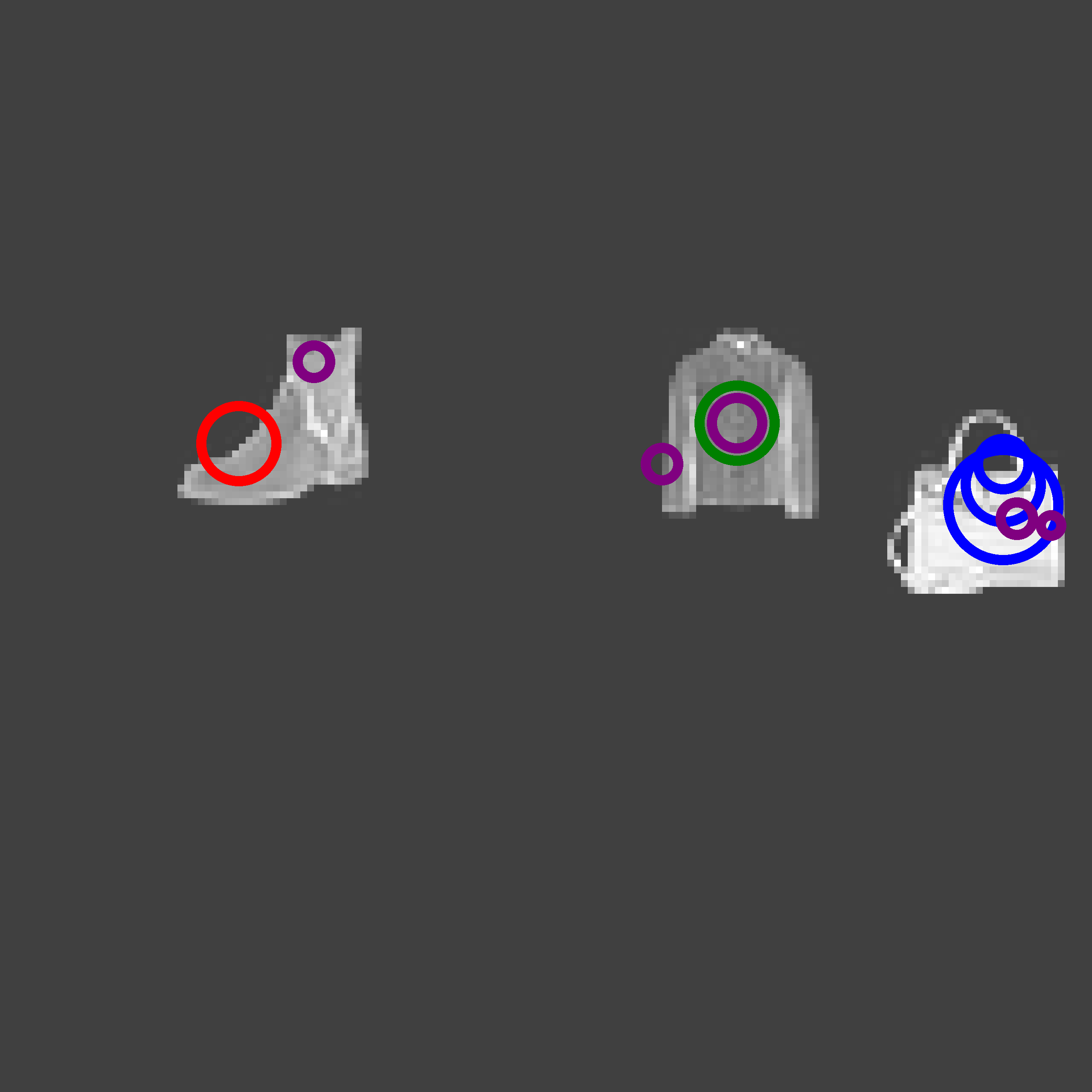}
          & \includegraphics[width=0.22\textwidth]{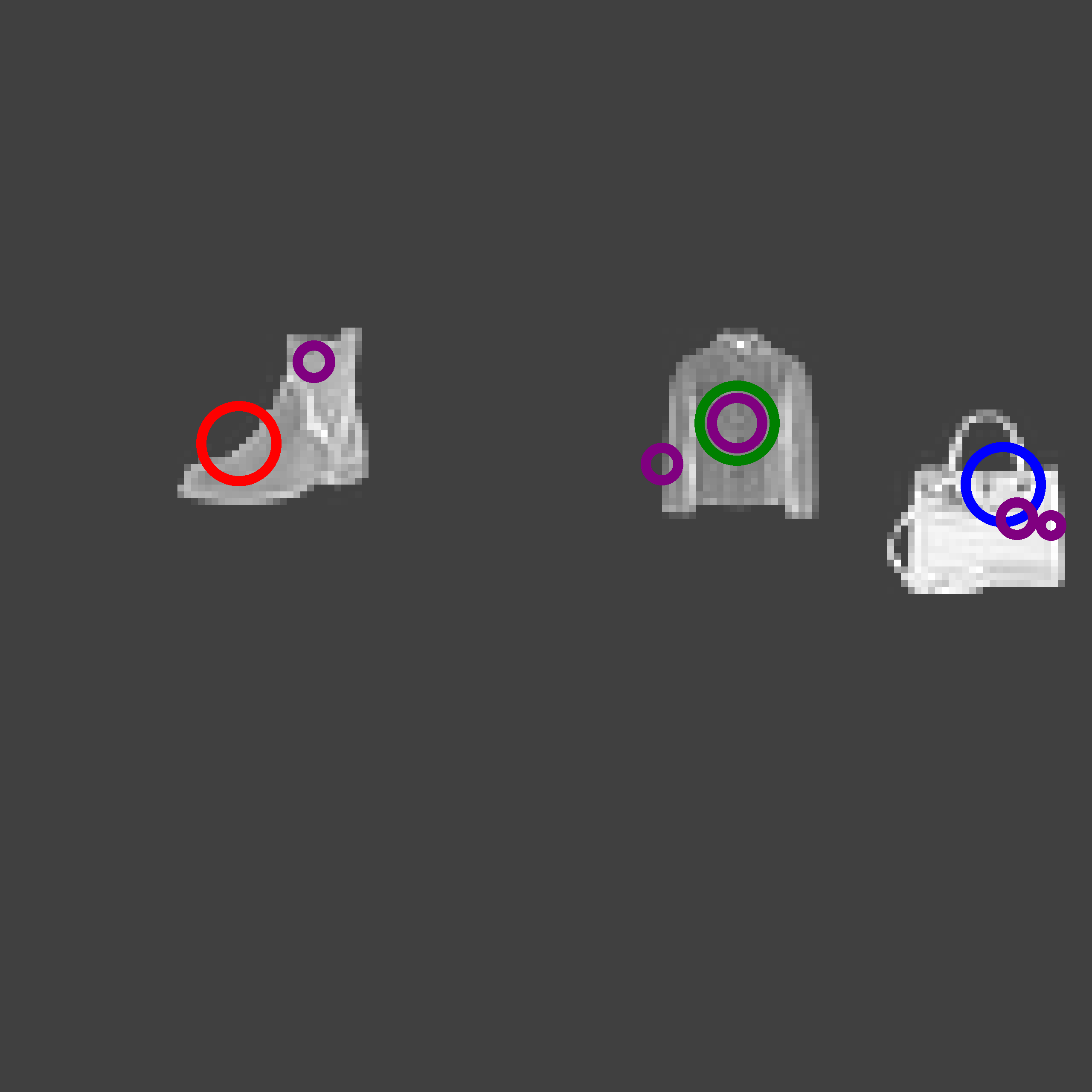}
          &  \includegraphics[width=0.22\textwidth]{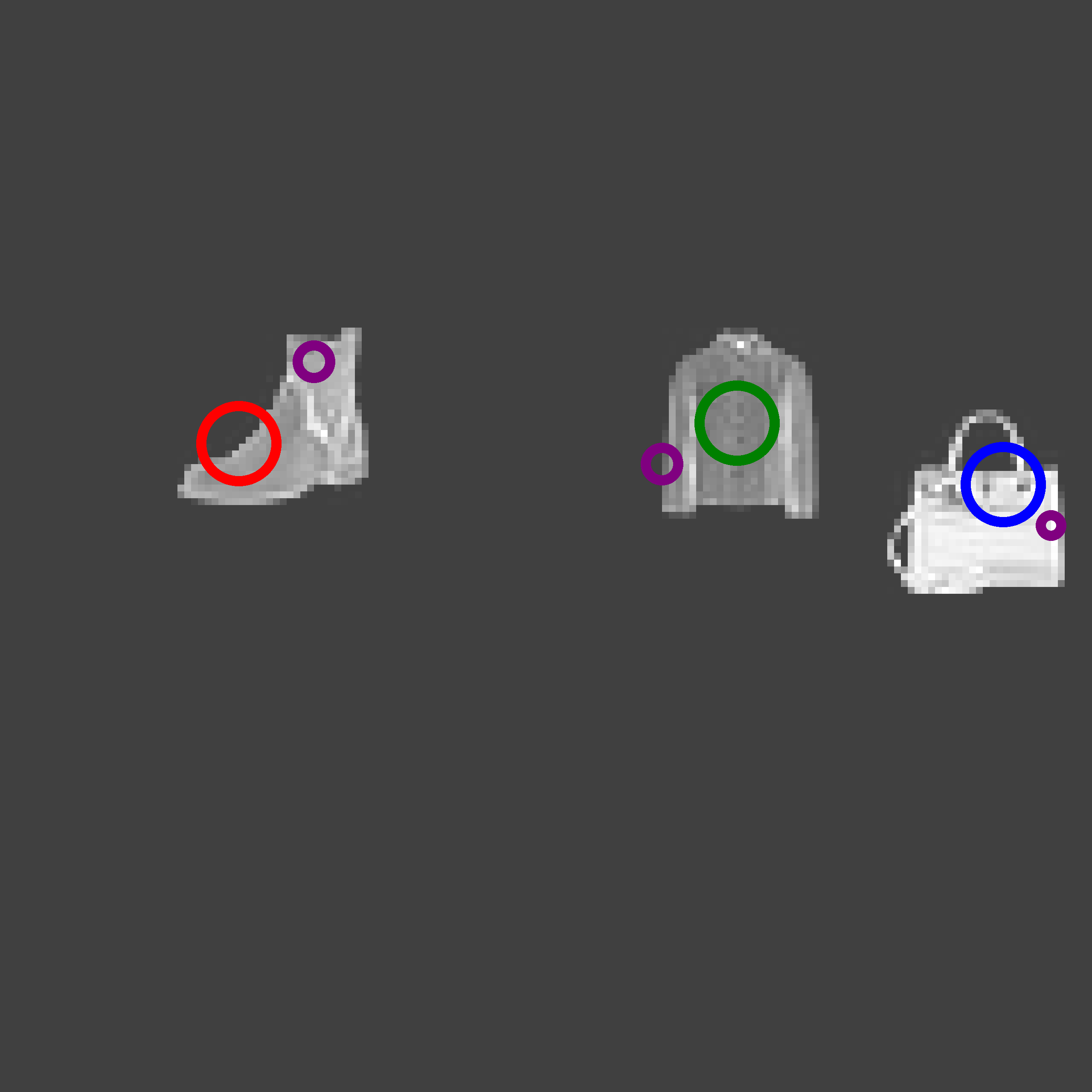}
      \end{tabular}
      \caption{Image data for the spatial size factor 1, with the
        predicted object hypotheses overlaid}
      \label{fig:subfig-mult-obj-sc-cov-sf1}
    \end{subfigure}

    \bigskip
    
    \begin{subfigure}{0.99\textwidth}      
      \begin{tabular}{cccc}
        \textit{Input image}
          & \textit{10 strongest maxima}
          & \textit{Intra-class overlaps merged}
          & \textit{Cross-class overlaps absorbed} \\
        \includegraphics[width=0.22\textwidth]{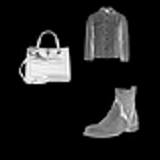}
          & \includegraphics[width=0.22\textwidth]{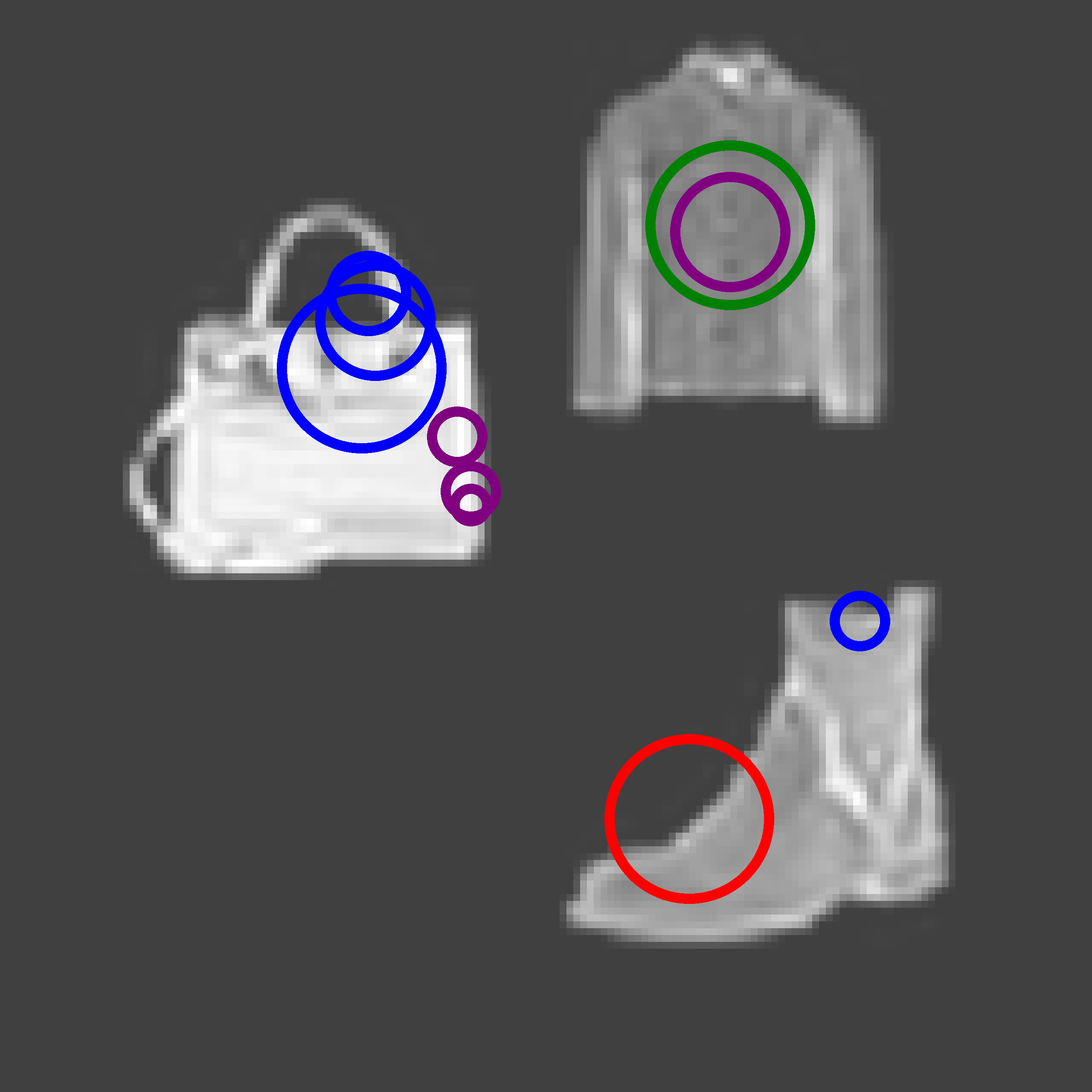}
          & \includegraphics[width=0.22\textwidth]{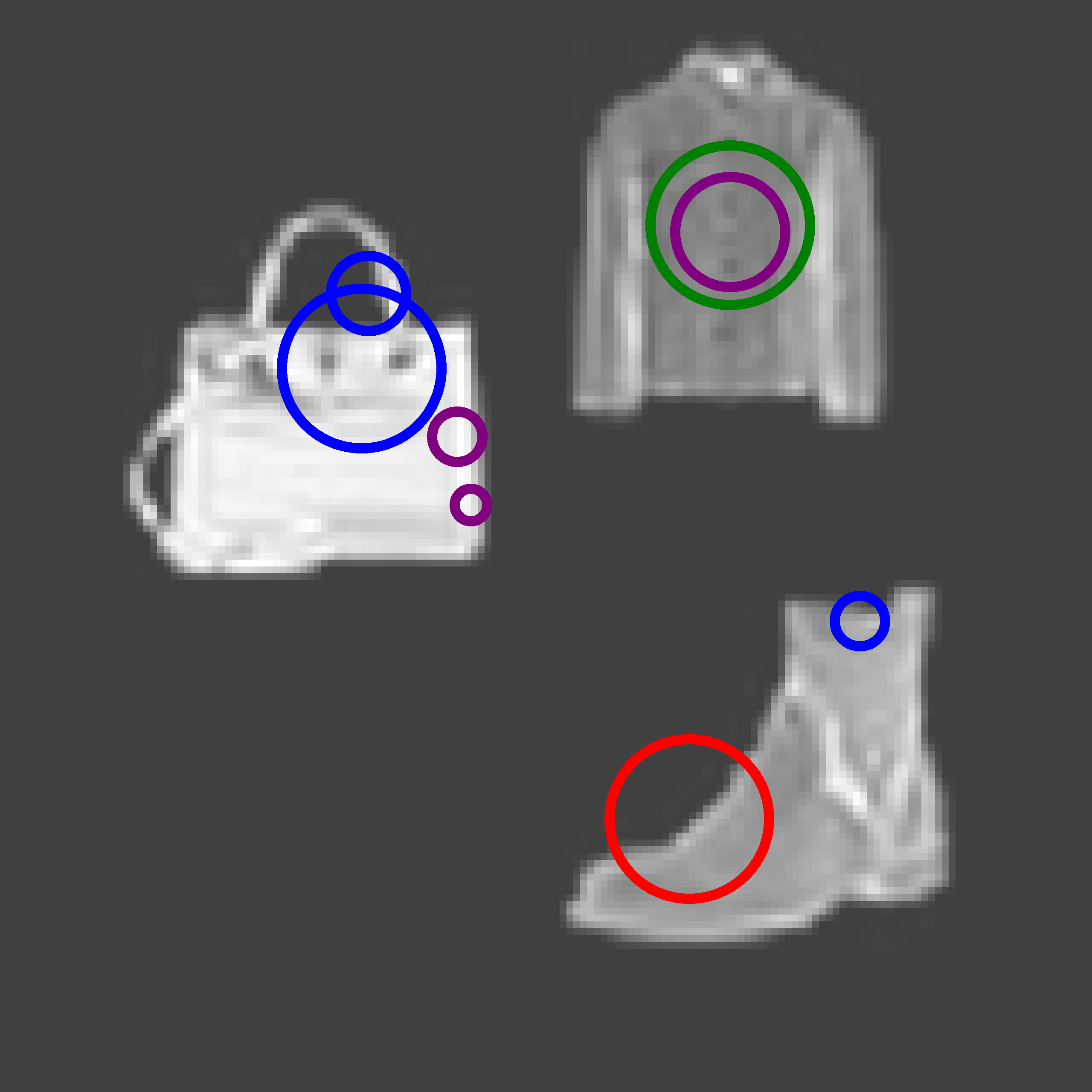}
          & \includegraphics[width=0.22\textwidth]{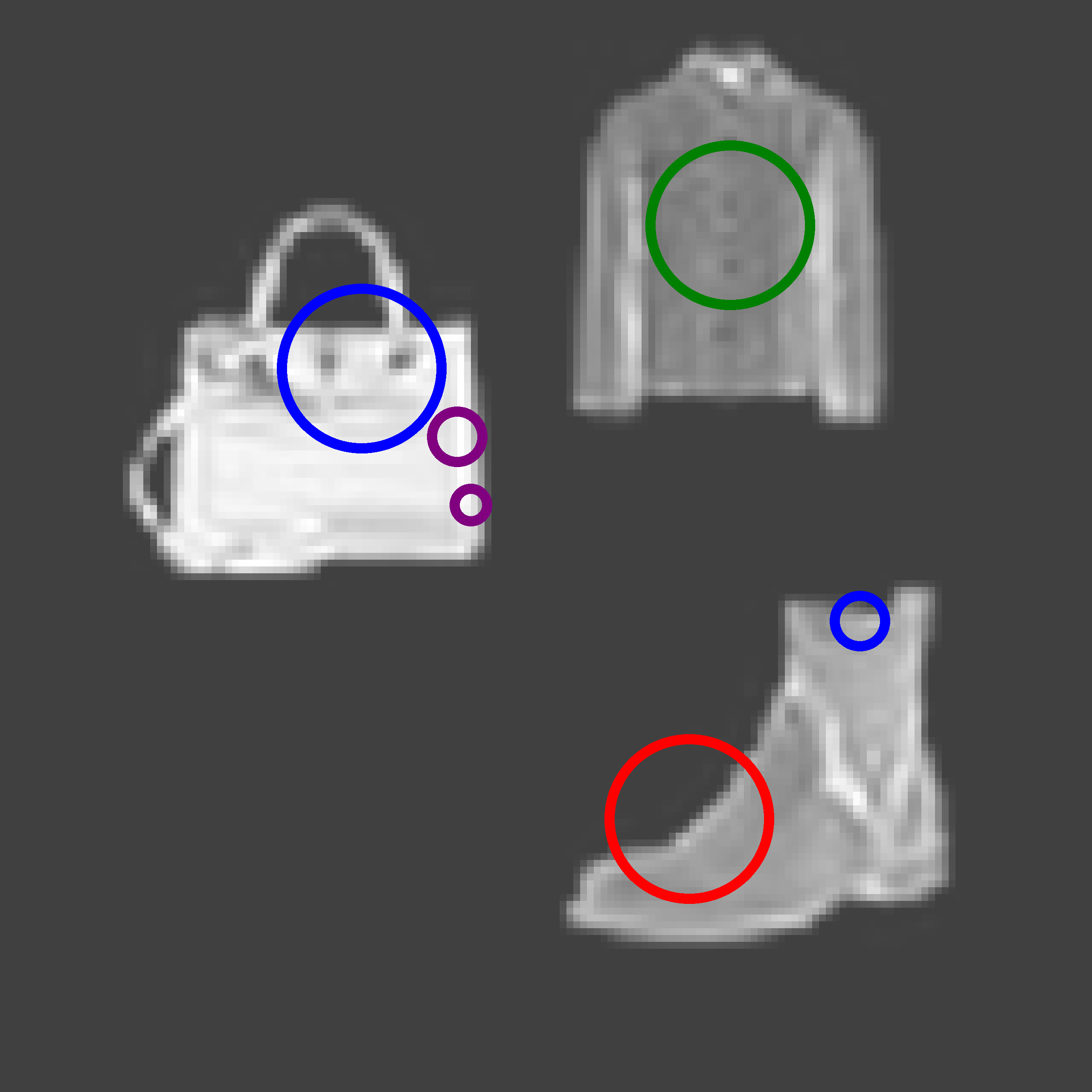}
      \end{tabular}
      \caption{Corresponding image data for the spatial size factor 2, with the predicted object hypotheses overlaid}
      \label{fig:subfig-mult-obj-sc-cov-sf2}
    \end{subfigure} 
  \end{center}
  \caption{Illustration of the scale-covariant and scale-invariant
    properties of the resulting method for generating object
    hypotheses from deep scale-space maxima from the deep scale-space
    representation obtained by combining the final layers for the
    different values of the initial scale parameter for each class.
    In the top row, images of three objects from the Fashion-MNIST
    test set have been put at random positions for the size factor 1.
    In the bottom row, the same objects have been put at other random
    positions for the size factor 2.
    As can be seen from the illustrations, the sizes of the circles largely
    follow the amount of spatial scaling of the input data by the
    factor of 2 between the first and the second rows.
    The colours of the circles indicate the classes of the object
    hypotheses, with \textcolor{red}{red = ``ankle boot''}, 
    \textcolor{green!60!black}{green = ``coat''},
    \textcolor{blue}{blue = ``bag''} and 
    \textcolor{violet}{purple for all the other classes}. 
    With the exception of the leftmost images, the underlying image 
    data have been brightened to improve clarity.}
  \label{fig-multi-obj-fashion-sc-cov-main}
\end{figure*}

\subsection{Proof-of-concept experiments regarding multi-object detection}

In the following, we will explore properties of the proposed methodology for joint
local spatial and scale selection as a mechanism for generating object hypotheses in
multi-object detection and recognition scenarios.

\subsubsection{Experimental methodology based on image data from the Fashion-MNIST dataset}

To investigate the properties of the resulting local-extended
GaussDerResNets with joint spatial and scale selection when applied
to multi-object detection tasks, we first
\begin{itemize}
\item
  trained a regular GaussDerResNet with
  (i)~global spatial selection based on spatial max pooling and 
  (ii)~global scale pooling based on logsumexp pooling over scales
  on the training images for the size factor~1 in the rescaled
  Fashion-MNIST dataset.
\end{itemize}
Then, we
\begin{itemize}
\item
  generated a set of semi-synthetic images, by pasting together
  images of objects from the test images in the rescaled Fashion-MNIST
  dataset, with randomly selected objects from the different classes within 
  each image, at random spatial locations. Each such test image was of 
  size $120 \times 120$ pixels and contained 3 objects, one for each size 
  factor of $2^{-\frac{3}{4}}$, 1 and $2^{\frac{3}{4}}$,
\item
  used the weights from the trained GaussDerResNet for
  computing the deep scale-space representations $\Pi^c$ for each class $c$
  represented in the Fashion-MNIST dataset for the semi-synthetic test images
  according to above, by extracting the
  information from the final layers in the pure scale channels
  \begin{equation}
    \Pi(x;\; \sigma_i) = (F_{\sigma_i,K})(x) = \Lambda_{\sigma_i}^c(f)(x)
  \end{equation}
  for each scale channel
  $i \in [1, |I|]$ without any use of spatial pooling or scale pooling,
\item
  detected separate sets of discrete deep
  scale-space maxima by comparisons to all the
  26~neighbours in the post-smoothed discrete
  deep scale-space representations
  \begin{multline}
    (\Pi_{\gamma}^c)(\cdot;\; \sigma_i)
    = g(\cdot;\; \gamma \, \sigma_{i,0}) * F_{\sigma_i,K}(\cdot) = \\
    = g(\cdot;\; \gamma \, \sigma_{i,0}) * \Lambda_{\sigma_i}^c(f(\cdot)),
  \end{multline}
  with thresholding on the magnitudes of the responses,
  thereby 
  leading to a set of candidate object hypotheses $\{ (x_{\nu};\; \sigma_{\nu}) \}^c$
  for each class $c$, and
\item
  represented each deep scale-space maximum with a circle 
  according to Definition~\ref{def-circle-rep-dscsp-max}, with 
  the radius proportional to the scale at which the deep scale-space 
  maximum was detected. Specifically, this scale reflects the total 
  amount of Gaussian smoothing that has been used for computing 
  the representation in the final layer in the scale channel corresponding 
  to the detection scale of the deep scale-space maximum.
\end{itemize}
For this purpose, we used
(i)~the post-smoothing factor 
$\gamma = \sqrt{\sum_{k = 1}^K \left( r^{k-1} \right)^2} \, \gamma_{\text{tot}}$ 
with $\gamma_{\text{tot}} = 0.2$ according to the complementary 
definition in Footnote~\ref{footnote-def-gamma-tot},
(ii)~a conservative threshold value $\Pi_{\text{thresh}} = 2$ on the magnitude 
of the response with the same threshold used for all the classes, and
(iii)~the relative radius factor $R = 1$.

\subsubsection{Experiments with GaussDerResNet trained on the
  rescaled Fashion-MNIST dataset}

Figure~\ref{fig-multi-obj-fashion-standard-main} shows one of the input images
used for this experiment, with its associated post-smoothed deep scale-space 
maxima obtained (i)~while not using any merging of multiple responses, 
(ii)~when merging multiple responses within each class, and (iii)~when 
additionally removing the weaker of overlapping responses between different 
classes for visualisation purposes. The merging approach in (ii) is described 
in Section~\ref{sec-methods-merge-mult-responses} based on
selecting only the strongest response for each cluster of overlapping
circle representations for the deep scale-space maxima.

As can be seen from the results, the use of post-smoothing and the 
use of the following merging of spatially coinciding responses of 
deep scale-space maxima substantially reduces the number of
multiple responses to the same underlying object, together with 
thresholding serving as a form of non-maxima suppression, sharing 
some similarity with bounding box filtering algorithms in object 
detection networks.

By comparing the colour coding of the three classes represented in 
each image, we can specifically note that the correct class labels 
are among the dominant predictions, having high confidence scores. 
These results also demonstrate that GaussDerResNets, trained on images 
containing a single centred object at size factor 1, are capable of extending 
their predictions to tasks involving multiple objects of different sizes and 
spatial positions.

Furthermore, we can note that because certain classes in the Fashion-MNIST 
dataset share similar features between each other and the network was 
trained to classify a single object only, weaker responses 
corresponding to incorrect classifications can appear in distant regions 
of the same object, as seen for the handbag. 
In some of the cases, these false positives are very interpretable. 
For example, the central purple circle on the coat responds to set of 
buttons with a classification of ``shirt'', for which the buttons constitute 
a characteristic feature for this dataset.

The extent of the region covered by a particular maximum is dependent 
on the value of the relative radius factor $R$, which affects the merging 
and absorbing properties of the circles, and could be important when 
designing a potential object detection stage for the network.

To demonstrate the scale covariance properties, we additionally 
took the same trained network and 
\begin{itemize}
\item
  applied it to a different semi-synthetic dataset, where one set of 
  images contains 3 objects at size factor 1, and the other contains
  the same image data subject to spatial scaling transformations 
  corresponding to a spatial scaling factor of $S_x= 2$, with each test 
  image of size $160 \times 160$ pixels, and where
\item
  the same parameters and merging methods for the scale-space maxima  
  were used as in the previous experiment.
\end{itemize}
Figure~\ref{fig-multi-obj-fashion-sc-cov-main} shows
the results for a representative image from these
experiments. As can be seen from the figures, the 
scale covariance properties of the GaussDerResNets are clearly 
preserved in the multi-object case, with the corresponding class 
circles scaling roughly proportionally to the object size.

For each object in the images, the activations covering the objects 
corresponding to the correct object class are among the strongest ones, 
with most of the other predictions being weaker. We can 
also see that for the larger size factor, some of the maxima have 
become relatively weaker and are no longer in the top 10 of activations, 
which is likely due to the effects of random fluctuations
and the use of post-smoothing.

\begin{figure*}[hbt]
  \begin{center}
    \begin{subfigure}{0.99\textwidth}      
      \begin{tabular}{cccc}
        \textit{Input image}
          & \textit{10 strongest maxima}
          & \textit{Intra-class overlaps merged}
          & \textit{Cross-class overlaps absorbed} \\
          \includegraphics[width=0.22\textwidth]{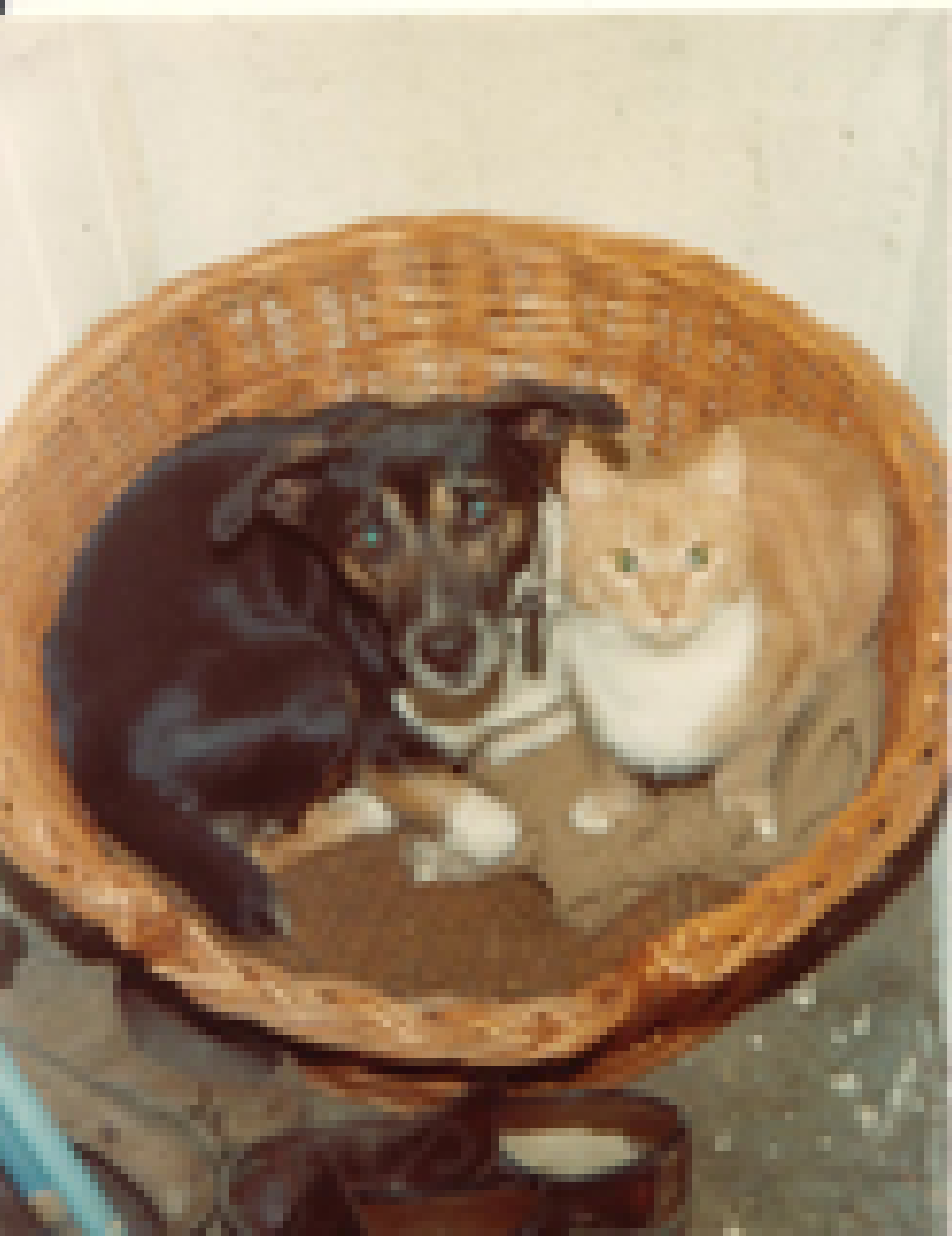}
          & \includegraphics[width=0.22\textwidth]{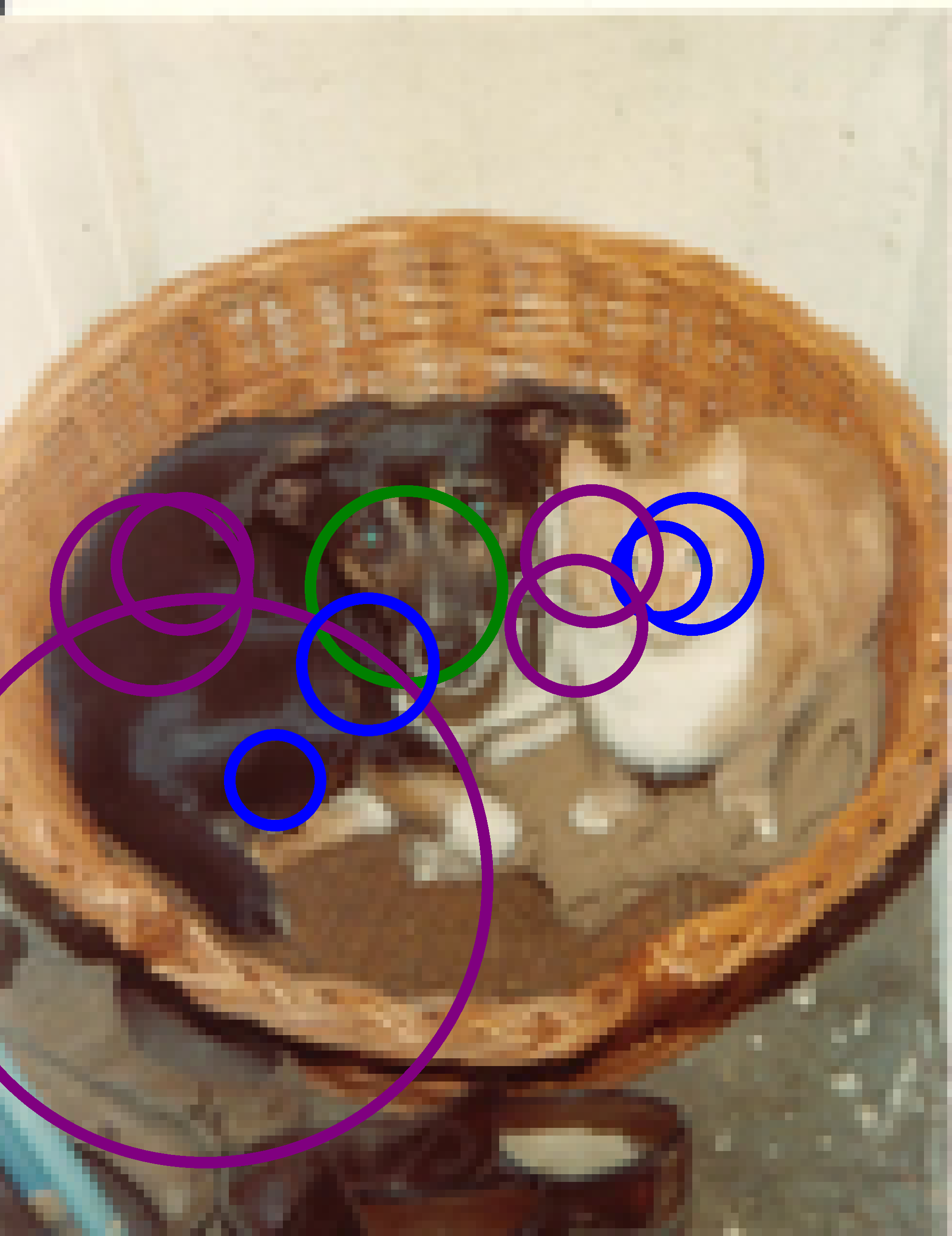}
          & \includegraphics[width=0.22\textwidth]{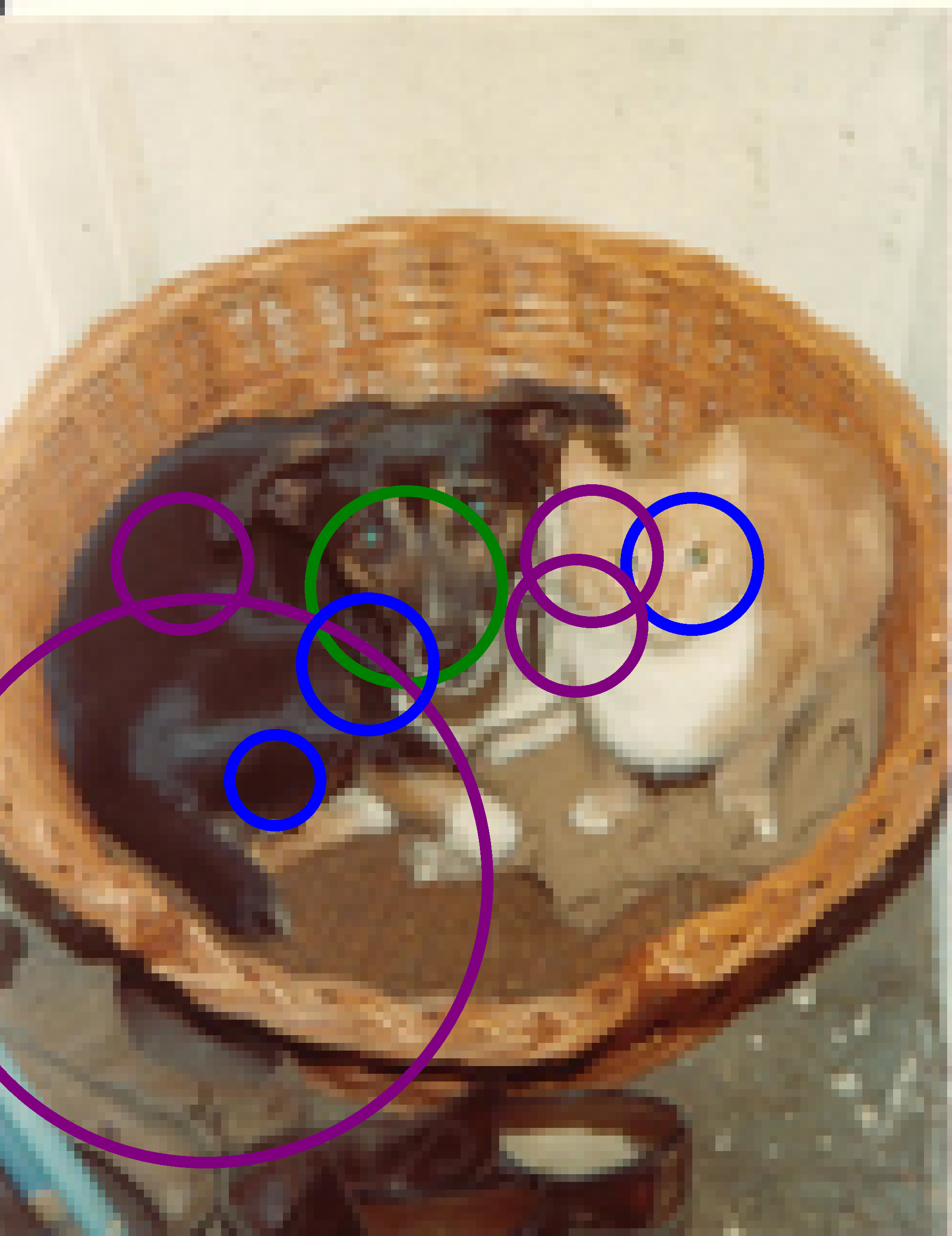}
          &  \includegraphics[width=0.22\textwidth]{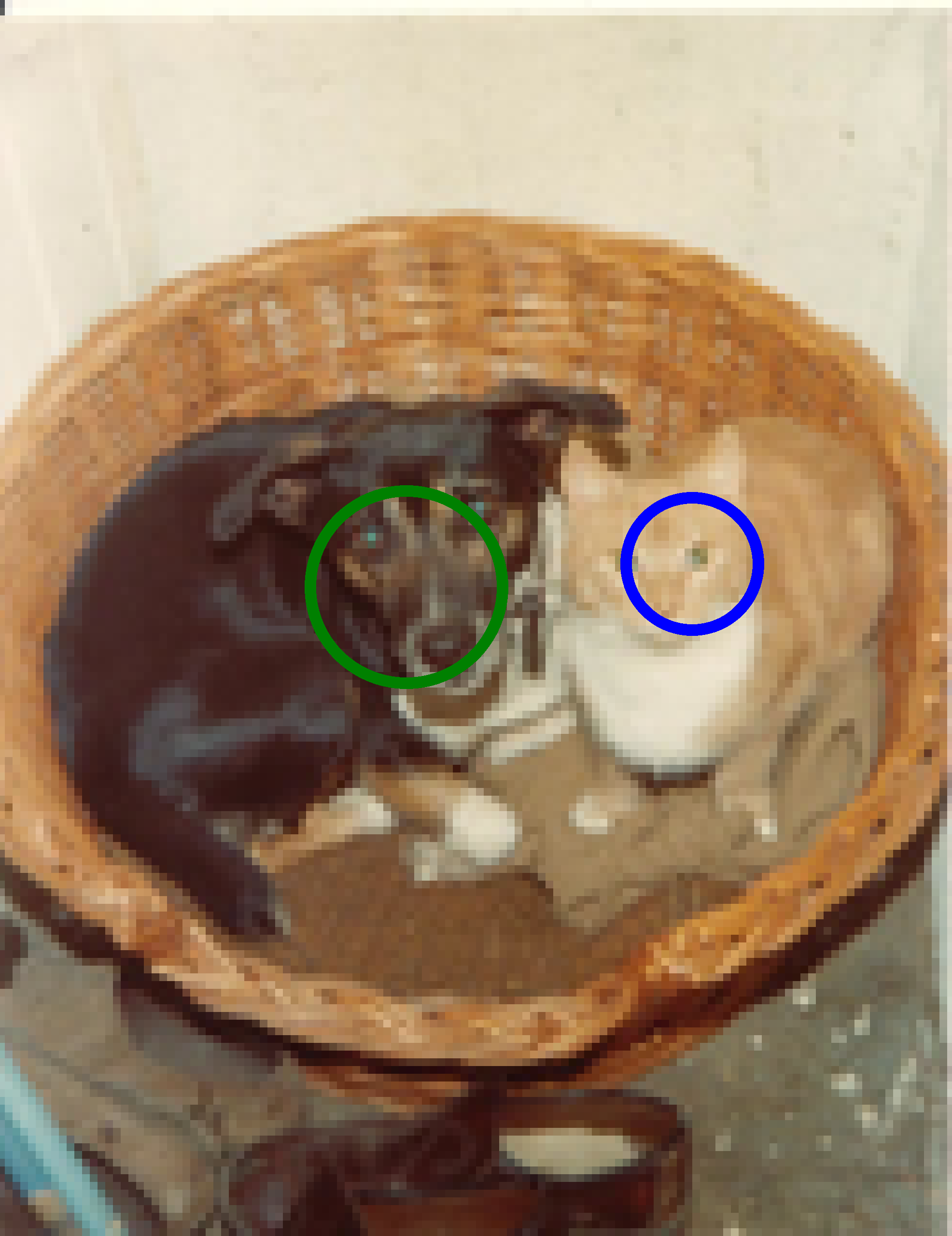}
      \end{tabular}
      \caption{First image of a cat and dog next to each other, with the predicted object hypotheses overlaid}
      \label{fig:subfig-mult-obj-sc-cov-sf1}
    \end{subfigure}

    \bigskip
    
    \begin{subfigure}{0.99\textwidth}      
      \begin{tabular}{cccc}
        \textit{Input image}
          & \textit{10 strongest maxima}
          & \textit{Intra-class overlaps merged}
          & \textit{Cross-class overlaps absorbed} \\
          \includegraphics[width=0.22\textwidth]{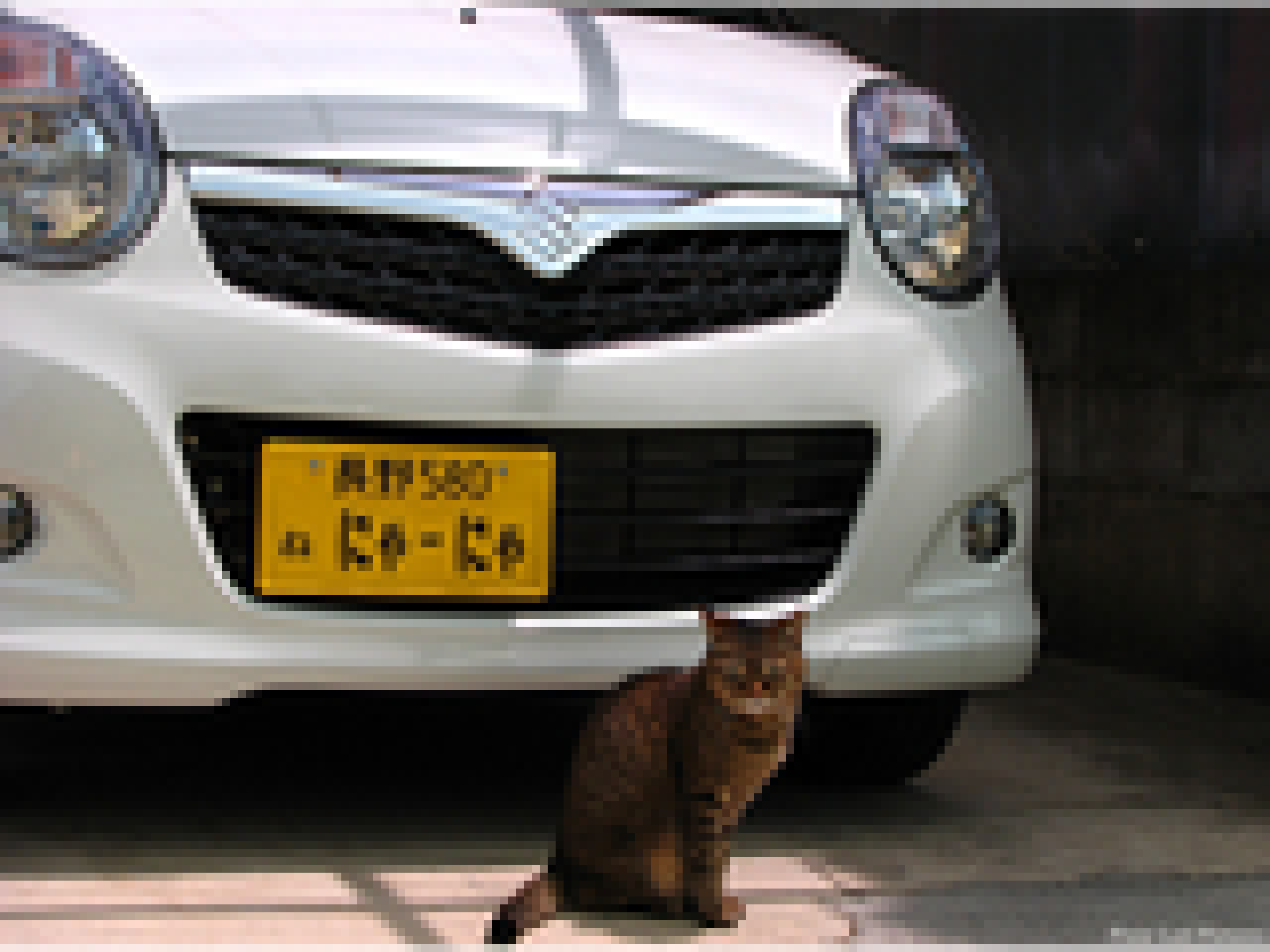}
          & \includegraphics[width=0.22\textwidth]{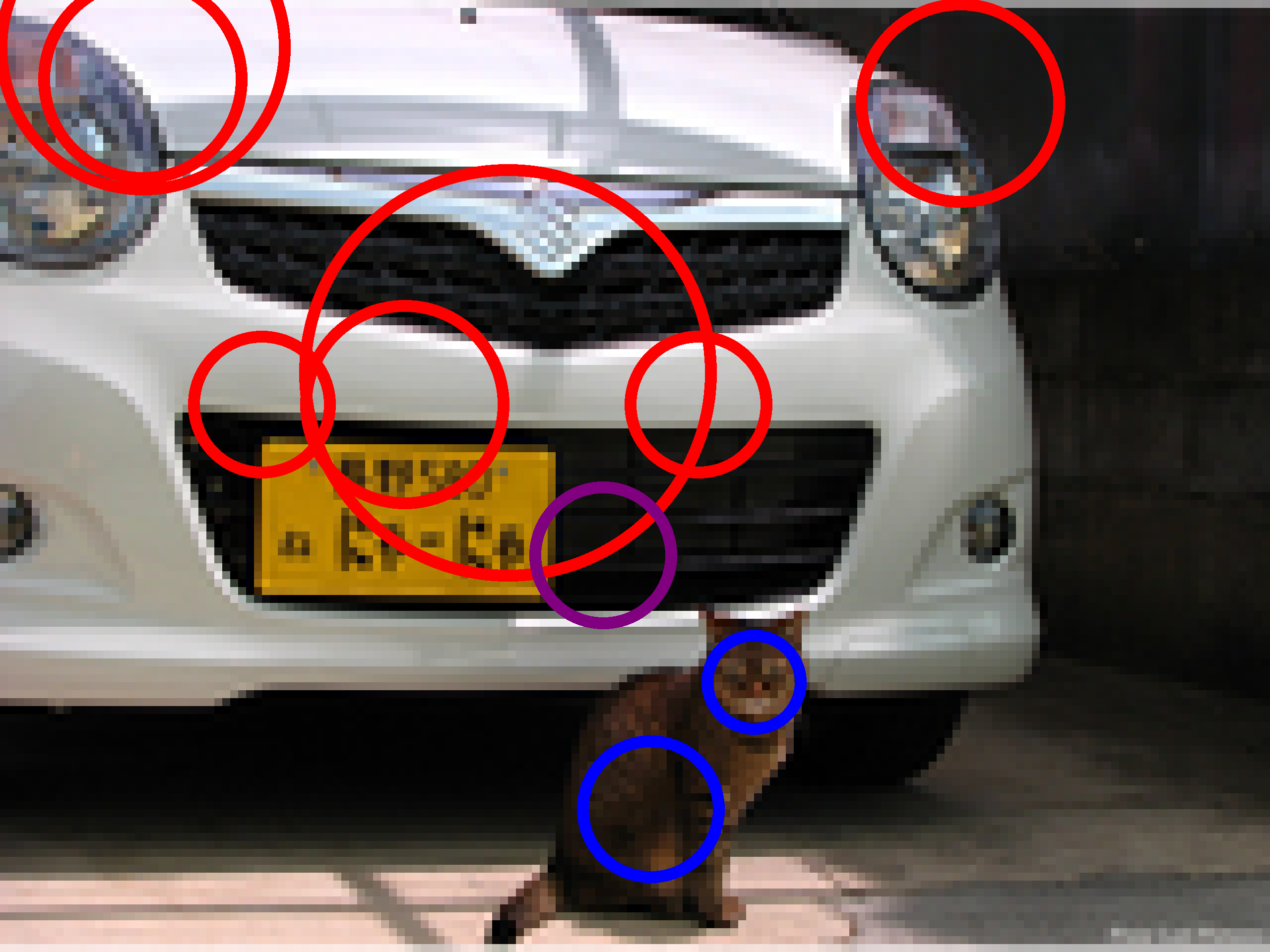}
          & \includegraphics[width=0.22\textwidth]{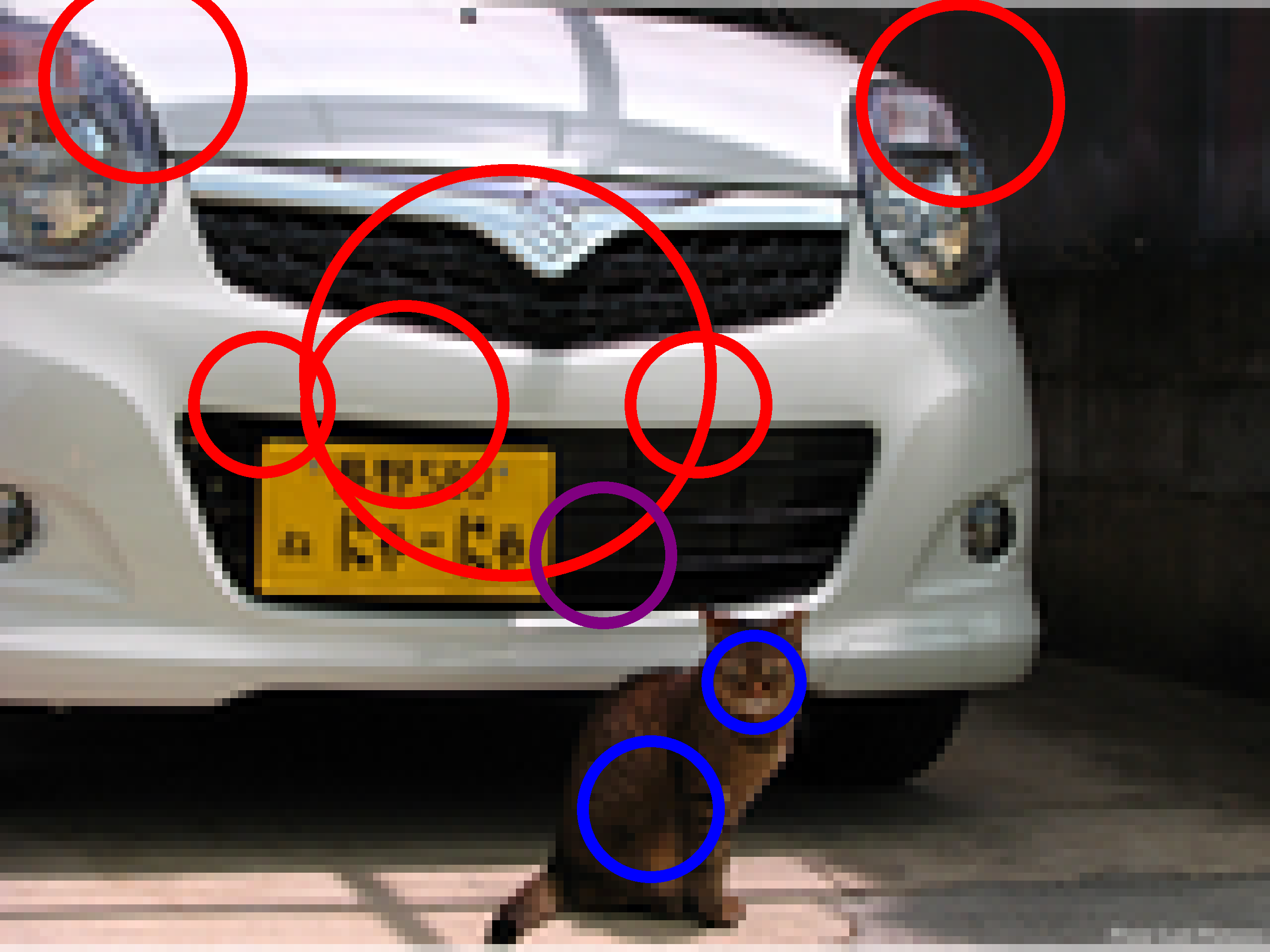}
          &  \includegraphics[width=0.22\textwidth]{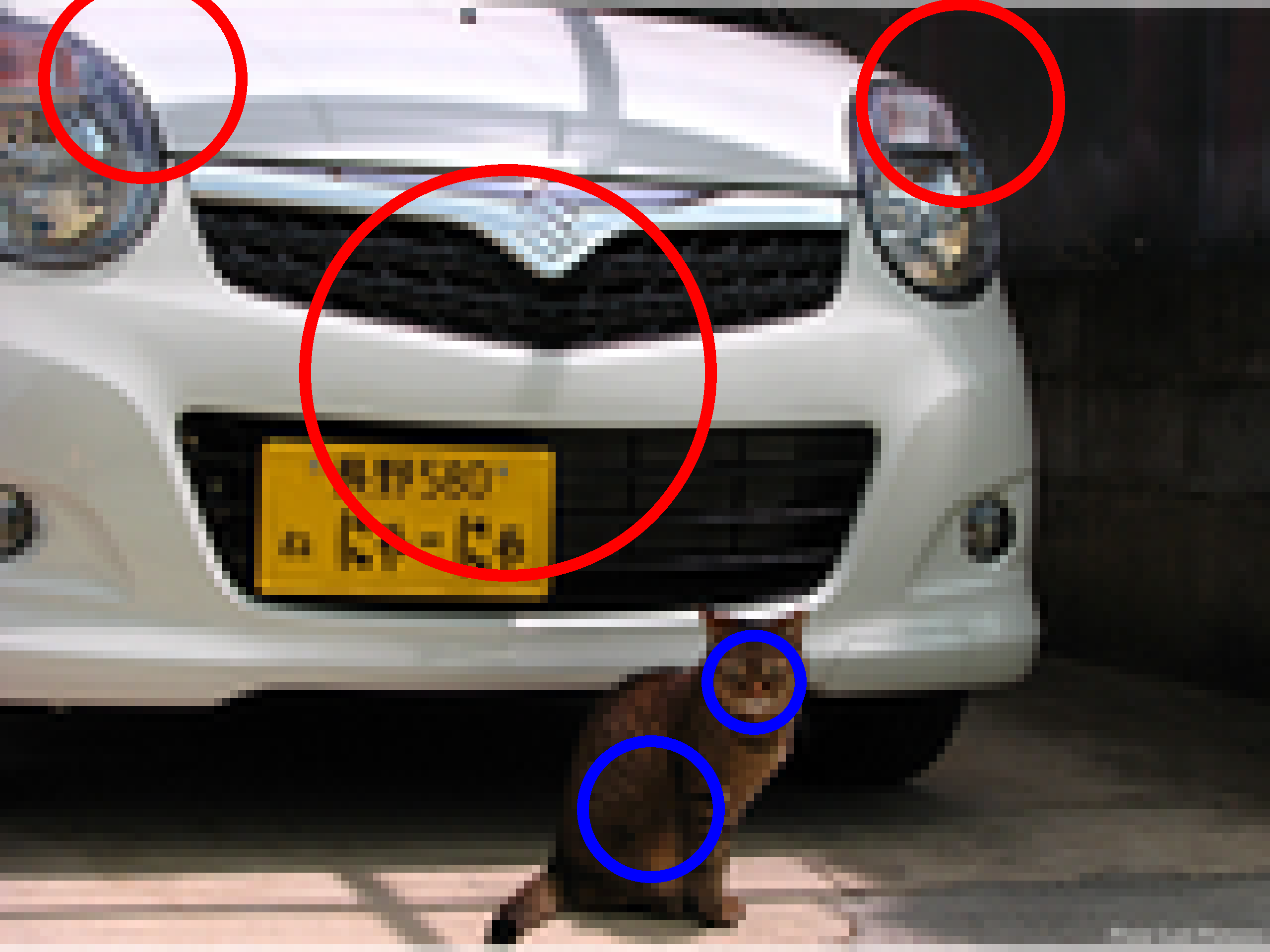}
      \end{tabular}
      \caption{Image of a cat in front of a car, with the predicted object hypotheses overlaid}
      \label{fig:subfig-mult-obj-sc-cov-sf2}
    \end{subfigure} 

    \bigskip
    
    \begin{subfigure}{0.99\textwidth}      
      \begin{tabular}{cccc}
        \textit{Input image}
          & \textit{10 strongest maxima}
          & \textit{Intra-class overlaps merged}
          & \textit{Cross-class overlaps absorbed} \\
          \includegraphics[width=0.22\textwidth]{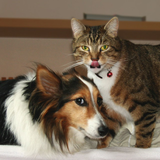}
          & \includegraphics[width=0.22\textwidth]{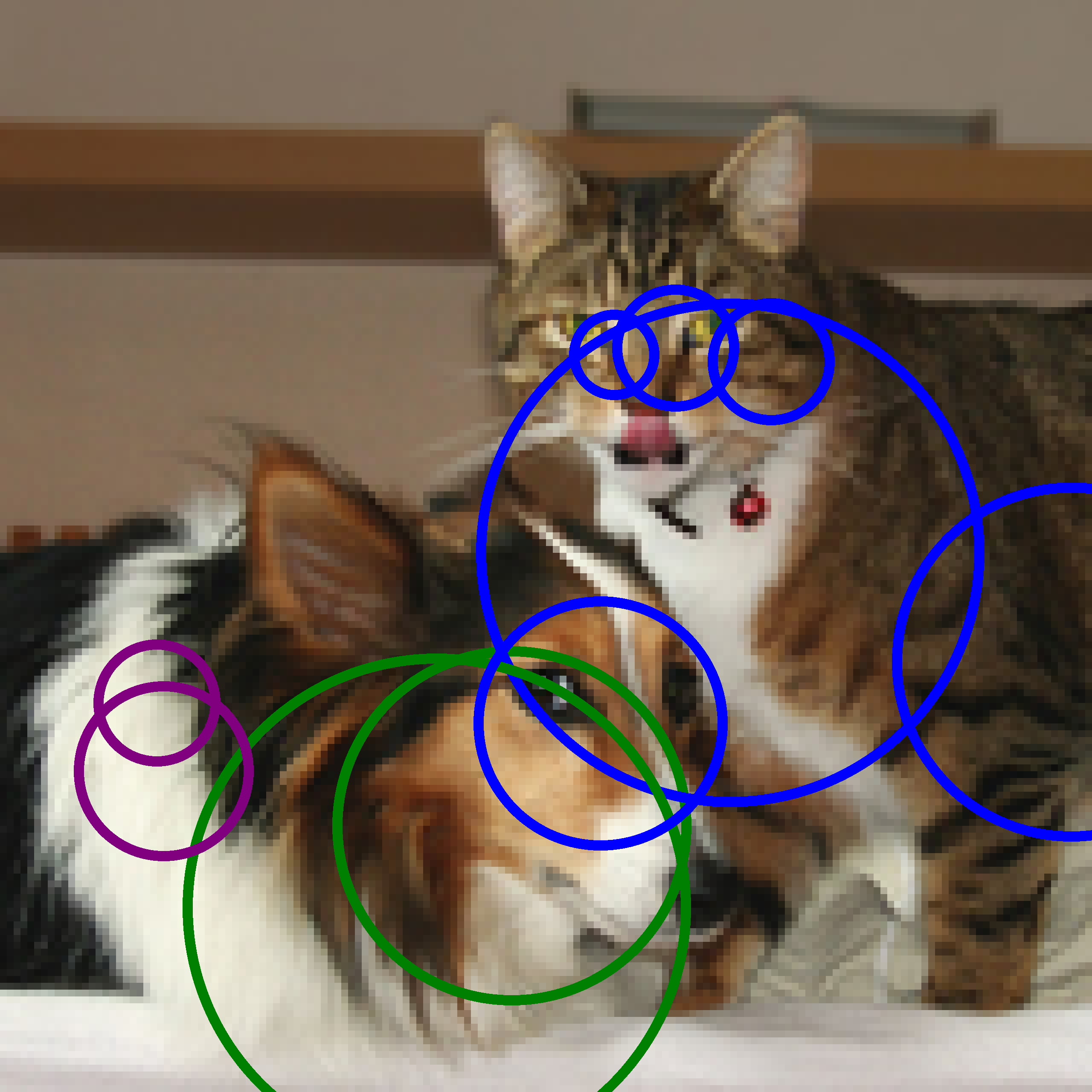}
          & \includegraphics[width=0.22\textwidth]{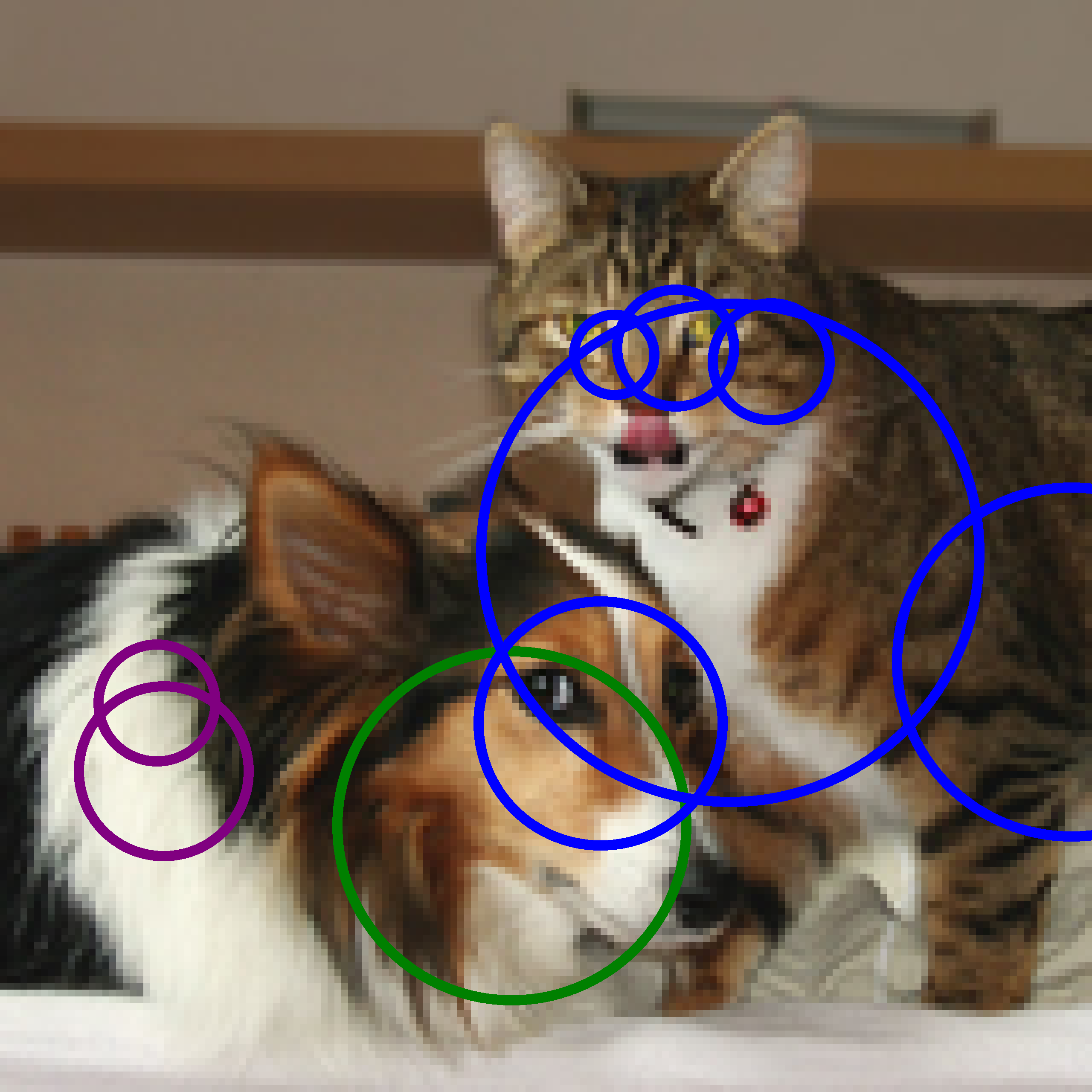}
          &  \includegraphics[width=0.22\textwidth]{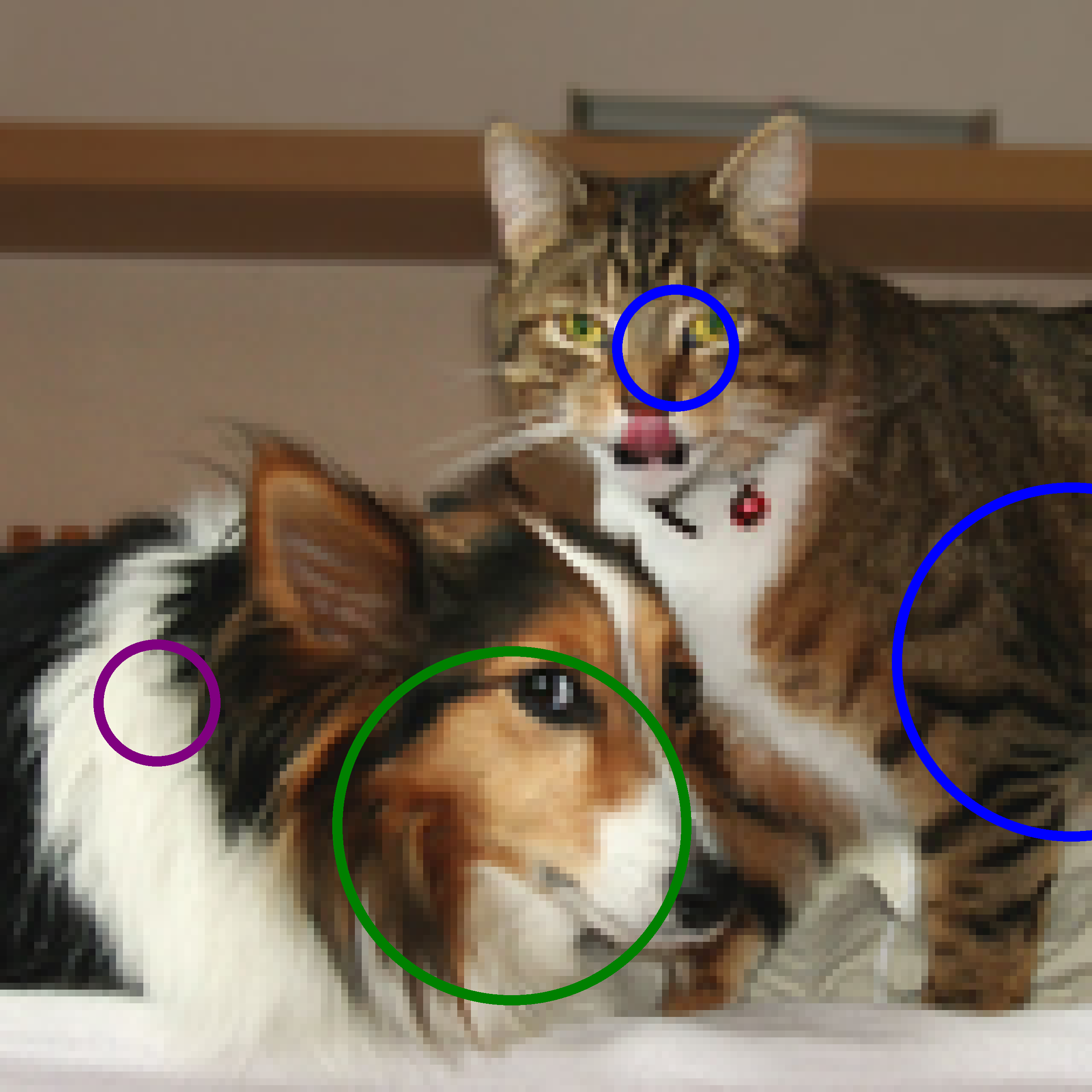}
      \end{tabular}
      \caption{Second image of a cat and dog next to each other, with the predicted object hypotheses overlaid}
      \label{fig:subfig-mult-obj-sc-cov-sf3}
    \end{subfigure} 
  \end{center}
  \caption{Illustration of the main steps in the generation of object
    hypotheses, using three images from the COCO dataset, from deep scale-space 
    maxima obtained by applying a GaussDerResNet based on max 
    pooling over scales, trained on the rescaled STL-10 dataset. 
    Subfigures (a) and (c) both show results for images with a cat 
    and dog seated next each other, while subfigure (b) shows a cat 
    in front of a car. From the resulting circles, we can see that that 
    the network is able to produce strong maxima over multiple objects 
    within real-world images, with the correct classes being the most 
    dominant over the corresponding objects.
    The colours of the circles indicate the classes of the object
    hypotheses, with \textcolor{red}{red = ``car''}, 
    \textcolor{green!60!black}{green = ``dog''}, \textcolor{blue}{blue = ``cat''} 
    and \textcolor{violet}{purple for all the other classes}.}
  \label{fig-multi-obj-COCO-main}
\end{figure*}

\subsubsection{Experiments with the GaussDerResNet trained on the
  rescaled STL-10 dataset}

Figure~\ref{fig-multi-obj-COCO-main} shows corresponding 
results of applying the GaussDerResNet with global spatial 
max pooling and global scale pooling from the maximum 
response over scales, trained on the rescaled STL-10 dataset,
to a set of real-world images from the COCO dataset (Lin et 
al. \citeyear{COCOdataset-14-ECCV}). Three images from 
COCO containing objects from the classes ``car'',  ``dog'' 
and ``cat'' were chosen, to demonstrate the generalisation 
ability of the network on common types of objects in a real-world 
setting. Interpolation was used to downsample the images by 
a factor of 4, to make them closer to the size of the images in the
rescaled STL-10 dataset that the network was trained on. A conservative threshold 
of $\Pi_{\text{thresh}} = 1$ was used.

These results show that the network can be extended to real-world 
images containing multiple objects of different sizes. We see that 
the GaussDerResNet successfully generalises to this new real-world 
dataset, with correct class labels once again being the dominant 
predictions, even in cases where the objects are positioned right 
next to each other, or if one object is partially occluding another 
object. Closely positioned objects do produce clusters that are 
adjacent or partially coinciding, however, the merging and absorption 
procedures are able to separate them in these cases.

From the circles in the images, we can see that the network 
focuses particularly strongly on the faces of the animals, as well as
on the headlights for the car and the bumper%
\footnote{Crucially, the GaussDerResNets are able to learn to 
detect characteristic features of interest for the different classes,
even though they were trained on images containing a single object, 
and were not provided with any information about the object 
during training other than its class label.}, with the network 
automatically selecting scale channels proportional to the size 
of the image structure of interest, which can be clearly seen in 
Figure~\ref{fig-multi-obj-COCO-main}(b), with mostly the coarser 
scale channels strongly activated by the large car and the finer
scale channels activated by the relatively small cat.

\subsubsection{Relationships to classical methods for object
  recognition based on interest points}

Structurally, the resulting approach for multi-object detection
and recognition can be seen as a hybrid between classical methods
for this purpose based on interest point detectors
(Lowe \citeyear{Low04-IJCV},
Arandjelovi{\'c} and Zisserman \citeyear{AraZis12-CVPR},
Lindeberg \citeyear{Lin12-Scholarpedia})
and deep learning.

While the SIFT descriptor first computes a set of interest points
based on scale-space extrema of a discrete approximation of
the scale-normalised Laplacian operator, and then computes
a scale-adapted and thereby scale-invariant image descriptor
at each such interest point, our approach performs these stages
jointly in the same step, using deep scale-space maxima of the
multi-scale feature channels in the deep scale-space representation
for both of these purposes, thereby providing interest points
that are more directly related to the different classes in the
object detection/recognition task than those produced by a mere class-agnostic
low-level Laplacian operator.

The output from the feature channels in the deep scale-space
representation do also ought to be more powerful than
the image descriptors in terms of orientation histograms
in the SIFT descriptor, and do therefore specifically do not
require any Hough-transform-based clustering approach
for generating the object hypotheses, as the object recognition
methods based on SIFT descriptors typically are based on.

In this way, we thereby obtain a set of object hypothesis
directly from deep scale-space maxima, that are extracted from
the deep scale-space representation constructed by
combining the set of final layers across the initial
scale levels $\sigma_{i,0}$ for each class.

\subsubsection{Conclusions from the extension to joint local spatial and
  scale selection}

In summary, the results of these experiments indicate that by
replacing the global spatial and scale pooling mechanisms in
the regular GaussDerResNet architecture described in
Section~\ref{sec-GaussDerResNets-theory} by the notion of local spatial
and scale selection, based on the detection of deep scale-space maxima
according to Definition~\ref{def-deep-scsp-max}, shows interesting
potential for extending the approach for single-object-detection
described in Section~\ref{sec-GaussDerResNets-theory} and
investigated in Section~\ref{sec-experiments} to the detection of
multiple objects in more complex scenes.
Specifically, our results indicate that:
\begin{itemize}
\item
  by this extension to local-extended GaussDerResNets,
  it is possible to obtain multiple responses of object hypotheses
  at different image positions and at different spatial scales
  for scenes that contain multiple objects,
  and that
\item
  the computed object hypotheses are both translationally covariant
  and scale covariant, with the circle representations transforming in
  a reasonable way under spatial scaling transformations.
\end{itemize}
In these ways, this proof-of-concept demonstrates the potential
of replacing the global spatial and scale pooling mechanisms,
that we have throughout used in the treatment of 
the regular GaussDerResNet architecture in
Section~\ref{sec-GaussDerResNets-theory} and
for the single-object-detection experiments in
Section 6, to adding detection of deep scale-space maxima
to the output from final layers in the pure scale channels in the
GaussDerResNet architecture,
to enable joint local spatial and scale selection,
which could then be used to address scenarios for
detecting and recognising multiple objects in the
same scene.

\section{Summary and conclusions}
\label{sec-conclusion}

We have proposed the notion of scale-invariant Gaussian derivative residual 
networks (GaussDerResNets) and 
presented an extensive investigation of their performance and scale 
generalisation properties, including architectural variants based on 
structural modifications. Additionally, we have examined
the effectiveness of  new training approaches on rescaled image datasets,
where the corresponding test sets contain spatial scaling variations
not encountered during training. The 
ability to handle different scales in image data is built in as prior knowledge 
into the GaussDerResNet architecture, thereby enabling the networks to generalise 
to scales not seen during training---a property that most other deep 
network architectures do not include in their design. This is achieved by
ensuring scale covariance of the receptive fields based on linear combinations 
of scale-normalised Gaussian derivatives coupled in cascade, with weight 
sharing between the scale channels and permutation invariant pooling over scales.

Building upon earlier work in Lindeberg (\citeyear{Lin22-JMIV}) and
Perzanowski and Lindeberg
(\citeyear{PerLin25-JMIV-ScSelGaussDerNets}),
where Gaussian derivative networks (GaussDerNets) without residual connections were
proposed and investigated experimentally, the GaussDerResNets
proposed in this paper substantially modernise and improve this architecture in 
several respects, such as the incorporation of scale covariant residual 
blocks and considerably increased network size and depth, which in this work 
is demonstrated to lead to significantly better performance and scale 
generalisation properties on natural image datasets.

For the GaussDerResNet architecture, we have described its theoretical
properties in Section~\ref{sec-GaussDerResNets-theory}, with formal
proofs of scale covariance and scale invariance properties, as well
as structural relations to semi-discretisations of the diffusion equation.
We have also described extensions to depthwise-separable 
convolutions and zero-order terms, which we then have investigated
experimentally by ablation studies in Appendix~C and in 
Section~\ref{sec-zero-order-term-study}.

To investigate how scaling variations affect the performance of
GaussDerResNets, we have in Section~\ref{sec-rescaled-stl-dataset}
defined a new rescaled STL-10 dataset, which incorporates systematic
spatial scaling variations. Compared to the previously existing
rescaled MNIST, Fashion-MNIST and CIFAR-10 datasets,
this dataset, as derived from a subset of the ImageNet dataset,
is substantially more challenging, has higher image resolution and contains far less 
training data.

In experimental investigations of the classification performance and
the scale generalisation properties of the new GaussDerResNet
architecture, we have in Section~\ref{sec-experiments} demonstrated
that this network architecture achieves noticeably better performance on
the rescaled Fashion-MNIST and STIR datasets, and significantly better 
performance on the rescaled CIFAR-10 dataset, compared to the
previously proposed GaussDerNets without residual connections.
We have also demonstrated that the here proposed GaussDerResNet
architecture allows for good classification performance on the regular
STL-10 dataset without scaling variations as well as good scale
generalisation properties on the new rescaled STL-10 dataset
with spatial scaling variations.

In these experiments, we notably
found that not including a zero-order term was preferable for the
rescaled Fashion-MNIST and CIFAR-10 datasets, while including
a zero-order term was strongly beneficial for the rescaled STL-10 dataset.
For the rescaled STL-10 dataset, we also included a spatial selection mechanism
based on spatial max pooling to handle non-centred objects in the
image domain. From our in-depth analysis of the scale selection properties 
of the networks, we specifically found that the selected scale levels
increased linearly with the spatial scaling factors in the underlying
test data, in agreement with the expected theoretical behaviour,
for all the studied types of scale pooling mechanisms in terms of
max pooling, logsumexp pooling, or average pooling over scales.

To gain a better understanding of the properties and behaviour of 
the GaussDerResNets under different settings and conditions,
we have in the appendix sections performed a set of additional ablation studies. 
There, we study the modernisation of the 
architecture into depthwise-separable convolutions, 
which significantly reduces the number of model parameters. 
Specifically, we demonstrated that this architecture retains 
provably scale covariant and scale invariant properties and can 
maintain good performance and scale generalisation 
properties. The depthwise-separable networks are, however, 
harder to train.
   
Furthermore, in the Appendix, we investigated the 
effectiveness of new training methods for 
GaussDerResNets, where we:
\begin{itemize}
\item
   demonstrated that pre-training a multi-scale-channel network by initialising 
   its weights with the weights from a trained single-scale-channel network 
   reduces the amount computation required to achieve good optimisation 
   during its training, and often improving the scale generalisation at fine
   scales, and
\item
   found that increasing the number of scale channels in a multi-scale-channel
   GaussDerResNet, through weight transfer after training, results in a slight 
   improvement of its scale generalisation properties.
\end{itemize}
With these extensions of our previous notion of GaussDerNets
(Lindeberg \citeyear{Lin22-JMIV},
Perzanowski and Lindeberg \citeyear{PerLin25-JMIV-ScSelGaussDerNets})
into deeper GaussDerResNets with residual skip connections,
we are now able to well handle substantially more complex datasets,
such as the Rescaled STL-10 dataset, which is harder to address
with regular GaussDerNets.
By the extension to depthwise-separable GaussDerResNets,
we additionally show that the framework allows
for computationally much more efficient implementations
in combined software and hardware environments that can
take advantage of the depthwise-separable convolutions in an
efficient manner.

Finally, in Section~\ref{sec-network-function-visualisation},
we have examined 
the activation maps, demonstrating that the GaussDerResNets
have good explainability properties in terms of the regions of
interest, producing strong responses at characteristic image 
structures that identify the objects in the images, and,
in Section~\ref{sec-loc-extended-gaussderresnets},
used the demonstrated combined feature- and scale-selective properties of
the GaussDerResNets for defining local-extended GaussDerResNets with
complementary local mechanisms for spatial and scale selection.
In this way, we are able to generate multiple hypotheses about different objects at
associated individually determined different scales, to detect and
recognise objects in more complex
scenes with multiple objects in the same scene.

To conclude, the purpose of this paper has been to show that the 
scale-invariant GaussDerResNet architecture, including its variants as
presented here, can be trained in a stable and successful manner, 
and can achieve very good classification performance and 
scale generalisation on natural image datasets with spatial scaling 
variations, thereby being able to handle test data that contain objects at scales
not present in the training data. 
We argue that such scale generalisation properties constitute
an important mechanism
for deep networks aimed at image classification under large scaling
variations. Specifically, provably scale-covariant and scale-invariant
deep network architectures, which enforce 
scaling symmetries, allow for a theoretically well-defined, consistent and 
interpretable treatment of the scaling variations commonly encountered in 
real-world data, thereby reducing reliance on data augmentation and/or
larger amounts of training data.

\section*{Acknowledgements}

We would like to thank the anonymous reviewers for valuable comments and
suggestions that have improved the presentation.

Computational resources were provided by the National Academic
Infrastructure for Supercomputing in Sweden (NAISS), funded by the
Swedish Research Council.
Complementary GPU resources were also
provided by the PDC Center for High Performance Computing at 
KTH Royal Institute of Technology in Sweden.

\appendix
\section*{Appendix}

This appendix addresses the following topics:
\begin{itemize}
\item
  Appendix~\ref{app-def-layer-ops-single-scale-channel-gaussderresnet}
  provides {\em formal mathematical definitions\/} of the cascade of
  (i)~a first regular Gaussian derivative layer,
  (ii)~a set of standard Gaussian derivative residual blocks in cascade, and
  (iii)~a final Gaussian derivative layer in a {\em single scale channel\/} of
  a Gaussian derivative residual network.
\item
  Appendix~\ref{sec-single-sc-net-gen-appendix} shows the {\em scale
  generalisation properties of single-scale-channel networks\/} applied
  to the rescaled Fashion-MNIST, CIFAR-10 and STL-10 data\-sets,
  demonstrating that the accuracy of the networks may decrease substantially
  with increasing differences between the scales in the training data
  vs.\ the testing data.
\item
  Appendix~\ref{sec-standard-vs-dwsc} explores the {\em extension of
  GaussDerResNets to depthwise-separable-convolution-based DSGaussDerResNets,
  thereby reducing both the number of parameters in the network and the
  computational cost\/}, by reusing the spatial components of the convolutions, 
  with reasonably maintained accuracy and scale generalisation properties.
\item
  Appendix~\ref{sec-effect-of-label-smoothing-appendix} shows
  the {\em influence of label smoothing on the scale generalisation and the
  scale selection properties\/}, demonstrating that label smoothing leads
  to better scale generalisation properties, while also leading to
  less sharp scale selection histograms.
\item
  Appendix~\ref{sec-pre-training} introduces a notion of {\em pre-training\/},
  where in a first phase a single-scale-channel network is trained on
  the training data, after which in a second phase the weights are
  transferred to training a multi-scale-channel network to achieve
  genuine multi-scale-channel training. In this way, {\em the training cost
  is substantially reduced, while at the same time often leading to better
  convergence properties\/} compared to training a multi-scale-channel
  network from scratch.
\item  
  Appendix~\ref{sec-semi-multi-training} explores a {\em mechanism of weight
  transfer, to achieve a denser spacing of the scale channels in the
  testing phase\/} than in the training phase, and demonstrates that this
  can lead to slight increases in the accuracy and the scale
  generalisation properties. In this way, we significantly reduce the amount of 
  computation during training compared to training a dense network from scratch, 
  while leading to a proportional increase in the amount of computations during inference.
\item
  Appendix~\ref{app-compare-disco} shows the results of a
  {\em comparison to the scale generalisation properties obtained
  with the DISCO network\/} by Sosnovik et al.\
  (\citeyear{SosMosSme21-BMVC}) applied on the rescaled
  Fashion-MNIST, CIFAR-10 and STL-10 datasets.
\item
  Appendix~\ref{app-compare-stir} shows the results of a
  {\em comparison to previously reported scale generalisation
  results on the STIR datasets\/} with traffic signs and
  aerial images reported by Altstidl et al.\
  (\citeyear{AltNguSchKoeMutEskZan23-IJCNN}).
\item
  Appendix~\ref{sec-filt-vis-subsection-appendix} shows {\em visualisations
  of the learned effective filters\/} in different layers of the Gaussian
  derivative ResNets based on zero-order terms applied to the rescaled
  STL-10 dataset, thereby visualising the types of receptive fields
  that the Gaussian derivative ResNet architecture is based on.
\end{itemize}

\section{Formal definitions of the layer operations in a
  single-scale-channel Gaussian derivative residual network with
  multiple feature channels in each layer}
\label{app-def-layer-ops-single-scale-channel-gaussderresnet}

Following Definition~9 in the main body of the paper,
a pure single scale channel in a GaussDerResNet is
defined by a cascade of:
\begin{itemize}
\item
  a first Gaussian derivative layer with a complementary
  batch normalisation stage,
\item
  a set of following standard Gaussian derivative blocks coupled in
  cascade, and
\item
  a final Gaussian derivative layer with an optional complementary
  batch normalisation stage (not used here).
\end{itemize}
The notion of a regular Gaussian derivative layer was defined in
Lindeberg (\citeyear{Lin22-JMIV}) Sections~4.1--4.2,
and extended to arbitrary dimensions in
Definition~4 in the main body of the paper.
The notion of a standard Gaussian derivative residual block with two
weight functions was defined in Definition~6 in the main body 
of the paper. In this section, we extend those
definitions with (i)~batch normalisation stages, and
(ii)~explicit notation for combining multiple feature channels
between the different levels in the deep hierarchy.

\subsection{Layer operations in a single scale channel over multiple
  feature channels}

\subsubsection{The first layer}

Given any input colour image $f \colon \bbbr^D \rightarrow \bbbr^{N_0}$, we
have that the first Gaussian derivative layer with the level index
$k=1$ is defined from an extension of the operator
${\cal G}_{w_1,\sigma_1}$ for computing a Gaussian
derivative layer at scale $\sigma_1$
according to Definition~4 in the main body of the paper into
an explicit expression of the form
\begin{multline}
  \label{def-gaussderresnet-layer-1}
  F_{\sigma_0,1}^{c_{\text{out}}} (\cdot) = \\
  \relu
  \left(
    \operatorname{BN}
    \left(
      \sum_{c_{\text{in}} \in [1, N_0]}
      w_1^{c_{\text{out}},c_{\text{in}}}(\cdot;\, \sigma_{1}) * f^{c_{\text{in}}}(\cdot)
                                   \right)
   \right),
\end{multline}
where
\begin{itemize}
\item
  $f^{c_{\text{in}}}(\cdot)$ denotes the channel with index
  $c_{\text{in}} \in [1, N_0]$ with $N_0 = 3$ 
  for the case of colour input data,
\item
  $w_1^{c_{\text{out}},c_{\text{in}}}(x;\; \sigma_1)$ denotes an extension
  of the Gaussian derivative weight
  function at scale $\sigma_1$, as defined according to
  Definition~1 in the main body of the paper,
  to using multiple input channels $c_{\text{in}}$ and multiple output
  channels $c_{\text{out}}$ in the input function and the first layer, respectively,
  and
\item
  the function $\operatorname{BN}(\cdot)$ denotes the batch
  normalisation operation.
\end{itemize}

\subsubsection{The intermediate layers}

For the intermediate levels, for $k \in [2, K-1]$,
given an extension of the definition of the operator
${\cal M}_{w_1,w_2,\sigma}$, for defining a residual block at scale $\sigma$
according to Equations~(21) and~(35) in the main body of the paper,
we can formally define the successive residual blocks of a
single-scale-channel GaussDerResNet in the following recursive manner
\begin{multline}
  \label{def-explicit-gaussderresnet-layer-k+1}
  F_{\sigma_0,k}^{c_{\text{out}}} (\cdot) = \\
  \relu \left( F^{c_{\text{in}}}_{\sigma_0,k-1}(\cdot)  + 
  \operatorname{BN} \left( \sum_{c_{\mu} \in [1, N_{k-1}]} 
    w_{2k-1}^{c_{\text{out}},c_{\mu}}(\cdot;\; \sigma_{k}) * \left(
      \vphantom{\sum_{c_{\text{in}} \in [1, N_{k-1}]} 
   {\cal G}_{\sigma_{k}}^{c_{\text{in}},c_{\text{in}}} } \right. \right. \right.\\
   \relu \left( \left. \left. \operatorname{BN} \left( \sum_{c_{\text{in}} \in [1, N_{k-1}]} 
    w_{2k-2}^{c_{\mu},c_{\text{in}}}(\cdot;\; \sigma_{k}) *
  \left( \vphantom{\sum} F^{c_{\text{in}}}_{\sigma_0,k-1}(\cdot) \right) 
  \right) \right) \left.  \vphantom{\sum_{c_{\text{in}} \in [1, N_{k-1}]} 
    {\cal G}_{w_{2k-1},\sigma_{k}}^{c_{\text{in}},c_{\text{in}}} } \right.\right.\right.\\
\left.\left.\left.
      \vphantom{\sum_{c_{\text{in}} \in [1, N_{k-1}]} 
   {\cal G}_{\sigma_{k}}^{c_{\text{in}},c_{\text{in}}} } 
    \right) \right) \right),
\end{multline}
where the intermediate feature channel index $c_{\mu}$ is typically set to $c_{\mu} = c_{\text{in}}$, and in the case when the 
feature channels $c_{\text{in}}$ and $c_{\text{out}}$ do not match, a 1$\times$1 convolution based projection is
applied to the skip connection, together with a batch normalisation
stage, to match its feature dimension with the output.

Note that the different indexing of the two Gaussian derivative layers reflects the fact that they use different weights, and similarly
to the definitions for the Gaussian derivative layers, the bias terms
are not explicitly shown here.

\subsubsection{The final layer}

Finally, for the final Gaussian derivative layer at level $k = K$, we have that
\begin{multline}
   \label{def-gaussderresnet-layer-final}
  F_{\sigma_0,K}^{c_{\text{out}}}(\cdot) = \\
  \relu
  \left(
    \operatorname{BN}
    \left(
      \sum_{c_{\text{in}} \in [1, N_0]}
      w_{2K-2}^{c,c_{\text{in}}}(\cdot;\, \sigma_{K}) * F_{\sigma_0,K-1}^{c_{\text{in}}}(\cdot)
    \right)
   \right),
 \end{multline}
where the corresponding size of the output feature channel over the channel
indices $c$ is defined to be equal to the number of classes.

\begin{figure*}[ht]
  \begin{center}
    \begin{subfigure}{0.32\textwidth}
        \centering
        \textit{\quad \quad Scale generalisation on rescaled Fashion-MNIST}
        \includegraphics[width=\linewidth]{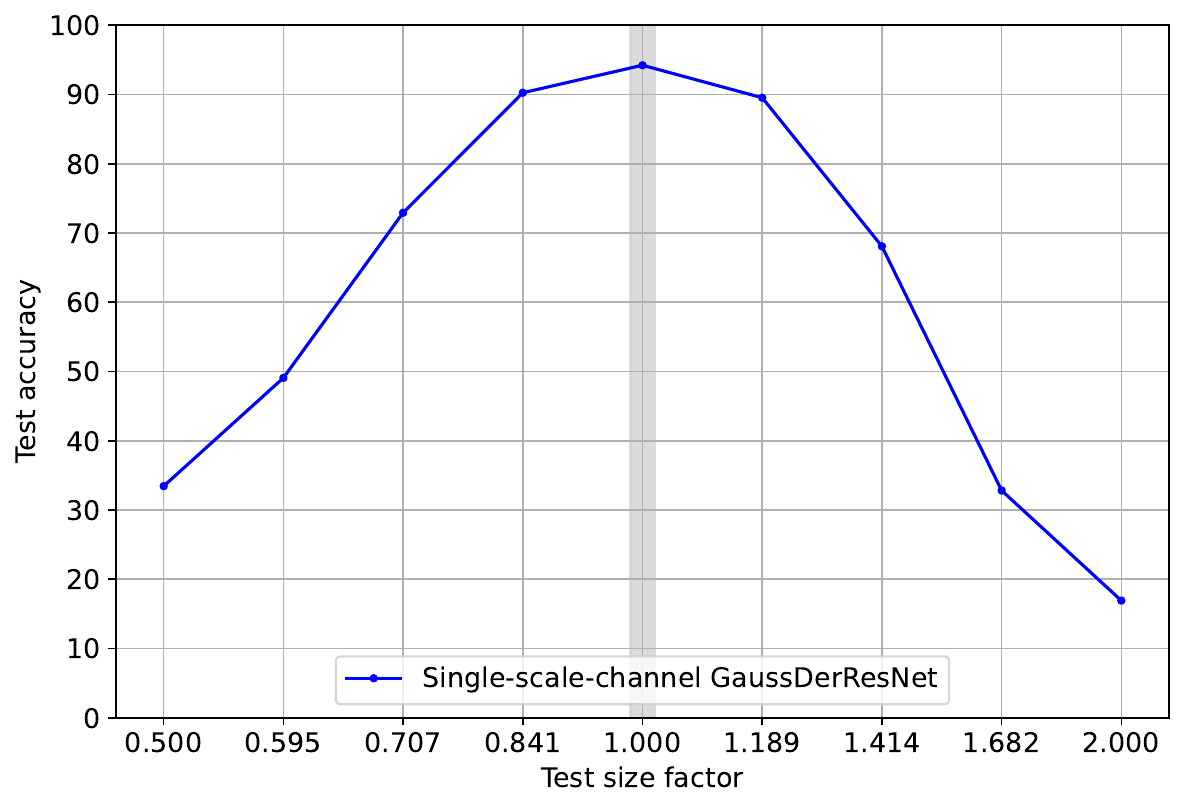}
        \label{fig:subfig-single-train-1}
    \end{subfigure}
    \hfill
    \begin{subfigure}{0.32\textwidth}
        \centering
        \textit{\quad \quad Scale generalisation on rescaled CIFAR-10}
        \includegraphics[width=\linewidth]{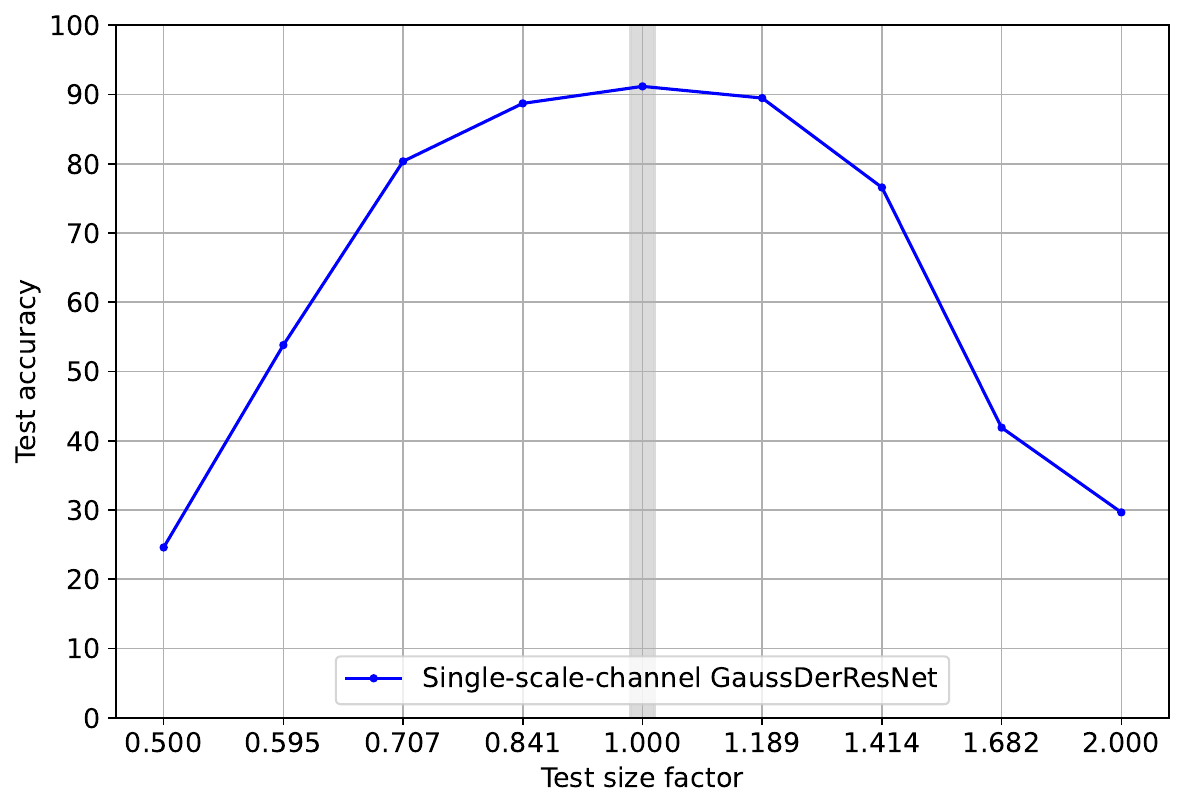}
        \label{fig:subfig-single-train-2}
    \end{subfigure}
    \hfill
    \begin{subfigure}{0.32\textwidth}
        \centering
        \textit{\quad \quad Scale generalisation on rescaled STL-10}
        \includegraphics[width=\linewidth]{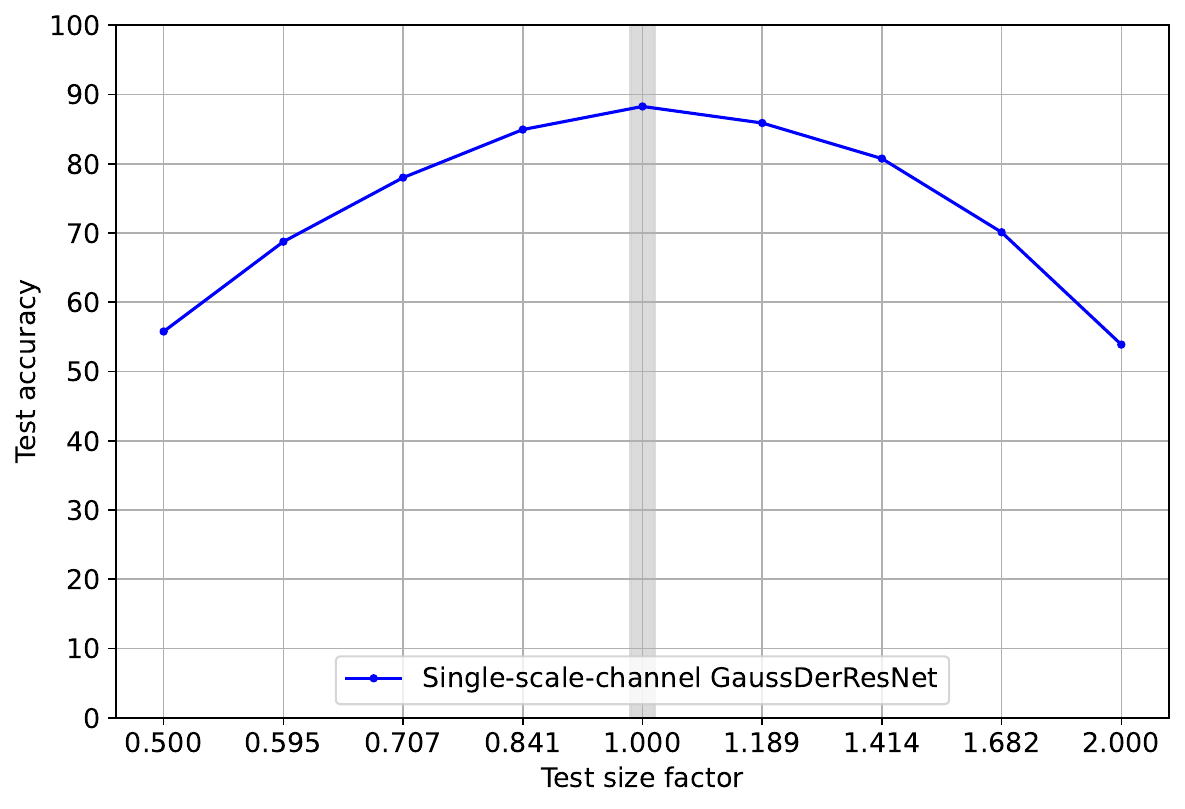}
        \label{fig:subfig-single-train-3}
    \end{subfigure}
  \end{center}
\caption{Scale generalisation curves for the single-scale-channel 
GaussDerResNets on the rescaled (left)~Fashion-MNIST, 
(middle)~CIFAR-10 and (right)~STL-10 datasets. Test data 
with the same size factor 1 as the training data is represented 
by the shaded grey region. Note that the vertical axes for the 
test accuracy are in the range 0--100\%, unlike 
the scale generalisation graphs in the main paper.}
\label{fig-gaussresnet-single-sc-appendix}
\end{figure*}

\section{Scale generalisation properties of single-scale-channel networks}
\label{sec-single-sc-net-gen-appendix}

Regular deep networks without in-built scale covariance or scale 
invariance properties will typically have very poor scale
generalisation properties.
We demonstrate this by training single-scale-channel GaussDerResNets, 
which lack the scale invariance property of their multi-scale-channel 
counterparts, on the rescaled Fashion-MNIST, CIFAR-10 and STL-10 
datasets. These single-scale-channel networks were trained in the same
manner as the multi-scale-channel networks according to
Sections~5 and~6 in the main body of the paper (although using larger 
batch sizes of 64 for the rescaled Fashion-MNIST and STL-10 datasets 
and 128 for the rescaled CIFAR-10 dataset), with each network 
based on the initial scale level $\sigma_{0}=1$.

Figure~\ref{fig-gaussresnet-single-sc-appendix} shows the results from
this experiment, where the training is performed 
for the single size factor 1, and testing is performed for all possible 
size factors between 1/2 and 2. From the graphs, we can see that for 
all the datasets the testing accuracy drops sharply for size factors other 
than 1, with the loss in performance becoming larger, the
further away the testing scale is from the training scale.
Additionally, from the results for the 
rescaled STL-10 dataset, we see that the scale generalisation curve is
somewhat flatter than for the other datasets, which may be due
to the presence of different scales within the original dataset
used for training, thus serving as a type of data augmentation.

\begin{table*}[hbtp]
  \centering
  \begin{tabular}{lccc}
       \hline
        & \textbf{rescaled Fashion-MNIST} & \textbf{rescaled CIFAR-10} & \textbf{rescaled STL-10}\\ 
       \hline
       \# network parameters (GaussDerResNets) &  295k & 1.78M & 2.07M \\ 
       \# network parameters (DSGaussDerResNets) & 120k & 547k & 669k \\ \hline
  \end{tabular}
  \caption{Number of network parameters for the multi-scale-channel 
  GaussDerResNets and DSGaussDerResNets employed in our experiments, 
  for the three rescaled datasets.}
  \label{tab:dwsc-num-network-param}
\end{table*}

\begin{figure*}[hbt]
  \begin{center}
    \begin{subfigure}{0.32\textwidth}
        \centering
        \includegraphics[width=\linewidth]{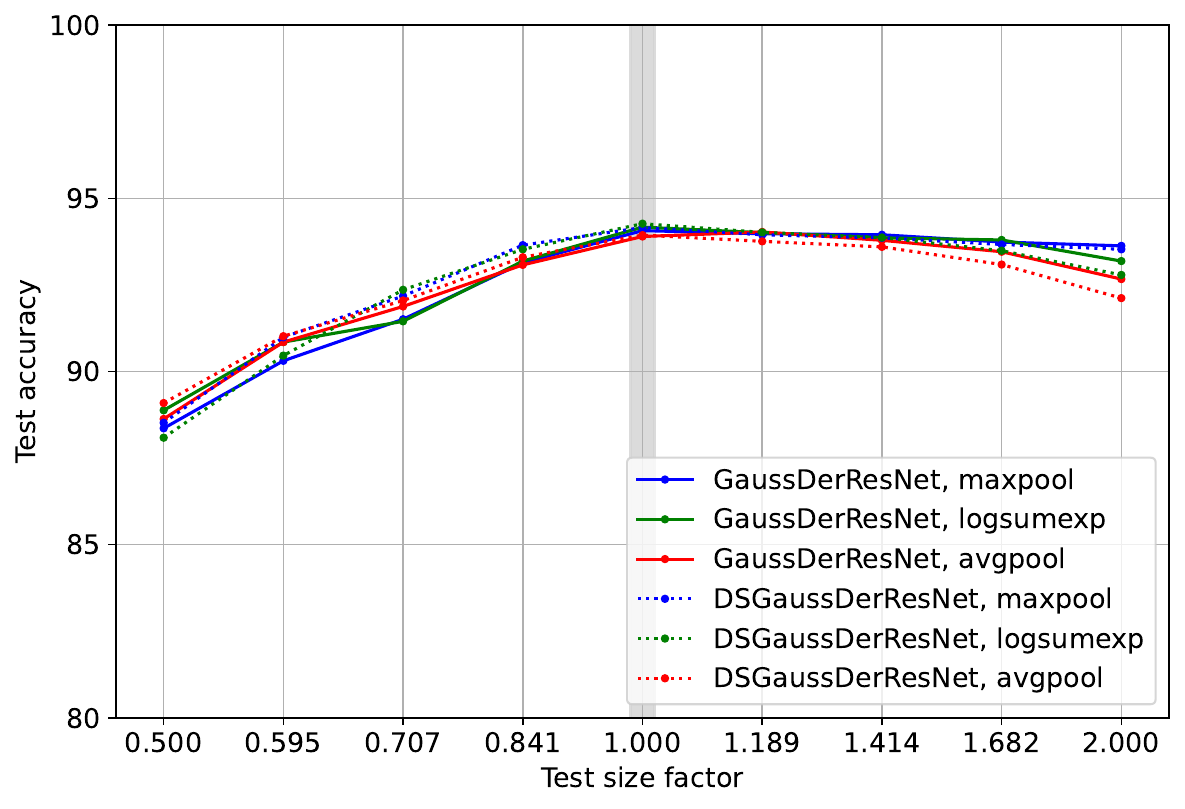}
        \caption{Rescaled Fashion-MNIST}
        \label{fig:subfiga-1-dwsc}
    \end{subfigure}
    \hfill
    \begin{subfigure}{0.32\textwidth}
        \centering
        \includegraphics[width=\linewidth]{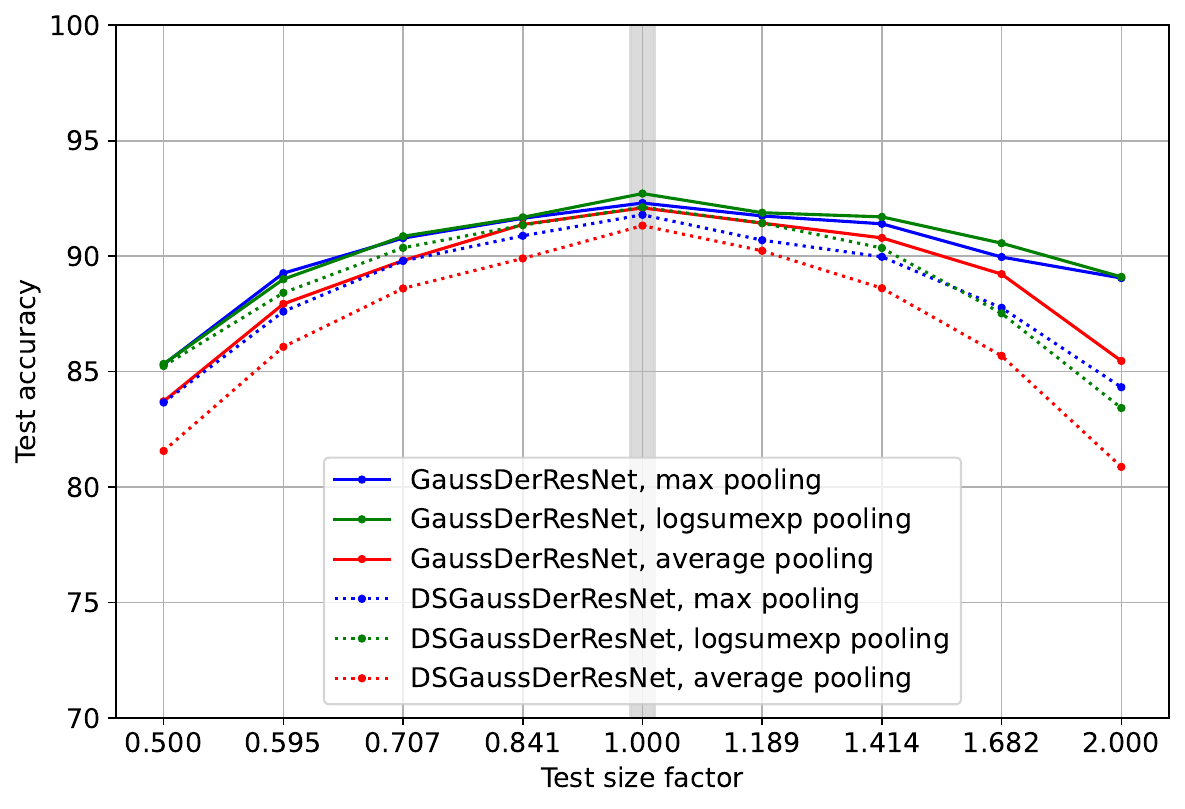}
        \caption{Rescaled CIFAR-10}
        \label{fig:subfigb-2-dwsc}
    \end{subfigure}
    \hfill
    \begin{subfigure}{0.32\textwidth}
        \centering
        \includegraphics[width=\linewidth]{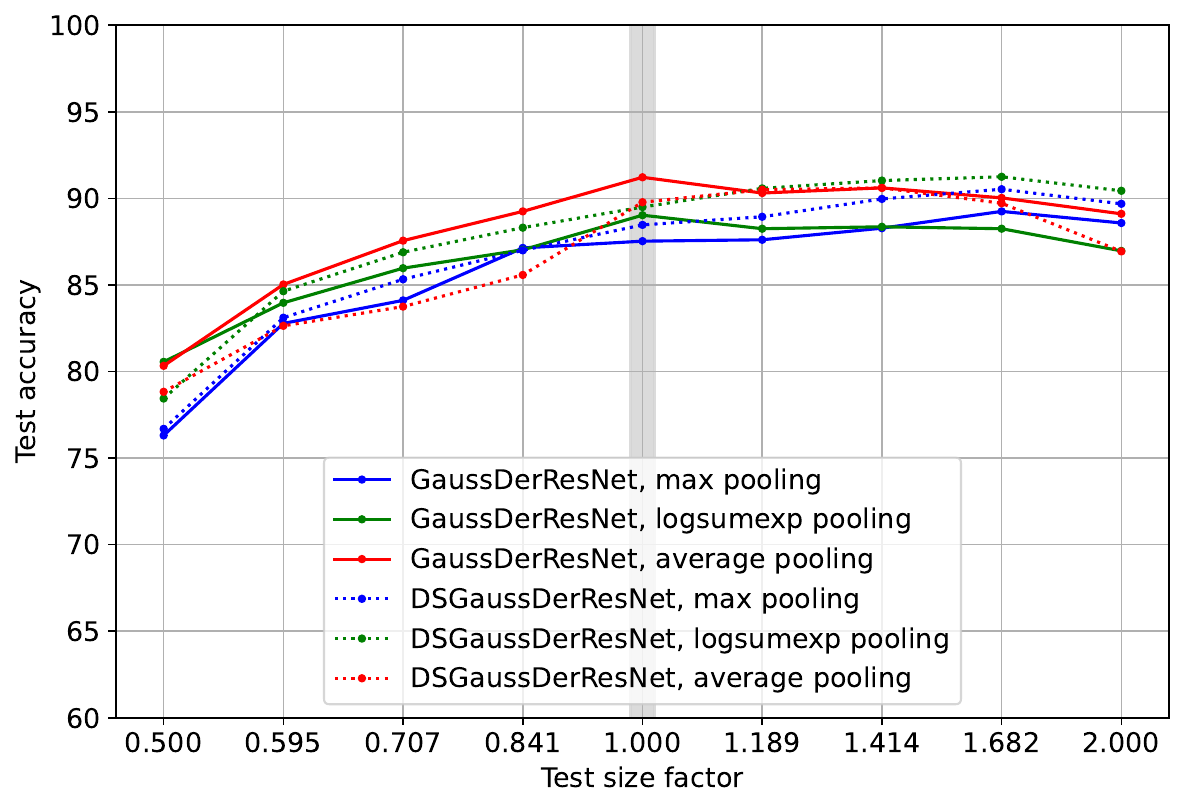}
        \caption{Rescaled STL-10}
        \label{fig:subfigb-3-dwsc}
    \end{subfigure}
  \end{center}
\caption{Comparison between the scale generalisation curves 
for the multi-scale-channel GaussDerResNets and the 
depthwise-separable GaussDerResNets (DSGaussDerResNets), 
with the networks based on scale selection using either max, 
logsumexp or average pooling over scales, and trained on the 
rescaled (a)~Fashion-MNIST, (b)~CIFAR-10 and (c)~STL-10 
datasets. The training of the models was performed on the 
training data for the size factor 1, while the evaluation was 
performed on the corresponding test datasets for all the size 
factors between 1/2 and 2. As can be seen from the graphs, 
for the rescaled Fashion-MNIST and STL-10 datasets, the performance
for the DSGaussDerResNets is similar to the performance of 
the GaussDerResNets. For the rescaled CIFAR-10 dataset, 
however, the performance for larger size factors is lower, probably 
caused by a less suitable choice of training parameters. (Note 
that the vertical axes for the test accuracy use different lower 
bounds for the different datasets.)}
\label{fig-gaussresnet-dwsc-scale-gen}
\end{figure*}

\begin{figure*}[ht]
  \begin{center}
    \begin{subfigure}{0.32\textwidth}
        \centering
        \textit{\quad \quad Scale selection on rescaled Fashion-MNIST}
        \includegraphics[width=\linewidth]{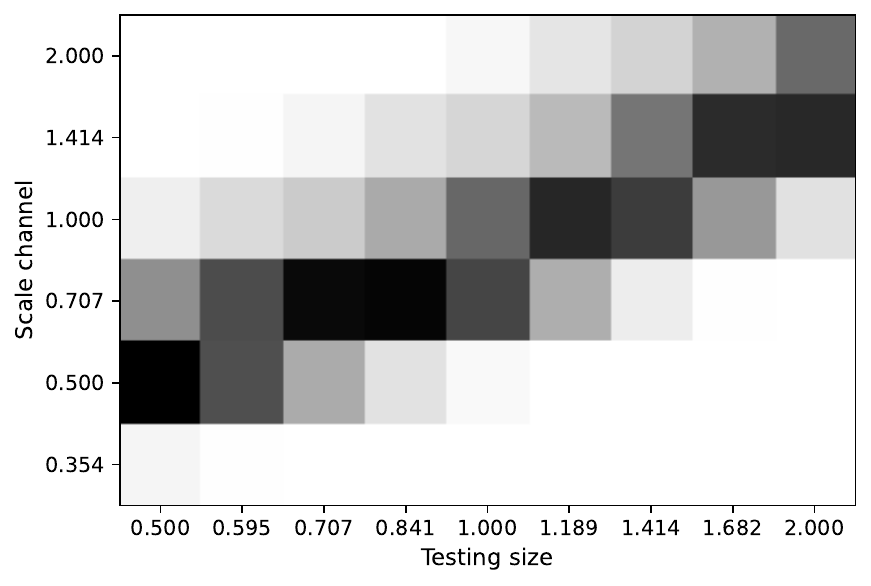}
        \label{fig:subfig-dwsc-scsel-1}
    \end{subfigure}
    \hfill
    \begin{subfigure}{0.32\textwidth}
        \centering
        \textit{\quad \quad Scale selection on rescaled CIFAR-10}
        \includegraphics[width=\linewidth]{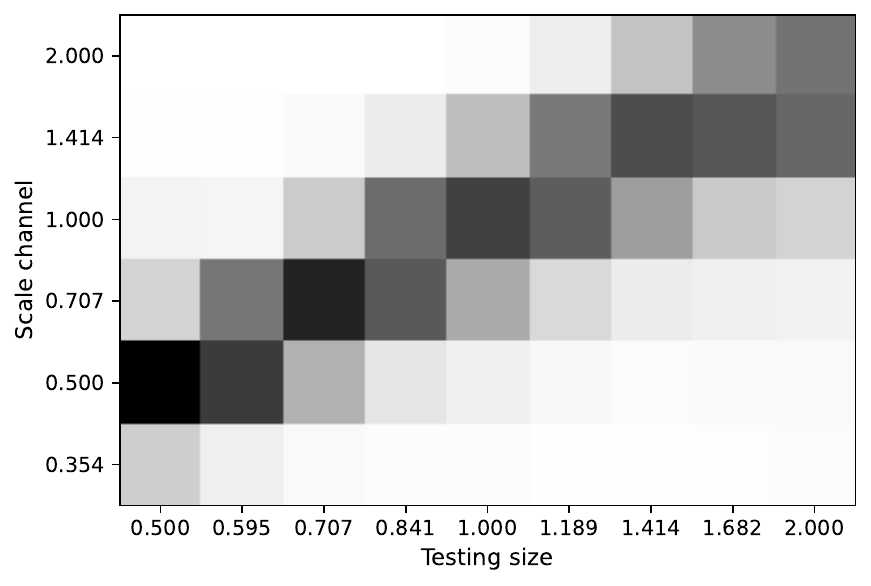}
        \label{fig:subfig-dwsc-scsel-2}
    \end{subfigure}
    \hfill
    \begin{subfigure}{0.32\textwidth}
        \centering
        \textit{\quad \quad Scale selection on rescaled STL-10}
        \includegraphics[width=\linewidth]{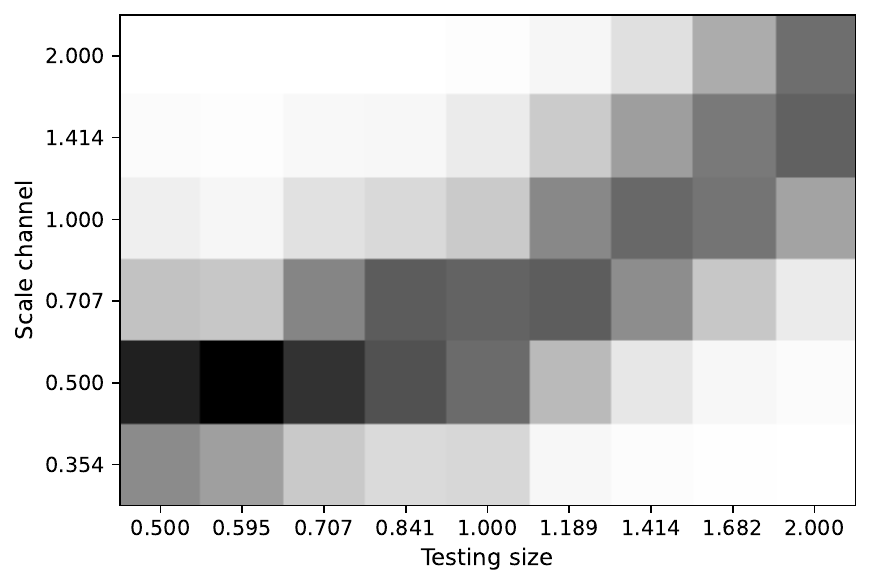}
        \label{fig:subfig-dwsc-scsel-3}
    \end{subfigure}
  \end{center}
\caption{Scale selection histograms for the multi-scale-channel 
DSGaussDerResNets based on max pooling over scales applied 
to the rescaled (a)~Fashion-MNIST, (b)~CIFAR-10 and (c)~STL-10 
datasets. These histograms visualise the relative contribution of each 
scale channel to the final classification. From these results we 
can see that, as expected from the theory, the GaussDerResNets 
based on depthwise-separable convolution operations have very 
similar scale selection properties as standard GaussDerResNets, 
with a rather clear linear trend obtained on all the 
rescaled datasets. This trend demonstrates that the same scale 
covariance properties hold between the scale channels based on 
standard or depthwise-separable convolutions, with the selected 
scale levels being proportional to the size of image features in the 
testing data.}
\label{fig-gaussresnet-dwsc-scale-selection}
\end{figure*}

\section{Gaussian derivative networks based on standard vs. depthwise-separable convolutions}
\label{sec-standard-vs-dwsc}

The GaussDerResNet architecture studied so far is based on standard
convolutions, however, it is also possible to construct scale-invariant 
GaussDerResNets based on depthwise-separable convolutions, referred 
to as DSGaussDerResNets, as outlined in Section~3.4.2 in the main 
body of the paper.

Incorporating depthwise-separable convolutions into the architecture 
can be beneficial, as such operations may achieve comparable 
performance to standard convolutions, while requiring significantly
fewer parameters as well as reducing the computational cost
by reusing the results from the spatial convolutions. Therefore, 
depthwise-separable convolutions are often included in modern
architectures for CNNs.

To investigate how depthwise-separable convolutions affect 
the performance and scale invariance properties of the GaussDerResNets,
we compared the test accuracy and scale generalisation of
DSGaussDerResNets and GaussDerResNets on the rescaled 
Fashion-MNIST, CIFAR-10 and STL-10 datasets.

The DSGaussDerResNet architecture is optimised for efficiency, 
by basing the first and last layer, as well as the first 4 residual blocks, 
on standard convolutions (meaning that no fusion with a pointwise 
convolution stage is used), while the remaining 5 blocks are based 
on depthwise-separable convolutions.
This design shares some similarity with the EfficientNetV2 architecture 
(Tan and Le \citeyear{TanLe21-ICML}). Specifically, it increases the 
efficiency, because while depthwise-separable convolutions are 
beneficial to include in the higher layers, they are computationally slow in the early layers.
Due to the use of depthwise-separable convolution based layers, 
the DSGaussDerResNets use several times fewer network parameters 
than the standard GaussDerResNets. The number of network parameters 
for the corresponding rescaled datasets used in our experiments is 
shown in Table~\ref{tab:dwsc-num-network-param}.

In the same manner as for the GaussDerResNets used in the 
experiments in Section~6 in the main body of the paper,
the DSGaussDerResNets in this study used, for the most part, 
the parameter values and the training configurations summarised 
in Section 5 in the main body of the paper, with the only difference 
being that for the rescaled CIFAR-10 dataset we used the relative
scale ratio $r=1.22$ between adjacent Gaussian derivative residual
blocks and for the rescaled STL-10 dataset we used an initial learning rate of 
0.005 and 170 training epochs, as this was found to improve the training stability. 
The multi-scale-channel networks were constructed using scale channels with 
$\sigma_{i,0} \in \{1/(2\sqrt{2}), 1/2, 1/\sqrt{2}, 1$, $\sqrt{2}, 2\}$, 
and the training process included label smoothing and learning 
rate warm-up. For the rescaled Fashion-MNIST and CIFAR-10 
datasets, central pixel extraction was used as the spatial selection 
method, while spatial max pooling was used for the rescaled STL-10 
dataset. Additionally, for the rescaled STL-10 dataset, the networks 
were once again based on layers that include a zero-order term.

Graphs showing the resulting scale generalisation curves obtained from 
these experiments are shown in Figure~\ref{fig-gaussresnet-dwsc-scale-gen}.
For the rescaled Fashion-MNIST dataset, which is a relatively simple dataset,
we can see that there are no significant differences between the scale generalisation performance for
the GaussDerResNets and the DSGaussDerResNets, with only a minor relative drop in 
performance found at larger size factors, for the DSGaussDerResNets, and a minor
relative performance increase found at small size factors.
 
For the rescaled CIFAR-10 dataset, we can see that the GaussDerResNets 
achieve higher performance on the size factor 1, of less than 1\%, 
compared to the DSGaussDerResNets. The scale generalisation of 
the GaussDerResNets is flatter compared to the DSGaussDerResNets, 
with the networks based on logsumexp and max pooling showing the 
best performance. A notable drop in performance for the larger size factors 
is observed for the DSGaussDerNets compared to the GaussDerResNets, 
up to $\sim$6~ppt, and only a minor drop for the smaller size factors, 
ranging between 0 and $\sim$2~ppt. This level of performance is 
good, considering that the DSGaussDerResNets use almost 4 times 
fewer network parameters compared to the GaussDerResNets.

Additionally, for the rescaled STL-10 dataset, the performance and the scale generalisation of
the average pooling based GaussDerResNet is for the most part better than the 
DSGaussDerResNets, with between 1.5 to 3~ppt improved performance at the 
size factor 1. We also see that for the larger size factors, the performance curve 
of the DSGaussDerResNets based on max and logsumexp pooling is slightly 
better relative to all the GaussDerResNets, and furthermore it outperforms the 
GaussDerResNets based on corresponding pooling methods across all size 
factors, by up to 2--4~ppt.

\begin{figure*}[ht]
  \begin{center}
    \begin{subfigure}{0.32\textwidth}
        \centering
        \textit{\quad \quad Scale generalisation on rescaled Fashion-MNIST}
        \includegraphics[width=\linewidth]{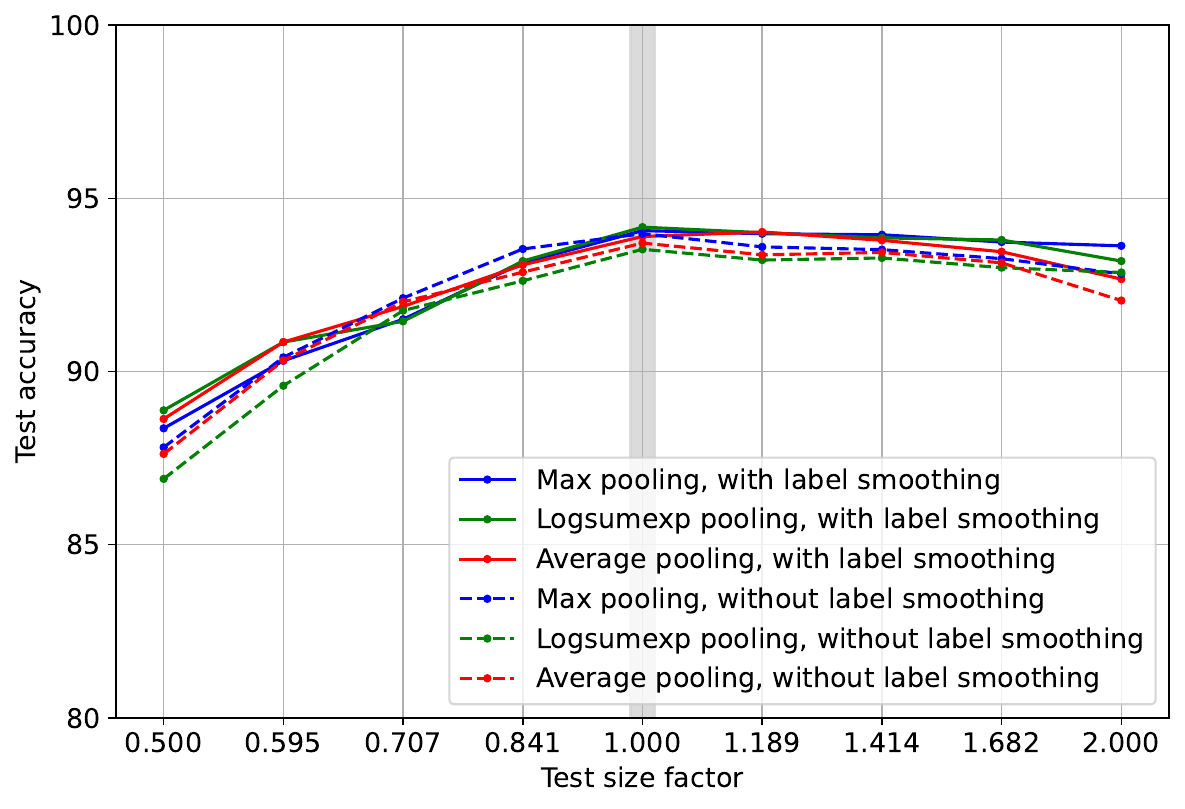}
        \caption{Fashion-MNIST comparison}
        \label{fig:subfig-labelsmooth-1}
    \end{subfigure}
    \hfill
    \begin{subfigure}{0.32\textwidth}
        \centering
        \textit{\quad \quad Scale generalisation on rescaled CIFAR-10}
        \includegraphics[width=\linewidth]{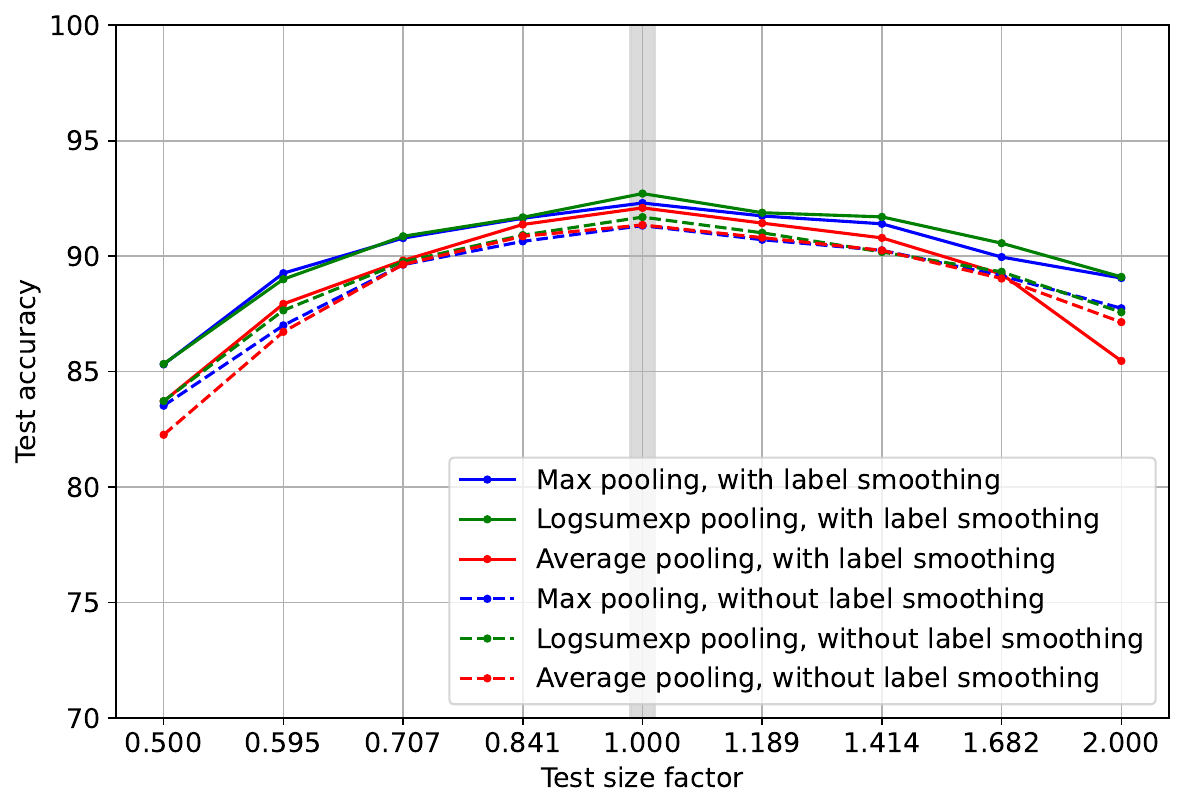}
        \caption{CIFAR-10 comparison}
        \label{fig:subfig-labelsmooth-2}
    \end{subfigure}
    \hfill
    \begin{subfigure}{0.32\textwidth}
        \centering
        \textit{\quad \quad Scale generalisation on rescaled STL-10}
        \includegraphics[width=\linewidth]{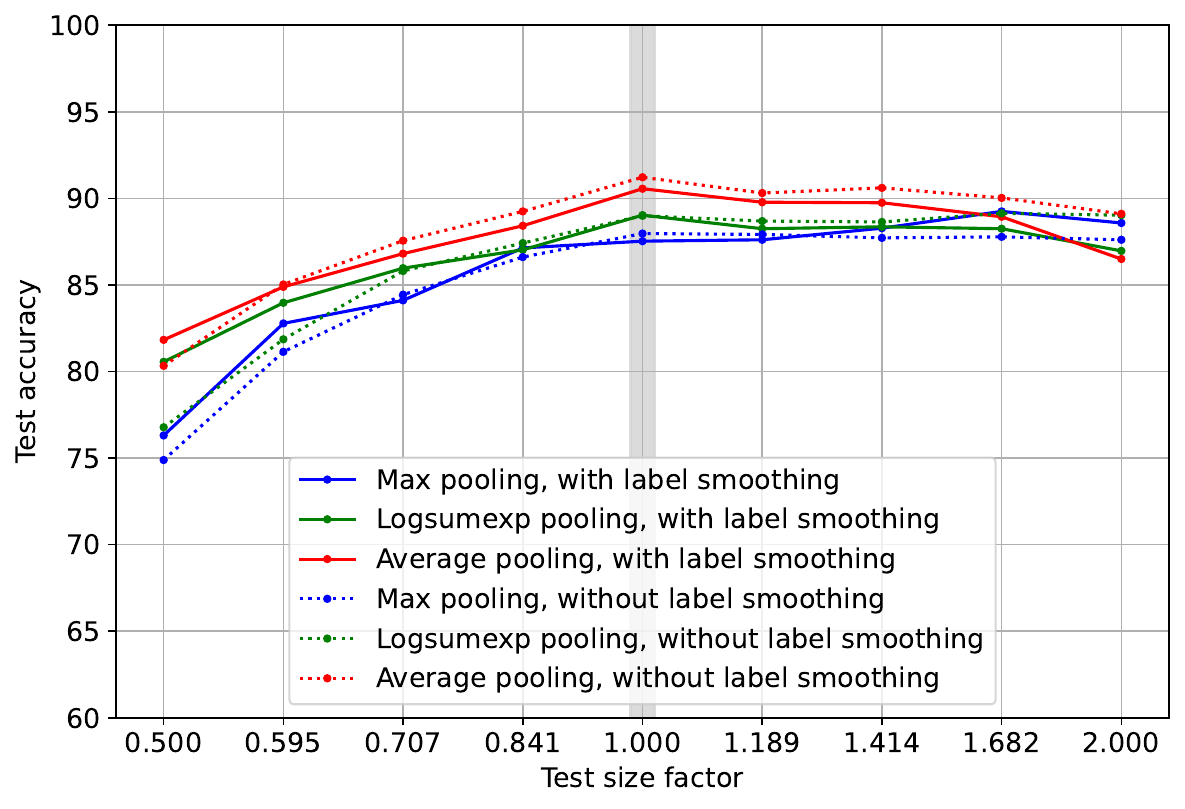}
        \caption{STL-10 comparison}
        \label{fig:subfig-labelsmooth-3}
    \end{subfigure}
  \end{center}
\caption{Scale generalisation curves for the multi-scale-channel networks 
  based on max, logsumexp or average pooling over scales, trained with 
  or without label smoothing during training, for the (left)~the rescaled 
  Fashion-MNIST dataset, (middle)~the rescaled CIFAR-10 dataset, and 
  (right)~the rescaled STL-10 dataset. The networks were then tested on 
  all the size factors between 1/2 and 2, after training on the single size 
  factor 1. The graphs show that using label smoothing as an additional 
  regularisation method during training typically results in better overall 
  performance, while retaining good scale generalisation properties. 
  Note that the vertical axes for the test accuracy use different lower 
  bounds for the different datasets.}
\label{fig-gaussresnet-sc-label-smooth-appendix}
\end{figure*}

\begin{figure*}[hbt]
  \begin{center}
    \begin{subfigure}{0.32\textwidth}
        \centering
        \includegraphics[width=\linewidth]{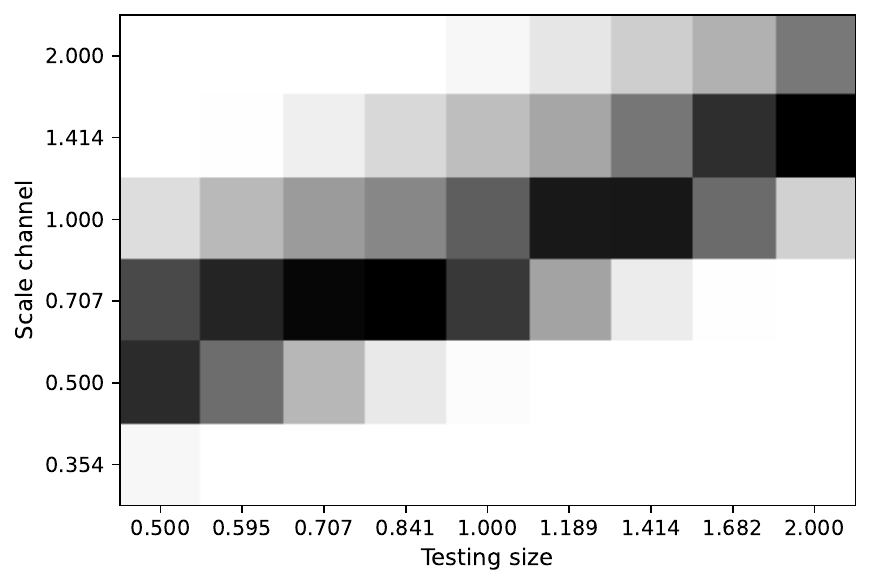}
        \caption{Max pooling, with label smoothing}
        \label{fig:subfig-fashion-lbl-hist-1}
    \end{subfigure}
    \hfill
    \begin{subfigure}{0.32\textwidth}
        \centering
        \includegraphics[width=\linewidth]{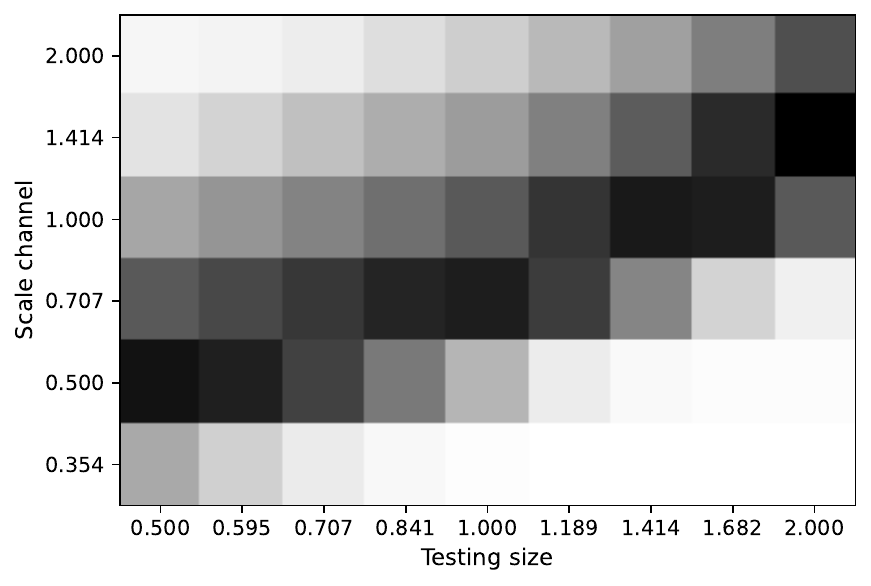}
        \caption{Logsumexp pooling, with label smoothing}
        \label{fig:subfig-fashion-lbl-hist-2}
    \end{subfigure}
    \hfill
    \begin{subfigure}{0.32\textwidth}
        \centering
        \includegraphics[width=\linewidth]{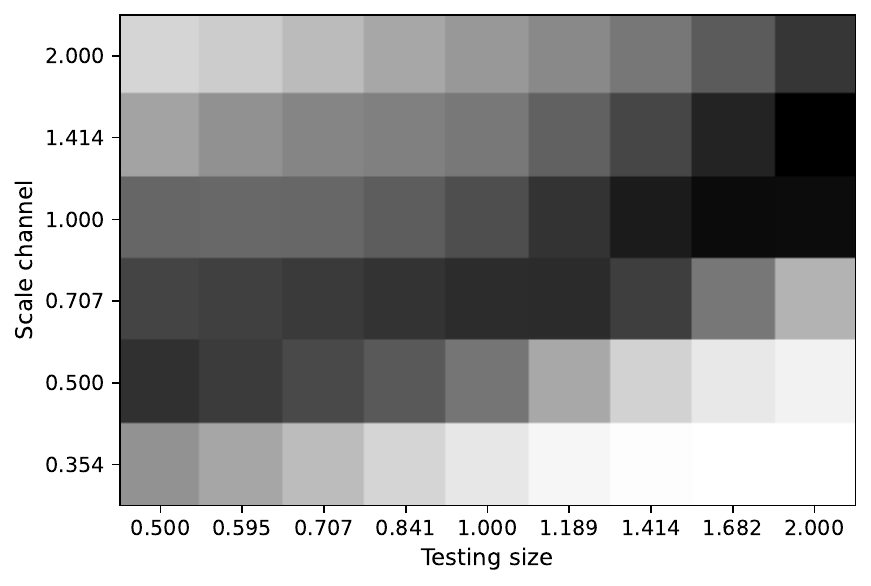}
        \caption{Average pooling, with label smoothing}
        \label{fig:subfig-fashion-lbl-hist-3}
    \end{subfigure}
    
    \vspace{0.4cm} 

    \begin{subfigure}{0.32\textwidth}
        \centering
        \includegraphics[width=\linewidth]{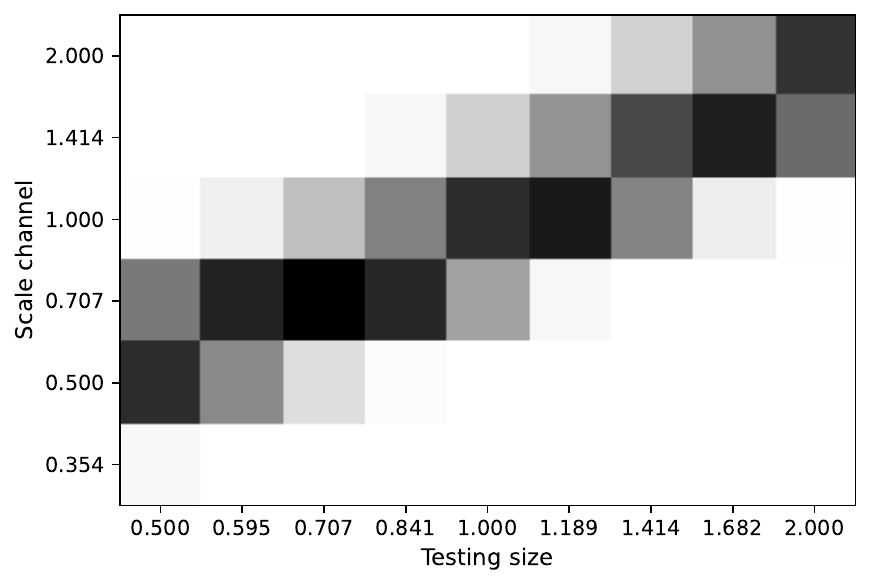}
        \caption{Max pooling, no label smoothing}
        \label{fig:subfig-fashion-lbl-hist-4}
    \end{subfigure}
    \hfill
    \begin{subfigure}{0.32\textwidth}
        \centering
        \includegraphics[width=\linewidth]{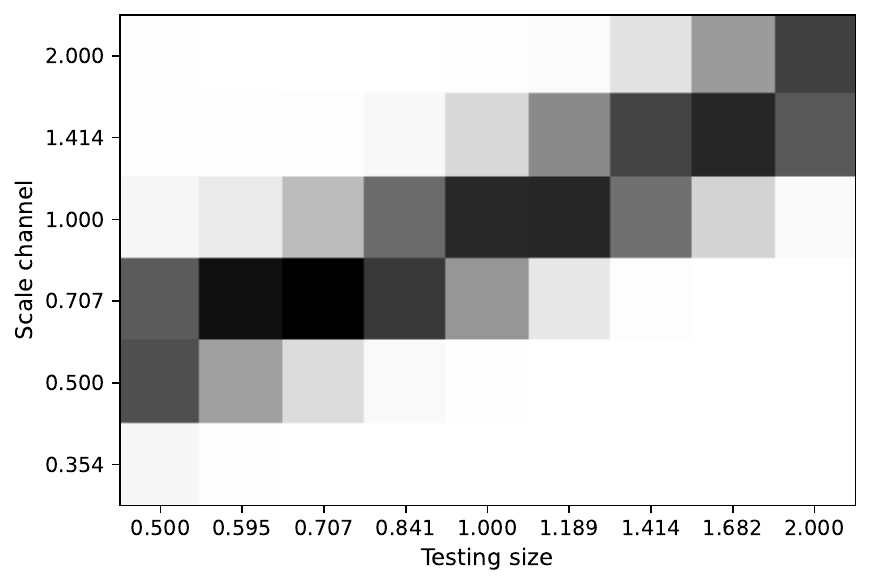}
        \caption{Logsumexp pooling, no label smoothing}
        \label{fig:subfig-fashion-lbl-hist-5}
    \end{subfigure}
    \hfill
    \begin{subfigure}{0.32\textwidth}
        \centering
        \includegraphics[width=\linewidth]{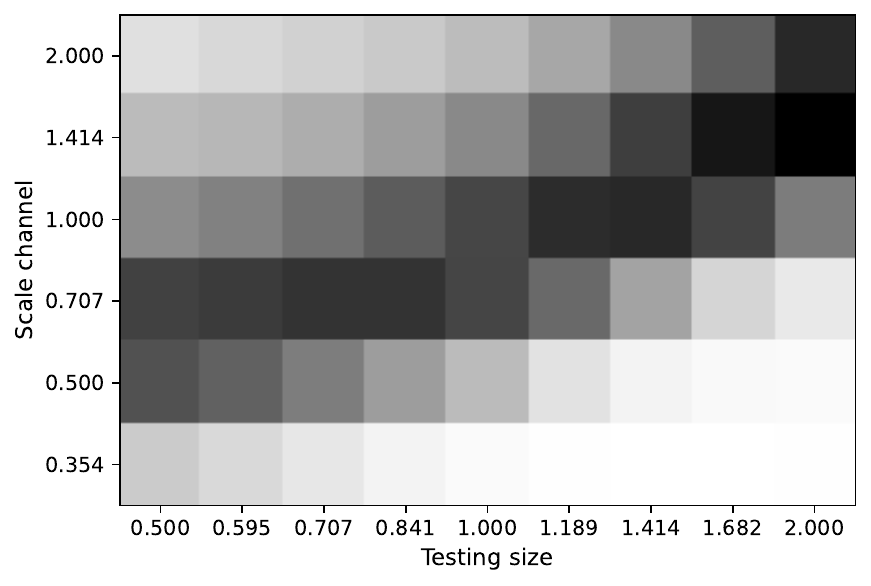}
        \caption{\small Average pooling, no label smoothing}
        \label{fig:subfig-fashion-lbl-hist-6}
    \end{subfigure}
  \end{center}
\caption{Scale selection histograms for the multi-scale-channel 
GaussDerResNets applied to the rescaled Fashion-MNIST dataset, 
trained (a)--(c) with label smoothing and trained (d)--(f) without label 
smoothing, for all the three permutation invariant pooling over
scale methods: max, logsumexp and average pooling. The scale 
selection histograms demonstrate a clear linear trend between 
the selected scale levels (the scale channels) and the size in the 
testing data. We can also see that the networks based on label 
smoothing create more spread out and less diagonal linear trends 
in the scale selection histograms.}
\label{fig-fashion-gaussresnet-scale-selection-appendix}
\end{figure*}

The property, that the DSGaussDerResNets attain comparable performance and 
scale generalisation on the rescaled CIFAR-10 and STL-10 datasets, 
with roughly a quarter the number of parameters, underscores the 
efficiency of their representations, and benefits of longer training.

Figure~\ref{fig-gaussresnet-dwsc-scale-selection} shows
the scale selection histograms for the DSGaussDerResNets 
trained on the rescaled Fashion-MNIST, CIFAR-10 and STL-10 
datasets. The histograms were obtained by accumulating the 
relative contributions of each scale channel to the predictions. 
For each dataset, we can see a clear linear trend, similar to the 
behaviour of classical methods for scale selection. The good 
scale generalisation properties and the linear trend showcased 
in the scale selection histograms obtained in the experimental 
results in this section empirically confirm that the scale-invariant 
properties of the GaussDerResNets also hold for the 
DSGaussDerResNets, in agreement with the theoretical analysis.

\begin{figure*}[btp]
  \begin{center}
    \begin{subfigure}{0.32\textwidth}
        \centering
        \includegraphics[width=\linewidth]{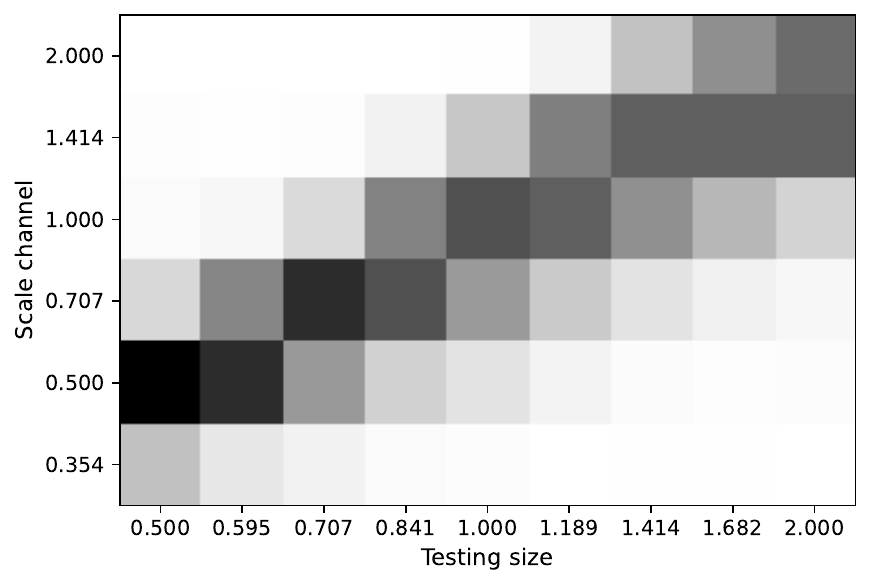}
        \caption{Max pooling with label smoothing}
        \label{fig:subfig-cifar-lbl-hist-1}
    \end{subfigure}
    \hfill
    \begin{subfigure}{0.32\textwidth}
        \centering
        \includegraphics[width=\linewidth]{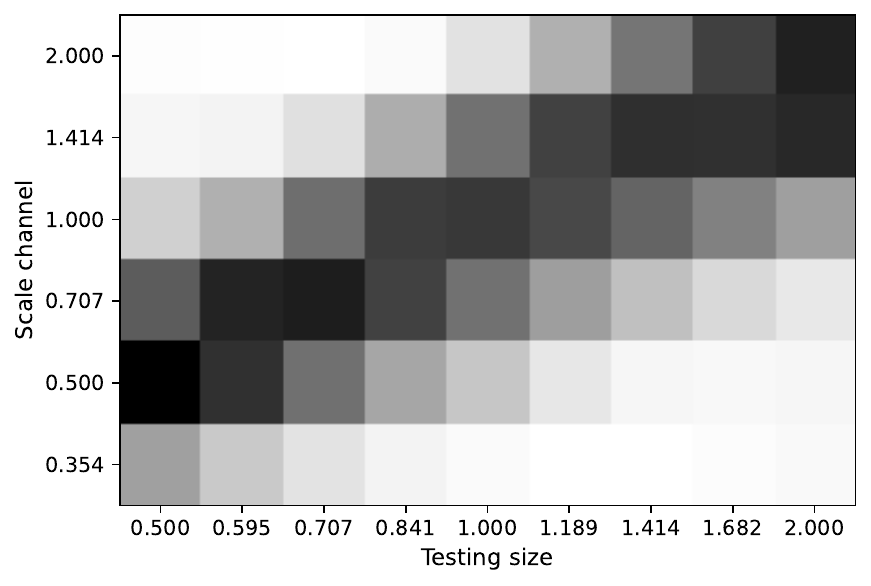}
        \caption{Logsumexp pooling with label smoothing}
        \label{fig:subfig-cifar-lbl-hist-2}
    \end{subfigure}
    \hfill
    \begin{subfigure}{0.32\textwidth}
        \centering
        \includegraphics[width=\linewidth]{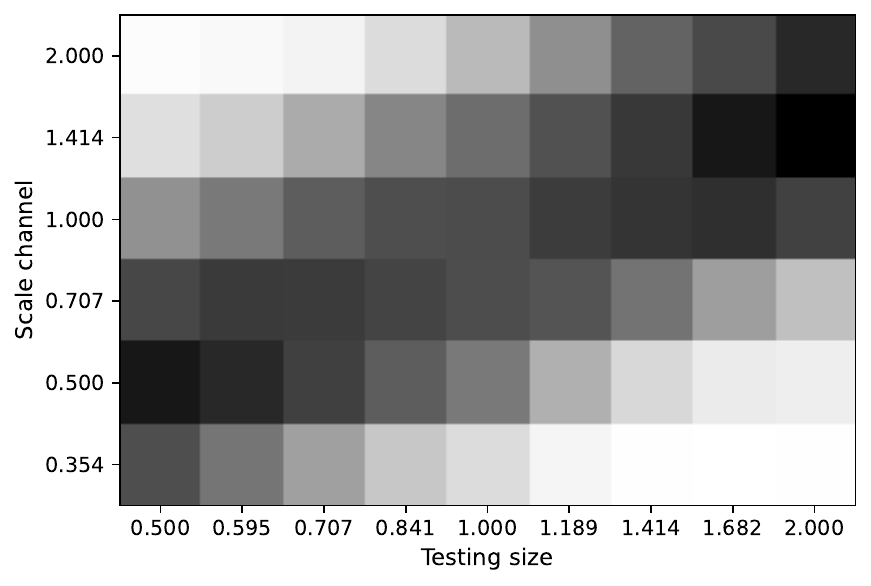}
        \caption{Average pooling with label smoothing}
        \label{fig:subfig-cifar-lbl-hist-3}
    \end{subfigure}
    
    \vspace{0.4cm} 

    \begin{subfigure}{0.32\textwidth}
        \centering
        \includegraphics[width=\linewidth]{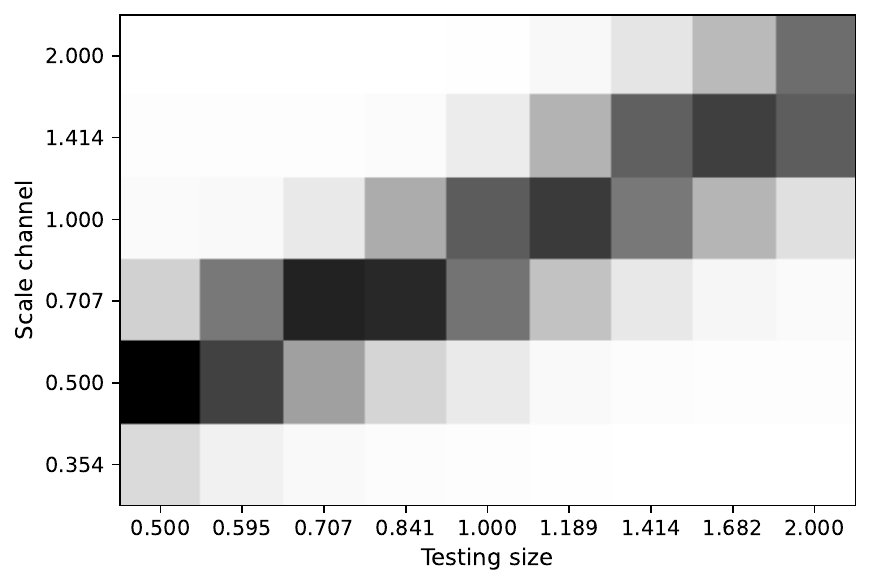}
        \caption{Max pooling, no label smoothing}
        \label{fig:subfig-cifar-lbl-hist-4}
    \end{subfigure}
    \hfill
    \begin{subfigure}{0.32\textwidth}
        \centering
        \includegraphics[width=\linewidth]{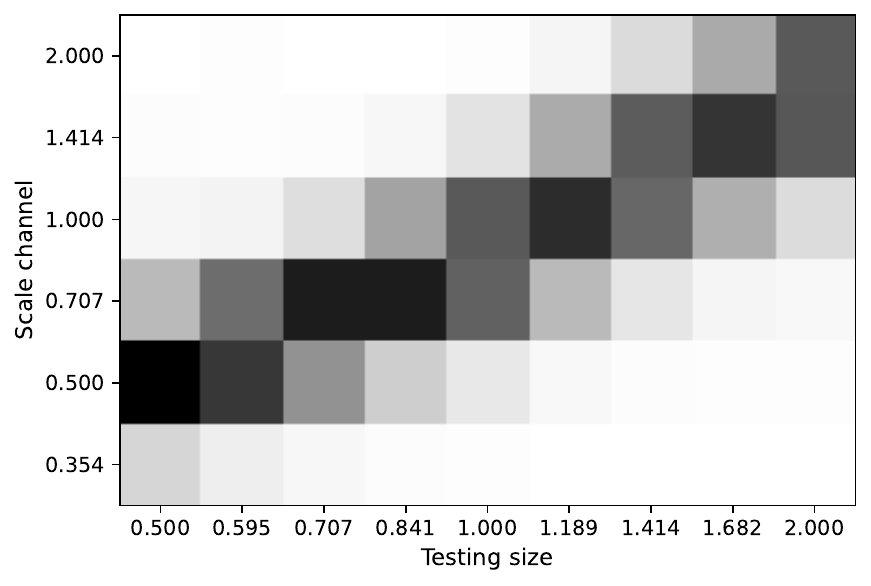}
        \caption{Logsumexp pooling, no label smoothing}
        \label{fig:subfig-cifar-lbl-hist-5}
    \end{subfigure}
    \hfill
    \begin{subfigure}{0.32\textwidth}
        \centering
        \includegraphics[width=\linewidth]{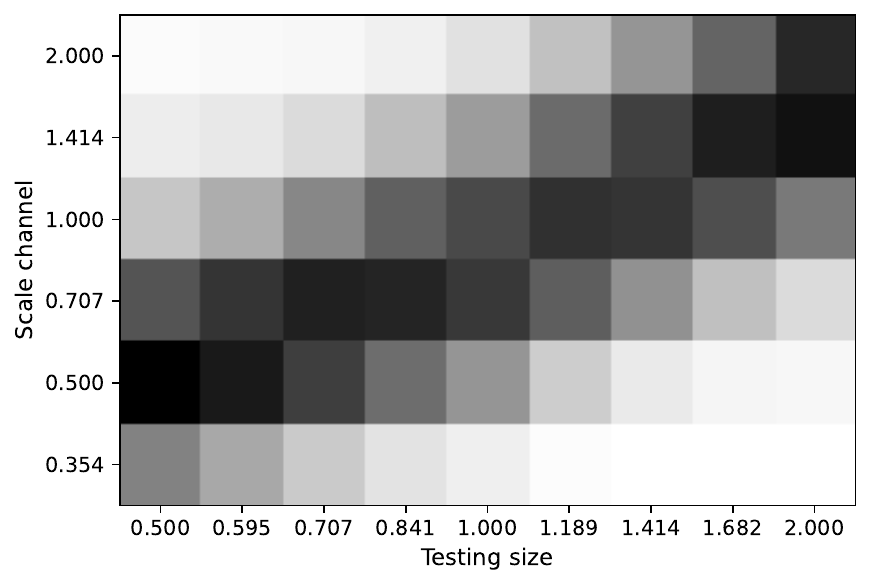}
        \caption{Average pooling, no label smoothing}
        \label{fig:subfig-cifar-lbl-hist-6}
    \end{subfigure}
  \end{center}
\caption{Scale selection histograms for the multi-scale-channel 
GaussDerResNets applied to the rescaled CIFAR-10 dataset, 
trained (a)--(c) with label smoothing and trained (d)--(f) without label 
smoothing, for all the three permutation invariant pooling over 
scale methods: max, smooth max (logsumexp) and average pooling. 
The scale selection histograms demonstrate a clear linear trend 
between the selected scale levels (the scale channels) and the 
size in the testing data. Once again we can see that the networks 
based on label smoothing create more spread out and less diagonal 
linear trends in the scale selection histograms.}
\label{fig-cifar10-gaussresnet-scale-selection-appendix}
\bigskip
\begin{center}
    \begin{subfigure}[t]{0.32\textwidth}
        \centering
        \includegraphics[width=\linewidth]{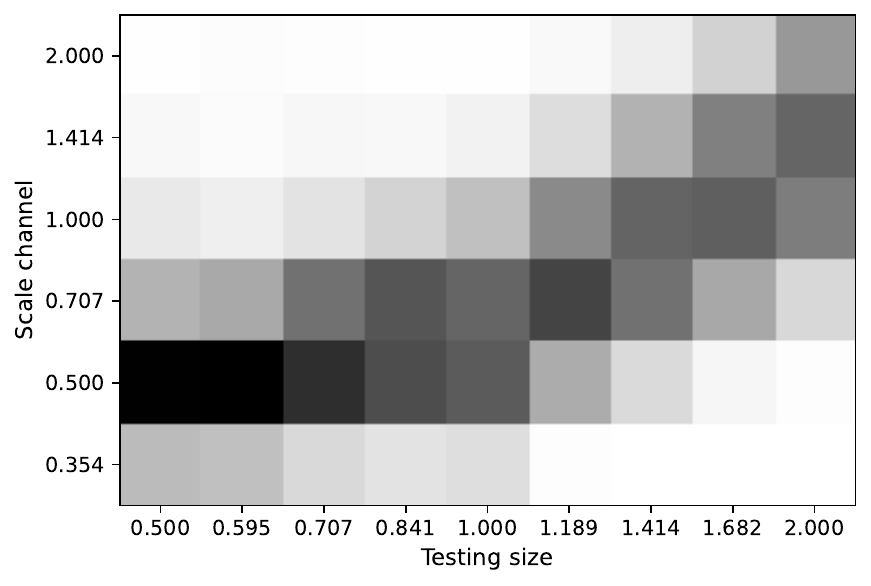}
        \caption{Max pooling with label smoothing}
        \label{fig:subfig-stl-lbl-hist-1}
    \end{subfigure}
    \hfill
    \begin{subfigure}[t]{0.32\textwidth}
        \centering
        \includegraphics[width=\linewidth]{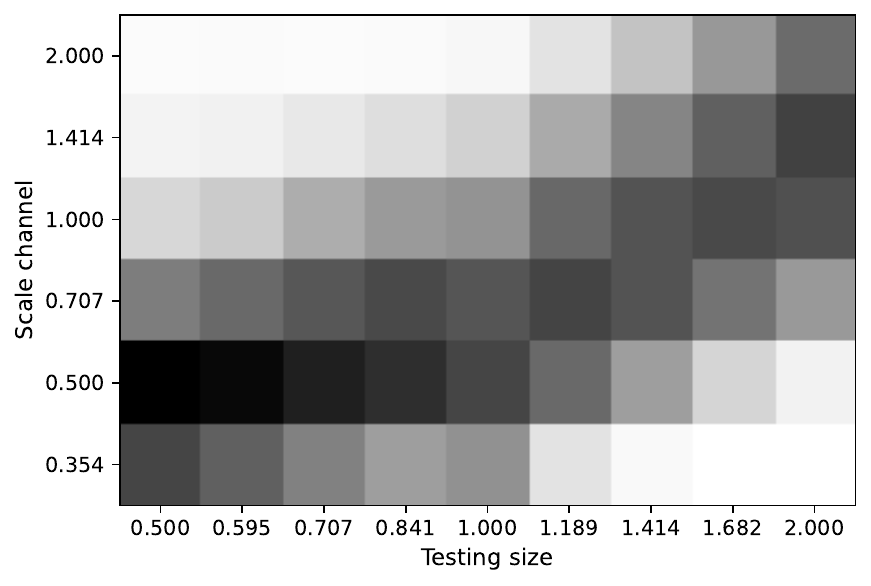}
        \caption{Logsumexp pooling with label smoothing}
        \label{fig:subfig-stl-lbl-hist-2}
    \end{subfigure}
    \hfill
    \begin{subfigure}[t]{0.32\textwidth}
        \centering
        \includegraphics[width=\linewidth]{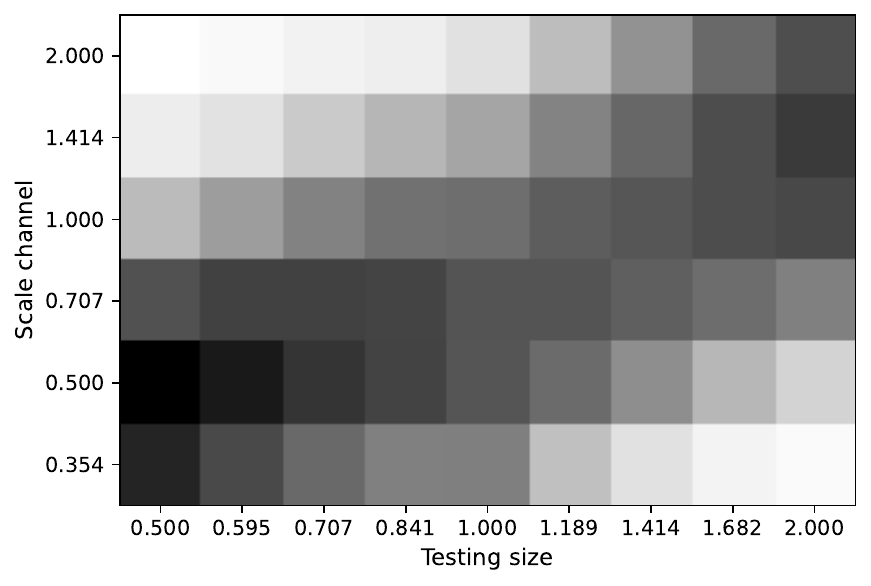}
        \caption{Average pooling with label smoothing}
        \label{fig:subfig-stl-lbl-hist-3}
    \end{subfigure}
    
    \vspace{0.4cm} 

    \begin{subfigure}{0.32\textwidth}
        \centering
        \includegraphics[width=\linewidth]{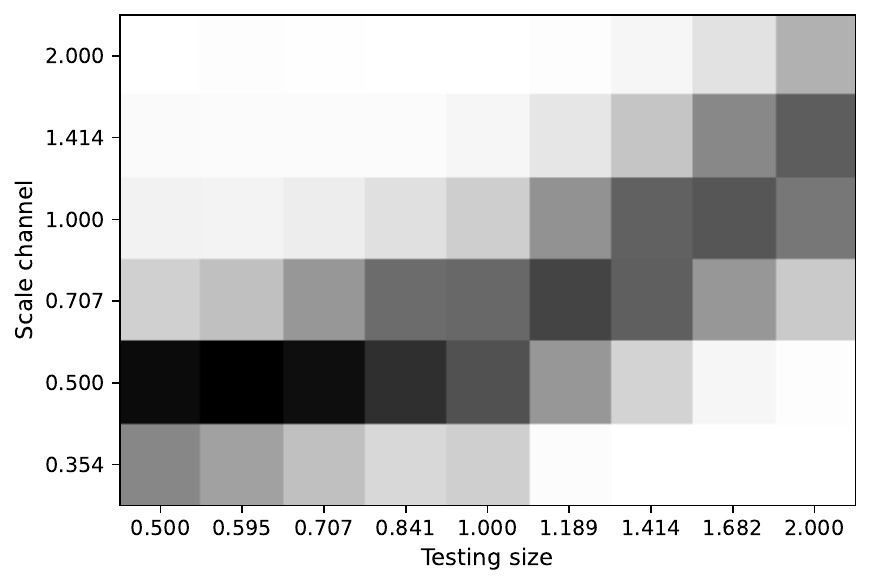}
        \caption{Max pooling, no label smoothing}
        \label{fig:subfig-stl-lbl-hist-4}
    \end{subfigure}
    \hfill
    \begin{subfigure}{0.32\textwidth}
        \centering
        \includegraphics[width=\linewidth]{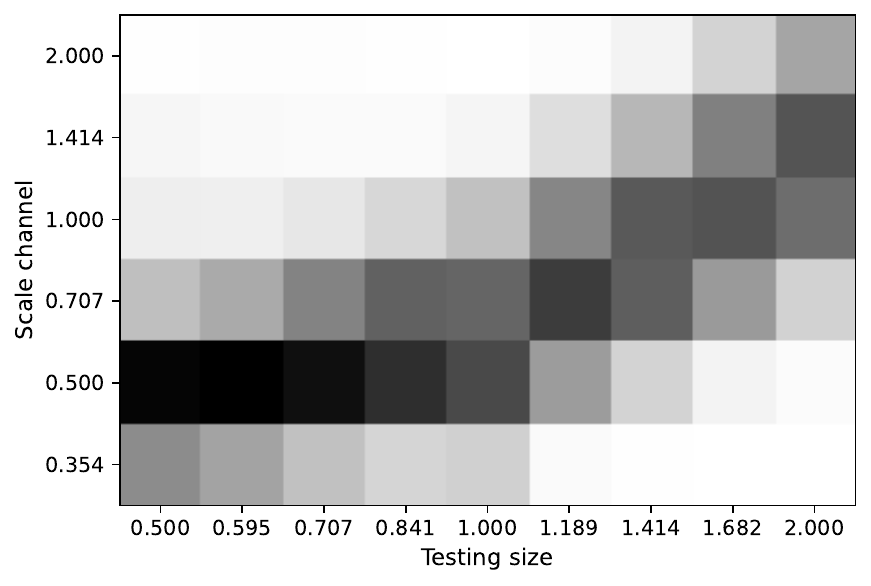}
        \caption{Logsumexp pooling, no label smoothing}
        \label{fig:subfig-stl-lbl-hist-5}
    \end{subfigure}
    \hfill
    \begin{subfigure}{0.32\textwidth}
        \centering
        \includegraphics[width=\linewidth]{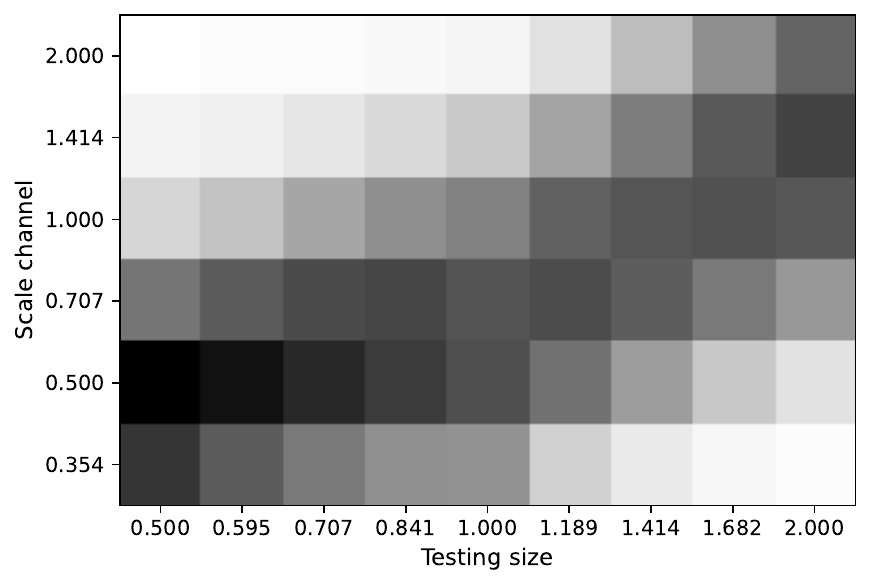}
        \caption{Average pooling, no label smoothing}
        \label{fig:subfig-stl-lbl-hist-6}
    \end{subfigure}
  \end{center}
\caption{Scale selection histograms for the multi-scale-channel 
GaussDerResNets applied to the rescaled STL-10 dataset, 
trained (a)--(c) with label smoothing and trained (d)--(f) without label 
smoothing, for all the three permutation invariant pooling over 
scale methods: max, smooth max (logsumexp) and average pooling. 
A fairly linear trend can be seen between the selected scale levels 
(the scale channels) and the size in the testing data, with only minor
flattening at small size factors. Furthermore, we can also see that the 
networks trained with label smoothing have a slightly more 
 We can also see that the networks based on label spread out diagonal 
 linear trends in the scale selection histograms.}
\label{fig-stl10-gaussresnet-scale-selection-appendix}
\end{figure*}

\section{Influence of label smoothing on scale generalisation and scale selection}
\label{sec-effect-of-label-smoothing-appendix}

To investigate how the use of label smoothing as a regularisation 
technique during training may affect the scale generalisation 
and scale selection properties of multi-scale-channel GaussDerResNets, 
we performed comparative studies on networks trained with 
and without label smoothing, on the rescaled Fashion-MNIST, 
CIFAR-10 and STL-10 datasets.

For each rescaled dataset, we trained a network for each possible 
permutation-invariant scale pooling approach: max pooling, 
logsumexp pooling or average pooling, respectively. The 
training was performed using the architectural parameters 
and the training protocol described in Section~5 in the main body 
of the paper, using scale channels based on the initial scale values
$\sigma_{i,0} \in \{1/(2\sqrt{2}), 1/2$, $1/\sqrt{2}, 1, \sqrt{2}, 2\}$. 
The only exceptions were the networks trained on the rescaled 
Fashion-MNIST dataset without the use of label smoothing, as 
they use slightly different training parameters, namely a weight 
decay value of 0.05 and scale channel dropout with $q=0.1$.

Figure~\ref{fig-gaussresnet-sc-label-smooth-appendix} shows the
resulting scale generalisation curves for each rescaled dataset, 
for networks trained on training data for the the size factor 1 and evaluated 
on copies of the test set, with each copy rescaled to a single 
distinct spatial size factor, ranging from 1/2 to 2.
As we can see from the results, using label smoothing during training 
results in, for the most part, better scale generalisation 
performance for the networks based on max or logsumexp pooling. For
the networks based on average pooling, we see that on the rescaled 
Fashion-MNIST and CIFAR-10 dataset using label smoothing is 
beneficial, showing improvement in improvement at every size factor, 
with the only exception being the performance for the size factor 2
for the rescaled CIFAR-10 dataset. 
For the rescaled STL-10 dataset, however, we see that not using label smoothing 
leads to improved performance, with only the performance for
the single size factor 1/2 showing a minor drop. For both the rescaled CIFAR-10 
and rescaled STL-10 datasets, there is, however, a certain drop in performance 
for the largest size factors, possibly caused by the training not having sufficiently converged.

Regarding the effect that label smoothing has on the scale selection properties, 
Figures~\ref{fig-fashion-gaussresnet-scale-selection-appendix}, 
\ref{fig-cifar10-gaussresnet-scale-selection-appendix} 
and~\ref{fig-stl10-gaussresnet-scale-selection-appendix}
depict the corresponding scale selection histograms, which 
visualise the contributions from each scale channel to the final 
classification. There, we can clearly see that in each case,
a linear trend is present, meaning that the size of the automatically 
selected scale channel is directly proportional to the size of the 
image structure. For the rescaled STL-10 dataset,
we see that for both of the regularisation methods, the scale selection 
trend is shifted slightly downwards, as the use of spatial max 
pooling as a spatial selection methods leads to the network 
preferring to focus on smaller features.

We can also see that using label smoothing during training leads 
to a trade-off, in that the networks based on label smoothing create 
slightly more spread out scale selection trends, 
while the scale generalisation performance is generally slightly better, 
as already seen in Figure~\ref{fig-gaussresnet-sc-label-smooth-appendix}. 
This is not surprising, as the role of label smoothing is to prevent
overconfidence in predictions, meaning that a reduction in the certainty 
of the prediction also leads to a minor reduction in certainty about which 
scale channel is the most appropriate to select.

\begin{figure*}[ht]
  \begin{center}
    \begin{subfigure}{0.32\textwidth}
        \centering
        \textit{\quad \quad Scale generalisation on rescaled Fashion-MNIST}
        \includegraphics[width=\linewidth]{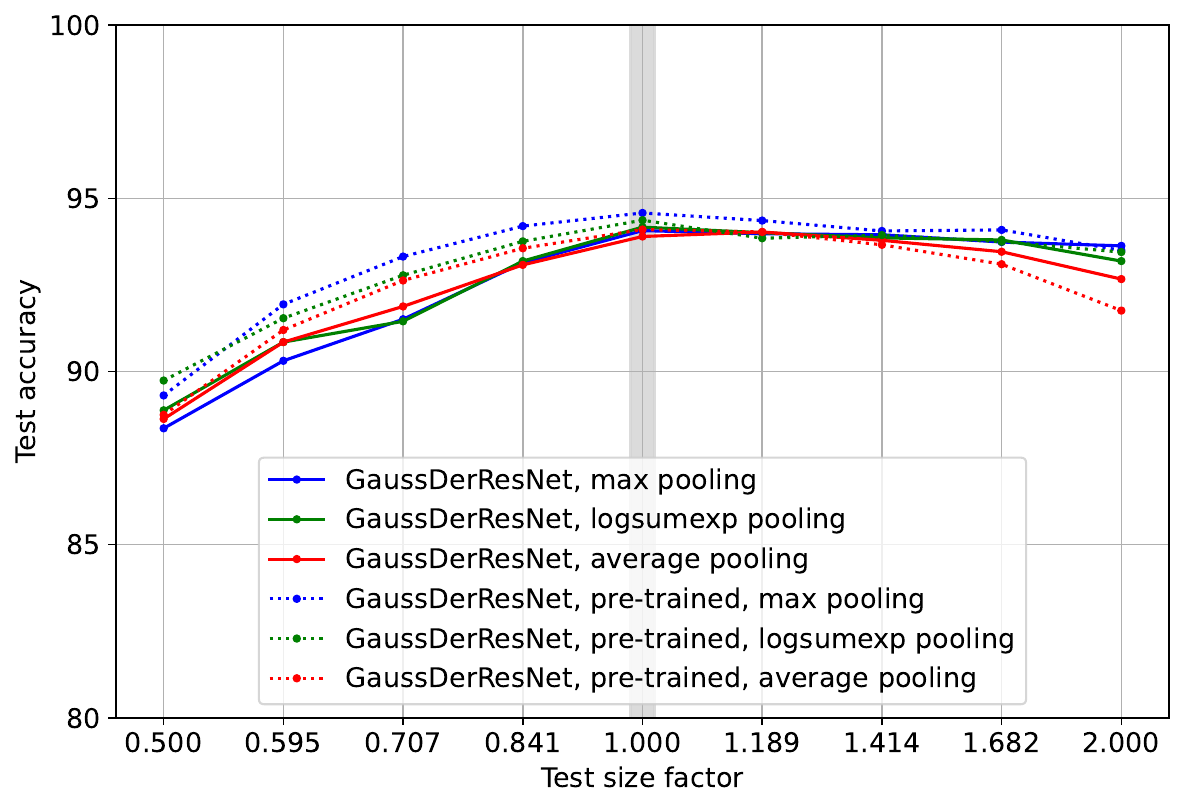}
        \label{fig:subfig-zeroth-1}
    \end{subfigure}
    \hfill
    \begin{subfigure}{0.32\textwidth}
        \centering
        \textit{\quad \quad Scale generalisation on rescaled CIFAR-10}
        \includegraphics[width=\linewidth]{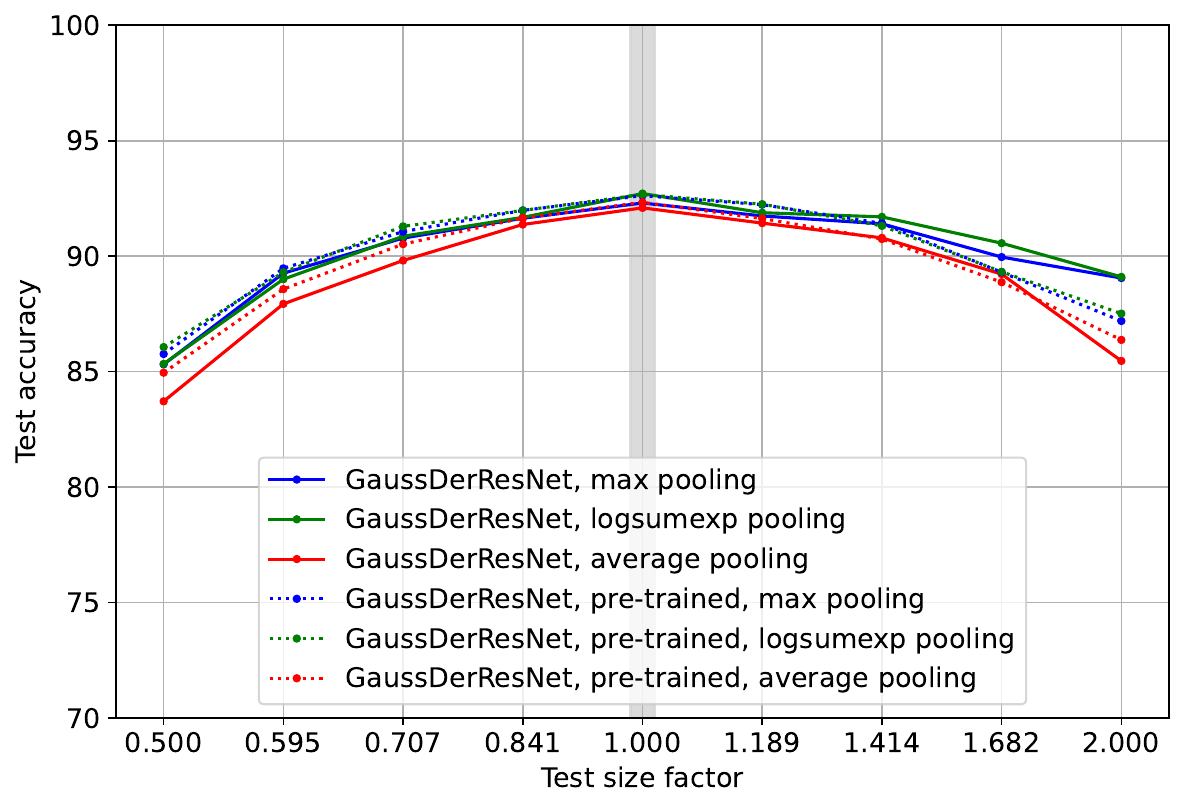}
        \label{fig:subfig-pre-train-2}
    \end{subfigure}
    \hfill
    \begin{subfigure}{0.32\textwidth}
        \centering
        \textit{\quad \quad Scale generalisation on rescaled STL-10}
        \includegraphics[width=\linewidth]{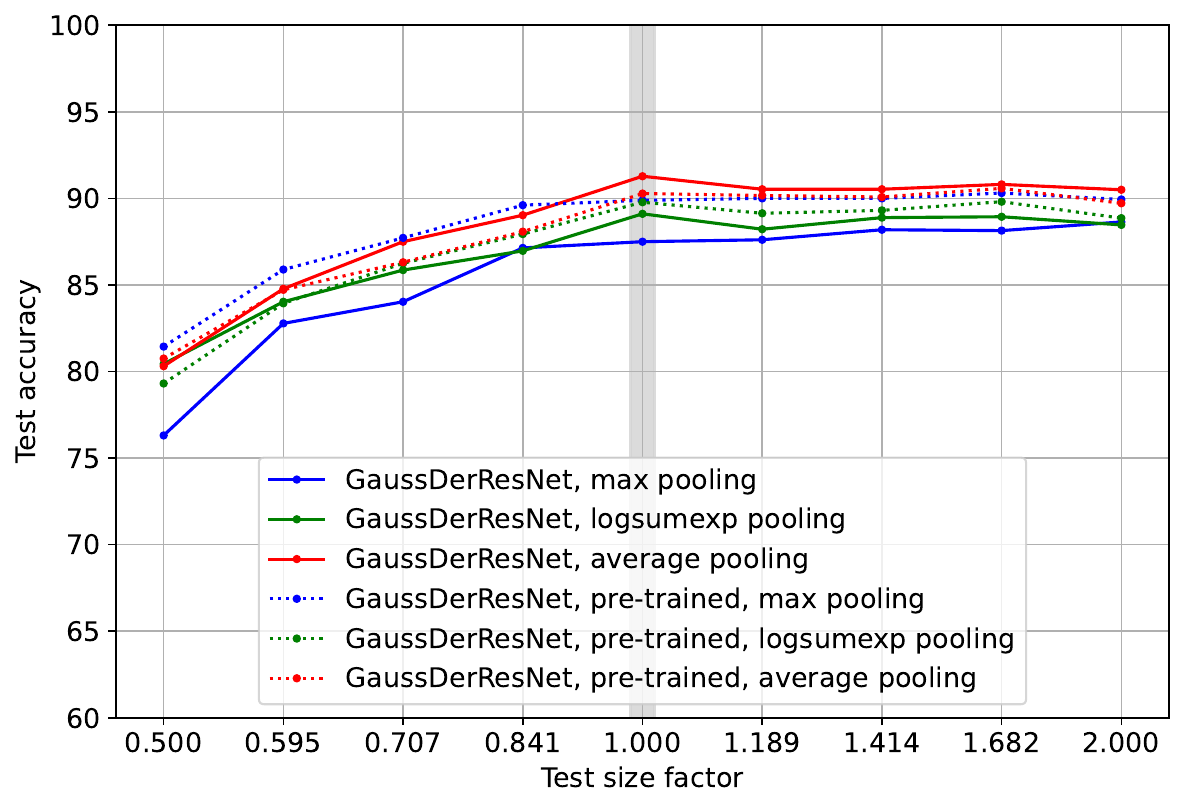}
        \label{fig:subfig-pre-train-3}
    \end{subfigure}
  \end{center}
\caption{Comparison between the scale generalisation curves 
for the multi-scale-channel GaussDerResNets trained using 
either pre-training or genuine multi-scale-channel training, with 
all the networks based on max pooling over scales, on the rescaled 
(a)~Fashion-MNIST, (b)~CIFAR-10 and (c)~STL-10 datasets. 
The networks were trained on size factor 1 training data, and 
then evaluated on rescaled test datasets for each of the size 
factors between 1/2 and 2. As can be seen from the graphs, 
pre-training improves the performance for the smaller size factors for
all the three 
rescaled datasets, while for the larger size factors the effect varies.
The pre-training improves the performance in some cases or reducing 
it slightly in other. Note that the vertical axes for the test accuracy use 
different lower bounds for the different datasets.}
\label{fig-gaussresnet-pre-training}
\end{figure*}

\begin{table*}[hbtp]
  \centering
  \begin{tabular}{lccc}
       \hline
       \textbf{Training Parameters} & \textbf{rescaled Fashion-MNIST} & \textbf{rescaled CIFAR-10} & \textbf{rescaled STL-10}\\ 
       \hline
       Number of first stage epochs & 16 & 40 & 50 \\
       Number of second stage epochs & 16 & 50 & 70 \\
       Batch size during first stage & 128 & 128 & 64 \\ 
       Batch size during second stage & 64 & 45 & 17 \\
       Initial learning rate (first stage) & 0.01 & 0.01 & 0.01 \\  
       Initial learning rate (second stage) & 0.005 & 0.005 & 0.005 \\ \hline
  \end{tabular}
  \caption{Training configurations for pre-training with single-scale-channel 
  of GaussDerResNets employed in our experiments, for the three rescaled datasets.}
  \label{tab:pre-training-parameters}
\end{table*}

\section{Effects of pre-training with a single-scale-channel network on scale generalisation}
\label{sec-pre-training}

The computational cost of training multi-scale-channel networks 
increases with the number of scale channels. Therefore, one could 
ask if it is possible to achieve similar or improved level of scale generalisation, 
while also requiring less training resources.

In Perzanowski and Lindeberg (\citeyear{PerLin25-JMIV-ScSelGaussDerNets}), 
it was demonstrated that transferring the weights of a single-scale-channel network, 
trained on training data with a fixed object size, to a multi-scale-channel network, 
greatly reduces the computational cost needed to train the network.
Barisin et al. (\citeyear{BarSchRed24-JMIV}) also used a related type
of weight transfer. Such weight transfer from a single-scale-channel network to a
multi-scale-channel network may, however, not reach the same level of 
scale generalisation performance as genuine multi-scale-channel training.

In this ablation study, we propose a modified version of this computationally 
efficient approach, where the training is divided into two stages: first pre-training 
with a single-scale-channel network,%
\footnote{In this case, the word "pre-training" 
  refers to the fact that single-scale-channel training is performed before later 
  fine-tuning with multi-scale-channel training, a process that does not involve 
  the use of any additional dataset.}
followed by post-training with a 
multi-scale-channel network initialised with the weights transferred from the 
pre-trained single-scale-channel network. This approach shares structural 
similarities with the scale-space methodology, in the sense that it starts 
at a single scale and then evolves into a multi-scale optimisation problem, 
thereby sharing similarities with Gaussian scale-space generation through 
smoothing or the linear diffusion paradigm, see Lindeberg (\citeyear{Lin93-Dis}).

We explore the viability of this 
approach on the rescaled Fashion-MNIST, CIFAR-10 and STL-10 datasets, 
by comparing the scale generalisation performance of networks trained 
with this method to networks fully trained using genuine multi-scale-channel 
training from Section~6.2 in the main body of the paper.

For these pre-training experiments, we used two training stages instead of one, 
each using different values of the batch size, the initial learning
rate and the number of epochs, as summarised in 
Table~\ref{tab:pre-training-parameters}. In the first stage (the pre-training stage), 
the network was a single-scale-channel network based on $\sigma_{0}=1$, 
while in the second stage (the post-training stage) the training process was then 
restarted, with the learned weights used to initialise the
corresponding multi-scale-channel networks.

\begin{figure*}[ht]
  \begin{center}
    \begin{subfigure}{0.32\textwidth}
        \centering
        \textit{\quad \quad Scale generalisation on rescaled Fashion-MNIST}
        \includegraphics[width=\linewidth]{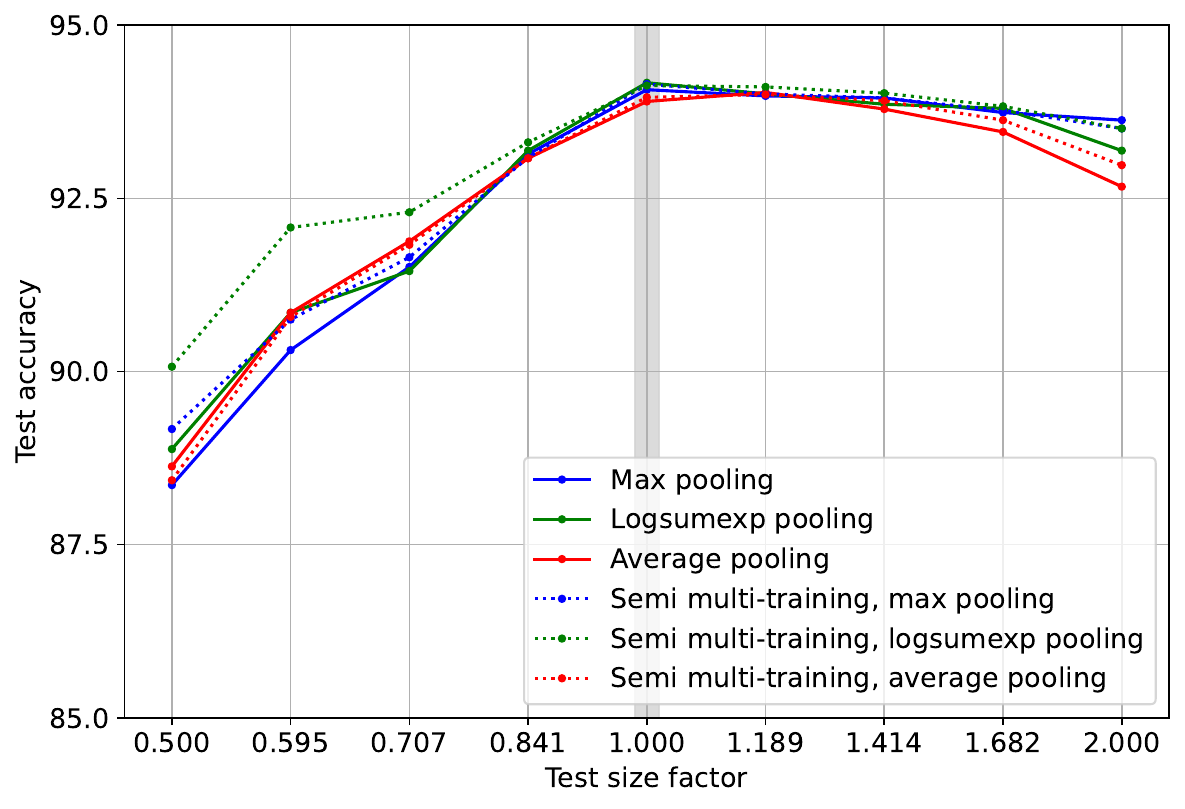}
        \label{fig:subfig-multi-train-1}
    \end{subfigure}
    \hfill
    \begin{subfigure}{0.32\textwidth}
        \centering
        \textit{\quad \quad Scale generalisation on rescaled CIFAR-10}
        \includegraphics[width=\linewidth]{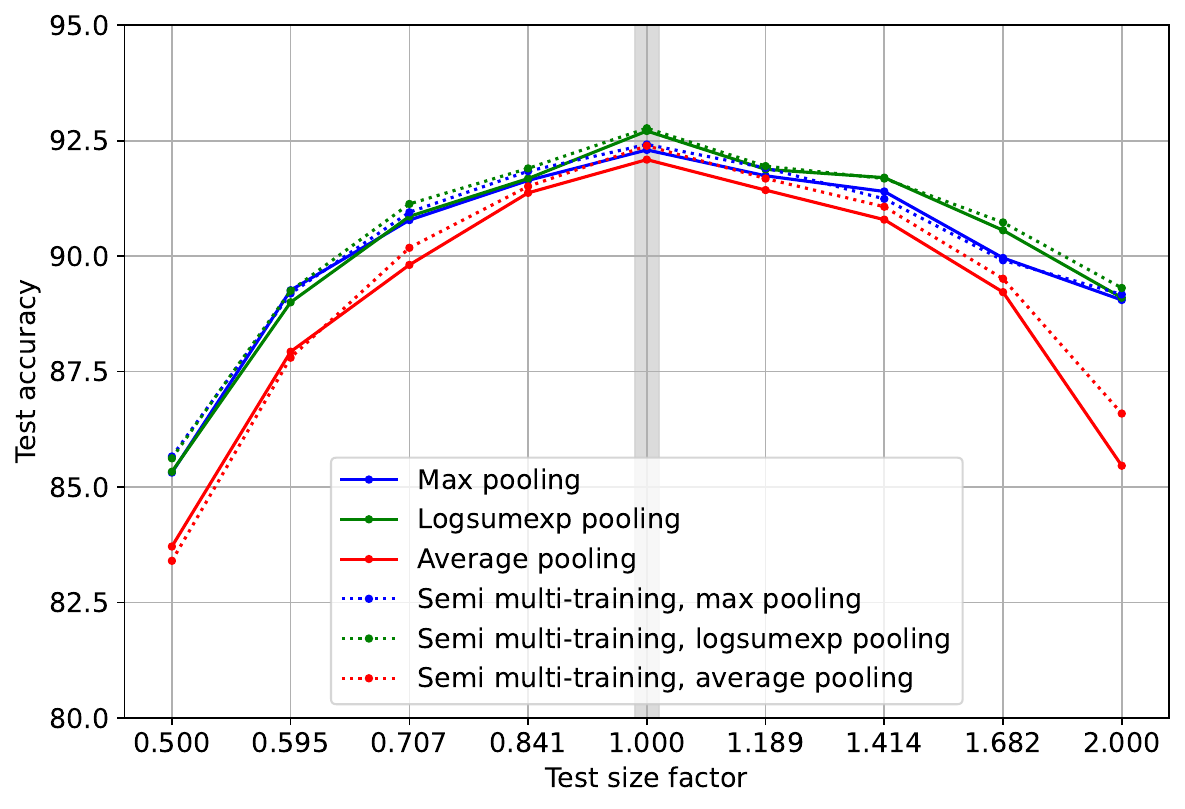}
        \label{fig:subfig-multi-train-2}
    \end{subfigure}
    \hfill
    \begin{subfigure}{0.32\textwidth}
        \centering
        \textit{\quad \quad Scale generalisation on rescaled STL-10}
        \includegraphics[width=\linewidth]{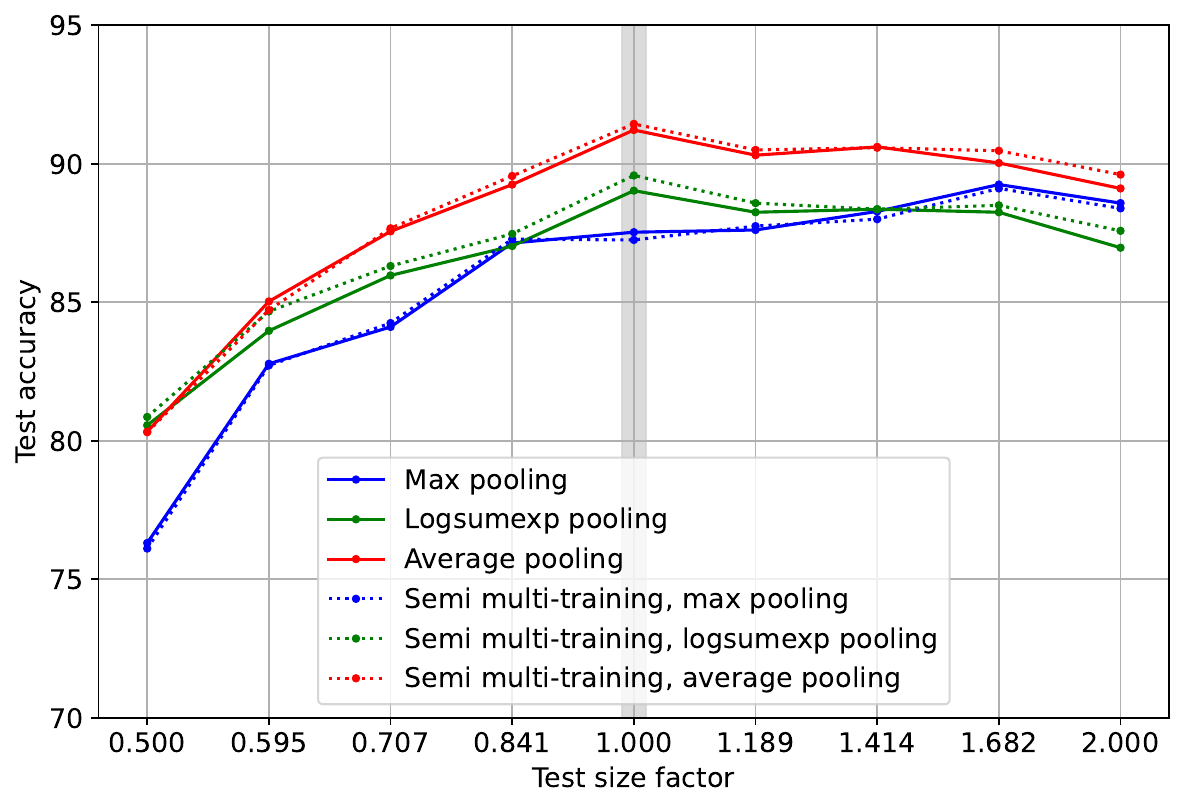}
        \label{fig:subfig-multi-train-3}
    \end{subfigure}
  \end{center}
\caption{Comparison between the zoomed in scale generalisation 
curves for the multi-scale-channel GaussDerResNets and for the 
GaussDerResNets extended after training using weight transfer 
to have additional scale channels that cover a denser scale sampling 
of the scale levels, with the networks based on max, logsumexp or 
average pooling over scales, on the rescaled (left)~Fashion-MNIST, 
(middle)~CIFAR-10 and (right)~STL-10 datasets. The training was 
performed on training data at size factor 1, and inference was 
performed on test data for each of the size factors between 1/2 
and 2. From the graphs, we can see that in general networks with 
more scale-channels tend to have marginally improved scale 
generalisation. Note that the vertical axes for the test accuracy 
use different lower bounds for the different datasets.}
\label{fig-gaussresnet-semi-multi-training-appendix}
\end{figure*}

Crucially, the pre-trained networks in this section used the same
total number of training epochs as the genuine multi-scale-channel 
trained networks in Section~6 in the main body of the paper,
to show the relative efficiency of this approach.%
\footnote{Probably, longer training could improve the results further.}
Furthermore, the first stage of training with a single-scale-channel 
network required far less GPU memory, thereby allowing for a larger 
batch size, which leads to faster training and improved generalisation.

The results of these experiments are shown in Figure~\ref{fig-gaussresnet-pre-training},
where the scale generalisation curves of the pre-trained networks are compared to the 
networks trained in the standard multi-scale-channel manner. 
During the evaluation of the network for the rescaled STL-10 dataset, 
the networks additionally used an extra boundary scale channel, 
based on the initial scale value $\sigma_{0}=2 \sqrt[\leftroot{0}\uproot{1}4]{2}$, because, 
as we will see in Appendix~F, this was found to improve 
the scale generalisation for the larger size factors.

For the rescaled Fashion-MNIST dataset, we can see clear improvements
in both the performance and the scale generalisation for the pre-trained 
GaussDerResNets, with the most notable improvement found for the size 
factors $\leq$ 1, especially for max and logsumexp pooling over scales. 
Similar results are seen for the rescaled CIFAR-10 dataset, 
with the pre-trained networks showing minor 
improvement in the test accuracy for the smaller size factors, and a slight drop 
in the performance for the larger size factors, here for the networks
based on max pooling and logsumexp pooling.

For the rescaled STL-10 dataset, we again find that pre-training results 
in greatly improved performance for the smaller size factors. The 
improvement is most notable for the network based on max pooling, 
achieving very good performance and scale generalisation. 
The network based on logsumexp pooling also benefits somewhat 
from pre-training, while the average pooling network works better when trained in 
a standard way, as it already has stable convergence.

These results thus show that this two-stage training approach not only 
requires much less computational resources during training, but that 
it can also improve the accuracy and the scale generalisation. It provides
a balanced middle ground between the efficient single-scale-channel training, 
where features at a single scale can be learned effectively, and 
multi-scale-channel training, where optimisation is done over several scales 
at once, which is crucial for the network to learn how to automatically adapt 
to and handle interactions between the different scale channels.

Let us finally remark that multi-scale-channel optimisation is non-trivial. 
Here, we have shown that including a single-scale-channel pre-training 
stage provides valuable initial guidance during the training process, 
which is especially beneficial for the smaller size factors, since it avoids suboptimal 
convergence during training, which would negatively affect the performance.

\section{Impact of semi-multi-scale-channel training}
\label{sec-semi-multi-training}

Multi-scale-channel training has been demonstrated to achieve better 
scale generalisation compared to weight transfer from a 
single-scale-channel network trained on fixed object size, but 
it can be computationally costly.

A middle ground between these two methods, which we refer 
to as semi-multi-scale-channel training, is when a 
multi-scale-channel network with a certain scale channel spacing 
is trained, and then during inference the weights are transferred 
to a multi-scale-channel network with a denser scale sampling 
of the scale levels. This approach retains the benefits of 
multi-scale-channel training, while reducing the required 
computational resources for optimising the GaussDerResNets 
with the denser distribution of scale channels. In this section, we 
compare the scale generalisation properties of the GaussDerResNets 
with a denser scale sampling of the scale levels trained using 
semi-multi-scale-channel training, to GaussDerResNets trained using 
standard scale-channel spacing.

We use $\lambda = \sqrt{2}$ as the spacing factor between 
neighbouring scale channels during training, and during 
inference we transfer the weights to a network with scale 
level sampling ratio of $\sqrt[4]{2}$. 
Specifically, we take a trained network based on 
$\sigma_{i,0} \in \{1/(2\sqrt{2}), 1/2, 1/\sqrt{2}, 1, \sqrt{2}, 2\}$, 
and transfer its weights to a network based on 
$\sigma_{i,0} \in \{1/(2\sqrt{2}), 1/(2 \sqrt[\leftroot{0}\uproot{1}4]{2})$, 
$..., 2, 2 \sqrt[\leftroot{0}\uproot{1}4]{2}\}$, which matches the
scale spacing of the test sets of the rescaled datasets we use.

The results are shown in Figure~\ref{fig-gaussresnet-semi-multi-training-appendix},
for the rescaled Fashion-MNIST, CIFAR-10 and STL-10 datasets.

For all the three datasets, we can see that semi-multi-scale-channel training
results in minor improvements to the scale generalisation performance.
In several cases, we can also see that the addition of the extra large 
boundary scale channel during the inference stage
clearly does help improve the generalisation properties for the larger size factors. 
One clear case, where the use of a denser scale channel 
sampling is found to be helpful, is in the leftmost plot for the rescaled 
Fashion-MNIST experiment, where the scale generalisation of the logsumexp 
based network improved significantly, especially for the smaller size factors.

Additionally, we can see that even for a less dense scale-channel 
 spacing, the scale channels cover the in-between range of scales quite
 well, and not only the exact scale they are defined for using the initial 
 scale level. This means that the scale generalisation curve does not 
 fluctuate (for the most part), despite the discrete sampled scale levels.

\begin{figure*}[hbt]
  \begin{center}
    \begin{subfigure}{0.32\textwidth}
        \centering
        \includegraphics[width=\linewidth]{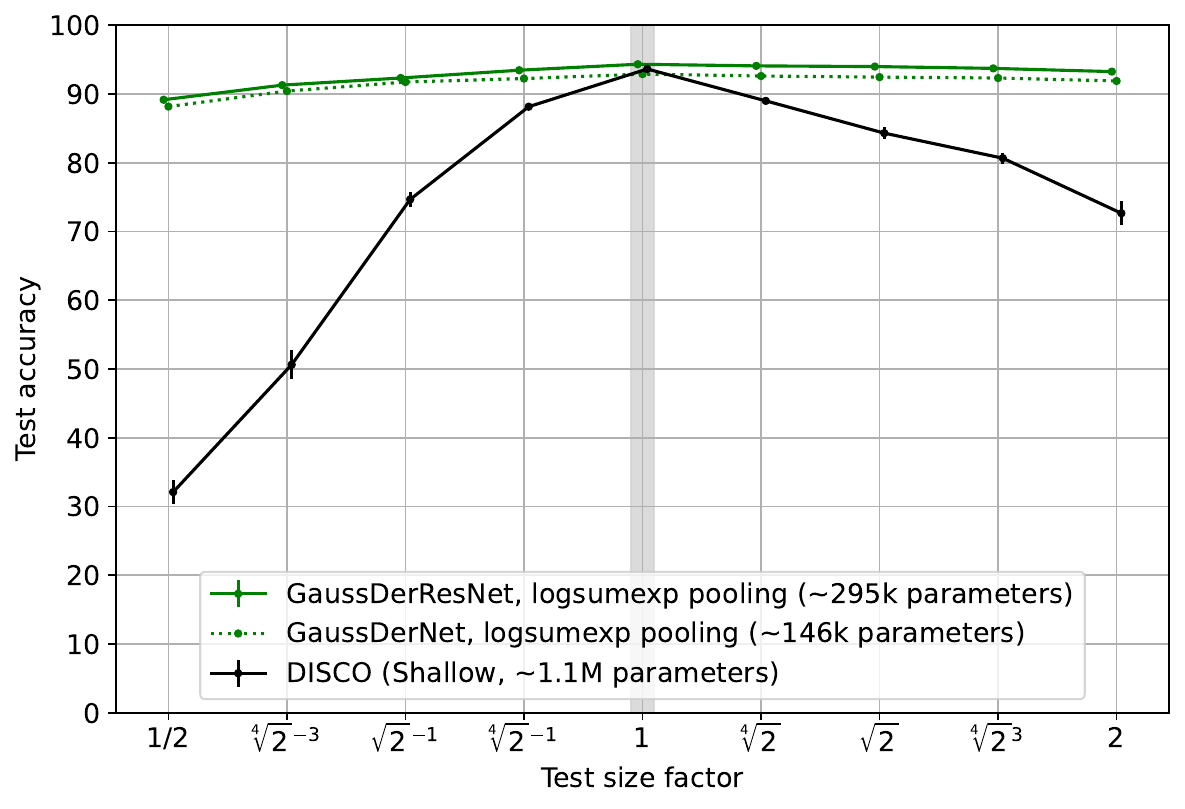}
        \caption{Rescaled Fashion-MNIST dataset}
        \label{fig:subfig-disco-gauss-1}
    \end{subfigure}
    \hfill
    \begin{subfigure}{0.32\textwidth}
        \centering
        \includegraphics[width=\linewidth]{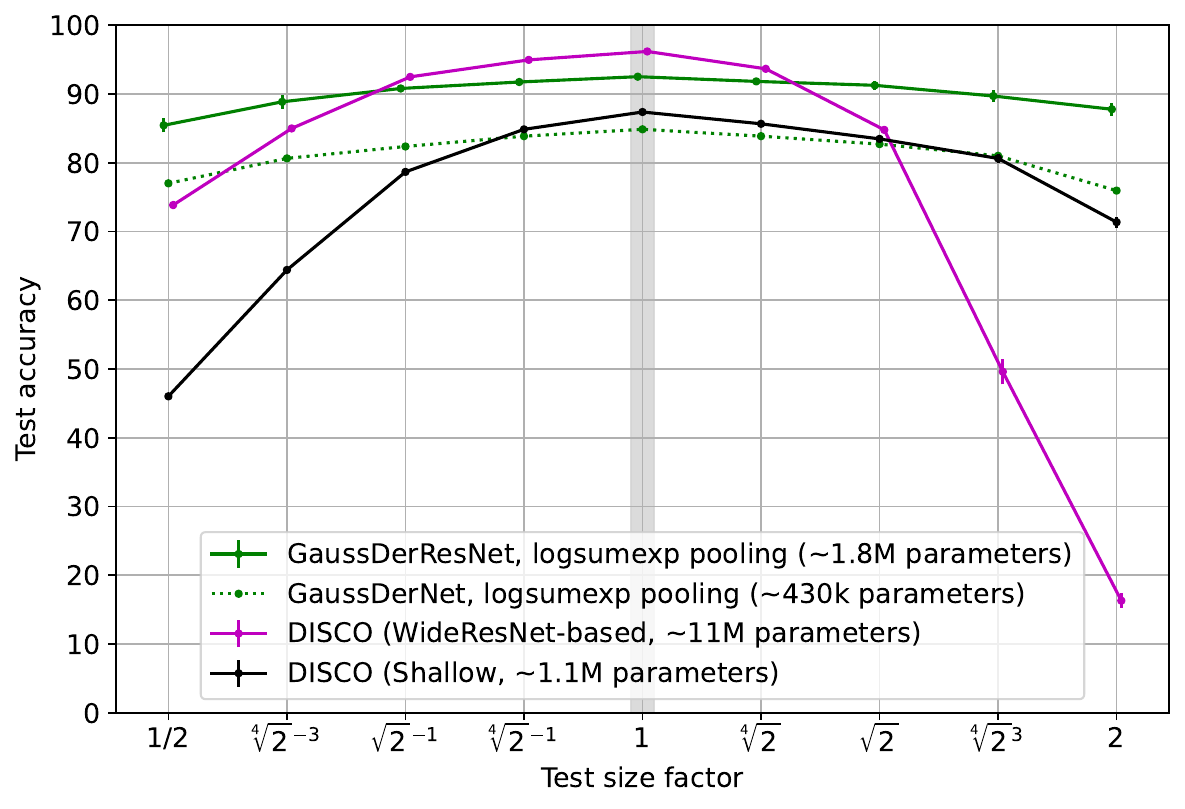}
        \caption{Rescaled CIFAR-10 dataset}
        \label{fig:subfig-disco-gauss-2}
    \end{subfigure}
    \hfill
    \begin{subfigure}{0.32\textwidth}
        \centering
        \includegraphics[width=\linewidth]{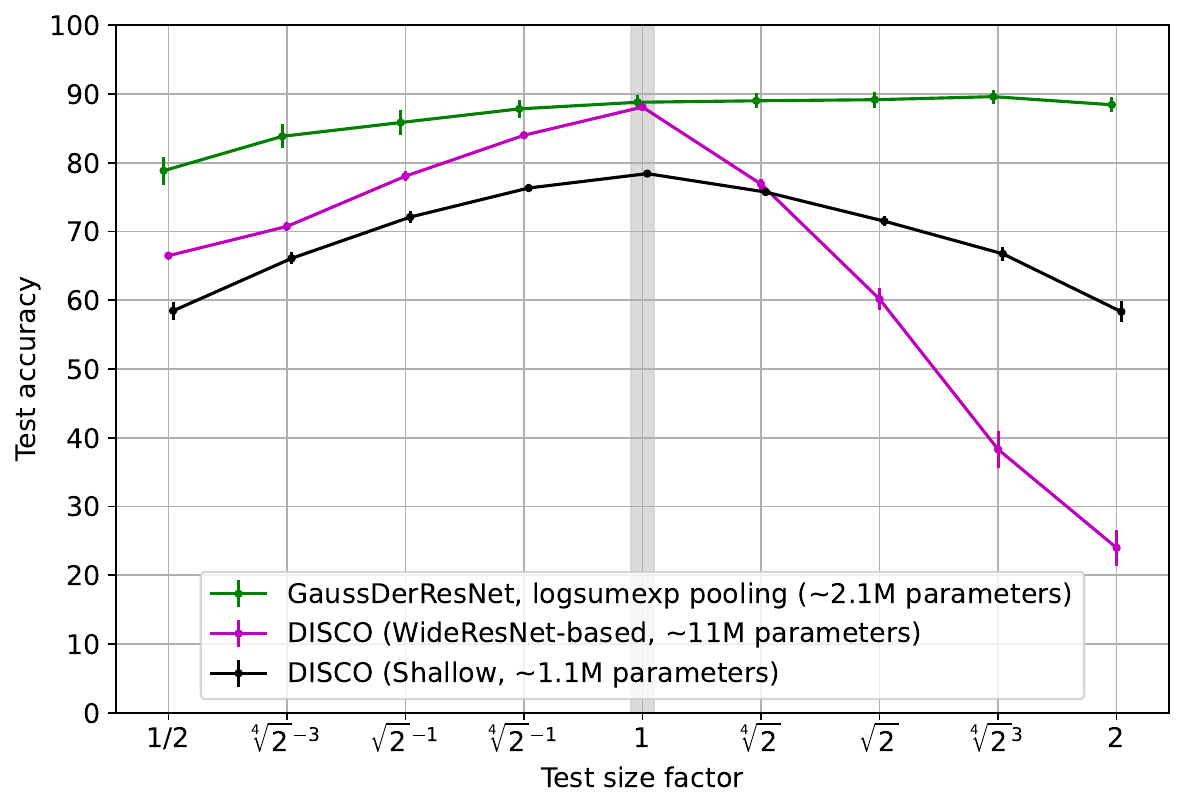}
        \caption{Rescaled STL-10 dataset}
        \label{fig:subfig-disco-gauss-3}
    \end{subfigure}
  \end{center}
  \caption{Scale generalisation curves for the rescaled (a)~Fashion-MNIST, 
  (b)~CIFAR-10, and (c)~STL-10 datasets, comparing both the mean accuracy 
  and the error bars for the standard deviation over 5 runs for the GaussDerResNet 
  based on logsumexp pooling over scales, and the Shallow and WideResNet-based 
  DISCO networks. 
  For the rescaled Fashion-MNIST and CIFAR-10 datasets, we 
  additionally compare to the scale generalisation curves of GaussDerNets 
  based on logsumexp pooling over scales, for only a single run (reproduced 
  from Perzanowski and Lindeberg (\citeyear{PerLin25-JMIV-ScSelGaussDerNets})). 
  The networks were all trained using training data for the size factor 1, 
  and then evaluated on single-scale test sets spanning size factors 
  between 1/2 and 2. From the graphs, we can see that the GaussDerResNets 
  have significantly better scale generalisation properties compared to 
  the DISCO networks, despite using several times fewer parameters. 
  The parameter counts in the legends have been computed in PyTorch.}
  \label{fig-DISCO-gaussresnet-comparison-appendix}
\end{figure*}

 \section{Comparison to the scale generalisation properties
   obtained with the DISCO network by Sosnovik et al.\
  (\citeyear{SosMosSme21-BMVC}) applied to the rescaled
  Fashion-MNIST, CIFAR-10 and STL-10 datasets}
\label{app-compare-disco}

The scale generalisation results for multi-scale-channel 
GaussDerResNets presented in Sections~6.2.1, 6.2.2 and 6.2.4
in the main body of the paper
demonstrate the scale invariance properties of the architecture,
however, it is useful to compare this ability to another scale equivariant 
deep network. In this section, we will compare scale generalisation 
abilities of GaussDerResNets to those of the scale-equivariant 
DISCO network by Sosnovik et al. (\citeyear{SosMosSme21-BMVC})
on the rescaled Fashion-MNIST, CIFAR-10 and STL-10 datasets.

\subsection{Experimental setup}

For all the three datasets, the default DISCO network, which we refer to as 
"Shallow DISCO", is used as a base comparison. This network uses 3 
convolutional layers and 2 fully connected (linear) layers, with mostly 
the same network and training settings as for the experiments on the 
MNIST-scale dataset in the original paper. However, since the input 
images now have a relatively much larger pixel count, we do not upsample 
them. An adaptive average pooling is also added to the pooling stage. 
We also increase the number of intermediate feature channels of the 3 
convolutional layers, from 32-63-95 to 32-95-159, as we have found it to 
be beneficial. For the rescaled CIFAR-10 and STL-10 dataset, random 
cropping and CutOut (DeVries and Taylor \citeyear{DevTay17-arXiv}) are 
used to augment the training process, similar to the WideResNet-based 
DISCO settings.

As the rescaled CIFAR-10 and STL-10 datasets are quite complex 
real-world RBG datasets, we additionally compare to the WideResNet-based 
DISCO implementation with the default parameters and training settings, 
identical to the implementation used in experiments on the regular STL-10 
dataset in the original paper%
\footnote{The experiments on DISCO presented here have been 
performed using the official code available at https://github.com/ISosnovik/disco/tree/master.}. 
For the rescaled CIFAR-10 dataset, the data augmentation parameters 
were adjusted slightly to better match the dataset and its sample dimensions.
The WideResNet-based DISCO has a similar depth as GaussDerResNets, 
as it is based on the WRN-16-8 architecture, however it uses 
approximately 5 times more parameters.

The scales in the scale convolutions are set to the default 
$\{1,2^{1/3},2^{2/3},2\}$ for the Shallow DISCO network and to $\{1,2^{1/2},2\}$
for the WideResNet-based DISCO network. We tried several different 
variations of these scale ranges, including making them denser 
or extending them to span a scale range over a factor of 4, but this 
did not lead to any meaningful improvements.

Finally, for the Fashion-MNIST and CIFAR-10 datasets we additionally 
show results for our shallow Gaussian derivative network (GaussDerNet) 
architecture which uses 6 convolution layers and no fully connected layers, 
as a relevant comparison to the Shallow DISCO network.

\subsection{Experimental results}

Figure~\ref{fig-DISCO-gaussresnet-comparison-appendix}
shows a comparison of the scale generalisation results obtained
for the DISCO networks in relation to the scale generalisation 
results obtained for the GaussDerResNets (and the GaussDerNets) 
with logsumexp pooling over scales. As can be seen from the graphs:
\begin{itemize}
\item
  The GaussDerResNets (and GaussDerNets) have significantly 
  better scale generalisation properties compared to both the 
  Shallow and WideResNet-based DISCO networks, in the sense 
  that the scale generalisation curves are much flatter across all 
  the datasets, despite using significantly fewer network parameters.
\item
  For the rescaled CIFAR-10 and STL-10 datasets, the scale 
  generalisation performance is significantly flatter for the 
  Shallow DISCO compared to WideResNet-based DISCO,
  demonstrating that combining network depth with good scale 
  generalisation, in the way GaussDerResNets are able to, is 
  not necessarily straightforward.
\item
  For the rescaled Fashion-MNIST dataset, both the 
  GaussDerResNet and GaussDerNet architectures achieve 
  similar peak test accuracy on the size factor 1 as the Shallow 
  DISCO network, with GaussDerResNet performance being the 
  highest by a tiny margin.
\item
  For the rescaled CIFAR-10 dataset, the WideResNet-based 
  DISCO network with about 5 times more parameters%
  \footnote{The GaussDerResNets can get closer to matching the peak 
   performance of the DISCO network for the CIFAR-10 dataset
   when using a similar number of parameters and fewer data augmentations 
   during training, as described in Footnote~18 in the main body 
   of the paper.} 
   achieves nearly 4~ppt higher peak performance at size factor 1 
  compared to GaussDerResNet. When the size factor differs by 
  about a factor of $\sqrt{2}$ from the training size 1, the accuracy 
  of the DISCO networks does, however, drop below the accuracy 
  of the GaussDerResNet network, most rapidly for larger size factors.
\item
  For the rescaled STL-10 dataset, the GaussDerResNet achieves 
  marginally higher peak test accuracy on the size factor 1 
  compared to the WideResNet-based DISCO networks, 
  despite using significantly fewer network parameters. The 
  drop in performance for the Shallow DISCO network is very 
  significant. These performance results are partly due to the 
  test images having their image size increased by a factor of 
  2 compared to the training images%
  \footnote{When the WideResNet-based DISCO network is evaluated on central cropped 
    test rescaled STL-10 data, meaning cropped from the size $192 \times 192$ to the 
    same size as the training data ($96 \times 96$), the performance improves by 
    about 6~ppt for the size factor 1. The scale generalisation properties, however, 
    still remain poor for the larger size factors, and are then also
    fundamentally limited by the fact that the rescaled images are
    no longer fully visible for the larger size factors. (For the
    GaussDerResNets, the performance 
    also improves for the size factor 1 when evaluated on cropped test data, but only 
    by about 3~ppt.)}; a change 
    that GaussDerResNets can handle well by default.  
\end{itemize}
Beginning with the results for the rescaled Fashion-MNIST dataset, 
the poor scale generalisation performance by DISCO on this dataset 
shows that the network does not by default generalise well to scales 
not seen during training, especially not for smaller size factors. Compared 
to the results for the DISCO network trained on the smaller STIR-MNIST 
dataset, shown in Figure~9 in Altstidl et al. (\citeyear{AltNguSchKoeMutEskZan23-IJCNN}), 
this level of scale generalisation is roughly similar at large size factors, 
while dropping substantially at the smallest size factors. 
The Shallow DISCO network may hence require additional parameter
tuning to improve this ability, while in contrast, no major tuning was 
needed for the GaussDerResNet and GaussDerNet architectures in 
order to achieve great scale generalisation on the rescaled Fashion-MNIST 
dataset.

For the rescaled CIFAR-10 and STL-10 datasets, the scale generalisation 
of the Shallow DISCO network improved, achieving flatter curves compared 
to those obtained for the rescaled Fashion-MNIST dataset,
while once again the GaussDerResNets and the GaussDerNets 
clearly have better generalisation properties.

During our testing, we have observed that incorporating an image 
upsampling stage or reducing the number of spatial subsampling layers 
can lead to a clear improvement in performance on the smaller size factors
for the Shallow DISCO network, in some cases on the order of 10~ppt.
That, however, typically reduces the performance for the larger size 
factors by a similar amount, meaning that balanced architecture tuning 
is required for good overall scale generalisation.

\begin{figure*}[hbt]
  \begin{center}
    \begin{subfigure}{0.99\textwidth}      
      \begin{tabular}{ccc}
	\hspace{3mm} \textit{Training with small object sizes}
          & \hspace{3mm} \textit{Training with medium object sizes}
          & \hspace{3mm} \textit{Training with large object sizes} \\
          \includegraphics[width=0.32\textwidth]{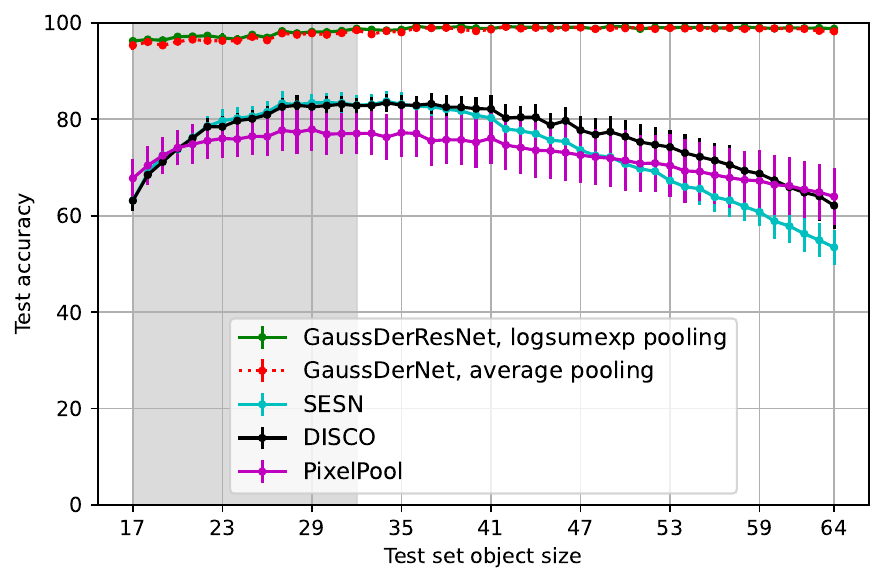}
          & \includegraphics[width=0.32\textwidth]{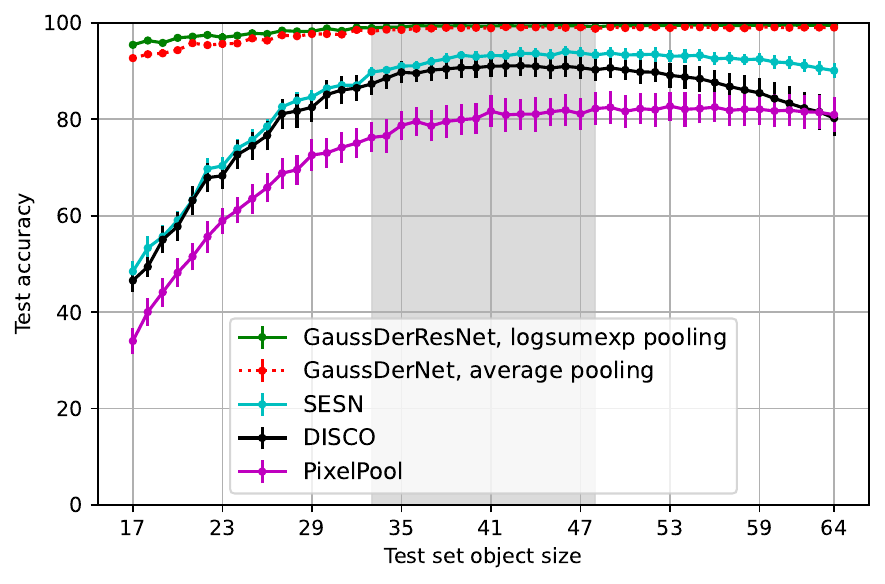}
          &  \includegraphics[width=0.32\textwidth]{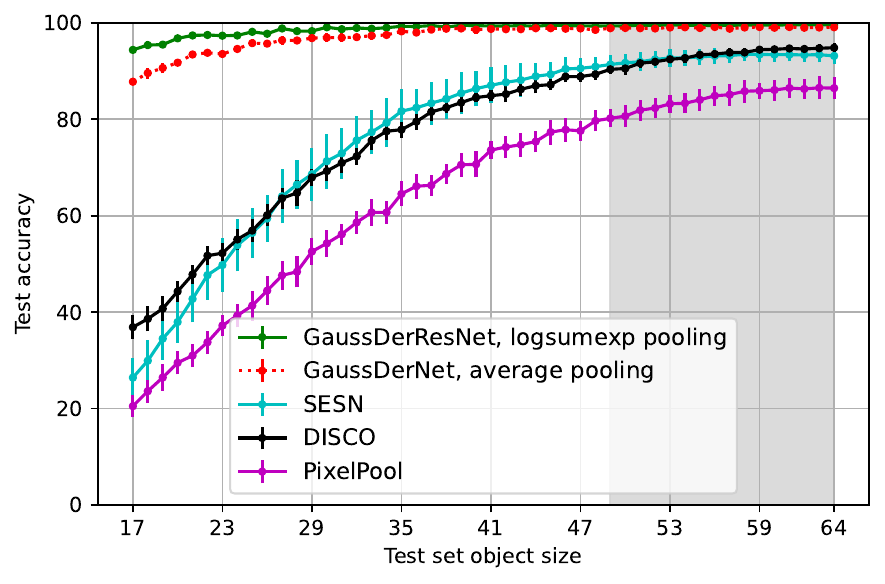}
      \end{tabular}
      \caption{Comparison of scale generalisation curves for the
        traffic sign subset of the STIR dataset}
      \label{fig:subfig-STIR-trafficsign-appendix}
    \end{subfigure}

    \bigskip
    
    \begin{subfigure}{0.99\textwidth}      
      \begin{tabular}{ccc}
          \hspace{3mm} \textit{Training with small object sizes}
          & \hspace{3mm} \textit{Training with medium object sizes}
          & \hspace{3mm} \textit{Training with large object sizes} \\
          \includegraphics[width=0.32\textwidth]{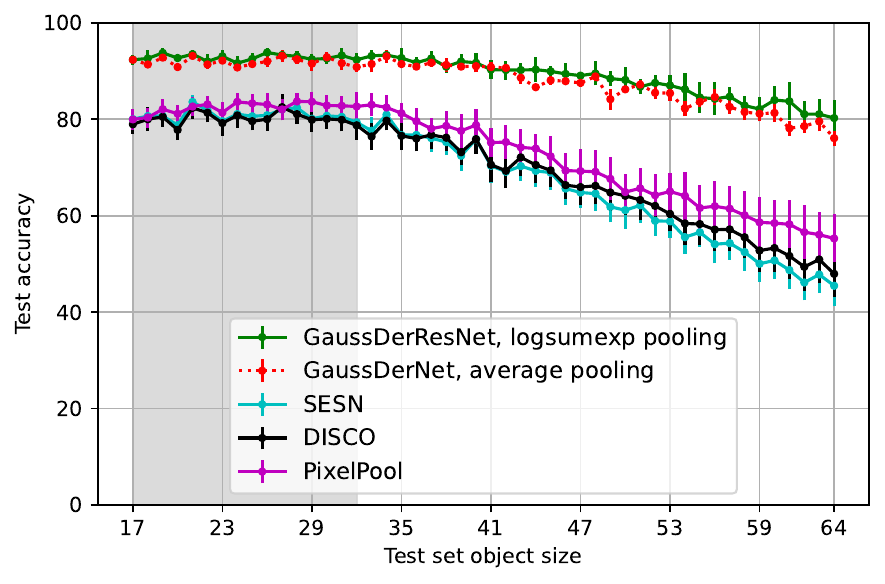}
          & \includegraphics[width=0.32\textwidth]{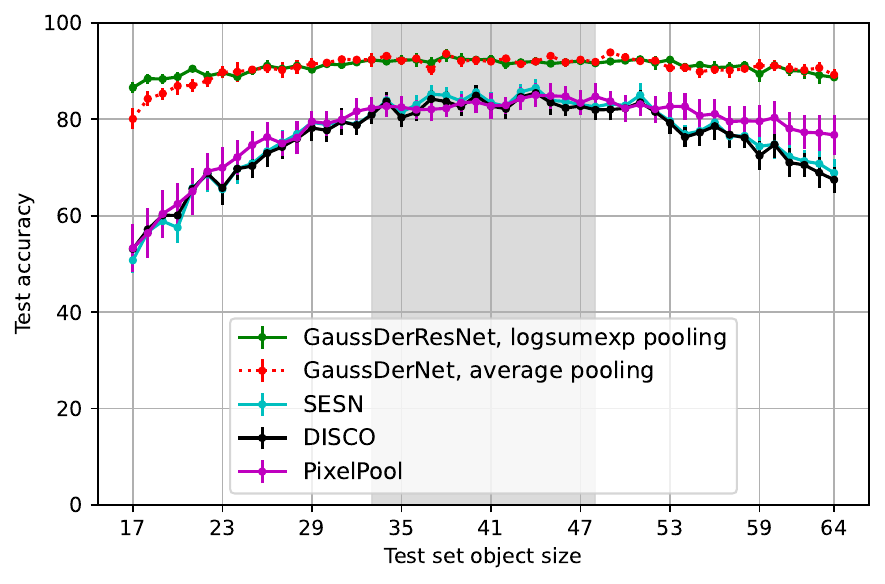}
          & \includegraphics[width=0.32\textwidth]{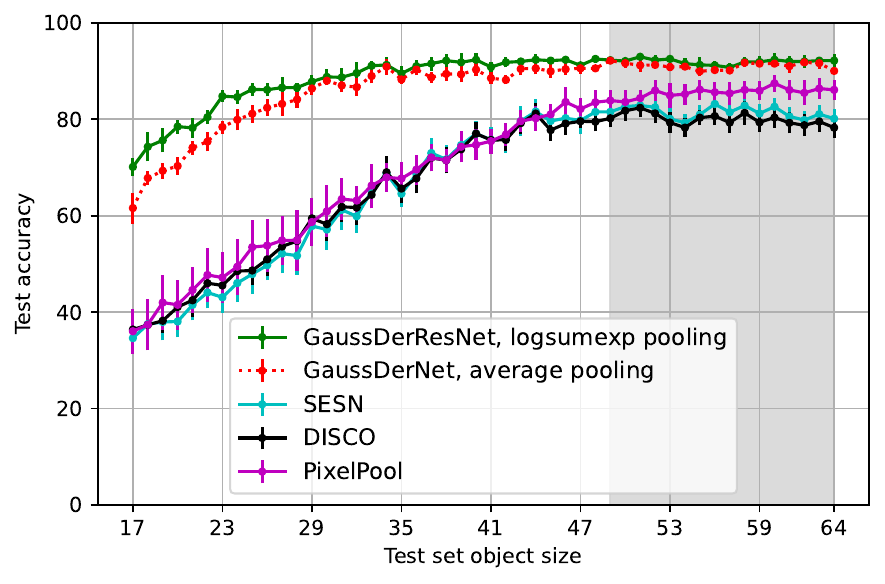}
      \end{tabular}
      \caption{Comparison of scale generalisation curves for the
        aerial image subset of the STIR dataset}
      \label{fig:subfig-STIR-aerial-appendix}
    \end{subfigure} 
  \end{center}
  \caption{Scale generalisation curves for the (a)~traffic sign and 
  (b)~aerial STIR dataset subsets, comparing both the mean accuracy 
  and the error bars for the standard deviation over 5 runs for the 
  multi-scale-channel GaussDerResNets based on logsumexp pooling over 
  scale and multi-scale-channel GaussDerNets based on average pooling 
  (reproduced from Perzanowski and Lindeberg (\citeyear{PerLin25-JMIV-ScSelGaussDerNets})). 
  These results are compared agains the scale generalisation curves of 
  (i)~the SESN, (ii)~the DISCO, and (iii)~the PixelPool networks, which are 
  some of the best performing models on the STIR benchmark.
  Three different training regimes are shown for each dataset, indicated 
  by the shaded grey region, with the networks trained on data containing 
  (left)~small scale objects (size [17,32]), (middle)~medium scale objects 
  (size [33,48]), and (right)~large scale objects (size [49,64]). All
  the networks 
  were evaluated on the test set containing all the possible object sizes, 
  $l \in [17,64]$. From the graphs, we see that GaussDerResNets have 
  the best scale generalisation properties among all these networks. 
  The curves for the SESN, DISCO and PixelPool models use results 
  averaged over 25 runs, based directly on results from experiments 
  performed by Altstidl et
  al. (\citeyear{AltNguSchKoeMutEskZan23-IJCNN}).}
  \label{fig-STIR-scale-selection-appendix}
\end{figure*}

As for the WideResNet-based DISCO network, its results on the rescaled CIFAR-10 
and STL-10 datasets demonstrate an interesting behaviour, in that 
the network obtains a decent performance on the smaller size factors, 
which can be partly explained by the fact that these datasets do 
contain a small span of scale variations in the training data, implying 
that the networks get to see some small objects during training,
thus serving as a mild form of data augmentation regarding
scale variations, which helps the DISCO network to make
better predictions at those scales due to its scale-equivariant design.

For the larger size factors, there is, however, a substantial drop
in performance for the rescaled CIFAR-10 and STL-10 datasets,
which is indeed dramatic for the CIFAR-10 dataset for size factors
greater than $\sqrt{2}$. This behaviour differs for the Shallow DISCO 
network and one may thus ask what could be the explanation for 
this dramatic drop in scale generalisation performance.

Among possible explanations for this, one may ask if the very tight
cropping of the original CIFAR-10 images in combination of
mirror-extension in the training data for the size factor 1 could
constitute contributing factors regarding the rescaled CIFAR-10 
dataset, where those mirror extensions gradually disappear with 
increasing size factors. In our experiments, we have observed that 
the presence and extent of these mirror extensions do have an effect 
on the performance of the DISCO network on large size factors. Alternatively, one may 
ask if the performance of the DISCO network is tied to a particular 
image size, which our GaussDerResNets are not dependent on.
Finally, one may ask if a different choice of kernel size, scale 
set or some other network settings may need to be adjusted to improve 
the performance for the larger size factors, and find the scale 
generalisation performance sweet spot between the default Shallow 
and ResNet-based DISCO networks. Such additional fine tuning is left
for future work.

These problems thereby indicate that the default WideResNet-based 
DISCO architecture may benefit from architectural changes 
and/or improvements in the training schedule to get better scale 
generalisation properties, as achieving simultaneous good scale
generalisation and high peak accuracy is not necessarily straightforward.

\begin{figure*}[htbp]
  \begin{center}
    \begin{tabular}{ccccccccccc}
      \raisebox{3\height}{Layer 17} &
      \hspace{-4mm}
      \includegraphics[width=0.09\textwidth]{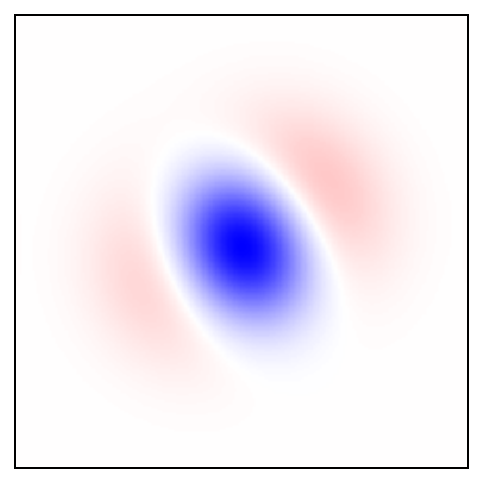} &
      \hspace{-5mm}
      \includegraphics[width=0.09\textwidth]{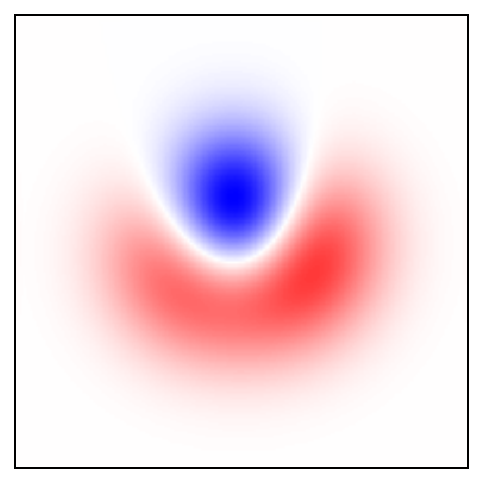} &
      \hspace{-5mm}
      \includegraphics[width=0.09\textwidth]{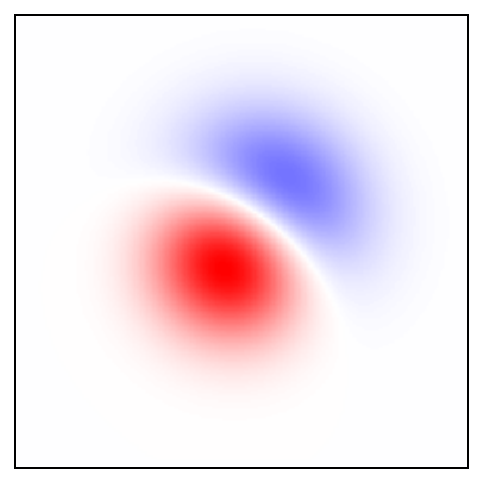} &
      \hspace{-5mm}
      \includegraphics[width=0.09\textwidth]{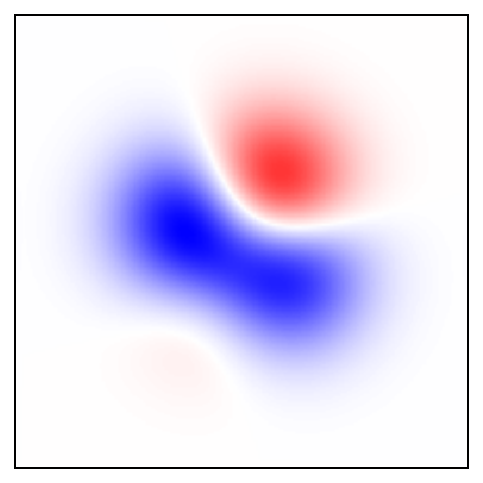} &
      \hspace{-5mm}
      \includegraphics[width=0.09\textwidth]{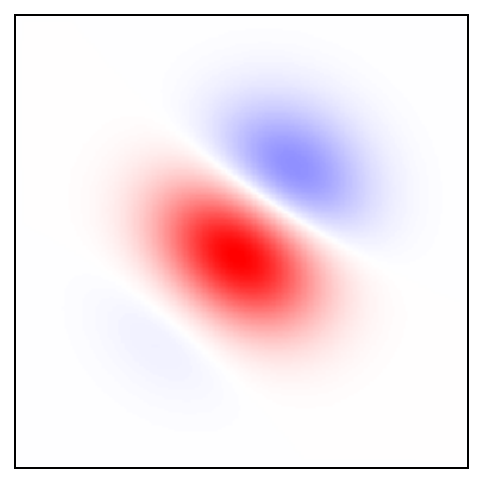} &
      \hspace{-5mm}
      \includegraphics[width=0.09\textwidth]{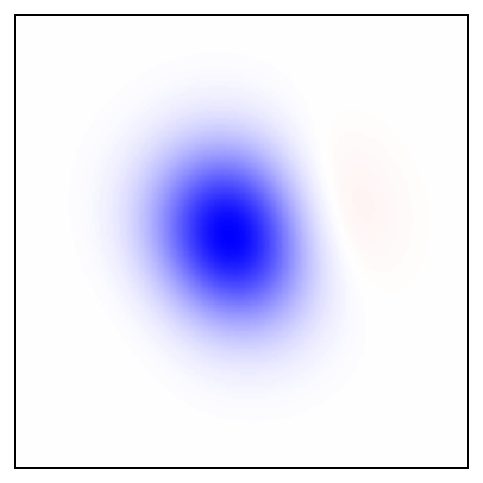} &
      \hspace{-5mm}
      \includegraphics[width=0.09\textwidth]{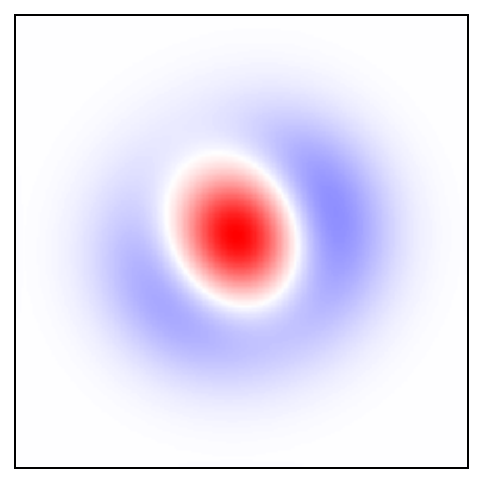} &
      \hspace{-5mm}
      \includegraphics[width=0.09\textwidth]{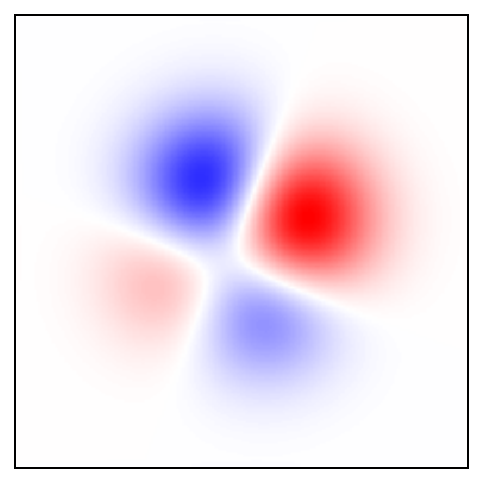} &
      \hspace{-5mm}
      \includegraphics[width=0.09\textwidth]{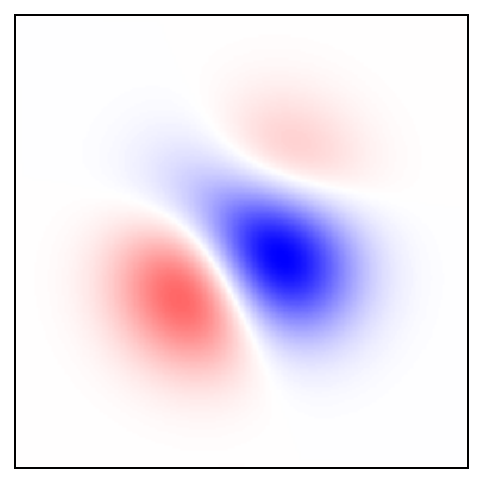} &
      \hspace{-5mm}
      \includegraphics[width=0.09\textwidth]{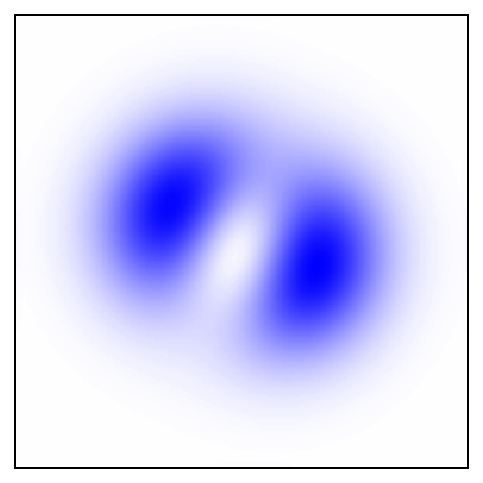}
      \\
    \end{tabular} 

  \vspace{-1.5mm}

    \begin{tabular}{ccccccccccc}
      \hline
      \raisebox{3\height}{Layer 11} &
      \hspace{-4mm}
      \includegraphics[width=0.09\textwidth]{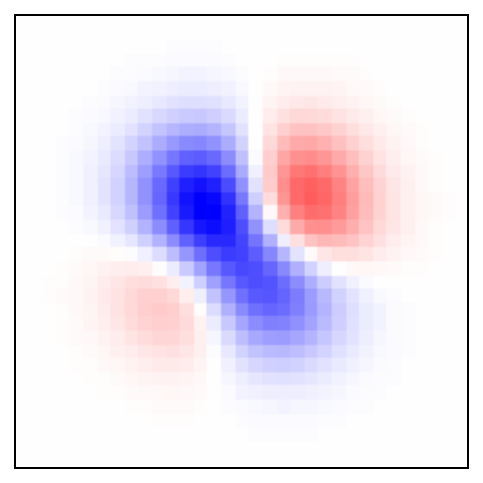} &
      \hspace{-5mm}
      \includegraphics[width=0.09\textwidth]{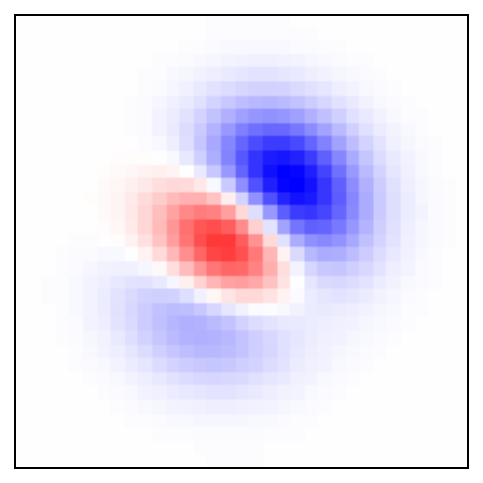} &
      \hspace{-5mm}
      \includegraphics[width=0.09\textwidth]{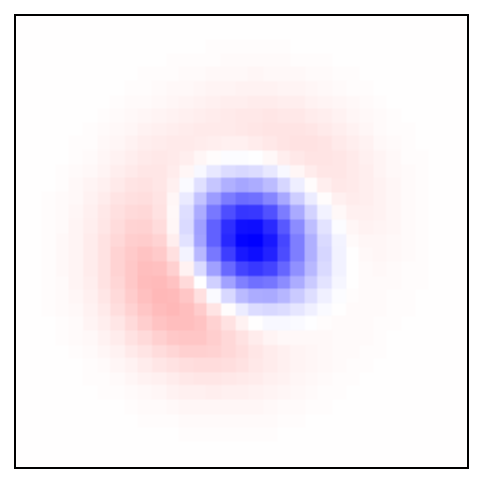} &
      \hspace{-5mm}
      \includegraphics[width=0.09\textwidth]{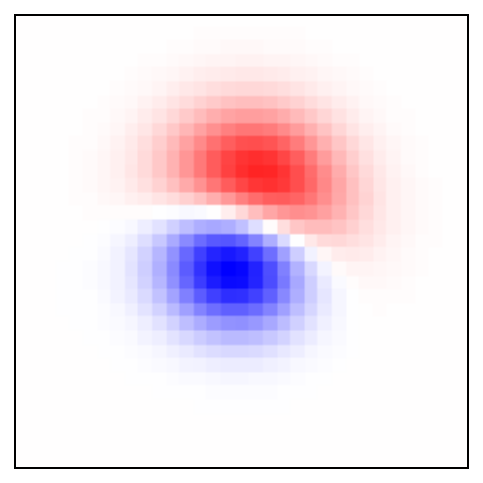} &
      \hspace{-5mm}
      \includegraphics[width=0.09\textwidth]{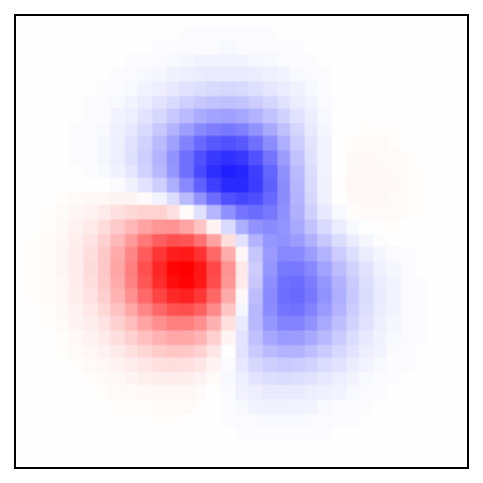} &
      \hspace{-5mm}
      \includegraphics[width=0.09\textwidth]{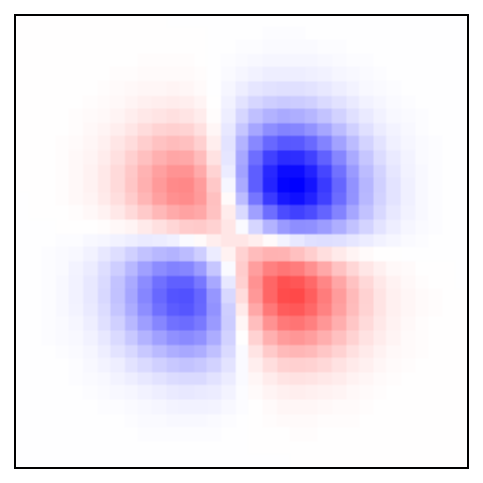} &
      \hspace{-5mm}
      \includegraphics[width=0.09\textwidth]{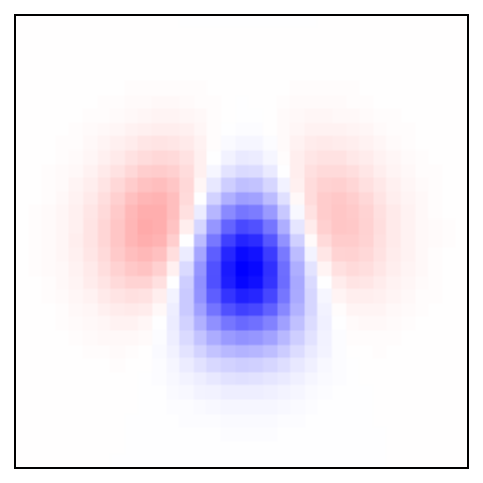} &
      \hspace{-5mm}
      \includegraphics[width=0.09\textwidth]{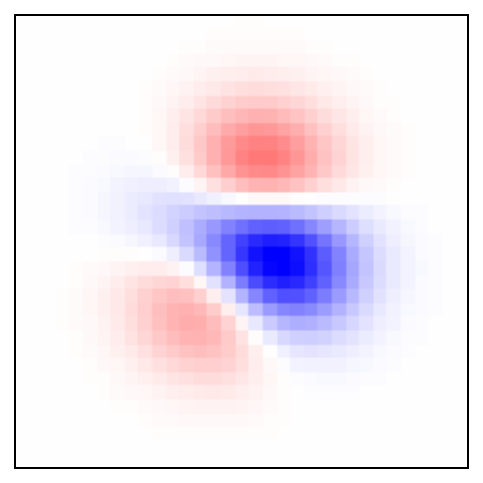} &
      \hspace{-5mm}
      \includegraphics[width=0.09\textwidth]{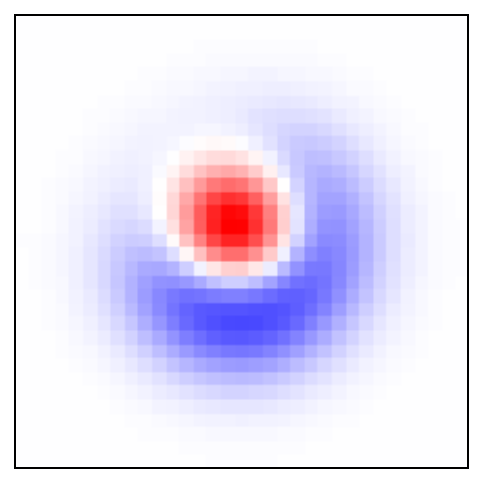} &
      \hspace{-5mm}
      \includegraphics[width=0.09\textwidth]{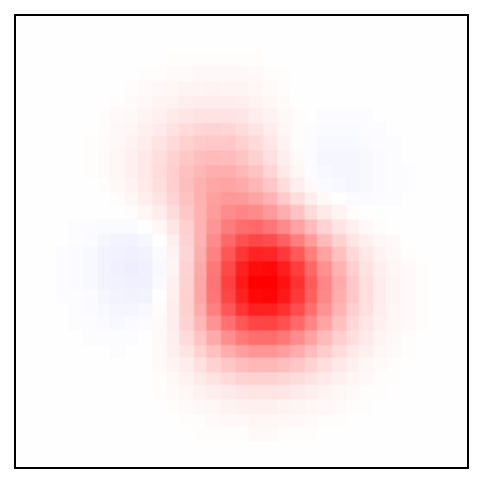}
      \\
    \end{tabular} 

  \vspace{-1.5mm}
  
    \begin{tabular}{ccccccccccc}
      \hline
      \raisebox{3\height}{Layer 10} &
      \hspace{-4mm}
      \includegraphics[width=0.09\textwidth]{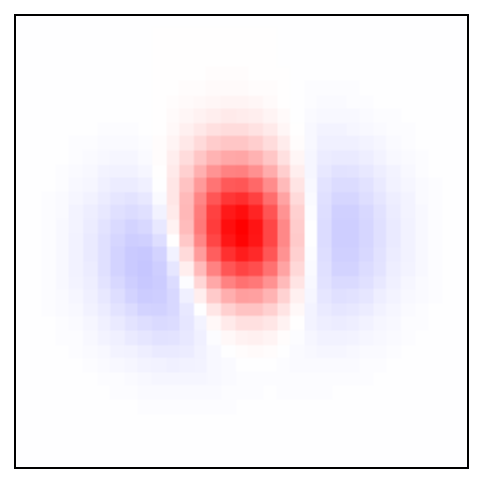} &
      \hspace{-5mm}
      \includegraphics[width=0.09\textwidth]{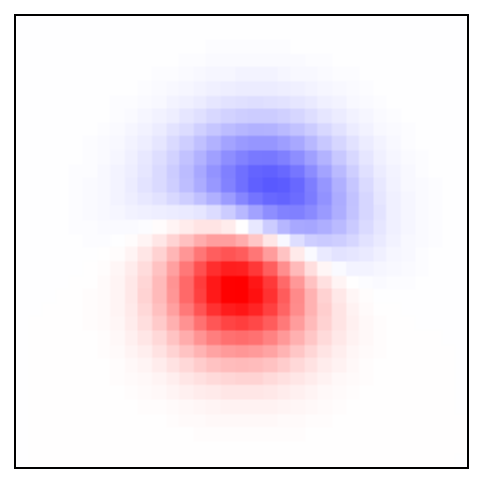} &
      \hspace{-5mm}
      \includegraphics[width=0.09\textwidth]{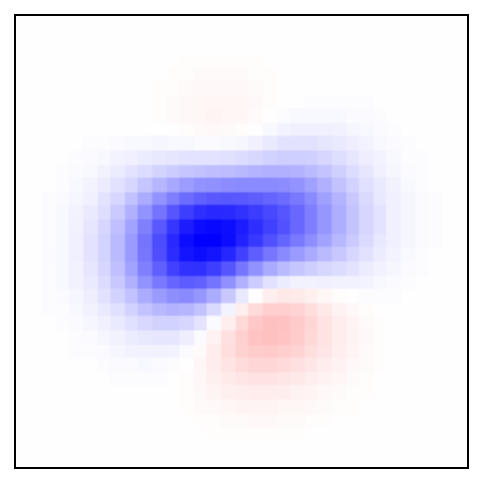} &
      \hspace{-5mm}
      \includegraphics[width=0.09\textwidth]{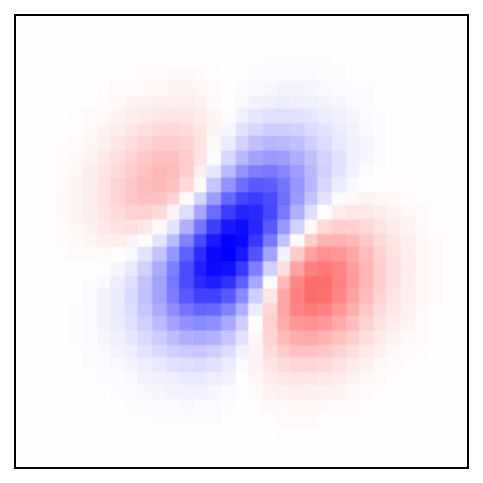} &
      \hspace{-5mm}
      \includegraphics[width=0.09\textwidth]{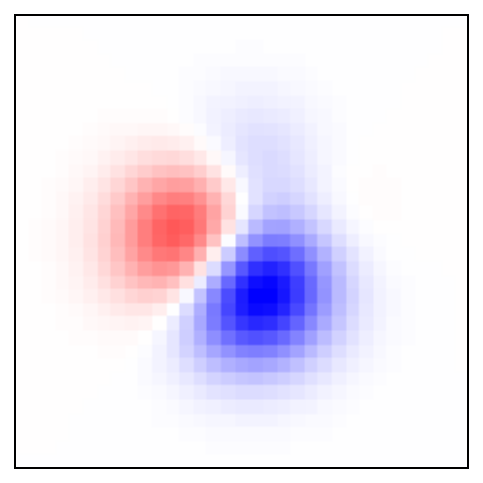} &
      \hspace{-5mm}
      \includegraphics[width=0.09\textwidth]{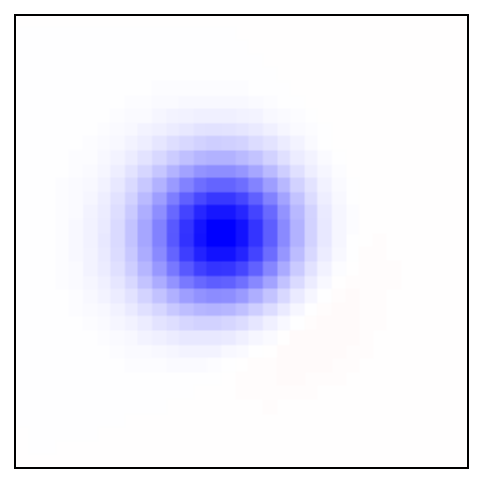} &
      \hspace{-5mm}
      \includegraphics[width=0.09\textwidth]{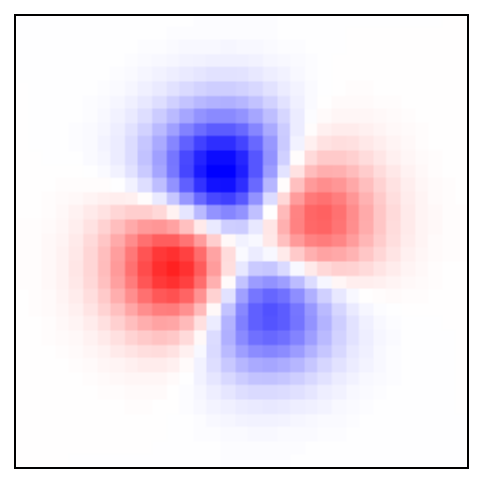} &
      \hspace{-5mm}
      \includegraphics[width=0.09\textwidth]{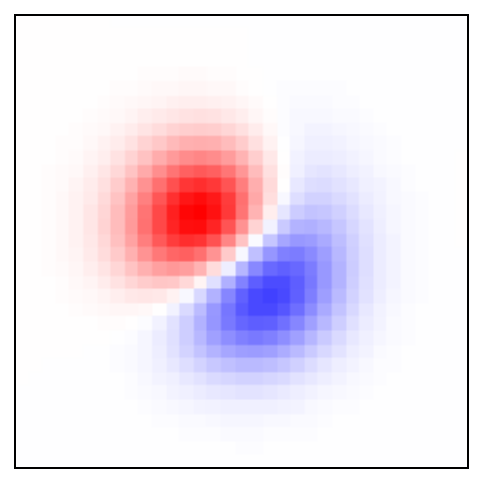} &
      \hspace{-5mm}
      \includegraphics[width=0.09\textwidth]{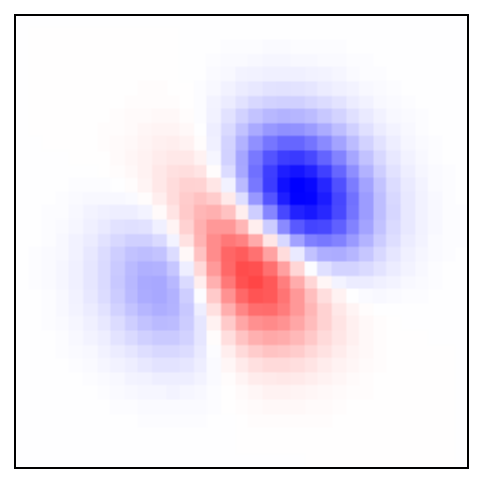} &
      \hspace{-5mm}
      \includegraphics[width=0.09\textwidth]{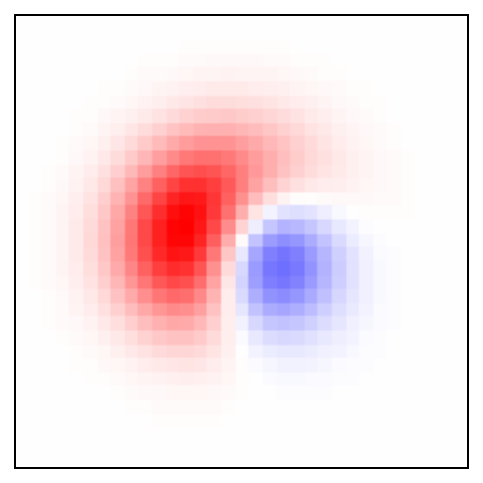}
      \\
    \end{tabular} 

  \vspace{-1.5mm}
  
    \begin{tabular}{ccccccccccc}
      \hline
      \raisebox{3\height}{Layer 2} &
      \hspace{-2.4mm}
      \includegraphics[width=0.09\textwidth]{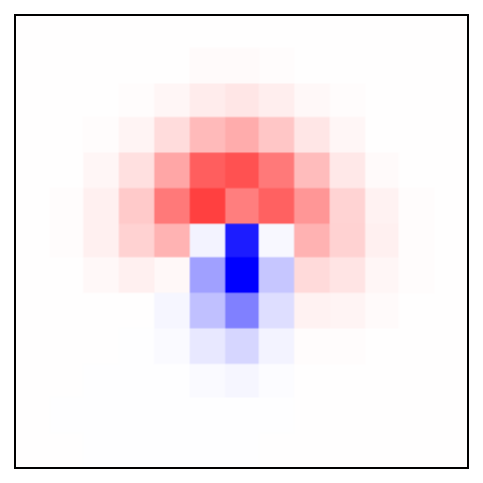} &
      \hspace{-5mm}
      \includegraphics[width=0.09\textwidth]{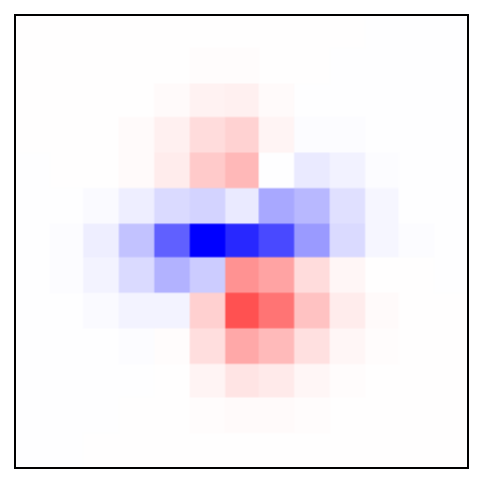} &
      \hspace{-5mm}
      \includegraphics[width=0.09\textwidth]{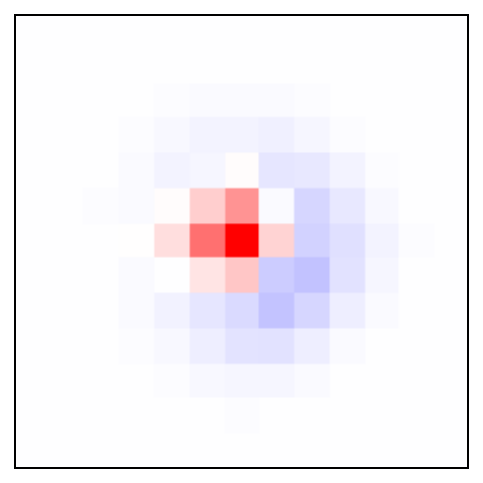} &
      \hspace{-5mm}
      \includegraphics[width=0.09\textwidth]{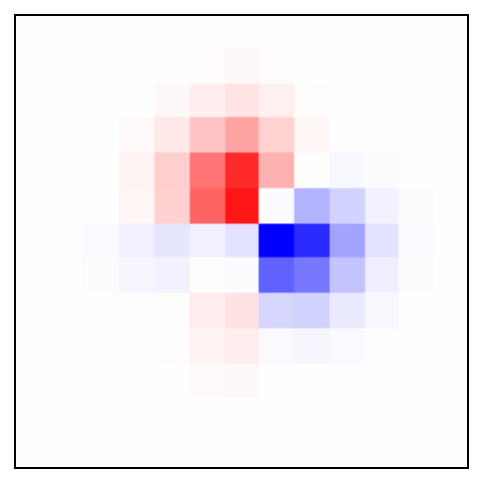} &
      \hspace{-5mm}
      \includegraphics[width=0.09\textwidth]{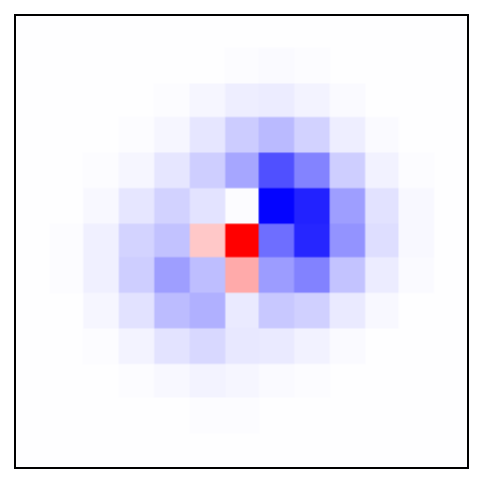} &
      \hspace{-5mm}
      \includegraphics[width=0.09\textwidth]{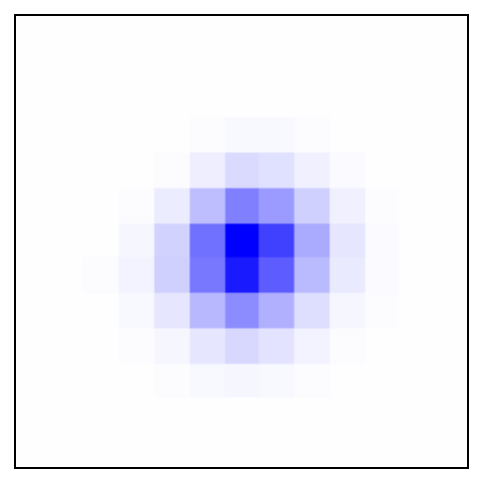} &
      \hspace{-5mm}
      \includegraphics[width=0.09\textwidth]{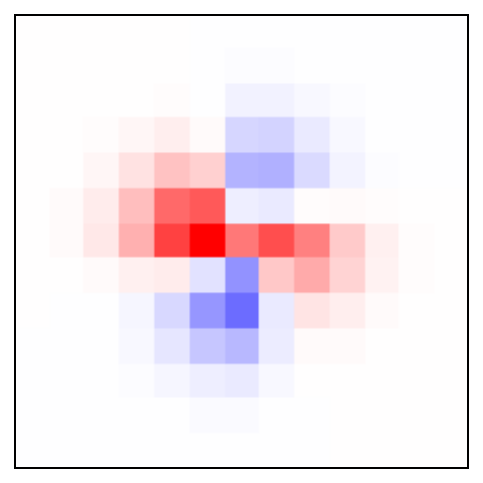} &
      \hspace{-5mm}
      \includegraphics[width=0.09\textwidth]{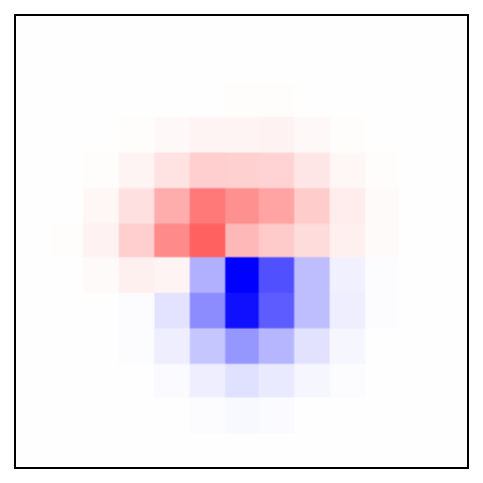} &
      \hspace{-5mm}
      \includegraphics[width=0.09\textwidth]{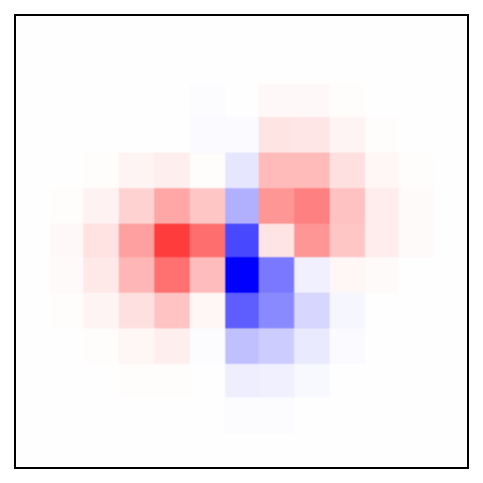} &
      \hspace{-5mm}
      \includegraphics[width=0.09\textwidth]{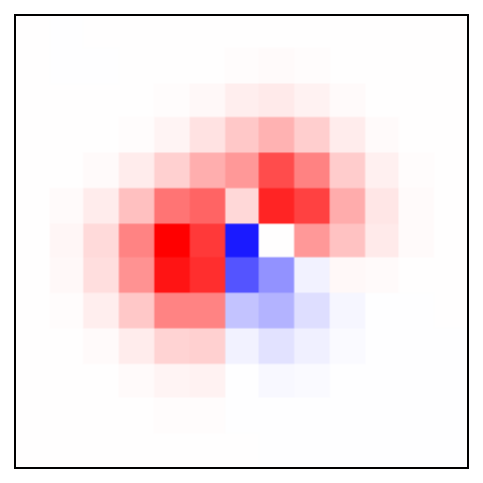}
    \end{tabular} 
  \end{center}
  \captionsetup{skip=-5pt}
  \caption{A representative showcase of 10 learned effective filters
    from an early layer, two intermediate layers from a single residual block 
    and the penultimate layer (specifically the 2nd, 10th, 11th and 17th layers) of 
  a single-scale-channel GaussDerResNet with a zero-order term included in the
  filter basis, and trained on the rescaled STL-10 dataset. The colour mapping 
  is configured such that zero values appear white, positive values are shown in red,
  and negative values are represented in blue. The size of these convolution kernels 
  becomes larger in the higher layers, as the consequence of the scale parameter 
  $\sigma$ increasing with depth according to Equation~(37).
  We can see that the learned filters are not always DC balanced, 
  and that some of them are similar to different directional derivatives or the 
  Laplacian-of-Gaussian.}
  \label{fig-learned-filters-standard-appendix}
\end{figure*}

\section{Comparison to previously reported scale generalisation
  results on the STIR datasets with traffic signs and
  aerial images by Altstidl et al.\
  (\citeyear{AltNguSchKoeMutEskZan23-IJCNN})}
\label{app-compare-stir}

To further evaluate the scale generalisation properties of the 
multi-scale-channel GaussDerResNets and GaussDerNets  
on the traffic sign and aerial image subsets of the STIR dataset in Section~6.2.3 
of the main body of the paper, we here compare our results with those obtained 
by the three best performing networks on this benchmark reported 
in Altstidl et al. (\citeyear{AltNguSchKoeMutEskZan23-IJCNN}). 
This comprises (i)~the SESN (scale-equivariant steerable) 
network (Sosnovik et al. \citeyear{SosSzmSme20-ICLR}), 
(ii)~the DISCO (discrete scale convolutions) network (Sosnovik 
et al. \citeyear{SosMosSme21-BMVC}), and (iii)~the PixelPool 
network (Altstidl et al. \citeyear{AltNguSchKoeMutEskZan23-IJCNN})%
\footnote{The scale generalisation experimental results for the SESN, DISCO and PixelPool 
  networks are the same as the ones presented in Altstidl et 
  al. (\citeyear{AltNguSchKoeMutEskZan23-IJCNN}),
  obtained from the first author's GitHub repository at 
  https://github.com/taltstidl/scale-equivariant-cnn/blob/main/plots/generalization.csv}.

From the results in Figure~\ref{fig-STIR-scale-selection-appendix},
we can see that the GaussDerResNet is the best performing network, 
achieving the flattest scale generalisation curves due to its provable
scale invariance properties. 
Note, however, that in the GaussDerResNets and GaussDerNets 
the number of parameters has been selected for best possible 
performance and has not been adjusted to keep a similar parameter 
count among the compared models, as used in the comparison by Altstidl et 
al. (\citeyear{AltNguSchKoeMutEskZan23-IJCNN}).

\section{Visualisation of the learned receptive fields}
\label{sec-filt-vis-subsection-appendix}

To provide further insight into the learned representations 
GaussDerResNets, we visualise the effective filters across 
the network layers for the GaussDerResNets with the 
computational primitives of the layers extended to also 
include zero-order terms and trained on the rescaled STL-10 
dataset, illustrating the progression from low-level to intermediate 
and deep hierarchical features. These receptive fields are 
computed as linear combinations of a scale-normalised 
Gaussian derivative kernel basis, here including the
zero-order term and using spatial max pooling in
the training stage.

The discrete kernels constituting the basis for the filters in that 
model are computed using the discrete analogue of Gaussian kernel, 
with the Gaussian derivatives computed using central difference 
operators, as described in Section~3.5 in the main body of the paper. While all the 
scale channels share the same weights, we showcase the filters 
specifically for the scale channel based on the initial scale value 
$\sigma_{0} = 1$. 

The extracted receptive fields are shown in 
Figure~\ref{fig-learned-filters-standard-appendix}.
As can be seen from the figure, some of the filters are similar to directional
derivatives of Gaussian kernels or Laplacians-of-Gaussians.
The variability in the shapes of the filters is, however,
substantially richer than plain Gaussian derivatives or
Laplacians-of-Gaussians. Furthermore, due to the 
inclusion of the zero-order term in the layers of the network, the receptive 
fields are not always DC-balanced, meaning that the masses of the
positive and the negative parts may not always be equal.

{\footnotesize
\bibliographystyle{Frontiers-Harvard}
\bibliography{defs,tlmac}}

\end{document}